\documentclass[12pt]{book}
\usepackage{float,amssymb,amsfonts,epsfig,makeidx}
\usepackage{amsmath,amsthm}
\usepackage{color}
\theoremstyle{plain}
\newtheorem{thm}{Theorem}[chapter]
\newtheorem{theorem}[thm]{Theorem}

\newtheorem{proposition}[thm]{Proposition}

\theoremstyle{definition}

\newtheorem{definition}{Definition}[chapter]

\input xy
\xyoption{matrix}
\xyoption{arrow}
\xyoption{rotate}
\xyoption{cmtip}
\xyoption{graph}
\xyoption{frame}
\def\red{\color{red}}
\def\blue{\color{blue}}

\def\reals{\mathbb{R}}

\def\co{\colon}
\def\mapdef#1#2#3{#1\co #2\rightarrow #3}
\def\transpos#1{#1^{\top}}
\def\s#1{{\cal #1}}
\def\norme#1{\left\|#1\right\|}
\def\remark{\bigskip\noindent{\bf Remark:}\enspace}
\def\smnorme#1{\|#1\|}
\def\lag{\left\langle}
\def\rag{\right\rangle}
\def\mfrac#1{{\mathfrak{#1}}}
\def\Ker{\mathrm{Ker}\,}
\title{\huge 
Spectral  Theory of Unsigned \\
and  Signed Graphs\\
Applications to Graph Clustering:  a Survey\\
}
\author{Jean Gallier\\
Department of Computer and Information Science\\
University of Pennsylvania\\
Philadelphia, PA 19104, USA\\
e-mail: {\tt jean@cis.upenn.edu}\\
\ \\
\copyright\ Jean Gallier}
\begin{document}
\maketitle
\ \vfill\eject\noindent
\def\Degsym{D}%

\
\vspace{3cm}

\noindent
{\bf Abstract:}
This is a survey of the method of graph cuts and its
applications to graph clustering of weighted unsigned and signed
graphs. I  provide a fairly  thorough treatment of the method
of normalized graph cuts, a
deeply original method due to Shi and Malik, including complete
proofs.
I also cover briefly the method of ratio cuts, and show how it can be
viewed as a special case of normalized cuts.
I include the necessary background on graphs and graph Laplacians.
I then explain in detail how the eigenvectors of the graph Laplacian can be used to draw
a graph. This is an attractive application of graph Laplacians. 
The main thrust of this paper is the method of normalized cuts.
I give a detailed account for $K = 2$ clusters, and also for $K > 2$ clusters, based 
on the work of Yu and Shi.
I also show how both graph drawing and normalized cut $K$-clustering
can be easily generalized to handle signed graphs, which
are weighted graphs in which the weight matrix $W$ may have negative
coefficients.  Intuitively, negative coefficients indicate distance or
dissimilarity.
The solution is to replace the degree matrix $\Degsym$
by the matrix $\overline{\Degsym}$ 
in which absolute values of the weights are used, and to replace the
Laplacian $L = \Degsym - W$ by
the signed Laplacian $\overline{L} = \overline{\Degsym} - W$.
The signed Laplacian $\overline{L}$ is always positive semidefinite,
and it may be positive definite (for unbalanced graphs, see Chapter \ref{chap4}).
As far as I know, 
the generalization of  $K$-way normalized clustering to signed graphs
is new.
Finally, I show how the method of ratio cuts, in which a cut is normalized by
the size of the  cluster rather than its volume,  is just  a special
case of normalized cuts. All that needs to be done is to replace
the normalized Laplacian $L_{\mathrm{sym}}$ by the unormalized
Laplacian $L$. This is also true for signed graphs (where we replace 
$\overline{L}_{\mathrm{sym}}$ by $\overline{L}$).

\medskip
Three points that do not appear to have been clearly articulated before are elaborated:
\begin{enumerate}
\item
The solutions of the main optimization problem should be viewed as tuples
in the $K$-fold cartesian product of projective space $\mathbb{RP}^{N-1}$.
\item
When $K > 2$, the solutions of the relaxed problem should be viewed as 
elements of the Grassmannian $G(K,N)$.
\item
Two possible Riemannian distances are available to compare the closeness of solutions:
(a) The distance on $(\mathbb{RP}^{N-1})^K$. (b) The distance on the Grassmannian.
\end{enumerate}

I also clarify what should be the necessary and sufficient conditions for a matrix to
represent a partition of the vertices of a  graph to be clustered.

\tableofcontents
\vfill\eject
\chapter{Introduction}
\label{chap-intro}
\def\Degsym{D}%
In the Fall of 2012, my friend Kurt Reillag suggested that I 
should be ashamed about knowing so little about 
graph Laplacians and normalized graph cuts.
These notes are the result of my efforts to rectify this 
situation.

\begin{figure}[http]
  \begin{center}
 \includegraphics[height=2truein,width=2truein]{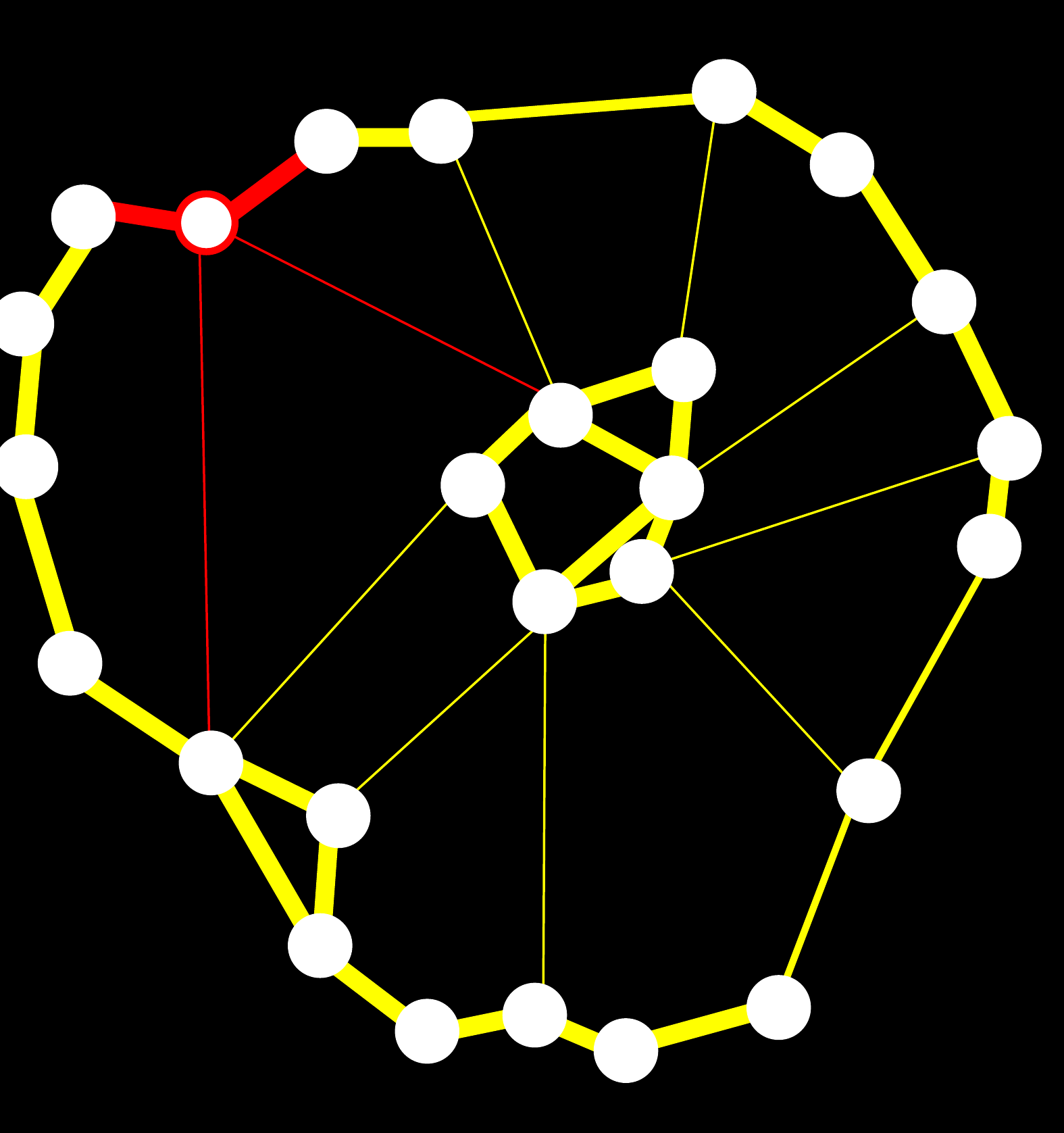}
\hspace{1cm}
 \includegraphics[height=2truein,width=2truein]{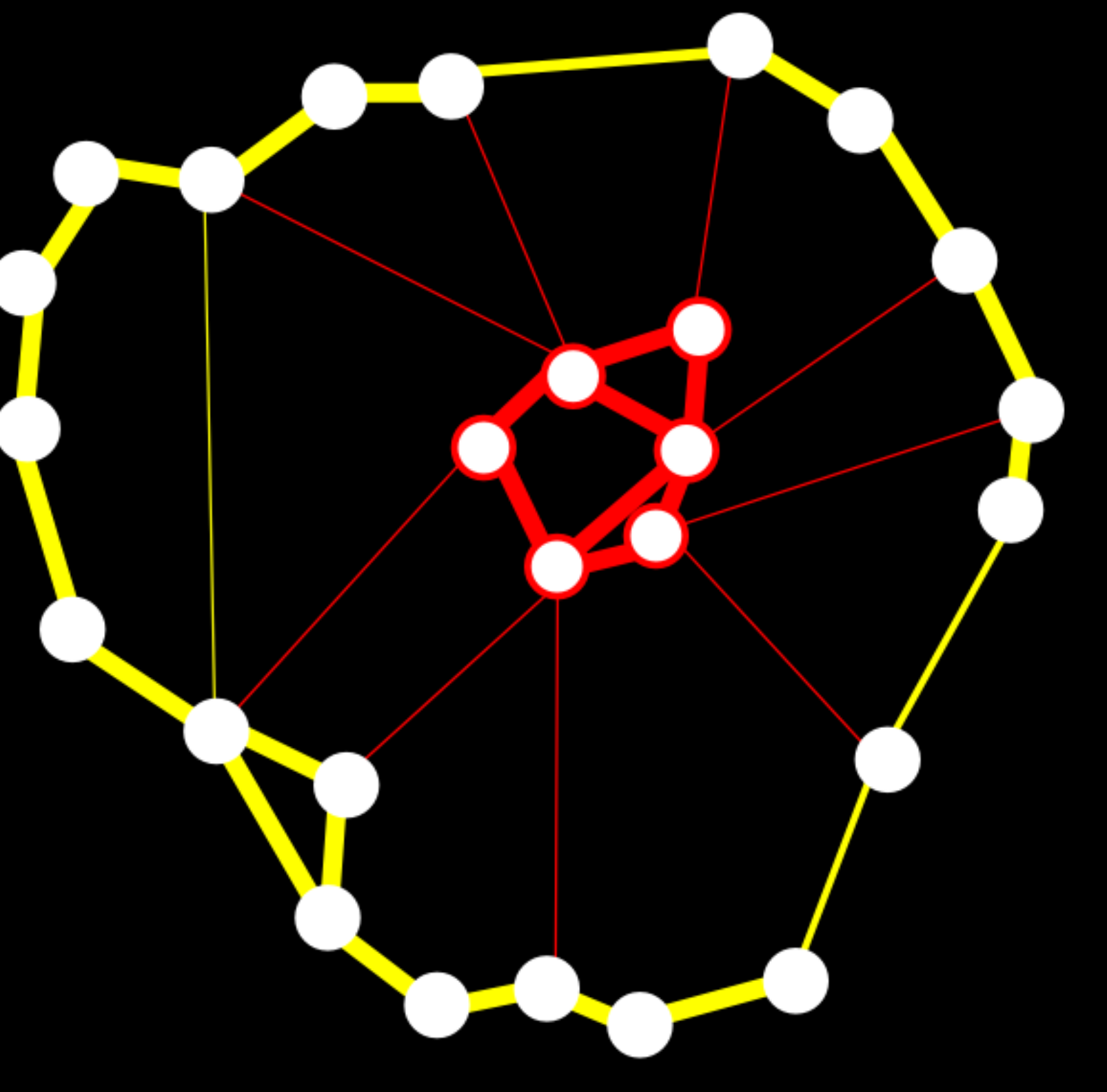}
  \end{center}
  \caption{Degree and volume.}
\label{ncg-fig2a}
\end{figure}

\medskip
I begin with a review of basic notions of graph theory.
Even though the graph Laplacian is fundamentally associated with an
undirected graph, I review the definition of both directed and
undirected graphs. For both directed and undirected graphs, I define
the degree matrix $D$, the incidence matrix $B$, and the
adjacency matrix $A$. 
Then, I define a {\it  weighted graph\/}. This is a pair $(V, W)$,
where $V$ is a finite set of nodes and $W$ is a $m\times m$ symmetric matrix with
nonnegative entries and zero diagonal entries (where $m = |V|$).
For every node $v_i\in V$, the {\it degree\/} $d(v_i)$ (or $d_i$) of $v_i$ is the sum
of  the weights of the  edges adjacent to $v_i$:
\[
d_i = d(v_i) = \sum_{j = 1}^m w_{i\, j}.
\]
The {\it degree matrix\/} is the diagonal matrix
\[
\Degsym = \mathrm{diag}(d_1, \ldots, d_m).
\] 
Given any subset of nodes $A \subseteq V$, we define the 
{\it  volume\/}
$\mathrm{vol}(A)$ of $A$ as the sum of the weights of all edges 
adjacent to nodes in $A$:
\[
\mathrm{vol}(A) =
\sum_{v_i \in A} \sum_{j =  1}^m w_{i\, j}.
\]
The notions of degree and volume are illustrated in Figure \ref{ncg-fig2a}.
\medskip
Given any two subset $A, B\subseteq V$ (not necessarily distinct), we
define
$\mathrm{links}(A, B)$ by
\[
\mathrm{links}(A, B) = \sum_{v_i\in A, v_j\in B} w_{i\, j}.
\]
The quantity
$ \mathrm{links}(A, \overline{A}) =
\mathrm{links}(\overline{A}, A)$
(where $\overline{A} = V - A$ denotes the complement of $A$ in $V$)
measures
how many links escape from $A$ (and $\overline{A}$). 
We define the {\it cut\/} of $A$ as
\[
\mathrm{cut}(A) =
\mathrm{links}(A, \overline{A}).
\]

\begin{figure}[H]
  \begin{center}
 \includegraphics[height=2truein,width=2truein]{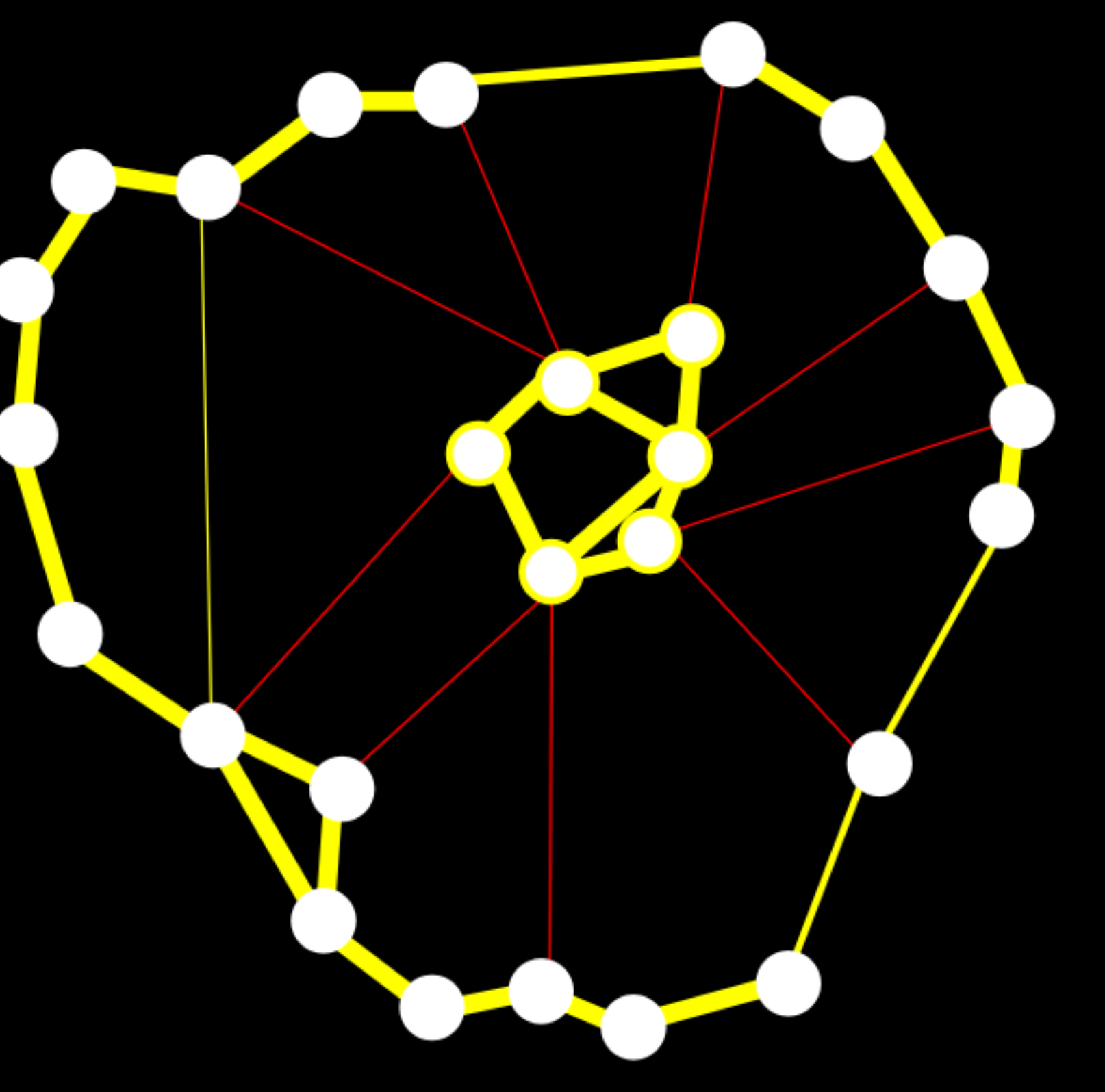}
  \end{center}
  \caption{A Cut involving the set of nodes in the center and the
    nodes on the perimeter.}
\label{ncg-fig3a}
\end{figure}
The notions of cut is illustrated in Figure \ref{ncg-fig3a}.
The above concepts play a crucial role in the theory of
normalized cuts. Then, I introduce the (unnormalized) {\it graph Laplacian\/} $L$ of
a directed graph $G$ in an  ``old-fashion,'' 
by showing that for any orientation of a graph
$G$, 
\[
B\transpos{B} = D - A = L
\]
is an invariant. I also define the (unnormalized) {\it graph
  Laplacian\/} $L$ of a weighted graph $G = (V, W)$ as $L = D - W$.
I show that the notion of incidence matrix can be generalized
to weighted graphs in a simple way.
For any graph $G^{\sigma}$ obtained by orienting the underlying
graph of  a weighted graph $G = (V, W$), there is an incidence matrix  
$B^{\sigma}$  such that
\[
B^{\sigma}\transpos{(B^{\sigma})} = D - W = L.
\]
I also prove that
\[
\transpos{x} L x =
\frac{1}{2}\sum_{i, j = 1}^m w_{i\, j} (x_i - x_j)^2
\quad\mathrm{for\ all}\> x\in \reals^m.
\]
Consequently, $\transpos{x} L x$ does not depend on the
diagonal entries in $W$, and if $w_{i\, j} \geq 0$ for all $i, j\in
\{1, \ldots,m\}$, then $L$ is positive semidefinite.
Then, if $W$ consists of nonnegative entries, 
the eigenvalues  $0 = \lambda_1 \leq \lambda_2 \leq  \ldots \leq 
\lambda_m$ of $L$ are real and nonnegative, and there is an
orthonormal basis of eigenvectors of $L$.
I show that the
number of connected components of the graph $G = (V,W)$
is equal to the dimension of the kernel of $L$, which is also equal to
the dimension of the kernel of the transpose $\transpos{(B^{\sigma})}$
of  any incidence matrix $B^{\sigma}$ obtained by orienting the
underlying graph of $G$.

\medskip
I also define the normalized graph Laplacians $L_{\mathrm{sym}}$ and $L_{\mathrm{rw}}$, given by
\begin{align*}
L_{\mathrm{sym}}& = \Degsym^{-1/2} L \Degsym^{-1/2} = I - \Degsym^{-1/2} W \Degsym^{-1/2}  \\
L_{\mathrm{rw}}& = \Degsym^{-1} L = I - \Degsym^{-1} W,
\end{align*}
and prove some simple properties relating the eigenvalues and the
eigenvectors of $L$, $L_{\mathrm{sym}}$ and $L_{\mathrm{rw}}$.
These normalized graph Laplacians show up when dealing with normalized cuts.

\medskip
Next, I turn to {\it graph drawings\/} (Chapter \ref{chap2}).
Graph drawing is a very
attractive application of so-called spectral techniques, which is a
fancy way of saying that that
eigenvalues and eigenvectors of the graph Laplacian are used.
Furthermore, it turns out that graph clustering using normalized cuts
can be cast as a certain type of graph drawing.

\medskip
Given an undirected graph $G = (V, E)$, with $|V| = m$, 
we would like to draw $G$ in $\reals^n$ for $n$ (much) smaller than
$m$. 
The idea is to assign a point $\rho(v_i)$ in $\reals^n$ to the vertex $v_i\in
V$, for every $v_i \in V$, 
and to draw a line segment between the points $\rho(v_i)$ and
$\rho(v_j)$.  Thus, a {\it graph drawing\/} is a function
$\mapdef{\rho}{V}{\reals^n}$.

\medskip
We define the  {\it matrix of a graph drawing $\rho$ 
(in  $\reals^n$)\/}  as a $m \times n$ matrix $R$ whose $i$th row consists
of the row vector $\rho(v_i)$ corresponding to the point representing $v_i$ in
$\reals^n$.
Typically, we want $n < m$; in fact $n$ should be much smaller than $m$.

\medskip
Since there are infinitely many graph drawings, it is desirable to
have some criterion to decide which graph is better than another.
Inspired by a physical model in which the edges are springs, 
it is natural to consider a representation to be better if it
requires the springs to be less extended. 
We can formalize this by
defining the {\it energy\/} of a drawing $R$ by
\[
\s{E}(R) = \sum_{\{v_i, v_j\}\in E} \norme{\rho(v_i) - \rho(v_j)}^2,
\]
where $\rho(v_i)$ is the $i$th row of $R$ and 
$\norme{\rho(v_i) - \rho(v_j)}^2$
is the square of the Euclidean length of the line segment
joining $\rho(v_i)$  and  $\rho(v_j)$.

\medskip
Then, ``good drawings''  are drawings that minimize the energy function
$\s{E}$.
Of course, the trivial representation corresponding to the zero matrix
is optimum, so we need to impose extra constraints to rule out the
trivial solution.

\medskip
We can consider the more general situation where the springs are not
necessarily identical. This can be modeled by a symmetric weight (or
stiffness)  matrix $W = (w_{i j})$, with $w_{i j} \geq
0$.
In this case,  our energy function becomes
\[
\s{E}(R) = \sum_{\{v_i, v_j\}\in E} w_{i j} \norme{\rho(v_i) - \rho(v_j)}^2.
\]

Following Godsil and Royle \cite{Godsil}, 
we prove that 
\[
\s{E}(R) = \mathrm{tr}(\transpos{R} L R),
\]
where 
\[
L = \Degsym - W,
\]
is the familiar unnormalized Laplacian matrix associated with $W$,
and where $\Degsym$ is the degree matrix associated with $W$.

\medskip
It can be shown that there is no loss in generality in 
assuming that the columns of $R$ are pairwise orthogonal
and that they have unit length. Such a matrix satisfies the equation
$\transpos{R} R = I$ and the corresponding drawing is called an
{\it orthogonal drawing\/}. This condition also rules out trivial
drawings.

\medskip
Then, I prove the main theorem about graph drawings
(Theorem \ref{graphdraw}), which essentially says that
the matrix $R$ of the desired graph drawing is constituted by the
$n$ eigenvectors of $L$ associated with the
smallest nonzero $n$ eigenvalues of $L$. 
We give a number examples of graph drawings, many of which are borrowed
or adapted from Spielman \cite{Spielman}.

\medskip
The next chapter (Chapter \ref{chap3}) 
contains  the ``meat'' of this document. This chapter is devoted to 
the method of normalized graph cuts for graph clustering.
This beautiful and deeply original method first published in
Shi and Malik \cite{ShiMalik},
has now come to be a ``textbook chapter''
of computer vision and machine learning. It was invented  by Jianbo Shi 
and Jitendra Malik, and was the main topic of Shi's dissertation.
This method was extended to $K \geq 3$ clusters by Stella
 Yu in her dissertation \cite{Yu}, and is also the subject of Yu and Shi
\cite{YuShi2003}.

\medskip
Given a set of data, the goal of clustering is to partition the data
into different groups according to their similarities. When the data
is given in terms of a similarity graph $G$, where the weight $w_{i\, j}$
between two nodes $v_i$ and $v_j$ is a measure of similarity of
$v_i$ and $v_j$, the problem can be stated as follows:
Find a partition $(A_1, \ldots, A_K)$ of the set of nodes $V$ into
different groups such   that the edges between different groups have
very low weight (which indicates that the points in different clusters
 are dissimilar),  and the edges within a group have high weight
(which indicates that points within the same cluster are similar).

\medskip
The above graph clustering problem can be formalized as an
optimization problem, using the notion of cut mentioned earlier.
If we want to partition $V$ into $K$ clusters, we can do so by finding
a partition ($A_1, \ldots, A_K$) that  minimizes the quantity
\[
\mathrm{cut}(A_1, \ldots, A_K) = \frac{1}{2} \sum_{i = 1}^K
\mathrm{cut}(A_i)
= \frac{1}{2} \sum_{i = 1}^K \mathrm{links}(A_i, \overline{A}_i).  
\]

For $K = 2$,  the mincut problem is a classical
problem that can be solved efficiently, but in practice, it does not
yield satisfactory partitions. Indeed, in many cases, the mincut
solution separates one vertex from the rest of the graph.  What we
need is to design our cost function in such a way that it keeps the
subsets $A_i$ ``reasonably large'' (reasonably balanced).

\begin{figure}[http]
  \begin{center}
 \includegraphics[height=2.5truein,width=2.8truein]{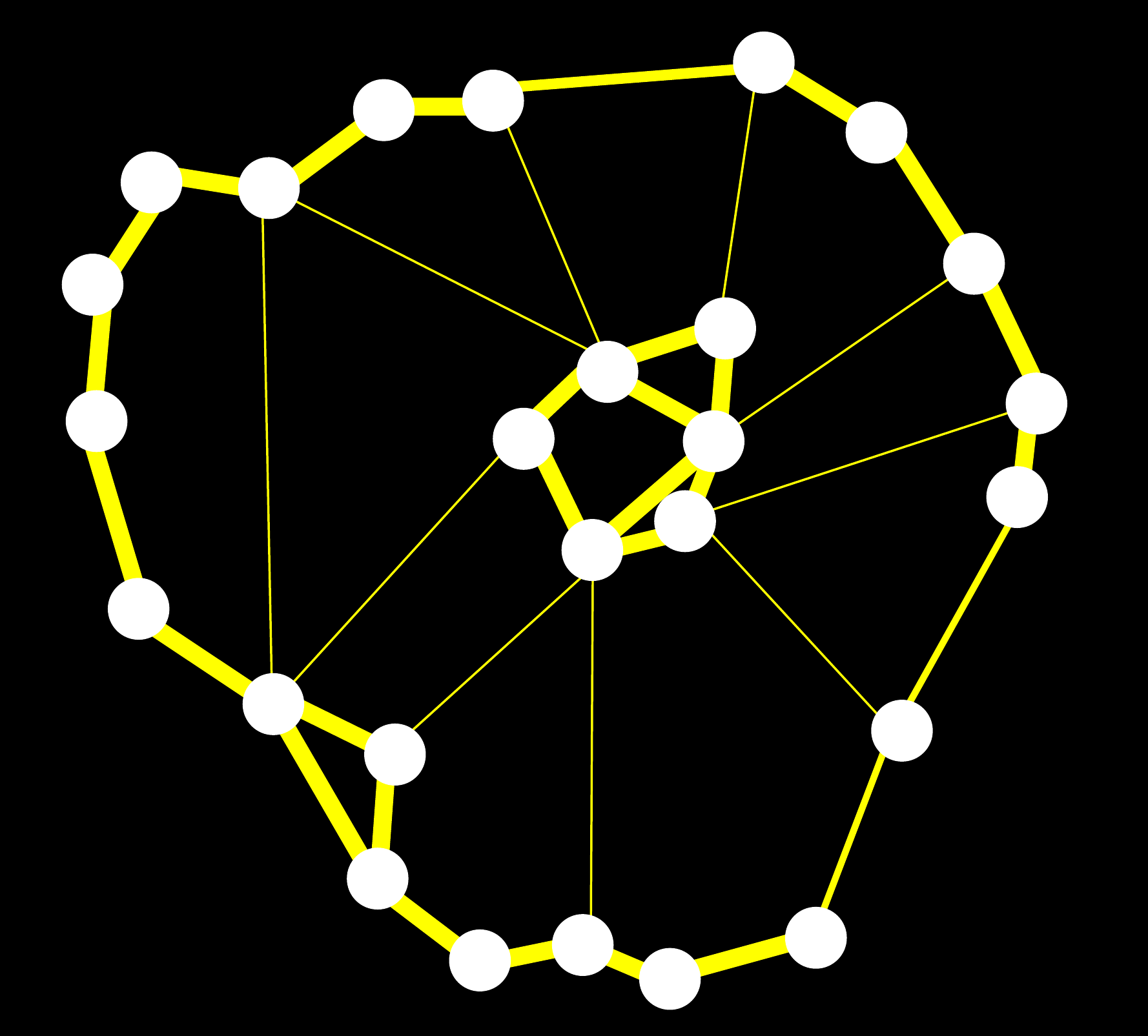}
\hspace{0.5cm}
 \includegraphics[height=2.5truein,width=2.8truein]{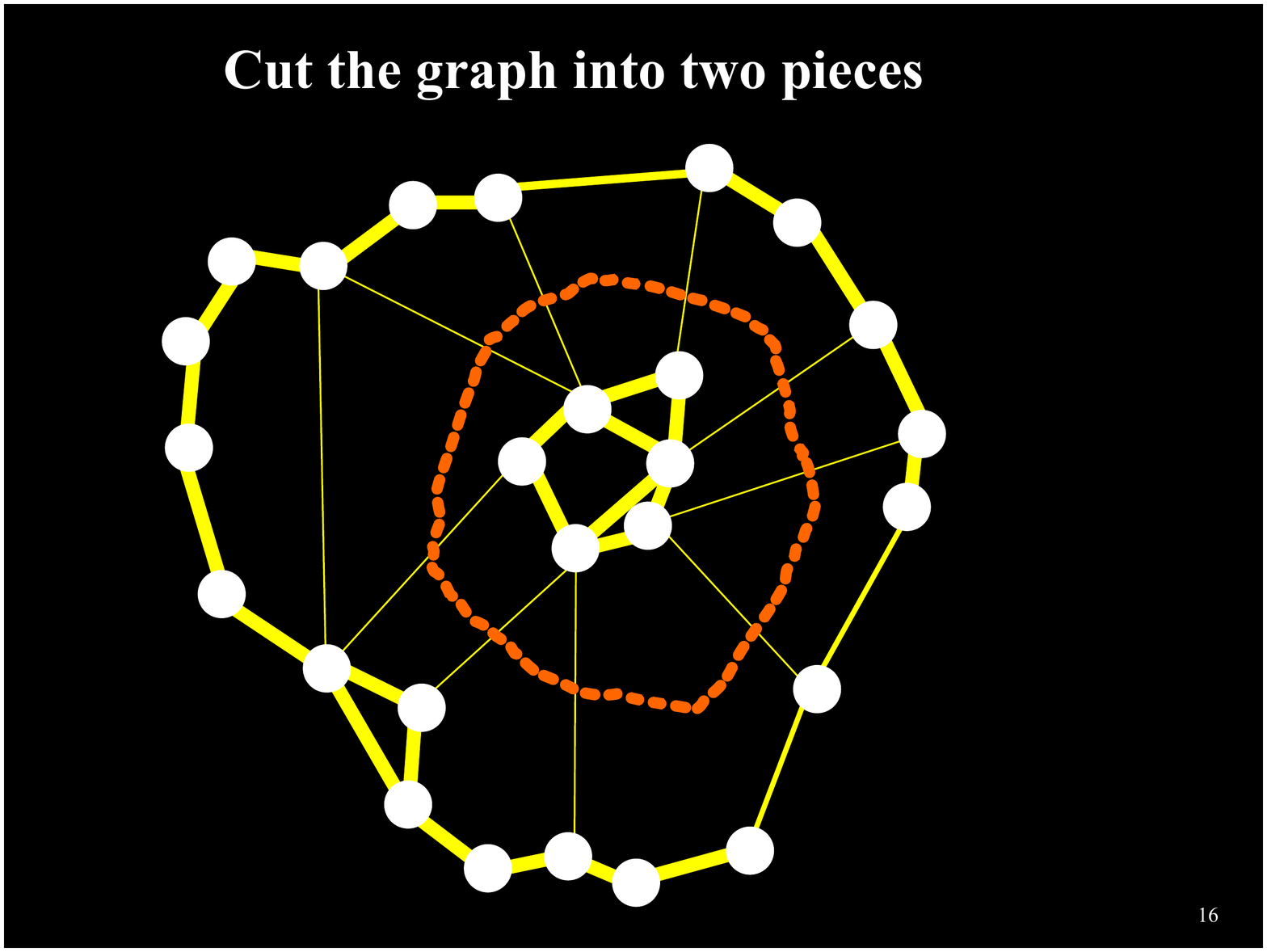}
  \end{center}
\caption{A weighted graph and its partition into two clusters.}
\label{ncg-fig4a}
\end{figure}

\medskip
An example of a weighted graph and a partition of
its nodes into two clusters is shown in Figure \ref{ncg-fig4a}.

\medskip
A way to get around this problem is to normalize the cuts by dividing
by some measure of each subset $A_i$. 
A solution using the volume $\mathrm{vol}(A_i)$ of $A_i$
(for $K = 2$) was proposed and
investigated in a seminal paper of  Shi and Malik \cite{ShiMalik}.
Subsequently, Yu (in her dissertation \cite{Yu}) and Yu and Shi
\cite{YuShi2003} extended the method to $K > 2$ clusters.
The idea is  to minimize the cost function
\[
\mathrm{Ncut}(A_1, \ldots, A_K) = 
 \sum_{i = 1}^K \frac{\mathrm{links}(A_i, \overline{A_i})}{\mathrm{vol}(A_i)}
= \sum_{i = 1}^K 
\frac{\mathrm{cut}(A_i, \overline{A_i})}{\mathrm{vol}(A_i)}.
\]

\medskip
The first step is to express our  optimization problem in matrix form.
In the case of two clusters, a single vector $Xx$ can be used to describe the
partition $(A_1, A_2)  = (A, \overline{A})$. We need to choose the structure of this vector
in such a way that 
\[
\mathrm{Ncut}(A, \overline{A}) = \frac{\transpos{X} L X}{\transpos{X} \Degsym X},
\]
where the term on the right-hand side is a Rayleigh ratio.

\medskip
After careful study of the original papers, I discovered various
facts that were implicit in these works, but I feel are important to
be pointed out explicitly.

\medskip
First, I realized 
that it is  important to pick a vector representation 
which is invariant under multiplication by a nonzero scalar,
because the Rayleigh ratio is scale-invariant, and it is crucial to take
advantage of this fact to make the denominator go away.
This implies that {\it the solutions $X$ are points in the projective
space $\mathbb{RP}^{N - 1}$\/}. This was my first revelation.

\medskip
Let $N = |V|$ be the number of nodes in the graph $G$.
In view of the desire for a scale-invariant representation,
it is natural to assume that the  vector $X$ is of the form
\[
X = (x_1, \ldots, x_N),
\]
where $x_i \in \{a, b\}$ for $i = 1, \ldots, N$, 
for any two distinct  real numbers $a, b$. 
This is an indicator vector in
the sense that, for $i = 1, \ldots, N$,
\[
x_i =
\begin{cases}
a & \text{if $v_i \in A$} \\
b & \text{if $v_i \notin A$} .
\end{cases}
\]

The  choice $a = +1, b = -1$ is natural, but premature.
The correct interpretation is really to 
view $X$ as a representative of a point 
in the real projective space $\mathbb{RP}^{N-1}$,  
namely the point $\mathbb{P}(X)$ of homogeneous
coordinates $(x_1\co \cdots \co x_N)$.

\medskip
Let  $d = \transpos{\mathbf{1}} \Degsym \mathbf{1}$ and $\alpha =
\mathrm{vol}(A)$, where $\mathbf{1}$ denotes the vector whose
components are all equal to $1$.
I prove that
\[
\mathrm{Ncut}(A, \overline{A}) = \frac{\transpos{X} L X}{\transpos{X} \Degsym X}
\]
holds iff the following condition holds:
\begin{equation}
a \alpha + b(d - \alpha) = 0.
\tag{$\dagger$}
\end{equation}
Note that condition $(\dagger)$ applied to a vector $X$ whose
components are $a$ or $b$ is equivalent to the fact that $X$
is orthogonal to $\Degsym \mathbf{1}$, since
\[
\transpos{X} \Degsym \mathbf{1} =  \alpha  a+ (d - \alpha) b,
\]
where $\alpha = \mathrm{vol}(\{v_i\in V \mid x_i = a\})$.

\medskip
If we let
\[
\s{X} = \big\{
(x_1, \ldots, x_N) \mid x_i \in \{a, b\}, \> a, b\in \reals,\> a,
b\not = 0
\big\},
\]
our solution set is
\[
\s{K}  = \big\{
X  \in\s{X}  \mid \transpos{X}  \Degsym\mathbf{1} = 0
\big\}.
\]
Actually, to be perfectly rigorous,  we are looking for solutions in
$\mathbb{RP}^{N-1}$, so our solution set is really
\[
\mathbb{P}(\s{K})  = \big\{
(x_1\co \cdots\co x_N) \in \mathbb{RP}^{N-1}\mid
(x_1, \ldots, x_N) \in \s{K}
\big\}.
\]
Consequently, our minimization problem can be stated as follows:

\medskip\noindent
{\bf Problem PNC1}
\begin{align*}
& \mathrm{minimize}     &  & 
\frac{\transpos{X} L X}{\transpos{X} \Degsym X} & &  &  &\\
& \mathrm{subject\ to} &  &
\transpos{X} \Degsym\mathbf{1} = 0,  & &  X\in \s{X}.      
\end{align*}

It is understood that the solutions are  points $\mathbb{P}(X)$
in $\mathbb{RP}^{N-1}$.

\medskip 
Since the Rayleigh ratio and the constraints 
$\transpos{X}\Degsym\mathbf{1} = 0$ and $X\in \s{X}$ 
are scale-invariant,
we are led to the following formulation of our problem:

\medskip\noindent
{\bf Problem PNC2}
\begin{align*}
& \mathrm{minimize}     &  & 
\transpos{X} L X & &  &  &\\
& \mathrm{subject\ to} &  & \transpos{X} \Degsym X = 1, &&
 \transpos{X} \Degsym\mathbf{1} = 0, && X\in \s{X}.   
\end{align*}

\medskip
Because  problem PNC2 requires the constraint 
$\transpos{X} \Degsym X =1$ to be satisfied,
it does not have the same set of solutions as  problem PNC1 , but
PNC2  and PNC1 are equivalent  in the sense that  they 
have the same set of minimal solutions as points $\mathbb{P}(X)
\in\mathbb{RP}^{N-1}$ given by their homogeneous coordinates $X$.
More precisely, if $X$ is any minimal solution of PNC1, then $X/(\transpos{X} \Degsym
X)^{1/2}$ is a minimal solution of PNC2 (with the same minimal value
for the objective functions), and if $X$ is a minimal solution of
PNC2, then $\lambda X$ is a minimal solution for PNC1 for all
$\lambda\not= 0$ (with the same minimal value
for the objective functions).

\medskip
Now, as in the classical papers, we consider the relaxation of  the
above problem obtained by dropping the condition that $X\in \s{X}$,
and proceed as usual. However,  having found a solution $Z$
to the relaxed problem, we need to find a discrete solution $X$ such
that $d(X, Z)$ is minimum in $\mathbb{RP}^{N-1}$.
All this is presented in Section \ref{ch3-sec2}.

\medskip
If the number of clusters $K$ is at least $3$, then we need
to choose  a matrix representation for partitions on the set of vertices.
It is important that such a representation be scale-invariant, and 
it is also necessary to state necessary and sufficient conditions
for such matrices to represent a partition
(to the best of our knowledge, these points are not clearly articulated 
in the literature).

\medskip
We describe a  partition $(A_1, \ldots, A_K)$ of the set of nodes $V$ by
an $N\times K$ matrix  $X = [X^1 \cdots X^K]$
whose columns $X^1, \ldots,
X^K$ are indicator vectors of the partition $(A_1, \ldots, A_K)$.
Inspired by what we did when  $K = 2$, 
we assume that the  vector $X^j$ is of the form
\[
X^j = (x_1^j, \ldots, x_N^j),
\]
where $x_i^j \in \{a_j, b_j\}$ for $j = 1, \ldots, K$ and
$i = 1, \ldots, N$, and
where $a_j, b_j$ are 
any  two distinct  real numbers.
The vector $X^j$  is an indicator vector for $A_j$ in
the sense that, for $i = 1, \ldots, N$,
\[
x_i^j =
\begin{cases}
a_j & \text{if $v_i \in A_j$} \\
b_j & \text{if $v_i \notin A_j$} .
\end{cases}
\]

The choice $\{a_j, b_j\} = \{0, 1\}$ for $j = 1, \ldots, K$ is
natural, but premature. I show that if we pick $b_i = 0$, then we have
\[
\frac{\mathrm{cut}(A_j, \overline{A_j})}{\mathrm{vol}(A_j)} 
= \frac{\transpos{(X^j)} L X^j}{\transpos{(X^j)}\Degsym X^j} \quad j =
1, \ldots, K, 
\]
which  implies that
\[
\mathrm{Ncut}(A_1, \ldots, A_K) 
= \sum_{j = 1}^K 
\frac{\mathrm{cut}(A_j, \overline{A_j})}{\mathrm{vol}(A_j)}
= \sum_{j = 1}^K 
\frac{\transpos{(X^j)} L X^j}{\transpos{(X^j)}\Degsym X^j}.
\]
Then, I give necessary and sufficient conditions for a matrix $X$ to
represent a partition. 

\medskip
If we let
\[
\s{X}  = \Big\{[X^1\> \ldots \> X^K] \mid
X^j = a_j(x_1^j, \ldots, x_N^j) , \>
x_i^j \in \{1, 0\},
 a_j\in \reals, \> X^j \not= 0
\Big\}
\]
(note that the condition $X^j \not= 0$ implies that $a_j \not= 0$),
then the set of matrices representing partitions of $V$ into $K$
blocks is
\begin{align*}
& & &\s{K}  = \Big\{ X = [X^1 \> \cdots \> X^K] \quad \mid & &  X\in\s{X},  &&\\
         & & &  & &  \transpos{(X^i)} \Degsym X^j = 0, \quad 1\leq i, j \leq K,\> 
i\not= j, && \quad\quad\quad\quad\quad\\
&  & & & & 
 X (\transpos{X} X)^{-1} \transpos{X} \mathbf{1} = \mathbf{1}\Big\}. && 
\end{align*}

As in the case $K = 2$, to be rigorous, the {\it solution are really
$K$-tuples of points in $\mathbb{RP}^{N-1}$\/}, so our solution set is
really
\[
\mathbb{P}(\s{K})  = \Big\{(\mathbb{P}(X^1), \ldots, \mathbb{P}(X^K)) \mid
[X^1 \> \cdots \> X^K] \in \s{K} 
\Big\}.
\]

\remark
For any $X\in \s{X}$, 
the condition  $X (\transpos{X} X)^{-1} \transpos{X} \mathbf{1} =
\mathbf{1}$ is redundant.
However, when we relax the problem and drop the condition 
$X\in \s{X}$, the condition  $X (\transpos{X} X)^{-1} \transpos{X} \mathbf{1} =
\mathbf{1}$ captures the fact $\mathbf{1}$ should be in the range of $X$. 

\medskip
In view of the above, we have our first formulation of $K$-way clustering
of a graph using normalized cuts, called problem PNC1 
(the notation PNCX  is used in  Yu \cite{Yu}, Section 2.1):

\medskip\noindent
{\bf $K$-way Clustering of a graph using Normalized Cut, Version 1: \\
Problem PNC1}

\begin{align*}
& \mathrm{minimize}     &  &  \sum_{j = 1}^K 
\frac{\transpos{(X^j)} L X^j}{\transpos{(X^j)}\Degsym X^j}& &  &  &\\
& \mathrm{subject\ to} &  & 
 \transpos{(X^i)} \Degsym X^j = 0, \quad 1\leq i, j \leq K,\> 
i\not= j,  & &  & & \\
& & & 
 X (\transpos{X} X)^{-1} \transpos{X} \mathbf{1} = \mathbf{1},  & & X\in \s{X}. & & 
\end{align*}

As in the case $K = 2$, the solutions that we are seeking are $K$-tuples 
$(\mathbb{P}(X^1), \ldots, \mathbb{P}(X^K))$ of points  in
$\mathbb{RP}^{N-1}$ determined by
their  homogeneous coordinates $X^1, \ldots, X^K$.

\medskip
Then, step by step, we transform problem PNC1 into an equivalent
problem PNC2. We eventually relax PNC1 into $(*_1)$ and PNC2 into
$(*_2)$, by dropping the condition that
$X\in \s{X}$.

\medskip
Our second revelation is that the relaxation $(*_2)$ of 
version 2 of  our minimization problem (PNC2), which is equivalent to
version 1, reveals that that the solutions of the relaxed problem
$(*_2)$ are members of the {\it Grassmannian\/} $G(K, N)$.

\medskip
This leads us to our third revelation:
{\it we have  two choices of metrics to compare solutions\/}:
(1) a metric on
$(\mathbb{RP}^{N - 1})^K$; (2) a metric on $G(K, N)$.
We  discuss the first choice, which is the choice implicitly
adopted by Shi and Yu. However, in approximating
a discrete solution $X$ by a solution $Z$ of problem $(*_1)$
we allow more general transformations of the form
$Q = R\Lambda$, where $R\in \mathbf{O}(K)$, and
$\Lambda$ is a diagonal invertible matrix. 
Thus we seek $R$ and $\Lambda$ to minimize 
$\norme{X - ZR\Lambda}_F$. This yields better 
discrete  solutions $X$. 

\medskip
In Chapter \ref{chap4}, I show how both the spectral method for graph
drawing and the normalized-cut method for $K$ clusters generalize to
signed graphs, which are graphs whose weight matrix $W$ may contain
negative entries. The intuition is that
negative weights indicate dissimilarity or distance. 

\medskip
The first obstacle is that the degree matrix may now contain negative
entries. As a consequence, the Laplacian $L$  may no longer be positive
semidefinite, and worse, $\Degsym^{-1/2}$ may not exist.

\medskip
A simple remedy is to use the absolute values of the  weights in the degree
matrix!  We denote this matrix by $\overline{\Degsym}$, and define
the {\it signed Laplacian\/} as
$\overline{L}  = \overline{\Degsym} - W$.
The idea to use positive degrees of nodes in the degree
matrix of a signed graph with weights $(-1, 0, 1)$ occurs in Hou
\cite{Hou}. The natural step of  using absolute values of weights in
the degree matrix is taken by Kolluri,  Shewchuk and  O'Brien \cite{Kolluri:2004:SSR} 
and Kunegis et al. \cite{kunegis:spectral}. 

\medskip
As we will see, this trick allows the whole machinery that we
have presented to be used to attack the problem of clustering signed
graphs using normalized cuts.  

\medskip
As in the case of unsigned weighted graphs, for any 
orientation $G^{\sigma}$ of the underlying graph of a signed graph $G = (V, W)$, 
there is an incidence  matrix $B^{\sigma}$ such that
\[
B^{\sigma} \transpos{(B^{\sigma})}= \overline{\Degsym} - W = \overline{L}.
\]
Consequently, $B^{\sigma}\transpos{(B^{\sigma})}$ is 
independent of the orientation of the underlying graph of  $G$ and
$\overline{L} = \overline{\Degsym} -W$ is symmetric and
positive semidefinite.
I also show that
\[
\transpos{x} \overline{L} x =
\frac{1}{2}\sum_{i, j = 1}^m |w_{i  j}| (x_i - \mathrm{sgn}(w_{i j}) x_j)^2
\quad\mathrm{for\ all}\> x\in \reals^m.
\]

\medskip
As in Section \ref{ch3-sec3}, given a partition of $V$ into $K$
clusters $(A_1, \ldots, A_K)$, if we represent the $j$th block of
this partition by a vector $X^j$ such that
\[
X^j_i = 
\begin{cases}
a_j & \text{if $v_i \in A_j$} \\
0 &  \text{if $v_i \notin A_j$} ,
\end{cases}
\]
for some $a_j \not= 0$, then the following result holds:
For any vector $X^j$ representing the $j$th block of a partition 
$(A_1, \ldots, A_K)$ of $V$, we have
\[
\transpos{(X^j)} \overline{L} X^j =
a_j^2(\mathrm{cut}(A_j, \overline{A_j}) + 2 \mathrm{links}^-(A_j, A_j)).
\]

The above  suggests defining the key notion of  signed normalized cut:
The {\it signed normalized cut\/}
$\mathrm{sNcut}(A_1, \ldots, A_K)$ of the
partition $(A_1, \ldots, A_K)$ is defined as
\[
\mathrm{sNcut}(A_1, \ldots, A_K) = \sum_{j = 1}^K
\frac{\mathrm{cut}(A_j, \overline{A_j})}{\mathrm{vol}(A_j)} + 
2 \sum_{j = 1}^K\frac{\mathrm{links}^-(A_j, A_j)}{\mathrm{vol}(A_j)}.
\] 
Our definition of a signed normalized cut appears to
be novel.

\medskip
Based on previous computations, we have
\[
\mathrm{sNcut}(A_1, \ldots, A_K) = 
\sum_{j = 1}^K \frac{\transpos{(X^j)} \overline{L} X^j} 
  {\transpos{(X^j)} \overline{\Degsym} X^j},
\]
where $X$ is the $N\times K$ matrix whose $j$th column is $X^j$.

\medskip
Observe that minimizing $\mathrm{sNcut}(A_1, \ldots, A_K)$ amounts to 
minimizing the number of positive and negative edges between clusters,
and also minimizing the number of negative edges within clusters.
This second minimization captures the intuition that nodes connected
by a negative edge should not be together  (they do not ``like''
each other; they should be far from each other). 
It would be preferable if the notion of signed cut only took into
account the contribution $\mathrm{links}^+(A_j, \overline{A_j})$ 
of the positively weighted edges between disjoint clusters, but 
we have not found a way to achieve this.

\medskip
Since 
\[
\mathrm{sNcut}(A_1, \ldots, A_K) =  
\sum_{j = 1}^K \frac{\transpos{(X^j)} \overline{L} X^j} 
  {\transpos{(X^j)} \overline{\Degsym} X^j},
\]
the whole machinery of Sections \ref{ch3-sec3} and \ref{ch3-sec5}  can
be applied with $\Degsym$ replaced by $\overline{\Degsym}$
and $L$ replaced by $\overline{L}$. However, there is a new
phenomenon, which is that $\overline{L}$ may be positive definite.
As a consequence, $\mathbf{1}$ is not always an eigenvector of
$\overline{L}$.  

\medskip
Following Kunegis et al. \cite{kunegis:spectral},
we show that the signed Laplacian $\overline{L}$ is positive definite
iff $G$ is unbalanced, which means that it contains some cycle with an
odd number of negative edges. We also characterize when a graph is
balanced in terms of  the kernel of the transpose $\transpos{B}$
of any of its incidence matrices. 

\medskip
To generalize the graph drawing method to signed graphs,
we explain that if  the energy function $\s{E}(R)$ of a graph drawing
is redefined to be
\[
\s{E}(R) = \sum_{\{v_i, v_j\}\in E} |w_{i j}| \norme{\rho(v_i) -
  \mathrm{sgn}(w_{i j})\rho(v_j)}^2,
\]
then we obtain orthogonal graph drawings of minimal energy,
and we give some examples.

\medskip
We conclude this survey with a short chapter on
graph clustering using ratio cuts.
The idea of ratio cut is to replace the volume $\mathrm{vol}(A_j)$ of
each block $A_j$ of the partition by its size, $|A_j|$
(the number of nodes in $A_j$).
Given an unsigned graph $(V, W)$, 
the {\it ratio cut\/}  $\mathrm{Rcut}(A_1, \ldots, A_K)$ of the 
partition $(A_1, \ldots, A_K)$ is defined as
\[
\mathrm{Rcut}(A_1, \ldots, A_K) = 
\sum_{i = 1}^K \frac{\mathrm{cut}(A_j, \overline{A}_j)}{|A_j|}.
\]
If we represent the $j$th block of
this partition by a vector $X^j$ such that
\[
X^j_i = 
\begin{cases}
a_j & \text{if $v_i \in A_j$} \\
0 &  \text{if $v_i \notin A_j$} ,
\end{cases}
\]
for some $a_j \not= 0$, then we obtain
\[
\mathrm{Rcut}(A_1, \ldots, A_K) = 
\sum_{i = 1}^K \frac{\mathrm{cut}(A_j, \overline{A}_j)}{|A_j|}
= \sum_{i = 1}^K \frac{\transpos{(X^j)} L X^j}{\transpos{(X^j)}  X^j}.
\]
On the other hand, the normalized cut is given by
\[
\mathrm{Ncut}(A_1, \ldots, A_K) = 
\sum_{i = 1}^K \frac{\mathrm{cut}(A_j, \overline{A}_j)}{\mathrm{vol}(A_j)}
= \sum_{i = 1}^K \frac{\transpos{(X^j)} L X^j}{\transpos{(X^j)}\Degsym  X^j}.
\]
Therefore, ratio cut is the special case of normalized cut where
$\Degsym = I$! 
Consequently, all that needs to be done is to replace
the normalized Laplacian $L_{\mathrm{sym}}$ by the unormalized
Laplacian $L$ (and omit  the step of considering Problem $(**_1)$). 

\medskip
In the case of signed graphs, we define the
{\it signed ratio cut\/}
$\mathrm{sRcut}(A_1, \ldots, A_K)$ of the
partition $(A_1, \ldots, A_K)$  as
\[
\mathrm{sRcut}(A_1, \ldots, A_K) = \sum_{j = 1}^K
\frac{\mathrm{cut}(A_j, \overline{A_j})}{|A_j|} + 
2 \sum_{j = 1}^K\frac{\mathrm{links}^-(A_j, A_j)}{|A_j|}.
\] 
Since we still have
\[
\transpos{(X^j)} \overline{L} X^j =
a_j^2(\mathrm{cut}(A_j, \overline{A_j}) + 2 \mathrm{links}^-(A_j, A_j)),
\]
we obtain
\[
\mathrm{sRcut}(A_1, \ldots, A_K) = 
\sum_{j = 1}^K \frac{\transpos{(X^j)} \overline{L} X^j} 
{\transpos{(X^j)}  X^j}.
\]
Therefore, this is similar to the case of unsigned graphs, with $L$
replaced with $\overline{L}$. The same algorithm applies, but as in
Chapter \ref{chap4}, the signed Laplacian $\overline{L}$
is positive definite iff $G$ is unbalanced.

\medskip
Some of the most technical material on the Rayleigh ratio, which is needed
for some proofs in Chapter \ref{chap2}, is the object of Appendix
\ref{Rayleigh-Ritz}. Appendix \ref{ch3-sec6} may seem a bit out of
place.
Its purpose is to explain how 
to define a metric on the projective space $\mathbb{RP}^n$. For this, we
need to review a few notions of differential geometry. 

\medskip
I hope that these notes will  make it easier for people to
become familiar with the wonderful theory of normalized graph cuts.
As far as I know, except for a short section in one of Gilbert
Strang's book, and  von Luxburg \cite{Luxburg} excellent survey
on spectral clustering, 
there is no comprehensive  writing on the topic
of graph cuts.

\chapter{Graphs  and Graph Laplacians; Basic Facts}
\label{chap1}
\section[Directed Graphs, Undirected Graphs, Weighted Graphs]
{Directed Graphs, Undirected Graphs, Incidence Matrices,
Adjacency Matrices, Weighted Graphs}
\label{ch1-sec1}%
\begin{definition}
\label{dirgraph}
A {\it directed graph\/} is a pair $G = (V, E)$, where 
$V = \{v_1,  \ldots, v_m\}$ is a set of
{\it nodes\/} or {\it vertices\/}, and $E\subseteq V \times V$ is a
set of ordered pairs of distinct nodes (that is, pairs  
$(u, v)\in V\times V$ with $u\not= v$), called {\it edges\/}.
Given any edge $e = (u, v)$, we let $s(e) = u$ be the {\it source\/}
of $e$ and $t(e) = v$ be the {\it target\/} of $e$.
\end{definition}

\medskip
\remark
Since an edge is a pair $(u, v)$ with $u\not= v$, self-loops are not
allowed.
Also, there is at most one edge from a node $u$ to a node $v$.
Such graphs are sometimes called {\it simple graphs\/}.

\begin{figure}[http]
  \begin{center}
   \includegraphics[height=1.7truein,width=2.3truein]{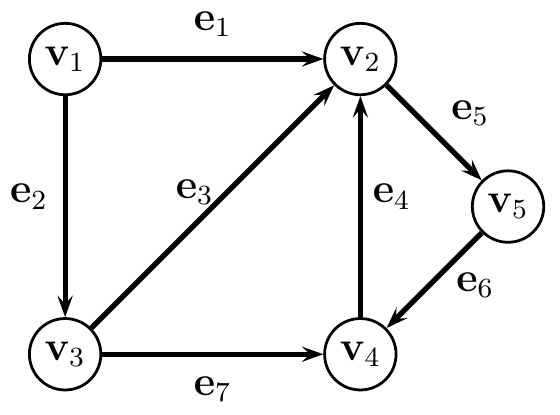}
  \end{center}
  \caption{Graph $G_1$.}
\label{graphfig17}
\end{figure}

\medskip
An example of a directed graph is shown in Figure \ref{graphfig17}.

\medskip
For every node $v\in V$, the {\it degree\/} $d(v)$ of $v$ is the number of
edges leaving or entering $v$:
\[
d(v) = |\{u\in V \mid (v, u)\in E\> \mathrm{or}\>  (u, v)\in E\}|.
\]
We abbreviate $d(v_i)$ as $d_i$. 
The {\it degree matrix\/} $\Degsym(G)$, is the diagonal matrix
\[
\Degsym(G) = \mathrm{diag}(d_1, \ldots, d_m).
\]
For example, for graph $G_1$, we have
\[
\Degsym(G_1) = 
\begin{pmatrix}
2 & 0 & 0 & 0 & 0 \\
0 & 4 & 0 & 0 & 0 \\
0 & 0 & 3 & 0 & 0 \\
0 & 0 & 0 & 3 & 0 \\
0 & 0 & 0 & 0 & 2
\end{pmatrix}.
\]
Unless confusion arises, we write $\Degsym$ instead of $\Degsym(G)$.

\begin{definition}
\label{dirpath}
Given a directed graph $G = (V, E)$, for any two nodes $u, v\in V$, a
{\it path from $u$ to $v$\/}  is a sequence of nodes
$(v_0, v_1, \ldots, v_k)$ such that $v_0 = u$,
$v_k = v$, and  $(v_i, v_{i+1})$ is an edge in $E$ for all $i$ with $0
\leq i \leq k - 1$. The integer $k$ is the {\it length\/} of the path.
A path is {\it closed\/} if $u = v$.
The graph $G$ is {\it strongly connected\/}
if for any two distinct node $u, v\in V$, there is a path from $u$ to
$v$ and there is a path from $v$ to $u$. 
\end{definition}

\remark
The terminology {\it walk\/} is often used instead of {\it path\/},
the word path being reserved to the case where the nodes $v_i$ are all
distinct, except that $v_0 = v_k$ when the path is closed.

\medskip
The binary relation on $V\times V$ defined so that
$u$ and $v$ are related iff there is a path from $u$  to $v$  and
there is a path from $v$ to $u$ is an equivalence relation whose
equivalence classes are called the {\it strongly connected
components\/} of $G$.

\begin{definition}
\label{incidence-matrix1}
Given a directed graph $G = (V, E)$, with $V = \{v_1, \ldots, v_m\}$, if 
$E = \{e_1, \ldots, e_n\}$,  then the {\it incidence matrix\/} $B(G)$
of $G$ is the $m\times n$
matrix whose entries $b_{i\, j}$ are given by
\[
b_{i\, j} = 
\begin{cases}
+1 & \text{if $s(e_j) = v_i$} \\
-1 & \text{if $t(e_j) =v_i$} \\
0 & \text{otherwise}.
\end{cases}
\]
\end{definition}

Here is the incidence matrix of the graph $G_1$:
\[
B = 
\begin{pmatrix}
1  & 1  & 0  & 0  & 0  & 0  & 0  \\
-1 & 0  & -1 & -1 & 1  & 0  & 0  \\ 
0  & -1 & 1  & 0  & 0  & 0  & 1  \\
0  & 0  & 0  & 1  & 0  & -1 & -1 \\
0  & 0  & 0  & 0  & -1 & 1  & 0
\end{pmatrix}.
\]

\medskip
Observe that every column of an incidence matrix contains
exactly two nonzero entries,  $+1$ and $-1$.
Again, unless confusion arises, we write $B$ instead of $B(G)$.

\medskip
When a directed graph has $m$ nodes  $v_1, \ldots, v_m$ and $n$ edges 
$e_1, \ldots, e_n$,  
a vector $x\in \reals^m$ can be
viewed as a function $\mapdef{x}{V}{\reals}$ assigning the value $x_i$
to the node $v_i$.   Under this interpretation, $\reals^m$ is viewed
as $\reals^V$.
Similarly, a vector $y\in \reals^n$ can be viewed as a function in $\reals^E$.
This point of view is often useful. For example, 
the incidence matrix $B$
can be interpreted  as a linear map from $\reals^E$ to $\reals^V$,
the {\it boundary map\/}, and 
$\transpos{B}$  can be interpreted  as a linear map 
from $\reals^V$ to $\reals^E$, the {\it coboundary map\/}.

\remark
Some authors adopt the opposite convention of sign in defining
the incidence matrix, which means that their incidence matrix is 
$-B$.

\medskip
Undirected graphs are obtained from directed graphs by forgetting the
orientation of the edges.

\begin{definition}
\label{graph}
A {\it  graph\/} (or {\it undirected graph\/})
is a pair $G = (V, E)$, where 
$V = \{v_1,  \ldots, v_m\}$ is a set of
{\it nodes\/} or {\it vertices\/}, and $E$ is a
set of two-element subsets of $V$  (that is, subsets  
$\{u, v\}$,  with $u, v\in V$ and $u\not= v$), called {\it edges\/}.
\end{definition}

\medskip
\remark
Since an edge is a set $\{u, v\}$,  we have  $u\not= v$, so self-loops are not
allowed.
Also, for every set of nodes $\{u, v\}$, there
is at most one edge between $u$ and $v$.
As in the case of directed graphs, 
such graphs are sometimes called {\it simple graphs\/}.

\begin{figure}
  \begin{center}
 \includegraphics[height=1.7truein,width=2.3truein]{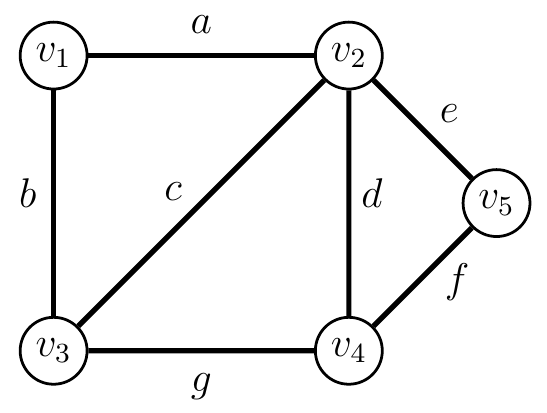}
  \end{center}
  \caption{The undirected graph $G_2$.}
\label{graphfig5bis}
\end{figure}

\medskip
An example of a  graph is shown in Figure \ref{graphfig5bis}.

\medskip
For every node $v\in V$, the {\it degree\/} $d(v)$ of $v$ is the number of
edges incident to $v$:
\[
d(v) = |\{u\in V \mid  \{u, v\}\in E\}|.
\]
The degree matrix $\Degsym$ is defined as before.

\begin{definition}
\label{path}
Given a  (undirected) graph $G = (V, E)$, for any two nodes $u, v\in V$, a
{\it path from $u$ to $v$\/}  is a sequence of nodes
$(v_0, v_1, \ldots, v_k)$ such that $v_0 = u$,
$v_k = v$, and  $\{v_i, v_{i+1}\}$ is an edge in $E$ for all $i$ with $0
\leq i \leq k - 1$. The integer $k$ is the {\it length\/} of the path.
A path is {\it closed\/} if $u = v$.
The graph $G$ is {\it connected\/}
if for any two distinct node $u, v\in V$, there is a path from $u$ to
$v$.
\end{definition}

\remark
The terminology {\it walk\/} or {\it chain\/} is often used instead of {\it path\/},
the word path being reserved to the case where the nodes $v_i$ are all
distinct, except that $v_0 = v_k$ when the path is closed.

\medskip
The binary relation on $V\times V$ defined so that
$u$ and $v$ are related iff there is a path from $u$  to $v$  
is an equivalence relation whose
equivalence classes are called the {\it connected
  components\/} of $G$.

\medskip
The notion of incidence matrix for an undirected graph is not as 
useful as in the case of directed graphs

\begin{definition}
\label{incidence-matrix2}
Given a graph $G = (V, E)$, with $V = \{v_1, \ldots, v_m\}$, if 
$E = \{e_1, \ldots, e_n\}$,  then the {\it incidence matrix\/} $B(G)$ 
of $G$ is the $m\times n$
matrix whose entries $b_{i\, j}$ are given by
\[
b_{i\, j} = 
\begin{cases}
+1 & \text{if $e_j = \{v_i, v_k\}$ for some $k$} \\
0 & \text{otherwise}.
\end{cases}
\]
\end{definition}

Unlike the case of directed graphs, the entries in the incidence
matrix of a graph (undirected) are nonnegative. 
We usually write $B$ instead of $B(G)$.

\medskip
The notion of adjacency matrix is basically the same for directed or undirected
graphs.

\begin{definition}
\label{adjacency}
Given a directed or undirected graph $G = (V, E)$, 
with $V = \{v_1, \ldots, v_m\}$,
the {\it adjacency matrix\/}  $A(G)$ of $G$ is the symmetric 
$m\times m$ matrix $(a_{i\, j})$ such that
\begin{enumerate}
\item[(1)] 
If $G$ is directed, then
\[
a_{i\, j} = 
\begin{cases}
1 & \text{if there is some edge $(v_i, v_j)\in E$ or some edge 
$(v_j,  v_i)\in E$} \\
0 & \text{otherwise}. 
\end{cases}
\]
\item[(2)]
Else if $G$ is undirected, then
\[
a_{i\, j} = 
\begin{cases}
1 & \text{if there is some edge $\{v_i, v_j\}\in E $} \\
0 & \text{otherwise}. 
\end{cases}
\]
\end{enumerate}
\end{definition}

\medskip
As usual, unless confusion arises, we write $A$ instead of $A(G)$.
Here is the adjacency matrix of both graphs $G_1$ and $G_2$:

\[
A = 
\begin{pmatrix}
0 & 1 & 1 & 0 & 0 \\
1 & 0 & 1 & 1 & 1 \\
1 & 1 & 0 & 1 & 0 \\
0 & 1 & 1 & 0 & 1 \\
0 & 1 & 0 & 1 & 0
\end{pmatrix}.
\]

\medskip
If $G = (V, E)$ is a directed or an undirected graph, given a node
$u\in V$, any node $v\in V$ such that there is an edge $(u, v)$ in the
directed case or $\{u, v\}$ in the undirected case is called {\it
  adjacent to $v$\/}, and we often use the notation
\[
u \sim v.
\]
Observe that the binary relation $\sim$ is symmetric when $G$ is an
undirected graph, but in general it is not symmetric when $G$ is a
directed graph.

\medskip
If $G = (V, E)$ is an undirected graph, the adjacency matrix $A$ of
$G$ can
be viewed as a linear map from $\reals^V$ to $\reals^V$, such
that for all $x\in \reals^m$, we have
\[
(Ax)_i = \sum_{j\sim i} x_j;
\]
that is, the value of  $Ax$ at $v_i$ is the sum of the values of $x$ 
at the nodes $v_j$ adjacent to $v_i$. The adjacency matrix can be
viewed as a {\it diffusion operator\/}.
This observation yields a geometric interpretation of what it means for a
vector $x\in \reals^m$ to be an eigenvector of $A$ associated with
some eigenvalue $\lambda$; we must have
\[
\lambda x_i  = \sum_{j\sim i} x_j, \quad i = 1, \ldots, m,
\]
which means that the the sum of the values of $x$
assigned to the nodes $v_j$ adjacent to $v_i$ is equal to $\lambda$ 
times the value of $x$ at $v_i$.

\begin{definition}
\label{orientG}
Given any undirected graph $G = (V, E)$, an 
{\it orientation\/} of $G$ is a function 
$\mapdef{\sigma}{E}{V\times V}$ 
assigning a source and a target to
every edge in $E$, which means that  for every edge $\{u, v\} \in E$,
either $\sigma(\{u, v\}) = (u, v)$ or  $\sigma(\{u, v\}) = (v, u)$. 
The {\it oriented  graph\/} $G^{\sigma}$ obtained
from $G$ by applying the orientation $\sigma$ is the directed graph
$G^{\sigma} = (V, E^{\sigma})$, with
$E^{\sigma} = \sigma(E)$.
\end{definition}

The following result shows how the number of connected
components of an undirected graph is related to the rank of the
incidence matrix of any oriented graph obtained from $G$.

\begin{proposition}
\label{ccnum}
Let $G = (V, E)$ be any undirected graph with $m$ vertices, $n$ edges,
and $c$ connected
components.  For any orientation $\sigma$ of $G$, if  $B$ is the
incidence matrix of the oriented graph $G^{\sigma}$, then
$c = \mathrm{dim}(\Ker(\transpos{B}))$, and 
$B$ has rank $m - c$. Furthermore, the nullspace of
$\transpos{B}$ has a basis consisting of indicator vectors of
the connected components of $G$; that is,
vectors $(z_1, \ldots, z_m)$ such that $z_j = 1$ iff $v_j$ is in the
$i$th component $K_i$ of $G$, and $z_j = 0$ otherwise.
\end{proposition}

\begin{proof}
(After Godsil and Royle \cite{Godsil}, Section 8.3).
We prove that the kernel of $\transpos{B}$ has dimension $c$. Since 
$\transpos{B}$ is a $n\times m$ matrix, we have
\[
m = \mathrm{dim}(\Ker ( \transpos{B})) + \mathrm{rank}(\transpos{B}),
\]
so $\mathrm{rank}(\transpos{B}) = m - c$. Since $B$ and
$\transpos{B}$ have the same rank,   $\mathrm{rank}(B) = m - c$, as claimed. 

\medskip
A vector $z\in \reals^m$ belongs to the kernel of $\transpos{B}$ iff
$\transpos{B} z = 0$ iff  $\transpos{z} B = 0$.
In view of the definition of $B$, for every edge $\{v_i, v_j\}$ of
$G$, the column of $B$ corresponding to the oriented edge
$\sigma(\{v_i, v_j\})$ has zero entries except  for a $+1$ and a $-1$
in position $i$ and position $j$ or vice-versa, so   
we have
\[
z_i  = z_j.
\]
An easy induction on the length of the path
shows that if there is a path from $v_i$ to $v_j$ in $G$
(unoriented), then  $z_i = z_j$. Therefore, $z$ has a constant
value on any connected component of $G$. It follows that
every vector $z \in \Ker(\transpos{B})$ can be written uniquely as a linear
combination
\[
z = \lambda_1 z^1 + \cdots + \lambda_c z^c,
\]
where the vector $z^i$ corresponds to the $i$th connected component
$K_i$ of $G$ and is defined such that
\[
z^i_j = 
\begin{cases}
1 & \text{iff $v_j \in K_i$} \\
0 & \text{otherwise}. 
\end{cases}
\]
This shows that
$\mathrm{dim}(\Ker ( \transpos{B})) = c$, and that 
$\Ker ( \transpos{B})$ has a basis consisting of indicator vectors.
\end{proof}

Following common practice, we denote by $\mathbf{1}$ the (column)
vector whose components are all equal to $1$. Since every column of
$B$ contains a single $+1$ and a single $-1$, the rows of
$\transpos{B}$ sum to zero, which can be expressed as
\[
\transpos{B}\mathbf{1} = 0.
\]
According to Proposition \ref{ccnum}, the graph $G$ is connected iff
$B$ has rank $m -1 $ iff 
the nullspace of $\transpos{B}$ is the one-dimensional space spanned
by $\mathbf{1}$.

\medskip
In many applications, the notion of graph needs to be generalized to
capture the intuitive idea that two nodes $u$ and $v$ are linked with
a degree of certainty (or strength). Thus, we assign a nonnegative weight $w_{i\, j}$ 
to an edge $\{v_i,  v_j\}$; the smaller $w_{i\, j}$ is, the weaker is
the link (or similarity) between $v_i$ and $v_j$, and the greater 
$w_{i\, j}$ is, the stronger is the link (or similarity) between $v_i$
and $v_j$.

\begin{definition}
\label{graph-weighted}
A {\it  weighted graph\/} 
is a pair $G = (V, W)$, where 
$V = \{v_1,  \ldots, v_m\}$ is a set of
{\it nodes\/} or {\it vertices\/}, and $W$ is a symmetric matrix
called the {\it weight matrix\/}, such that $w_{i\, j} \geq 0$
for all $i, j \in \{1, \ldots, m\}$, 
and $w_{i\, i} = 0$ for $i = 1, \ldots, m$.
We say that a set $\{v_i, v_j\}$  is an edge iff
$w_{i\, j} > 0$. The corresponding (undirected) graph $(V, E)$
with $E = \{\{v_i, v_j\} \mid w_{i\, j} > 0\}$, 
is called the {\it underlying graph\/} of $G$.
\end{definition}

\medskip
\remark
Since $w_{i\, i} = 0$, these graphs have no self-loops.
We can think of the matrix $W$ as a generalized adjacency matrix.
The case where $w_{i\, j} \in \{0, 1\}$ is equivalent to the notion
of a graph as in Definition \ref{graph}.

\begin{figure}[http]
  \begin{center}
 \includegraphics[height=2.5truein,width=2.8truein]{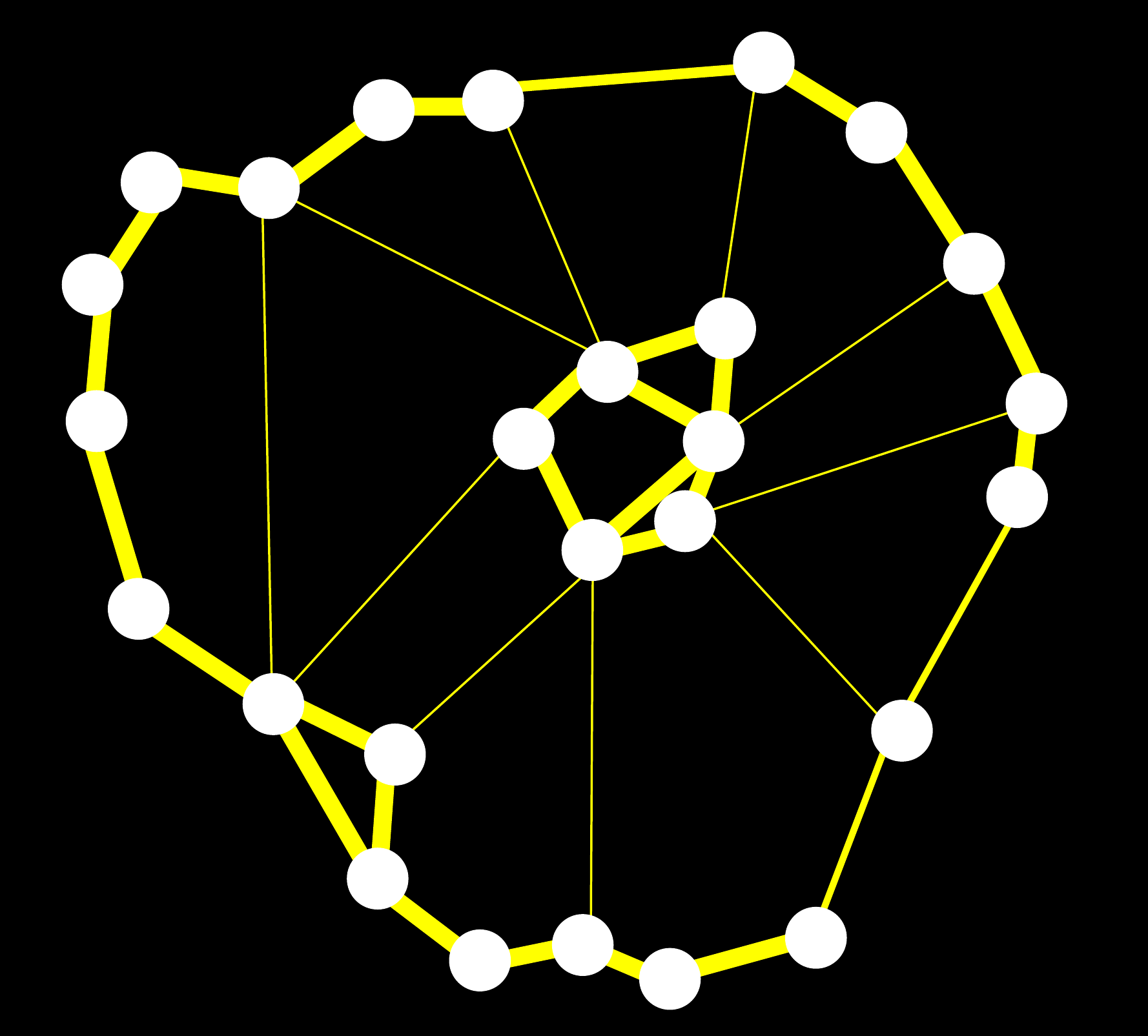}
  \end{center}
  \caption{A weighted graph.}
\label{ncg-fig1}
\end{figure}

\medskip
We can think of the weight $w_{i\, j}$ of an edge $\{v_i, v_j\}$ as a degree of
similarity (or affinity)  in an image, or a cost in a network.
An example of a weighted graph is shown in Figure \ref{ncg-fig1}.
The thickness of an edge corresponds to the magnitude of its weight.

\medskip
For every node $v_i\in V$, the {\it degree\/} $d(v_i)$ of $v_i$ is the sum
of  the weights of the 
edges adjacent to $v_i$:
\[
d(v_i) = \sum_{j = 1}^m w_{i\, j}.
\]
Note that in the above sum, only nodes $v_j$ such that there is an
edge $\{v_i, v_j\}$ have a nonzero contribution. Such nodes are said
to be  {\it adjacent\/} to $v_i$, and we write $v_i \sim v_j$.
The degree matrix $\Degsym$ is defined as before, namely by
$\Degsym = \mathrm{diag}(d(v_1), \ldots, d(v_m))$.

\medskip
The weight matrix  $W$ can be viewed as a linear map from $\reals^V$ to itself.
For all $x\in \reals^m$,  we have
\[
(Wx)_i = \sum_{j\sim i} w_{i j} x_j;
\]
that is, the value of  $Wx$ at $v_i$ is the weighted sum of the values of $x$ 
at the nodes $v_j$ adjacent to $v_i$.

\medskip
Observe that  $W \mathbf{1}$
is the (column) vector $(d(v_1), \ldots, d(v_m))$ consisting of the
degrees of the nodes
of the graph.

\medskip
Given any subset of nodes $A \subseteq V$, we define the 
{\it  volume\/}
$\mathrm{vol}(A)$ of $A$ as the sum of the weights of all edges 
adjacent to nodes in $A$:
\[
\mathrm{vol}(A) = \sum_{v_i\in A} d(v_i) = 
\sum_{v_i \in A} \sum_{j =  1}^m w_{i\, j}.
\]
\remark
Yu and Shi \cite{YuShi2003}
use the notation $\mathrm{degree}(A)$ instead of $\mathrm{vol}(A)$. 

\begin{figure}[http]
  \begin{center}
 \includegraphics[height=2truein,width=2truein]{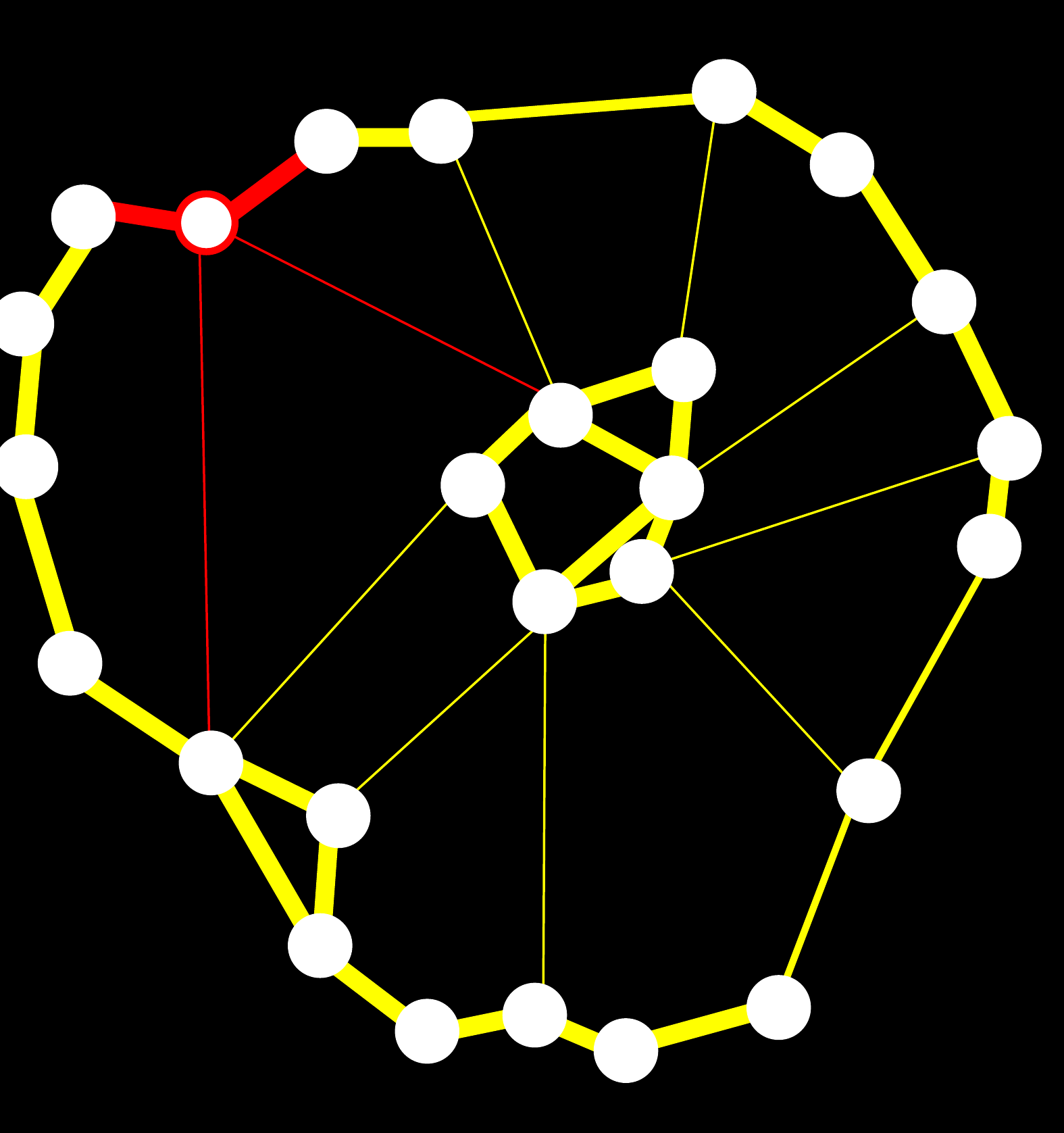}
\hspace{1cm}
 \includegraphics[height=2truein,width=2truein]{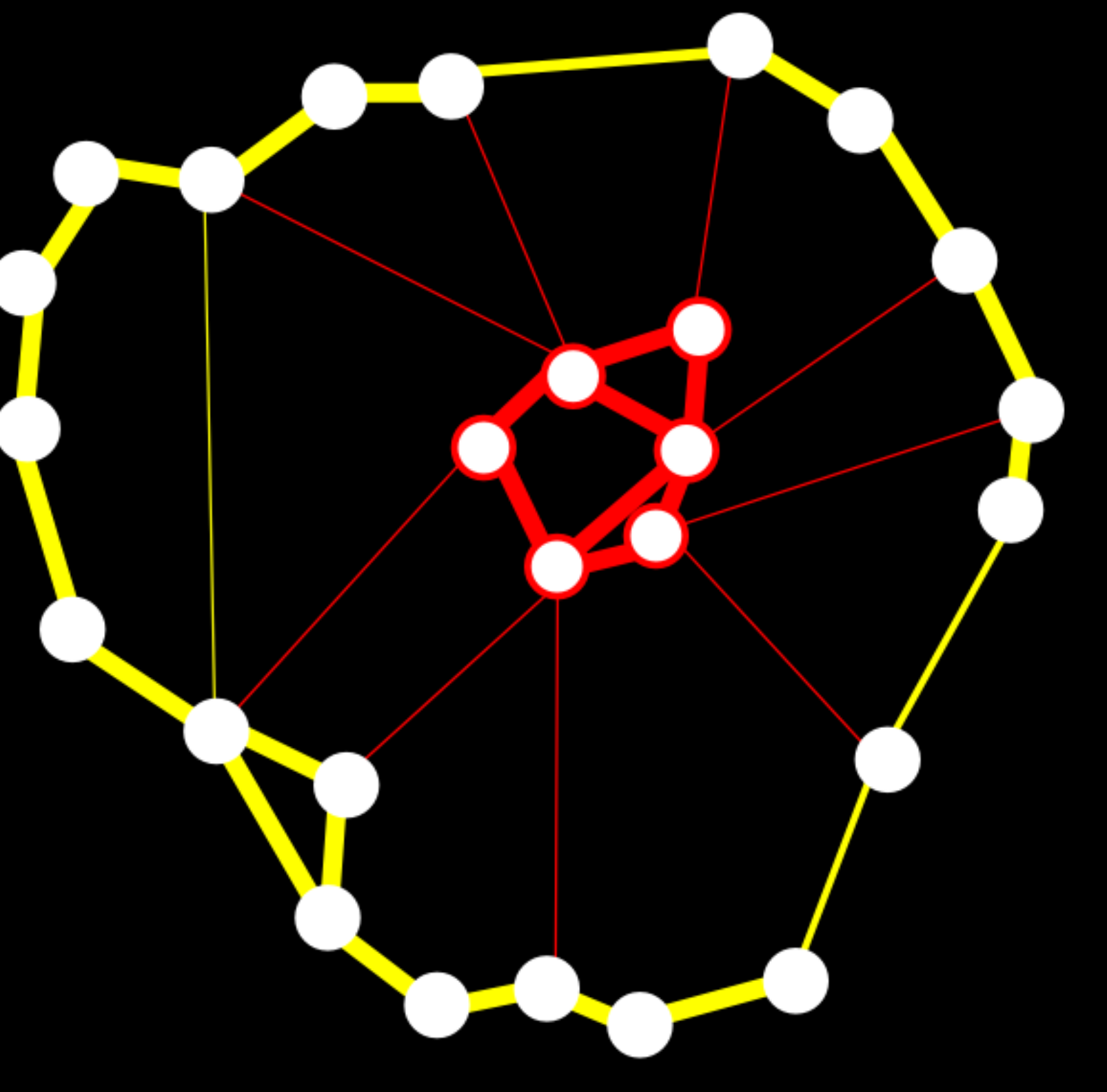}
  \end{center}
  \caption{Degree and volume.}
\label{ncg-fig2}
\end{figure}

The notions of degree and volume are illustrated in Figure \ref{ncg-fig2}.

\medskip
Observe that $\mathrm{vol}(A) = 0$ if $A$ consists of isolated vertices,
that is, if $w_{i\, j} = 0$ for all $v_i\in A$.  Thus, it is best to
assume that $G$ does not have isolated vertices.

\medskip
Given any two subset $A, B\subseteq V$ (not necessarily distinct), we
define
$\mathrm{links}(A, B)$ by
\[
\mathrm{links}(A, B) = \sum_{v_i\in A, v_j\in B} w_{i\, j}.
\]
Since the matrix $W$ is symmetric, we have
\[
\mathrm{links}(A, B) = \mathrm{links}(B, A),  
\]
and observe that
$\mathrm{vol}(A) = \mathrm{links}(A, V)$.

\begin{figure}[http]
  \begin{center}
 \includegraphics[height=2.2truein,width=2.2truein]{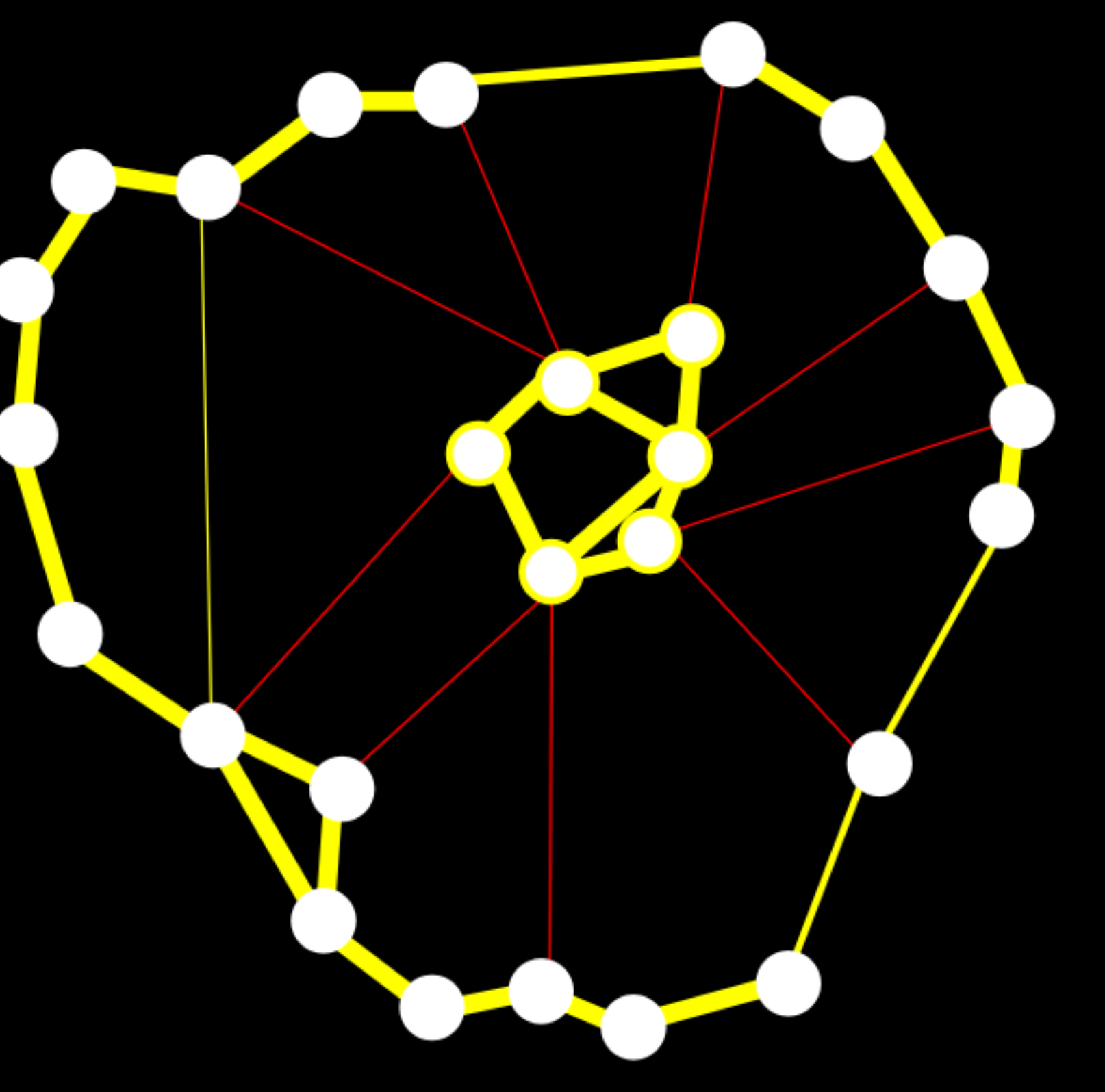}
  \end{center}
  \caption{A Cut involving the set of nodes in the center and the
    nodes on the perimeter.}
\label{ncg-fig3}
\end{figure}

\medskip
The quantity
$ \mathrm{links}(A, \overline{A}) =
\mathrm{links}(\overline{A}, A)$
(where $\overline{A} = V - A$ denotes the complement of $A$ in $V$)
measures
how many links escape from $A$ (and $\overline{A}$), and the quantity
$\mathrm{links}(A,A)$ 
measures how many links stay within $A$ itself.
The quantity  
\[
\mathrm{cut}(A) =
\mathrm{links}(A, \overline{A})
\]
is often called the {\it cut\/} of $A$, and the quantity 
\[
\mathrm{assoc}(A) =
\mathrm{links}(A,A)
\]
is often called the {\it association\/} of $A$.
Clearly,
\[
\mathrm{cut}(A) + \mathrm{assoc}(A) = \mathrm{vol}(A).
\]
The notions of cut is illustrated in Figure \ref{ncg-fig3}.

\medskip
We now define the most important concept of these notes: The Laplacian
matrix of a graph. Actually, as we will see, it comes in several
flavors.

\section{Laplacian Matrices of Graphs}
\label{ch1-sec2}
Let us begin with directed graphs, although as we will see, graph
Laplacians are fundamentally associated with undirected graph.
The key proposition below shows how 
$B\transpos{B}$ relates to the adjacency matrix $A$.
We reproduce the proof in Gallier
\cite{GallDiscmath}
(see also Godsil and Royle \cite{Godsil}).

\begin{proposition}
\label{adjp2}
Given any directed graph $G$ if $B$ is the incidence
matrix of $G$, $A$ is the adjacency matrix of $G$,
and $\Degsym$ is the degree matrix such
that $\Degsym_{i\, i} = d(v_i)$,
then
\[
B\transpos{B}= \Degsym - A.
\]
Consequently, $B\transpos{B}$ is 
independent of the orientation of $G$ and
$\Degsym - A$ is symmetric and
positive semidefinite; that is, the eigenvalues of
$\Degsym - A$ are real and nonnegative.
\end{proposition}

\begin{proof}
The entry
$B\transpos{B}_{i\, j}$ is the inner
product of the $i$th row  $b_i$, and the $j$th row $b_j$ of $B$.
If $i = j$, then as 
\[
b_{i\, k} = 
\begin{cases}
+1 & \text{if $s(e_k) = v_i$} \\
-1 & \text{if $t(e_k) = v_i$} \\
0 & \text{otherwise}
\end{cases}
\]
we see that $b_i\cdot b_i =  d(v_i)$.
If $i \not= j$, then  $b_i\cdot b_j \not= 0$ iff
there is some edge $e_k$ with 
$s(e_k) = v_i$ and $t(e_k) = v_j$
or vice-versa,
in which case, $b_i\cdot b_j = -1$.
Therefore,
\[
B\transpos{B} = \Degsym - A,
\]
as claimed.

\medskip
For every $x\in \reals^m$, we have
\[
\transpos{x} L x = \transpos{x} B \transpos{B} x=
\transpos{(\transpos{B} x)} \transpos{B} x = \norme{\transpos{B}
  x}_2^2 \geq 0, 
\]
since the Euclidean norm $\norme{\>}_2$ is positive (definite).
Therefore, $L = B\transpos{B}$ is positive semidefinite. 
It is well-known that a real symmetric matrix is positive semidefinite
iff its eigenvalues are nonnegtive.
\end{proof}

\medskip
The matrix $L = B\transpos{B} = \Degsym - A$ is called the 
{\it  (unnormalized) graph Laplacian\/} of the graph $G$. For example, the
graph Laplacian of graph $G_1$ is
\[
L = 
\begin{pmatrix}
2 & -1 & -1 & 0 & 0 \\
-1 & 4 & -1 & -1 & -1 \\
-1 & -1 & 3 & -1 & 0 \\
0 & -1 & -1 & 3 & -1 \\
0 & -1 & 0 & -1 & 2
\end{pmatrix}.
\]

\medskip
The {\it  (unnormalized) graph Laplacian\/} of an undirected graph 
$G = (V, E)$ is defined by
\[
L = \Degsym - A.
\]
Observe that each row of $L$ sums to zero (because 
$\transpos{B}\mathbf{1} = 0$).
Consequently, the vector
$\mathbf{1}$ is in the nullspace of $L$.

\remark
With the unoriented version of the incidence matrix
(see Definition \ref{incidence-matrix2}), it can be shown that
\[
B\transpos{B} = \Degsym + A.
\]

\medskip
The natural generalization of the notion of graph Laplacian to
weighted graphs is this:

\begin{definition}
\label{graphLaplacian}
Given any weighted graph $G = (V, W)$ with
$V = \{v_1, \ldots,v_m\}$, the {\it
  (unnormalized) graph Laplacian $L(G)$ of $G$\/} is defined by
\[
L(G) = \Degsym(G) - W,
\]
where $\Degsym(G) = \mathrm{diag}(d_1, \ldots,d_m)$ is the degree matrix
of $G$ (a diagonal matrix), with
\[
d_i = \sum_{j = 1}^m w_{i \, j}.
\]
As usual, unless confusion arises, we write $L$ instead of $L(G)$. 
\end{definition}

The graph Laplacian can be interpreted as a linear map from 
$\reals^V$ to itself. For all  $x  \in \reals^V$,  we have
\[
(Lx)_i  = \sum_{j \sim i} w_{i j}(x_i - x_j).
\]

\medskip
It is clear that each row of $L$ sums to $0$, so the vector
$\mathbf{1}$ is the nullspace of $L$, but it is less obvious that $L$ is
positive semidefinite.  One way to prove it is to generalize slightly
the notion of incidence matrix.

\begin{definition}
\label{incidence-matrix-w}
Given a weighted graph $G = (V, W)$, with $V = \{v_1, \ldots, v_m\}$, 
if $\{e_1, \ldots, e_n\}$ are the edges of the underlying graph of $G$
(recall that $\{v_i, v_j\}$ is an edge of this graph iff $w_{i j} >
0$), for any oriented graph $G^{\sigma}$ obtained by giving an
orientation to 
the underlying graph of $G$,
the {\it incidence matrix\/} $B^{\sigma}$ of
$G^{\sigma}$ is the $m\times n$
matrix whose entries $b_{i\, j}$ are given by
\[
b_{i\, j} = 
\begin{cases}
+\sqrt{w_{i j}} & \text{if $s(e_j) = v_i$}\\
-\sqrt{w_{i j}} & \text{if $t(e_j) = v_i$} \\
0 & \text{otherwise}.
\end{cases}
\]
\end{definition}

For example,  given the weight matrix
\[
W = 
\begin{pmatrix}
0  &   3  &   6  &   3 \\
     3  &   0  &   0  &   3 \\
     6  &   0  &   0  &   3 \\
     3  &   3  &   3  &   0
\end{pmatrix},
\]
the incidence matrix $B$ corresponding to the orientation
of the underlying graph of $W$ where an edge $(i, j)$ is oriented positively iff $i < j$ is
\[
B =
\begin{pmatrix}
1.7321  &  2.4495  &  1.7321    &     0   &      0 \\
   -1.7321   &      0    &     0 &  1.7321   &      0 \\
         0  &  -2.4495    &     0   &      0  &  1.7321 \\
         0   &      0  &  -1.7321  &  -1.7321  & -1.7321
\end{pmatrix}.
\]
The reader should verify that $B\transpos{B} = D - W$.
This is true in general, see Proposition \ref{adjp2w}.

\medskip
It is easy to see that Proposition \ref{ccnum}
applies to the underlying graph of 
$G$. For any oriented graph $G^{\sigma}$ obtained from the underlying graph of $G$,
the rank of the incidence matrix
 $B^{\sigma}$ is equal to 
$m - c$, where $c$ is the number of connected components of the 
underlying graph of $G$, and we have $\transpos{(B^{\sigma})}\mathbf{1} = 0$.
We also have the following version of Proposition \ref{adjp2}
whose proof is immediately adapted.

\begin{proposition}
\label{adjp2w}
Given any weighted  graph $G = (V, W)$  with $V = \{v_1, \ldots, v_m\}$, 
if $B^{\sigma}$ is the incidence
matrix of any oriented graph $G^{\sigma}$ obtained from the underlying
graph of $G$ and $\Degsym$ is the degree matrix of $W$,
then
\[
B^{\sigma} \transpos{(B^{\sigma})}= \Degsym - W = L.
\]
Consequently, $B^{\sigma}\transpos{(B^{\sigma})}$ is 
independent of the orientation of the underlying graph of  $G$ and
$L = \Degsym -W$ is symmetric and
positive semidefinite; that is, the eigenvalues of
$L = \Degsym - W$ are real and nonnegative.
\end{proposition}

\remark
Given any orientation $G^{\sigma}$ of the underlying graph of a weighted
graph $G = (V, W)$, if $B$ is the incidence matrix of $G^{\sigma}$,
then $\transpos{B}$ defines  a kind of discrete covariant derivative
$\mapdef{\nabla}{ \reals^V\times \s{X}(G)}{\reals^V}$ on the set of
$0$-forms, which is just 
the set of functions $\reals^V$.
For every vertex $v_i\in V$, we view the set of edges with source or
endpoint $v_i$,
\[
T_{v_i} G = \{(v_i, v_j) \mid w_{i j} \not= 0\} \cup \{(v_h, v_i) \mid
w_{h i} \not= 0\},
\]
as a kind of discrete tangent space at $v_i$. The disjoint union of
the tangent spaces $T_{v_i} G$ is the discrete tangent bundle $T G$.
A discrete vector field is then a function $\mapdef{X}{V}{T G}$ 
that assigns to every vertex
$v_i\in V$ some edge $X(v_i) = e_k \in T_{v_i} G$, and we denote the set of all
discrete vectors fields by $\s{X}(G)$.
For every function $f\in\reals^V$ and for every vector field $X\in
\s{X}(G)$,  we define the function $\nabla_X f$, a discrete analog
of the covariant derivative of the function $f$ with respect to the
vector field $X$,  by
\[
(\nabla_X f)(v_i) =\transpos{B}(f)(X(v_i));
\]
that is, if $X(v_i)$ is  the $k$th edge  $e_k = (v_i, v_j)$, then 
\[
(\nabla_{X} f)(v_i)  = \sqrt{w_{i j}} (f_i - f_j),
\]
else if $X(v_i)$ is  the $k$th edge  $e_k = (v_j, v_i)$, then 
\[
(\nabla_{X} f)(v_i)  = \sqrt{w_{i j}} (f_j - f_i).
\]
Then, the graph Laplacian $L$ is given by
\[
L = B\transpos{B};
\]
for every node $v_i$, we have
\[
(Lx)_i = \sum_{j \sim i} w_{i j} (x_i - x_j).
\]
Thus, $L$ appears to be a discrete analog of the {\it connection
  Laplacian\/}
(also known as {\it Bochner Laplacian\/}),
rather than a discrete analog of  the Hodge (Laplace--Beltrami)  Laplacian;
see Petersen \cite{Petersen}.
To make the above statement precise, we need to view $\nabla f$ as
the function from $\s{X}(G)$ to $\reals^V$ given by
\[
(\nabla f)(X) = \nabla_X f.
\]
The set of functions from $\s{X}(G)$ to $\reals^V$ is in bijection
with the set of functions $\reals^{\s{X}(G)\times V}$ from
$\s{X}(G)\times V$ to $\reals$, and  we can view the discrete connection
$\nabla$ as a linear map $\mapdef{\nabla}{\reals^V}{\reals^{\s{X}(G)\times V}}$.
Since both $\s{X}(G)$ and $V$ are
finite, we can use the inner product on the vector space
$\reals^{\s{X}(G)\times V}$ (and the inner product on $\reals^V$)
to define the adjoint $\mapdef{\nabla^*}{\reals^{\s{X}(G)\times V}}{\reals^V}$
of $\mapdef{\nabla}{\reals^V}{\reals^{\s{X}(G)\times V}}$ by 
\[
\lag \nabla^* F, f\rag = \lag F, \nabla f\rag,
\] 
for all $f\in \reals^V$ and all $F\in \reals^{\s{X}(G)\times V}$.
Then, the connection Laplacian $\mapdef{\nabla^*\nabla}{\reals^V}{\reals^V}$ is indeed equal to $L$.

\medskip
Another way to prove that $L$ is positive semidefinite
is to evaluate the
quadratic form $\transpos{x} L x$.

\begin{proposition}
\label{Laplace1}
For any  $m\times m$ symmetric matrix $W = (w_{i j})$, if we let $L = \Degsym - W$
where $\Degsym$ is the degree matrix associated with  $W$,
then  we have
\[
\transpos{x} L x =
\frac{1}{2}\sum_{i, j = 1}^m w_{i\, j} (x_i - x_j)^2
\quad\mathrm{for\ all}\> x\in \reals^m.
\]
Consequently, $\transpos{x} L x$ does not depend on the
diagonal entries in $W$, and if $w_{i\, j} \geq 0$ for all $i, j\in
\{1, \ldots,m\}$, then $L$ is positive semidefinite.
\end{proposition}
\begin{proof}
We have
\begin{align*}
\transpos{x} L x & = \transpos{x} \Degsym x - \transpos{x} W x \\
& = \sum_{i = 1}^m d_i x_i^2 - \sum_{i, j = 1}^m w_{i\, j} x_i x_j \\
& = \frac{1}{2}\left( \sum_{i = 1}^m d_i x_i^2  
- 2 \sum_{i, j = 1}^m  w_{i\, j} x_i x_j +  \sum_{i = 1}^m d_i x_i^2  
\right) \\
& = 
\frac{1}{2}\sum_{i, j = 1}^m w_{i\, j} (x_i - x_j)^2.
\end{align*}
Obviously, the quantity on the right-hand side does not depend on the
diagonal entries in $W$, and  if $w_{i\, j} \geq 0$ for all $i, j$,
then this quantity is nonnegative.
\end{proof}

\medskip
Proposition \ref{Laplace1} immediately implies the following facts:
For any weighted graph $G = (V, W)$, 
\begin{enumerate}
\item
The eigenvalues  $0 = \lambda_1 \leq \lambda_2 \leq  \ldots \leq 
\lambda_m$ of $L$ are real and nonnegative, and there is an
orthonormal basis of eigenvectors of $L$.
\item
The smallest eigenvalue $\lambda_1$ of $L$ is equal to $0$, and
$\mathbf{1}$ is a corresponding eigenvector.
\end{enumerate}

\medskip
It turns out that the dimension of the nullspace of $L$ (the
eigenspace of $0$)  is equal to the number of  connected components
of the underlying graph of $G$.
  
\begin{proposition}
\label{Laplace2}
Let $G = (V, W)$ be a weighted graph. The number $c$ of connected components
$K_1, \ldots, K_c$ of the underlying graph of $G$ is equal to the
dimension of the nullspace of $L$, which is equal to the multiplicity
of the eigenvalue $0$. Furthermore, the nullspace of $L$ has a basis
consisting of indicator vectors of the connected components of $G$,
that is, vectors $(f_1, \ldots, f_m)$
such that $f_j = 1$ iff $v_j\in K_i$ and $f_j  = 0$ otherwise.
\end{proposition}

\begin{proof}
Since $L = B\transpos{B}$ for the incidence matrix $B$ 
associated with any oriented graph obtained from $G$,
and since $L$ and $\transpos{B}$ have the same nullspace,
by Proposition \ref{ccnum}, the dimension of the nullspace of $L$
is equal to the number $c$ of connected components of $G$
and the indicator vectors of  the connected components of $G$
form a basis of $\Ker(L)$.
\end{proof}

\medskip
Proposition \ref{Laplace2} implies that if the underlying graph of $G$
is connected, 
then the second eigenvalue $\lambda_2$ of $L$  is strictly positive.

\medskip
Remarkably, the eigenvalue $\lambda_2$ contains a lot of information
about the graph $G$ (assuming that $G = (V, E)$ is an undirected graph).
This was first discovered by Fiedler in 1973, and
for this reason, $\lambda_2$ is often referred to as the {\it Fiedler
  number\/}.  For more on the properties of the Fiedler number,
see Godsil and Royle \cite{Godsil} (Chapter 13) and Chung \cite{Chung}.
More generally, the spectrum $(0, \lambda_2, \ldots, \lambda_m)$ of
$L$ contains a lot of information about the combinatorial structure of
the graph $G$. Leverage of this information is the object of 
{\it spectral graph theory\/}.

\medskip
It turns out that  normalized variants of the graph Laplacian are
needed, especially in applications to graph clustering. 
These variants make sense only if $G$ has no isolated vertices, which
means that every row of $W$ contains some strictly positive entry.
In this case, the degree matrix $\Degsym$ contains positive entries,
so it is invertible and
$\Degsym^{-1/2}$ makes sense; namely
\[
\Degsym^{-1/2} = \mathrm{diag}(d_1^{-1/2}, \ldots, d_m^{-1/2}),
\]
and similarly for any real exponent $\alpha$.

\begin{definition}
\label{graphLaplacian2}
Given any weighted directed graph $G = (V, W)$ 
with no isolated vertex and with
$V = \{v_1, \ldots,v_m\}$,
the {\it (normalized) graph Laplacians $L_{\mathrm{sym}}$ and
  $L_{\mathrm{rw}}$  of $G$\/} are defined by
\begin{align*}
L_{\mathrm{sym}} & = \Degsym^{-1/2} L \Degsym^{-1/2} = I - \Degsym^{-1/2} W \Degsym^{-1/2}  \\
L_{\mathrm{rw}} & = \Degsym^{-1} L = I - \Degsym^{-1} W.
\end{align*}
\end{definition}

\medskip
Observe that the Laplacian $L_{\mathrm{sym}}  = \Degsym^{-1/2} L \Degsym^{-1/2}$
is a symmetric matrix (because $L$ and $\Degsym^{-1/2}$ are symmetric)
and that
\[
L_{\mathrm{rw}} = \Degsym^{-1/2} L_{\mathrm{sym}} \Degsym^{1/2}.
\]
The reason for the notation $L_{\mathrm{rw}}$ is that this matrix
is closely related to a random walk on the graph $G$.

\medskip
Since the unnormalized Laplacian $L$ can be written as
$L = B\transpos{B}$, where $B$ is the incidence matrix of any oriented
graph obtained from the underlying graph of $G = (V, W)$, if we let
\[
B_{\mathrm{sym}} = \Degsym^{-1/2} B,
\]
we get
\[
L_{\mathrm{sym}}= B_{\mathrm{sym}}\transpos{B_{\mathrm{sym}}}. 
\]
In particular, for any singular decomposition
$B_{\mathrm{sym}} = U \Sigma \transpos{V}$ of $B_{\mathrm{sym}}$
(with $U$ an $m\times m$ orthogonal matrix, $\Sigma$ a 
``diagonal'' $m\times n$ matrix of singular values,  and $V$ an
$n\times n$ orthogonal matrix), the eigenvalues of $L_{\mathrm{sym}}$
are the squares of the top $m$ singular values of 
$B_{\mathrm{sym}}$, and the vectors in $U$ are orthonormal
eigenvectors of $L_{\mathrm{sym}}$ with respect to these eigenvalues
(the  squares of the  top $m$ diagonal entries of $\Sigma$).
Computing the SVD of $B_{\mathrm{sym}}$ generally yields
more accurate results than diagonalizing  $L_{\mathrm{sym}}$,
especially when  $L_{\mathrm{sym}}$ has eigenvalues with
high multiplicity.

\medskip
There are simple relationships between the eigenvalues and the eigenvectors of 
$L_{\mathrm{sym}}$, and $L_{\mathrm{rw}}$. There is also a simple
relationship with  the generalized eigenvalue problem $Lx = \lambda
\Degsym x$.

\begin{proposition}
\label{Laplace3}
Let $G = (V, W)$ be a weighted graph without isolated vertices. The
graph Laplacians, $L, L_{\mathrm{sym}}$, and $L_{\mathrm{rw}}$ satisfy
the following properties:
\begin{enumerate}
\item[(1)]
The matrix $ L_{\mathrm{sym}}$ is symmetric and positive
semidefinite. In fact, 
\[
\transpos{x} L_{\mathrm{sym}} x =
\frac{1}{2}\sum_{i, j = 1}^m w_{i\, j} \left(\frac{x_i}{\sqrt{d_i}} - \frac{x_j}{\sqrt{d_j}}\right)^2
\quad\mathrm{for\ all}\> x\in \reals^m.
\]
\item[(2)]
The normalized graph Laplacians $L_{\mathrm{sym}}$ and $L_{\mathrm{rw}}$ 
have the same spectrum \\
$(0 = \nu_1 \leq  \nu_2 \leq  \ldots\leq
\nu_m)$, and a vector $u\not= 0$ is an eigenvector of
$L_{\mathrm{rw}}$ for $\lambda$ iff $\Degsym^{1/2} u$ is an eigenvector of
$L_{\mathrm{sym}}$ for $\lambda$.
\item[(3)]
The graph Laplacians, $L, L_{\mathrm{sym}}$, and $L_{\mathrm{rw}}$ are
symmetric and positive semidefinite.
\item[(4)]
A vector $u\not = 0$ is a solution of the generalized eigenvalue
problem $L u = \lambda \Degsym u$ iff $\Degsym^{1/2} u$ is an eigenvector of
$L_{\mathrm{sym}}$ for the eigenvalue $\lambda$ iff
$u$ is  an eigenvector of
$L_{\mathrm{rw}}$ for the eigenvalue $\lambda$.
 \item[(5)]
The graph Laplacians, $L$ and  $L_{\mathrm{rw}}$ have the same
nullspace. For any vector $u$, 
we have $u\in \Ker(L)$ iff $\Degsym^{1/2} u \in \Ker(L_{\mathrm{sym}})$.  
 \item[(6)]
The vector $\mathbf{1}$ is in the nullspace of $L_{\mathrm{rw}}$, and 
$\Degsym^{1/2} \mathbf{1}$ is in the nullspace of $L_{\mathrm{sym}}$.
\item[(7)]
For every eigenvalue $\nu_i$ of the normalized graph Laplacian
$L_{\mathrm{sym}}$, we have
$0 \leq \nu_i \leq 2$. Furthermore, $\nu_m = 2$ iff the underlying
graph of $G$ contains a nontrivial connected bipartite component.
\item[(8)]
If $m \geq 2$ and if
the underlying graph of $G$ is not a complete graph, then $\nu_2
\leq 1$. Furthermore the underlying graph of $G$ is a
complete graph iff
$\nu_2 = \frac{m}{m - 1}$.
\item[(9)]
If $m \geq 2$ and if
the underlying graph of $G$ is connected  then $\nu_2 > 0$. 
\item[(10)]
If $m \geq 2$ and if
the underlying graph of $G$ has no isolated vertices, then
$\nu_m \geq \frac{m}{m - 1}$.
\end{enumerate}
\end{proposition}

\begin{proof}
(1)
We have $L_{\mathrm{sym}} = \Degsym^{-1/2} L \Degsym^{-1/2}$, and
$\Degsym^{-1/2}$ is a symmetric invertible matrix (since it is an
invertible diagonal matrix). It is a well-known fact of linear algebra
that if $B$ is an invertible matrix, then
a matrix $S$ is symmetric, positive semidefinite iff 
$B S\transpos{B}$ is  symmetric, positive semidefinite.
Since $L$ is symmetric,  positive semidefinite, so is
$L_{\mathrm{sym}}  = \Degsym^{-1/2} L \Degsym^{-1/2}$. The formula
\[
\transpos{x} L_{\mathrm{sym}} x =
\frac{1}{2}\sum_{i, j = 1}^m w_{i\, j} \left(\frac{x_i}{\sqrt{d_i}} - \frac{x_j}{\sqrt{d_j}}\right)^2
\quad\mathrm{for\ all}\> x\in \reals^m
\]
follows immediately from Proposition \ref{Laplace1} by replacing $x$
by $\Degsym^{-1/2} x$, and also shows that $L_{\mathrm{sym}}$ is
positive semidefinite.

\medskip
(2)
Since 
\[
L_{\mathrm{rw}}= \Degsym^{-1/2} L_{\mathrm{sym}} \Degsym^{1/2},
\]
the matrices $L_{\mathrm{sym}}$ and $L_{\mathrm{rw}}$ are similar,
which implies that they have the same spectrum. In fact, since 
$\Degsym^{1/2}$ is invertible, 
\[
L_{\mathrm{rw}} u =
\Degsym^{-1} L u = \lambda u
\]
iff
\[
\Degsym^{-1/2} L u = \lambda \Degsym^{1/2} u
\]
iff
\[
\Degsym^{-1/2} L \Degsym^{-1/2} \Degsym^{1/2} u =  L_{\mathrm{sym}}
\Degsym^{1/2} u = \lambda \Degsym^{1/2} u,
\]
which shows that a vector $u\not= 0$ is an eigenvector of
$L_{\mathrm{rw}}$ for $\lambda$ iff $\Degsym^{1/2} u$ is an eigenvector of
$L_{\mathrm{sym}}$ for $\lambda$.

\medskip
(3)
We already know that $L$ and $L_{\mathrm{sym}}$ are positive
semidefinite,
and (2) shows that $L_{\mathrm{rw}}$ is also positive semidefinite. 

\medskip
(4)
Since $\Degsym^{-1/2}$ is invertible, we have
\[
Lu = \lambda \Degsym u
\]
iff
\[
\Degsym^{-1/2} Lu = \lambda \Degsym^{1/2} u
\]
iff
\[
\Degsym^{-1/2} L\Degsym^{-1/2} \Degsym^{1/2} u = L_{\mathrm{sym}}
\Degsym^{1/2} u = \lambda \Degsym^{1/2} u,
\]
which shows that a vector $u\not = 0$ is a solution of the generalized eigenvalue
problem $L u = \lambda \Degsym u$ iff $\Degsym^{1/2} u$ is an eigenvector of
$L_{\mathrm{sym}}$ for the eigenvalue $\lambda$. The second part of
the statement follows from (2).

\medskip
(5)
Since $\Degsym^{-1}$ is invertible, we have $L u = 0$ iff 
$\Degsym^{-1} L u = L_{\mathrm{rw}} u = 0$. Similarly, since 
$\Degsym^{-1/2}$ is invertible, we have $Lu = 0$ iff
$\Degsym^{-1/2}L \Degsym^{-1/2}\Degsym^{1/2}u = 0 $
iff $  \Degsym^{1/2}u  \in \Ker(L_{\mathrm{sym}})$.

\medskip
(6)
Since $L\mathbf{1} = 0$, we get  $L_{\mathrm{rw}} \mathbf{1} = \Degsym^{-1} L
\mathbf{1} = 0$. That $\Degsym^{1/2} \mathbf{1}$ is in the nullspace of
$L_{\mathrm{sym}}$
follows from (2).
Properties (7)--(10) are proved in Chung \cite{Chung} (Chapter 1).
\end{proof}

\medskip
A version of Proposition \ref{Laplace2} also holds for the graph Laplacians
$L_{\mathrm{sym}}$ and $L_{\mathrm{rw}}$.  
This follows easily from the fact that Proposition \ref{ccnum}
applies to the underlying graph of a weighted graph.
The proof is left as an exercise.

\begin{proposition}
\label{Laplace4}
Let $G = (V, W)$ be a weighted graph. The number $c$ of connected components
$K_1, \ldots, K_c$ of the underlying graph of $G$ is equal to the
dimension of the nullspace of both  $L_{\mathrm{sym}}$ and
$L_{\mathrm{rw}}$,  which is equal to the multiplicity
of the eigenvalue $0$. Furthermore, the nullspace of $L_{\mathrm{rw}}$ has a basis
consisting of indicator vectors of the connected components of $G$,
that is, vectors $(f_1, \ldots, f_m)$
such that $f_j = 1$ iff $v_j\in K_i$ and $f_j  = 0$ otherwise.
For  $L_{\mathrm{sym}}$, a basis of the nullpace is obtained by
multiplying the above basis of the nullspace of $L_{\mathrm{rw}}$ by $\Degsym^{1/2}$.
\end{proposition}

\chapter{Spectral Graph  Drawing}
\label{chap2}
\section{Graph  Drawing and Energy Minimization}
\label{ch2-sec1}
Let $G = (V, E)$ be some undirected graph. It is often desirable to
draw a graph, usually in the plane but possibly in 3D, and it turns
out that the  graph Laplacian can be used to design 
surprisingly good methods. Say $|V| = m$.
The idea is to assign a point $\rho(v_i)$ in $\reals^n$ to the vertex $v_i\in
V$, for every $v_i \in V$, 
and to draw a line segment between the points $\rho(v_i)$ and
$\rho(v_j)$ iff there is an edge $\{v_i, v_j\}$.
Thus, a {\it graph drawing\/} is a function
$\mapdef{\rho}{V}{\reals^n}$.

\medskip
We define the  {\it matrix of a graph drawing $\rho$ 
(in  $\reals^n$)\/}  as a $m \times n$ matrix $R$ whose $i$th row consists
of the row vector $\rho(v_i)$ corresponding to the point representing $v_i$ in
$\reals^n$.
Typically, we want $n < m$; in fact $n$ should be much smaller than $m$.
A representation is {\it balanced\/} iff the sum of the entries of
every column is zero, that is,
\[
\transpos{\mathbf{1}} R = 0.
\]
If a representation is not balanced, it can be made balanced 
by a suitable translation.  We may also assume that the columns of $R$
are linearly independent, since any basis of the column space also
determines the drawing.  Thus, from now on,  we may assume that $n \leq m$.

\medskip
\remark
A graph drawing $\mapdef{\rho}{V}{\reals^n}$ is not required to be injective,
which may result in degenerate drawings where distinct vertices are
drawn as the same point. For this reason, we prefer not to use the
terminology
{\it graph embedding\/}, which is often used in the literature.  This
is because in differential geometry, an embedding always refers to  an injective map.
The term {\it graph immersion\/} would be more appropriate.

\medskip
As explained in Godsil and Royle \cite{Godsil}, we can imagine building
a physical model of $G$ by connecting adjacent vertices (in
$\reals^n$) by identical
springs.
Then, it is natural to consider a representation to be better if it
requires the springs to be less extended. We can formalize this by
defining the {\it energy\/} of a drawing $R$ by
\[
\s{E}(R) = \sum_{\{v_i, v_j\}\in E} \norme{\rho(v_i) - \rho(v_j)}^2,
\]
where $\rho(v_i)$ is the $i$th row of $R$ and 
$\norme{\rho(v_i) - \rho(v_j)}^2$
is the square of the Euclidean length of the line segment
joining $\rho(v_i)$  and  $\rho(v_j)$.

\medskip
Then, ``good drawings''  are drawings that minimize the energy function
$\s{E}$.
Of course, the trivial representation corresponding to the zero matrix
is optimum, so we need to impose extra constraints to rule out the
trivial solution.

\medskip
We can consider the more general situation where the springs are not
necessarily identical. This can be modeled by a symmetric weight (or
stiffness)  matrix $W = (w_{i j})$, with $w_{i j} \geq
0$.
Then our energy function becomes
\[
\s{E}(R) = \sum_{\{v_i, v_j\}\in E} w_{i j} \norme{\rho(v_i) - \rho(v_j)}^2.
\]
It turns out that this function can be expressed in terms of the
Laplacian $L = \Degsym - W$.
The following proposition is shown in
Godsil and Royle \cite{Godsil}.
We give a slightly more direct proof.

\begin{proposition}
\label{energyprop1}
Let $G = (V, W)$ be a weighted graph, with $|V| = m$
and $W$  an $m\times m$ symmetric  matrix, and let $R$ be the
matrix of a graph drawing $\rho$ of  $G$ in $\reals^n$ 
(a $m\times n$ matrix).
If $L = \Degsym - W$ is 
the  unnormalized Laplacian matrix associated with $W$, then
\[
\s{E}(R) = \mathrm{tr}(\transpos{R} L  R).
\]
\end{proposition}

\begin{proof}
Since $\rho(v_i)$ is the $i$th row of $R$ (and $\rho(v_j)$ is the
$j$th row of $R$), if we denote the $k$th column
of $R$ by $R^k$, using Proposition \ref{Laplace1},
we have
\begin{align*}
\s{E}(R)  & = \sum_{\{v_i, v_j\}\in E} w_{i j} \norme{\rho(v_i) - \rho(v_j)}^2\\
 & = \sum_{k = 1}^n\sum_{\{v_i, v_j\}\in E} w_{i j} (R_{i k} - R_{j k})^2\\
 & = \sum_{k = 1}^n\frac{1}{2}\sum_{i, j = 1}^m  w_{i j} (R_{i k} - R_{j k})^2\\
 & = \sum_{k = 1}^n \transpos{(R^k)}L R^k =  \mathrm{tr}(\transpos{R} L  R), 
\end{align*}
as claimed.
\end{proof}

Note that
\[
L \mathbf{1} = 0,
\]
as we already observed.  

\medskip
Since the matrix  $\transpos{R} L R $ is
symmetric, it has real eigenvalues. Actually, since $L$ is
positive semidefinite, so is $\transpos{R} L R$.
Then, the trace of $\transpos{R} L  R$ is
equal to the sum of its positive eigenvalues, and this is the
energy $\s{E}(R)$ of the graph drawing.

\medskip
If $R$ is the matrix of a graph drawing in $\reals^n$, then for any
invertible matrix $M$, the map that assigns $\rho(v_i)M$ to $v_i$ is
another graph drawing of $G$, and these two drawings convey
the same amount of information. From this point of view, a graph
drawing is determined by the column space of $R$. Therefore, it is
reasonable to assume that the columns of $R$ are pairwise orthogonal
and that they have unit length. Such a matrix satisfies the equation
$\transpos{R} R = I$, and the corresponding drawing is called an
{\it orthogonal drawing\/}. This condition also rules out trivial
drawings.
The following result tells us how to find minimum energy 
orthogonal balanced graph
drawings, provided the graph is connected.

\begin{theorem}
\label{graphdraw}
Let $G = (V, W)$ be a weigted graph with $|V| = m$.
If $L = \Degsym - W$ is the (unnormalized) 
Laplacian of  $G$, and if the eigenvalues of $L$ are
$0 = \lambda_1 < \lambda_2 \leq \lambda_3 \leq \ldots \leq \lambda_m$, 
then the minimal energy of any balanced orthogonal graph drawing of
$G$ in $\reals^n$ is equal to $\lambda_2 + \cdots + \lambda_{n + 1}$
(in particular, this implies that $n < m$). The $m \times n$ matrix 
$R$ consisting of any unit eigenvectors $u_2, \ldots, u_{n+1}$
associated with $\lambda_2 \leq \ldots \leq \lambda_{n + 1}$ 
yields a balanced orthogonal graph drawing of minimal energy; it satisfies the
condition $\transpos{R} R = I$.
\end{theorem}

\begin{proof}
We present the  proof  given in  Godsil and Royle \cite{Godsil} (Section 13.4,
Theorem 13.4.1). The key point is that
the sum of the  $n$ smallest eigenvalues of $L$  is a lower bound for
$\mathrm{tr}(\transpos{R} L R)$. This can be shown using an argument
using the Rayleigh ratio; see Proposition \ref{interlace} (the
Poincar\'e separation theorem).
Then, any $n$ eigenvectors $(u_1, \ldots,
u_n)$ associated with
$\lambda_1, \ldots, \lambda_n$ achieve this bound.
Because the first eigenvalue of $L$ is
$\lambda_1 = 0$ and because we are assuming that $\lambda_2 > 0$, 
we have $u_1 = \mathbf{1}/\sqrt{m}$. 
Since the $u_j$ are pairwise
orthogonal  for $i = 2, \ldots, n$ and
since $u_i$ is orthogonal to $u_1 =  \mathbf{1}/\sqrt{m}$,
the entries in $u_i$ add up to $0$. Consequently, 
for any $\ell$ with $2 \leq \ell \leq n$,
by deleting $u_1$ and using $(u_2, \ldots, u_{\ell})$, 
we obtain a balanced orthogonal graph drawing in $\reals^{\ell - 1}$ with the  same
energy as the orthogonal graph drawing in  $\reals^{\ell}$
using $(u_1, u_2,\ldots,u_{\ell})$.
Conversely, from any balanced orthogonal drawing in $\reals^{\ell - 1}$
using  $(u_2, \ldots, u_{\ell})$, we obtain an
orthogonal graph drawing in $\reals^{\ell}$
using $(u_1, u_2,\ldots,u_{\ell})$ with the same energy.
Therefore, the minimum energy of a balanced orthogonal graph
drawing in $\reals^n$ is equal to the minimum energy of an orthogonal
graph drawing in $\reals^{n+1}$, and this minimum is
$\lambda_2 + \cdots + \lambda_{n + 1}$. 
\end{proof}

\medskip
Since $\mathbf{1}$  spans the nullspace of $L$, using $u_1$
(which belongs to $\Ker L$) as one of the vectors in $R$ 
would have the effect that all points
representing vertices of $G$ would have the same first coordinate.
This would mean that the drawing lives in a hyperplane in $\reals^{n}$,
which is undesirable, especially when $n = 2$, where all vertices
would be collinear. This is why we omit the first eigenvector $u_1$.

\medskip
Observe that for any orthogonal $n\times n$ matrix $Q$, since
\[
\mathrm{tr}(\transpos{R} L R) = \mathrm{tr}(\transpos{Q}\transpos{R} L
R Q),
\]
the matrix $RQ$ also yields a minimum orthogonal graph drawing. 
This amounts to applying the rigid motion $\transpos{Q}$ to the
rows of $R$.

\medskip
In summary, if $\lambda_2 > 0$, an automatic method for drawing a
graph in $\reals^2$ is this:

\begin{enumerate}
\item
Compute the two smallest nonzero eigenvalues $\lambda_2 \leq \lambda_3$ of 
the graph Laplacian $L$ (it is possible that
$\lambda_3 = \lambda_2$ if $\lambda_2$ is a multiple eigenvalue);
\item
Compute two unit eigenvectors $u_2, u_3$ associated
with $\lambda_2$ and $\lambda_3$, and let $R = [u_2\> u_3]$
be the $m\times 2$ matrix having $u_2$ and $u_3$ as columns.
\item
Place vertex $v_i$ at the point whose coordinates is the $i$th row of
$R$, that is, $(R_{i 1}, R_{i 2})$.
\end{enumerate}

\medskip
This method generally gives pleasing results, but beware that
there is no guarantee that distinct nodes are assigned distinct images,
because $R$ can have identical rows. This does not seem to happen
often in practice.

\section{Examples of Graph  Drawings}
\label{ch2-sec2}
We now give a number of examples using {\tt Matlab}. Some of these are
borrowed or adapted from Spielman \cite{Spielman}.

\begin{figure}[http]
  \begin{center}
   \includegraphics[height=2.2truein,width=2.4truein]{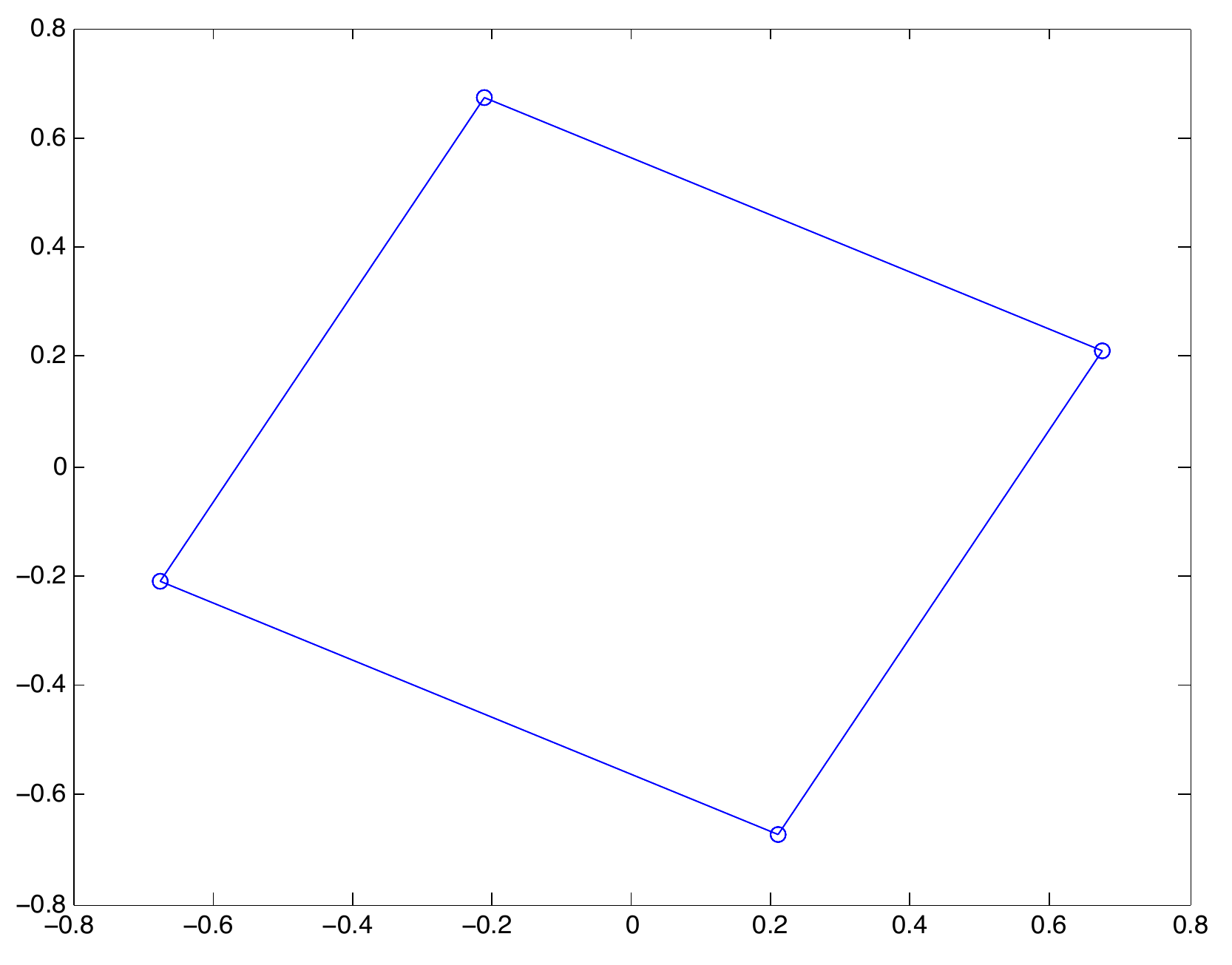}
  \end{center}
  \caption{Drawing of the graph from Example 1.}
\label{graph1}
\end{figure}

\medskip\noindent
{\it Example\/} 1.
Consider the graph with four nodes whose  adjacency matrix is
\[
A = 
\begin{pmatrix}
 0 & 1 & 1 & 0 \\
1 & 0 & 0 & 1 \\
1 & 0 & 0 & 1 \\
0 & 1 & 1 & 0
\end{pmatrix}.
\]
We use the following program to compute $u_2$ and $u_3$:
\begin{verbatim}
A = [0 1 1 0; 1 0 0 1; 1 0 0 1; 0 1 1 0];
D = diag(sum(A));
L = D - A;
[v, e] = eigs(L);
gplot(A, v(:,[3 2]))
hold on;
gplot(A, v(:,[3 2]),'o')
\end{verbatim}

The graph of Example 1 is shown in  Figure \ref{graph1}.
The function  {\tt eigs(L)}  computes the six largest eigenvalues of $L$ in decreasing
order, and corresponding eigenvectors. It turns out that $\lambda_2 =
\lambda_3 = 2$ is a double eigenvalue.

\begin{figure}[http]
  \begin{center}
   \includegraphics[height=2.2truein,width=2.4truein]{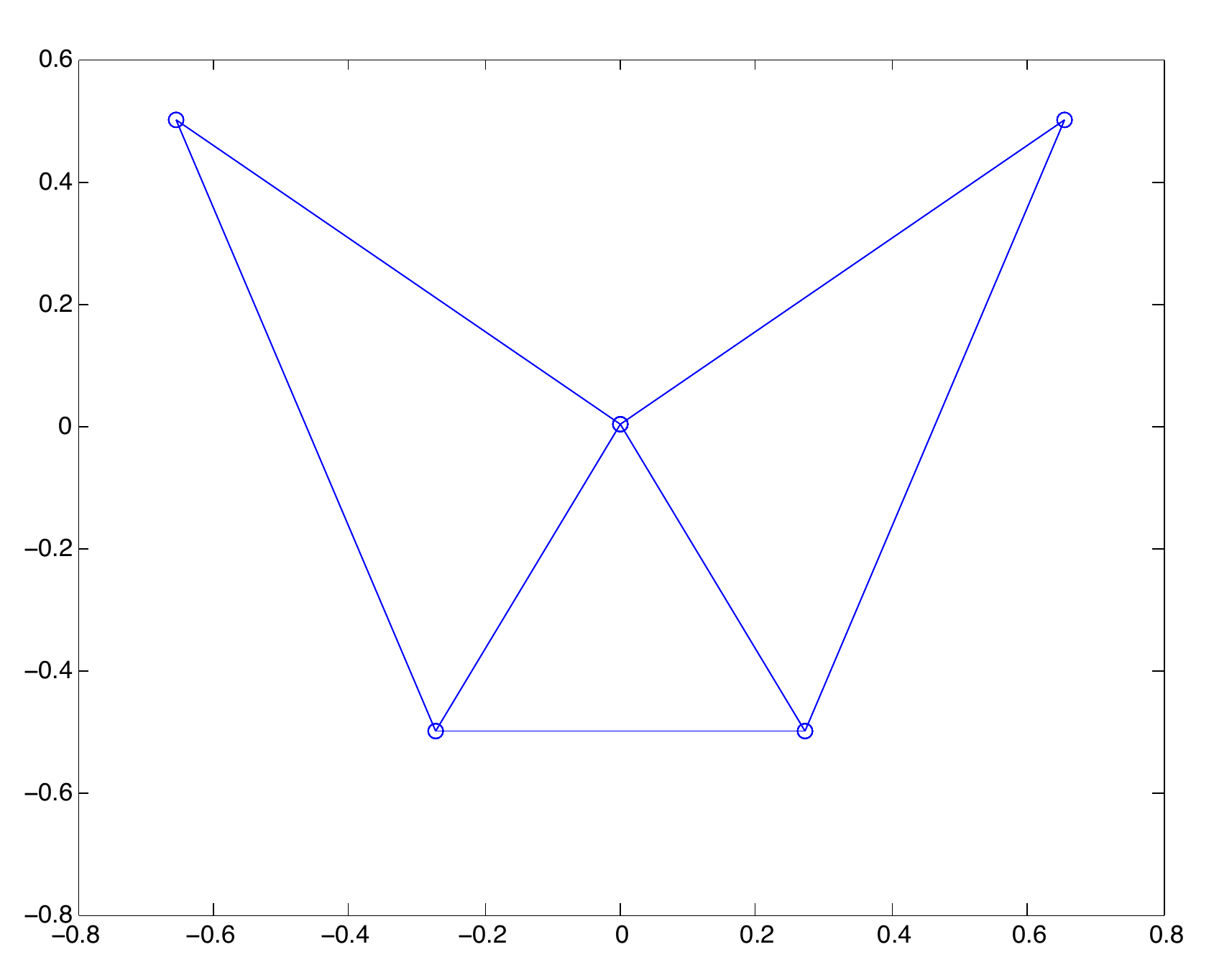}
  \end{center}
  \caption{Drawing of the graph from Example 2.}
\label{graph1b}
\end{figure}

\medskip\noindent
{\it Example\/} 2.
Consider the  graph $G_2$ shown in Figure \ref{graphfig5bis} given 
by the adjacency matrix 
\[
A = 
\begin{pmatrix}
0 & 1 & 1 & 0 & 0 \\
1 & 0 & 1 & 1 & 1 \\
1 & 1 & 0 & 1 & 0 \\
0 & 1 & 1 & 0 & 1 \\
0 & 1 & 0 & 1 & 0
\end{pmatrix}.
\]
We use the following program to compute $u_2$ and $u_3$:

\begin{verbatim}
A = [0 1 1 0 0; 1 0 1 1 1; 1 1 0 1 0; 0 1 1 0 1; 0 1 0 1 0];
D = diag(sum(A));
L = D - A;
[v, e] = eig(L);
gplot(A, v(:, [2 3]))
hold on
gplot(A, v(:, [2 3]),'o')
\end{verbatim}

The function  {\tt eig(L)}  (with no {\tt s} at the end) computes the  eigenvalues of $L$ in increasing
order. The result of drawing the graph is shown in Figure \ref{graph1b}.
Note that node $v_2$ is
assigned to the point $(0, 0)$, so the difference between this drawing
and the drawing  in Figure \ref{graphfig5bis}  is that the
drawing of Figure \ref{graph1b} is not convex.

\medskip\noindent
{\it Example\/} 3.
Consider the ring graph defined by the adjacency matrix $A$ given in
the {\tt Matlab} program shown below:

\begin{verbatim}
A = diag(ones(1, 11),1);
A = A + A';
A(1, 12) = 1; A(12, 1) = 1;
D = diag(sum(A));
L = D - A;
[v, e] = eig(L);
gplot(A, v(:, [2 3]))
hold on
gplot(A, v(:, [2 3]),'o')
\end{verbatim}

\begin{figure}[http]
  \begin{center}
   \includegraphics[height=2.2truein,width=2.4truein]{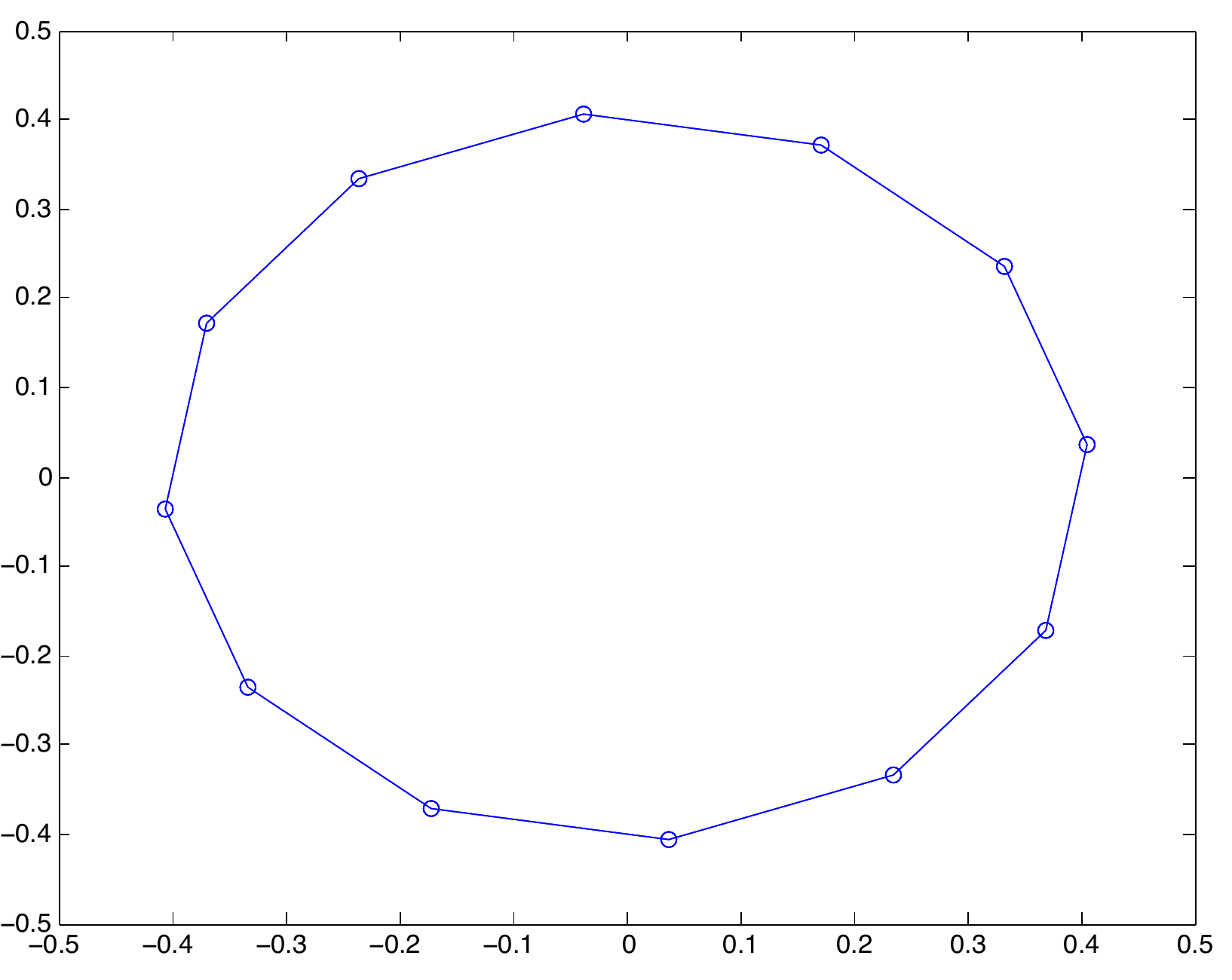}
  \end{center}
  \caption{Drawing of the graph from Example 3.}
\label{graph2}
\end{figure}

Observe that we get a very nice ring; see Figure \ref{graph2}.
Again $\lambda_2 =
0.2679$ is a double eigenvalue (and so are the next pairs of
eigenvalues, except the last, $\lambda_{12} = 4$).

\medskip\noindent
{\it Example\/} 4.
In this example adpated from Spielman, we generate $20$ randomly
chosen points in the unit square, compute their Delaunay
triangulation, then the adjacency matrix of the corresponding graph,
and finally draw the graph using the second and third eigenvalues of
the Laplacian.

\begin{verbatim}
A = zeros(20,20);
xy = rand(20, 2); 
trigs = delaunay(xy(:,1), xy(:,2));
elemtrig = ones(3) - eye(3);
for i = 1:length(trigs),
 A(trigs(i,:),trigs(i,:)) = elemtrig;
end
A = double(A >0);
gplot(A,xy)
D = diag(sum(A));
L = D - A;
[v, e] = eigs(L, 3, 'sm');
figure(2)
gplot(A, v(:, [2 1]))
hold on
gplot(A, v(:, [2 1]),'o')
\end{verbatim}

\begin{figure}[http]
  \begin{center}
   \includegraphics[height=2.2truein,width=2.4truein]{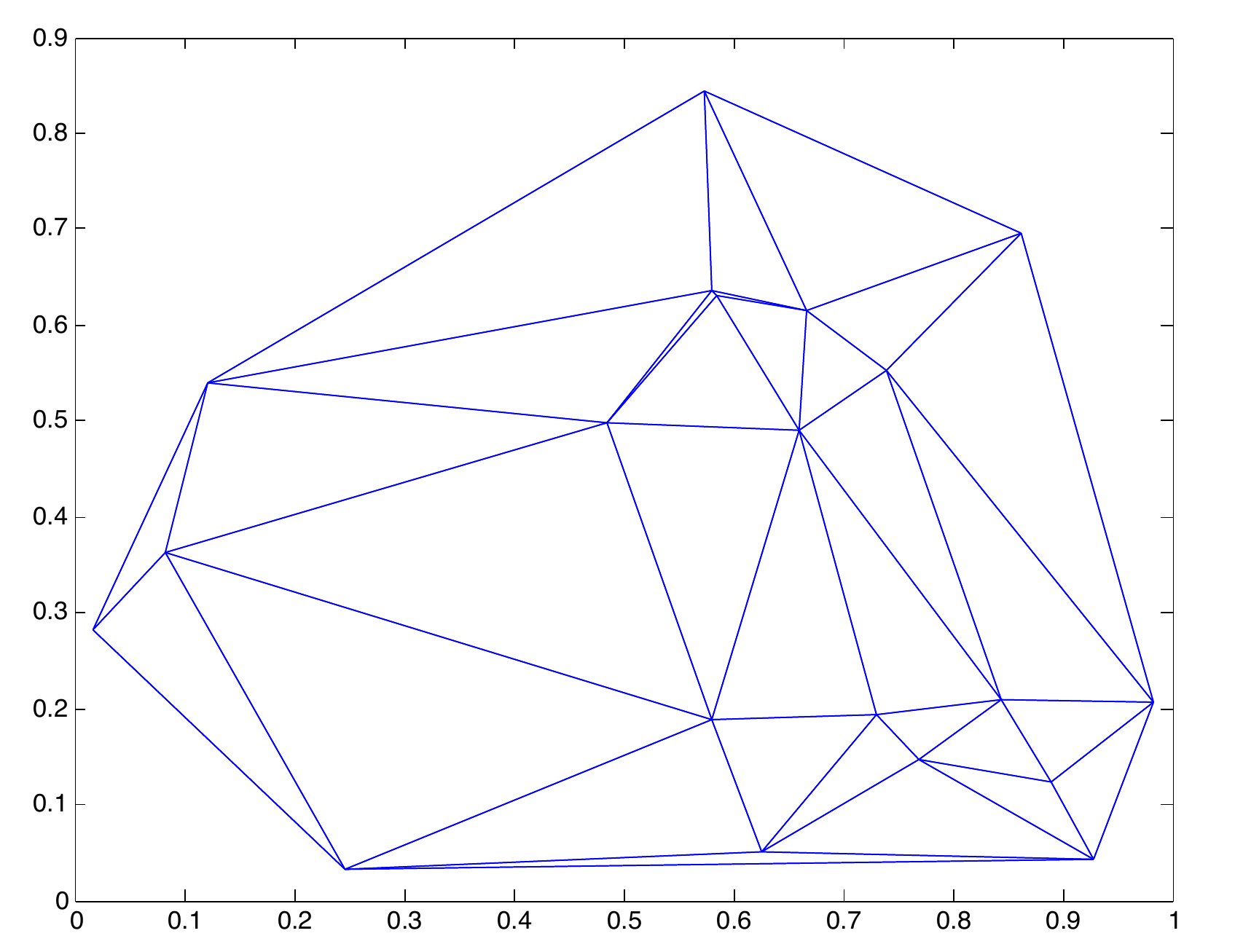}
\hspace{1cm}
   \includegraphics[height=2.2truein,width=2.4truein]{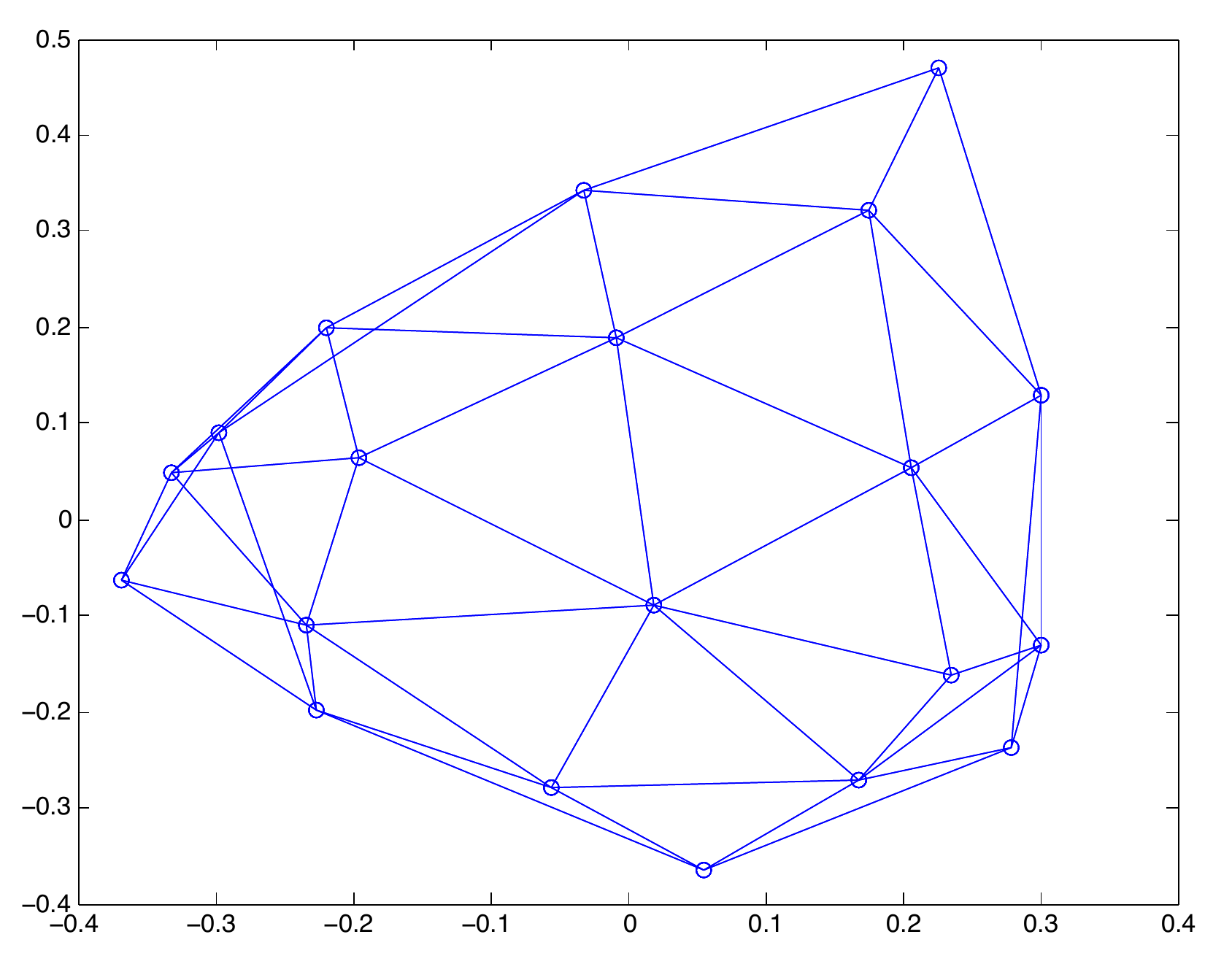}
  \end{center}
  \caption{Delaunay triangulation (left) and
drawing of the graph from Example 4 (right).}
\label{graph3}
\end{figure}

The Delaunay triangulation of the set of $20$ points and the drawing
of the corresponding graph are shown in Figure \ref{graph3}.
The graph drawing on the right looks nicer than the graph on the left
but is is no longer planar.

\medskip\noindent
{\it Example\/} 5.
Our last example, also borrowed from Spielman \cite{Spielman}, corresponds to  the 
skeleton of the ``Buckyball,''  a geodesic dome 
invented by the architect Richard Buckminster Fuller (1895--1983).
The Montr\'eal Biosph\`ere is an example of a geodesic dome designed by
 Buckminster Fuller.

\begin{verbatim}
A = full(bucky);
D = diag(sum(A));
L = D - A;
[v, e] = eig(L);
gplot(A, v(:, [2 3]))
hold on;
gplot(A,v(:, [2 3]), 'o')
\end{verbatim}

\begin{figure}[http]
  \begin{center}
\hspace{1cm}
   \includegraphics[height=2truein,width=2.2truein]{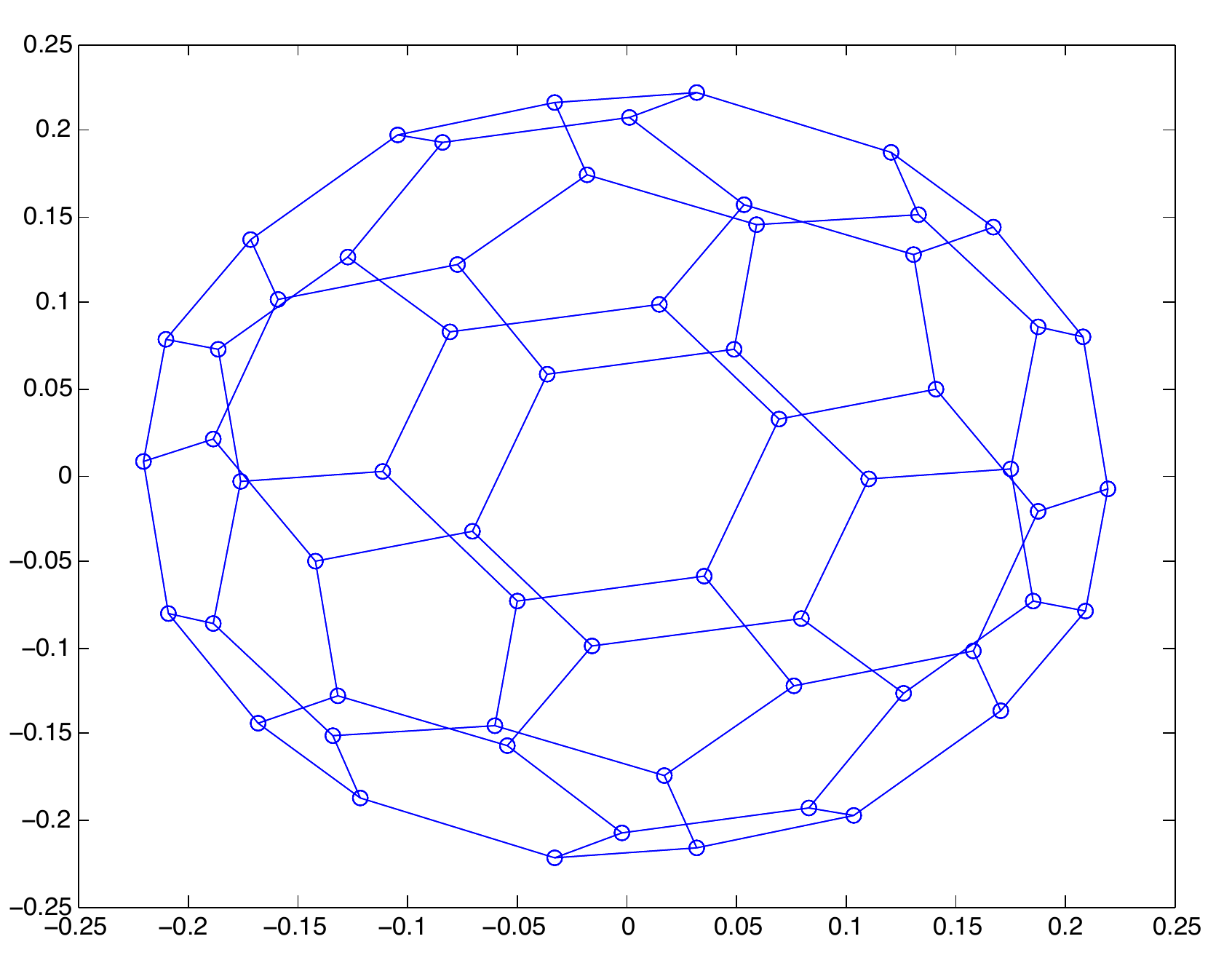}
  \end{center}
  \caption{Drawing of the graph of the Buckyball.}
\label{graph4}
\end{figure}

Figure \ref{graph4} shows a  graph drawing of the Buckyball. This
picture seems a bit squashed for two reasons. First, it is really a
$3$-dimensional graph; second, $\lambda_2 = 0.2434$ is a triple
eigenvalue.
(Actually, the Laplacian of $L$  has many multiple eigenvalues.)
What we should really do is to plot this graph in $\reals^3$ using
three orthonormal eigenvectors associated with $\lambda_2$.

\begin{figure}[http]
  \begin{center}
\hspace{1cm}
   \includegraphics[height=2.7truein,width=4truein]{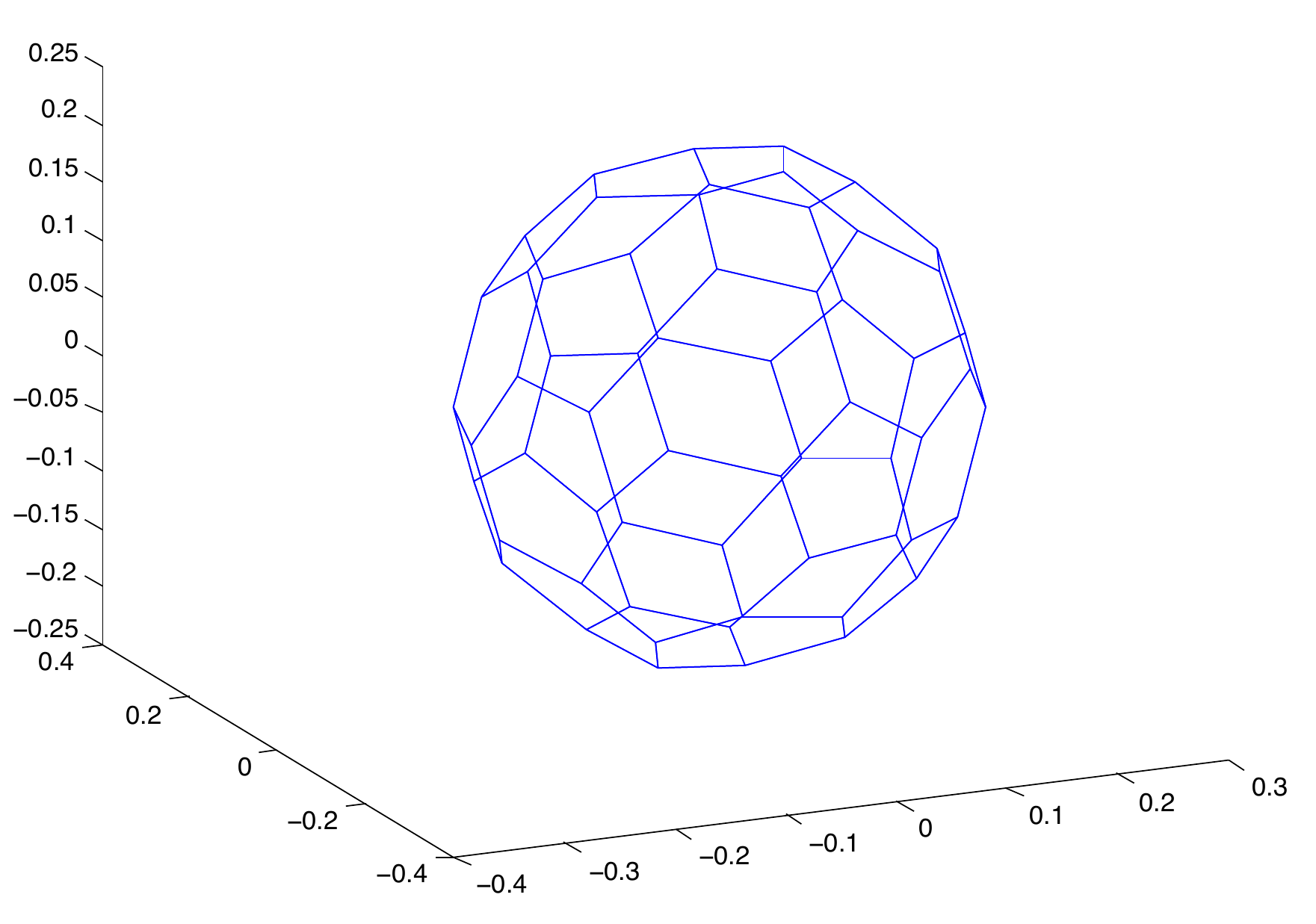}
  \end{center}
  \caption{Drawing of the graph of the Buckyball in $\reals^3$.}
\label{graph5}
\end{figure}

\medskip
A $3$D picture of the  graph of the Buckyball is produced by the
following {\tt Matlab} program, and its image is shown in Figure  \ref{graph5}.
It looks better!

\begin{verbatim}
[x, y] = gplot(A, v(:, [2 3]));
[x, z] = gplot(A, v(:, [2 4]));
plot3(x,y,z)
\end{verbatim}

\chapter{Graph Clustering}
\label{chap3}
\section{Graph Clustering Using Normalized  Cuts}
\label{ch3-sec1}
Given a set of data, the goal of clustering is to partition the data
into different groups according to their similarities. When the data
is given in terms of a similarity graph $G$, where the weight $w_{i\, j}$
between two nodes $v_i$ and $v_j$ is a measure of similarity of
$v_i$ and $v_j$, the problem can be stated as follows:
Find a partition $(A_1, \ldots, A_K)$ of the set of nodes $V$ into
different groups such   that the edges between different groups have
very low weight (which indicates that the points in different clusters
 are dissimilar),  and the edges within a group have high weight
 (which indicates that points within the same cluster are similar).

\medskip
The above graph clustering problem can be formalized as an
optimization problem, using the notion of cut mentioned at the 
end of Section \ref{ch1-sec1}.

\medskip
Given a subset $A$ of the set of vertices $V$, recall that we define $\mathrm{cut}(A)$ by
\[
\mathrm{cut}(A) = \mathrm{links}(A, \overline{A})= 
\sum_{v_i\in A, v_j\in \overline{A}} w_{i\, j},
\]
and that
\[
\mathrm{cut}(A)= \mathrm{links}(A, \overline{A}) = 
\mathrm{links}(\overline{A}, A) = \mathrm{cut}(\overline{A}).
\]
If we want to partition $V$ into $K$ clusters, we can do so by finding
a partition ($A_1, \ldots, A_K$) that  minimizes the quantity
\[
\mathrm{cut}(A_1, \ldots, A_K) = \frac{1}{2} \sum_{i = 1}^K \mathrm{cut}(A_i). 
\]
The reason for introducing the factor $1/2$ is to avoiding counting
each edge twice.  In particular,
\[
\mathrm{cut}(A,\overline{A}) = \mathrm{links} (A, \overline{A}).
\]
For $K = 2$,  the mincut problem is a classical
problem that can be solved efficiently, but in practice, it does not
yield satisfactory partitions. Indeed, in many cases, the mincut
solution separates one vertex from the rest of the graph.  What we
need is to design our cost function in such a way that it keeps the
subsets $A_i$ ``reasonably large'' (reasonably balanced).

\begin{figure}[http]
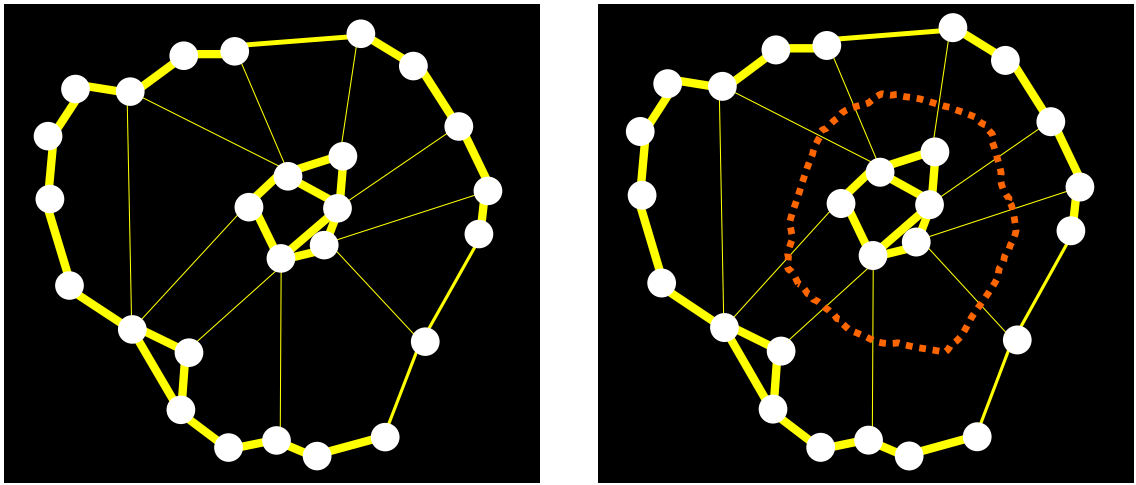

  \begin{center}
 \includegraphics[height=2.5truein,width=2.8truein]{ncuts-figs/ncuts-g-fig1.pdf}
\hspace{0.5cm}
 \includegraphics[height=2.5truein,width=2.8truein]{ncuts-figs/ncuts-g-fig2.pdf}
  \end{center}
  \caption{A weighted graph and its partition into two clusters.}
\label{ncg-fig4}
\end{figure}

\medskip
A example of a weighted graph and a partition of
its nodes into two clusters is shown in Figure \ref{ncg-fig4}.

\medskip
A way to get around this problem is to normalize the cuts by dividing
by some measure of each subset $A_i$. One possibility is to use the
size (the number of elements) of $A_i$. Another is to use the volume
$\mathrm{vol}(A_i)$ of $A_i$. A solution using the second measure (the volume)
(for $K = 2$) was proposed and
investigated in a seminal paper of  Shi and Malik \cite{ShiMalik}.
Subsequently, Yu (in her dissertation \cite{Yu}) and Yu and Shi
\cite{YuShi2003} extended the method to $K > 2$ clusters.
We will describe this  method later.
The idea is  to minimize the cost function
\[
\mathrm{Ncut}(A_1, \ldots, A_K) = 
 \sum_{i = 1}^K \frac{\mathrm{links}(A_i, \overline{A_i})}{\mathrm{vol}(A_i)}
= \sum_{i = 1}^K 
\frac{\mathrm{cut}(A_i, \overline{A_i})}{\mathrm{vol}(A_i)}.
\]

\medskip
We begin with the case $K = 2$, which is easier to handle.

\section{Special Case: $2$-Way Clustering Using Normalized  Cuts}
\label{ch3-sec2}
Our goal is to express our  optimization problem in matrix form.
In the case of two clusters, a single vector $X$ can be used to describe the
partition $(A_1, A_2)  = (A, \overline{A})$. We need to choose the structure of this vector
in such a way that $\mathrm{Ncut}(A, \overline{A})$ 
is equal to the Rayleigh ratio
\[
\frac{\transpos{X} L X}{\transpos{X} \Degsym X}.
\]
It is also important to pick a vector representation 
which is invariant under multiplication by a nonzero scalar,
because the Rayleigh ratio is scale-invariant, and it is crucial to take
advantage of this fact to make the denominator go away.

\medskip
Let $N = |V|$ be the number of nodes in the graph $G$.
In view of the desire for a scale-invariant representation,
it is natural
to assume that the  vector $X$ is of the form
\[
X = (x_1, \ldots, x_N),
\]
where $x_i \in \{a, b\}$ for $i = 1, \ldots, N$, 
for any two distinct  real numbers $a, b$. 
This is an indicator vector in
the sense that, for $i = 1, \ldots, N$,
\[
x_i =
\begin{cases}
a & \text{if $v_i \in A$} \\
b & \text{if $v_i \notin A$} .
\end{cases}
\]

The correct interpretation is really to 
view $X$ as a representative of a point 
in the real projective space $\mathbb{RP}^{N-1}$,  
namely the point $\mathbb{P}(X)$ of homogeneous
coordinates $(x_1\co \cdots \co x_N)$.
Therefore,  from now on, we  view $X$ as a vector of homogeneous
coordinates representing  the point  $\mathbb{P}(X)\in \mathbb{RP}^{N-1}$.

\medskip
Let  $d = \transpos{\mathbf{1}} \Degsym \mathbf{1}$ and $\alpha =
\mathrm{vol}(A)$.  
Then, $\mathrm{vol}(\overline{A}) = d - \alpha$.
By Proposition \ref{Laplace1}, we have
\[
\transpos{X} L X = (a - b)^2\, \mathrm{cut}(A, \overline{A}),
\]
and we easily check that
\[
\transpos{X} \Degsym  X = \alpha a^2 + (d - \alpha)b^2.
\]
Since $\mathrm{cut}(A, \overline{A}) = \mathrm{cut}(\overline{A}, A)$,
we have
\[
\mathrm{Ncut}(A, \overline{A}) = 
\frac{\mathrm{cut}(A,  \overline{A})}{\mathrm{vol}(A)} +
\frac{\mathrm{cut}(\overline{A}, A)}{\mathrm{vol}(\overline{A})} 
= 
\left(\frac{1}{\mathrm{vol}(A)}  +
  \frac{1}{\mathrm{vol}(\overline{A})} \right) \mathrm{cut}(A, \overline{A}),
\]
so we obtain
\[
\mathrm{Ncut}(A, \overline{A}) 
= \left(\frac{1}{\alpha}  +
  \frac{1}{d - \alpha} \right) \mathrm{cut}(A, \overline{A})
=
\frac{d}{\alpha(d - \alpha)}\,\mathrm{cut}(A, \overline{A}).
\]
Since
\[
\frac{\transpos{X} L X}{\transpos{X} \Degsym X} =
\frac{(a - b)^2}{\alpha a^2 + (d - \alpha)b^2}\, \mathrm{cut}(A, \overline{A}),
\]
in order to have
\[
\mathrm{Ncut}(A, \overline{A}) = \frac{\transpos{X} L X}{\transpos{X} \Degsym X},
\]
we need to find $a$ and $b$ so that
\[
\frac{(a - b)^2}{\alpha a^2 + (d - \alpha)b^2} = 
\frac{d}{\alpha(d - \alpha)}.
\]
The above is equivalent to
\[
(a - b)^2 \alpha(d - \alpha) =  \alpha d a^2 + (d - \alpha)d b^2,
\]
which can be rewritten as
\[
a^2(\alpha d - \alpha(d - \alpha)) + b^2(d^2 - \alpha d - \alpha(d -
\alpha)) + 2 \alpha(d - \alpha) a b = 0.
\]
The above yields
\[
a^2 \alpha^2 + b^2(d^2 - 2\alpha d + \alpha^2) + 2\alpha(d - \alpha) a
b = 0,
\]
that is,
\[
a^2\alpha^2 + b^2(d - \alpha)^2 + 2\alpha(d - \alpha) a b = 0,
\]
which reduces to
\[
(a \alpha + b(d - \alpha))^2 = 0.
\]
Therefore, we get the condition
\begin{equation}
a \alpha + b(d - \alpha) = 0.
\tag{$\dagger$}
\end{equation}
Note that condition $(\dagger)$ applied to a vector $X$ whose
components are $a$ or $b$ is equivalent to the fact that $X$
is orthogonal to $\Degsym \mathbf{1}$, since
\[
\transpos{X} \Degsym \mathbf{1} =  \alpha  a+ (d - \alpha) b,
\]
where $\alpha = \mathrm{vol}(\{v_i\in V \mid x_i = a\})$.

\medskip
We claim the following two facts.
For any nonzero vector $X$ whose components are $a$ or
$b$, if  $\transpos{X} \Degsym \mathbf{1} =  \alpha  a+ (d - \alpha) b
= 0$, then
\begin{enumerate}
\item[(1)]
$\alpha\not= 0$ and $\alpha \not= d$ iff 
$a\not= 0$ and  $b\not= 0$.
\item[(2)]
if $a, b\not= 0$, then $a\not = b$. 
\end{enumerate}

\medskip
(1)
First assume that $a \not= 0$ and $b \not= 0$. If $\alpha = 0$, then
$\alpha  a + (d - \alpha) b = 0$
yields $d  b = 0$ with $d\not= 0$, which implies $b = 0$, a contradiction.
If $d - \alpha = 0$, then we get $d a = 0$ with $d\not= 0$, which implies $a = 0$, a contradiction.

\medskip
Conversely, assume that $\alpha\not= 0$ and $\alpha \not= d$. If
$a = 0$, then from  $\alpha  a+ (d - \alpha) b = 0$ we get
$(d - \alpha) b = 0$, which implies $b = 0$, contradicting the fact
that $X\not = 0$. 
Similarly, if $b = 0$,  then we get $\alpha  a = 0$, 
which implies $a = 0$, contradicting the fact
that $X\not = 0$. 

\medskip
(2) 
If $a, b\not= 0$, $a = b$ and $\alpha  a+ (d - \alpha) b = 0$, then 
$\alpha  a+ (d - \alpha) a = 0$,  and since $a\not= 0$, we deduce that
$d = 0$, a contradiction.

\medskip
If  $\transpos{X} \Degsym \mathbf{1} =  \alpha  a+ (d - \alpha) b = 0$
and $a, b\not= 0$, then
\[
b = -\frac{\alpha}{(d - \alpha)}\, a, 
\]
so we get
\begin{align*}
\alpha a^2 + (d - \alpha)b^2& = \alpha \frac{(d -
  \alpha)^2}{\alpha^2} b^2+ (d - \alpha)b^2 \\
& = (d - \alpha)\left( \frac{d - \alpha}{\alpha} + 1 \right)b^2 
 = \frac{(d - \alpha) d b^2}{\alpha},
\end{align*}
and
\begin{align*}
(a - b)^2 & = \left( -\frac{(d - \alpha)}{\alpha}\, b - b \right)^2 \\
& = \left( \frac{d - \alpha}{\alpha} + 1 \right)^2b^2 
 = \frac{d^2 b^2}{\alpha^2}.
\end{align*}
Since 
\begin{align*}
\transpos{X} \Degsym X & = \alpha a^2 + (d - \alpha)b^2 \\
\transpos{X} L X & =  (a - b)^2\, \mathrm{cut}(A, \overline{A}),
\end{align*}
we obtain
\begin{align*}
\transpos{X} \Degsym X & =  \frac{(d - \alpha)d b^2}{\alpha}
=  \frac {\alpha d  a^2}{(d - \alpha)} \\
\transpos{X} L X & =  
\frac{d^2 b^2}{\alpha^2} \, \mathrm{cut}(A, \overline{A}) 
= \frac{d^2 a^2}{(d - \alpha)^2} \, \mathrm{cut}(A, \overline{A}) .
\end{align*} 
If we wish to make $\alpha$ disappear, we   pick
\[
a = \sqrt{\frac{d - \alpha}{\alpha}}, \quad
b = -\sqrt{\frac{\alpha}{d - \alpha}},
\]
and then
\begin{align*}
\transpos{X} \Degsym X & =   d  \\
\transpos{X} L X & =  
\frac{d^2}{\alpha(d - \alpha)} \,
\mathrm{cut}(A, \overline{A})
= d\, \mathrm{Ncut}(A, \overline{A}).
\end{align*} 
In this case, we are  considering indicator vectors of the form
\[
\left\{
(x_1, \ldots, x_N) \mid
x_i \in \left\{
\sqrt{\frac{d - \alpha}{\alpha}},
-\sqrt{\frac{\alpha}{d - \alpha}} \right\},
\alpha = \mathrm{vol}(A)  \right\},
\]
for any nonempty proper subset $A$
of $V$. This is the choice adopted in 
von Luxburg \cite{Luxburg}.
Shi and Malik \cite{ShiMalik} use
\[
a = 1, \quad b = -\frac{\alpha}{d - \alpha} = - \frac{k}{1 - k},
\]
with
\[
k = \frac{\alpha}{d}.
\]
Another choice found in the literature (for example, in Belkin and Niyogi
\cite{Belkin-Niyogi}) is
\[
a  = \frac{1}{\alpha}, \quad
b  = -\frac{1}{d - \alpha} .
\]
However, there is no need to restrict solutions to be of either of these forms. 
So, let
\[
\s{X} = \big\{
(x_1, \ldots, x_N) \mid x_i \in \{a, b\}, \> a, b\in \reals,\> a,
b\not = 0
\big\},
\]
so that our solution set is
\[
\s{K}  = \big\{
X  \in\s{X}  \mid \transpos{X}  \Degsym\mathbf{1} = 0
\big\},
\]
because by previous observations, since  vectors
$X\in \s{X}$ have nonzero components,  $\transpos{X}\Degsym\mathbf{1} = 0$ 
implies that $\alpha \not= 0$,  $\alpha \not= d$, and $a\not= b$, where
$\alpha = \mathrm{vol}(\{v_i\in V \mid x_i = a\})$.
Actually, to be perfectly rigorous, we are looking for solutions in
$\mathbb{RP}^{N-1}$, so our solution set is really
\[
\mathbb{P}(\s{K})  = \big\{
(x_1\co \cdots\co x_N) \in \mathbb{RP}^{N-1}\mid
(x_1, \ldots, x_N) \in \s{K}
\big\}.
\]
Consequently, our minimization problem can be stated as follows:

\medskip\noindent
{\bf Problem PNC1}
\begin{align*}
& \mathrm{minimize}     &  & 
\frac{\transpos{X} L X}{\transpos{X} \Degsym X} & &  &  &\\
& \mathrm{subject\ to} &  &
\transpos{X} \Degsym\mathbf{1} = 0,  & &  X\in \s{X}.      
\end{align*}

It is understood that the solutions are  points $\mathbb{P}(X)$
in $\mathbb{RP}^{N-1}$.

\medskip 
Since the Rayleigh ratio and the constraints 
$\transpos{X}\Degsym\mathbf{1} = 0$ and $X\in \s{X}$ 
are scale-invariant 
(for any $\lambda \not= 0$, 
the Rayleigh ratio does not change if $X$ is
replaced by $\lambda X$, $X\in \s{X}$ iff $\lambda X \in \s{X}$,
and $\transpos{ (\lambda X)}\Degsym\mathbf{1} = \lambda \transpos{
  X}\Degsym\mathbf{1} = 0$),
we are led to the following formulation of our problem:

\medskip\noindent
{\bf Problem PNC2}
\begin{align*}
& \mathrm{minimize}     &  & 
\transpos{X} L X & &  &  &\\
& \mathrm{subject\ to} &  & \transpos{X} \Degsym X = 1, &&
 \transpos{X} \Degsym\mathbf{1} = 0, && X\in \s{X}.   
\end{align*}

\medskip
Because  problem PNC2 requires the constraint 
$\transpos{X} \Degsym X =1$ to be satisfied,
it does not have the same set of solutions as  problem PNC1.
Nevertherless, 
problem PNC2 is equivalent to problem PNC1, in the sense that 
if $X$ is any minimal solution of PNC1, then $X/(\transpos{X} \Degsym
X)^{1/2}$ is a minimal solution of PNC2 (with the same minimal value
for the objective functions), and if $X$ is a minimal solution of
PNC2, then $\lambda X$ is a minimal solution for PNC1 for all
$\lambda\not= 0$ (with the same minimal value
for the objective functions). Equivalently, problems PNC1 and PNC2
have the same set of minimal solutions as points $\mathbb{P}(X)
\in\mathbb{RP}^{N-1}$ given by their homogeneous coordinates $X$.

\medskip
Unfortunately, this is an NP-complete problem, as
shown by Shi and Malik \cite{ShiMalik}.
As often with hard combinatorial problems, we can look for a {\it
  relaxation\/} of our problem, which means looking for an optimum in
a larger continuous domain. After doing this, the problem is to find
a discrete solution which is close to a continuous optimum of the
relaxed problem.

\medskip
The natural relaxation of this problem is to allow $X$ to be any nonzero
vector in $\reals^N$, and we get the problem:

\[
\mathrm{minimize} \quad \transpos{X} L X
\quad\mathrm{subject\ to} \quad \transpos{X} \Degsym X = 1, \quad
\transpos{X} \Degsym\mathbf{1} = 0.
\]

\medskip
In order to apply Proposition   \ref{PCAlem1}, we make the change of variable 
$Y = \Degsym^{1/2} X$, so that $X = \Degsym^{-1/2} Y$. Then, 
the condition   $\transpos{X}\Degsym X = 1$ becomes
\[
\transpos{Y} Y = 1,
\]
the condition 
\[
\transpos{X} \Degsym \mathbf{1} = 0
\]
becomes
\[
\transpos{Y} \Degsym^{1/2} \mathbf{1} = 0,
\]
and 
\[
\transpos{X} L X = \transpos{Y} \Degsym^{-1/2} L \Degsym^{-1/2} Y. 
\]
We obtain the problem:
\[
\mathrm{minimize} \quad \transpos{Y} \Degsym^{-1/2} L \Degsym^{-1/2}  Y
\quad\mathrm{subject\ to} \quad \transpos{Y}  Y = 1, \quad
\transpos{Y} \Degsym^{1/2}\mathbf{1} = 0.
\]

\medskip
Because $L\mathbf{1} = 0$, the vector $ \Degsym^{1/2} \mathbf{1}$
belongs to
the nullspace of the symmetric Laplacian $L_{\mathrm{sym}} =
\Degsym^{-1/2} L \Degsym^{-1/2}$.
By Proposition   \ref{PCAlem1}, 
minima are achieved by any unit eigenvector $Y$ of the
second eigenvalue $\nu_2 > 0$ of $L_{\mathrm{sym}}$.
Since $0$ is the smallest eigenvalue of $L_{\mathrm{sym}}$
and  since $ \Degsym^{1/2} \mathbf{1}$ belongs to the nullspace of
$L_{\mathrm{sym}}$, as the eigenvectors associated with distinct
eigenvalues are orthogonal, 
the vector $Y$ is orthogonal to $ \Degsym^{1/2}\mathbf{1}$,
so the constraint $\transpos{Y} \Degsym^{1/2}\mathbf{1} = 0$ is satisfied.
Then, $Z = \Degsym^{-1/2} Y$ is a solution of our original relaxed
problem.
Note that because $Z$ is nonzero and  orthogonal to $\Degsym \mathbf{1}$, a vector
with positive entries, it must have negative and positive entries.

\medskip
The next question is to figure how close is $Z$  to an exact  solution in
$\s{X}$.  
Actually, because solutions are points in
$\mathbb{RP}^{N-1}$, the correct statement of the question is:
Find an exact solution $\mathbb{P}(X) \in\mathbb{P}(\s{X})$
which is the closest (in a suitable sense) to the approximate
solution $\mathbb{P}(Z)\in \mathbb{RP}^{N-1}$. 
However, because $\s{X}$ is closed under the antipodal map, as
explained in Appendix \ref{ch3-sec6}, minimizing the distance
$d(\mathbb{P}(X), \mathbb{P}(Z))$ on $\mathbb{RP}^{N-1}$ is equivalent
to minimizing the Euclidean distance $\norme{X - Z}_2$,
where $X$ and $Z$ are representatives of $\mathbb{P}(X)$ and $\mathbb{P}(Z)$
on the unit sphere
(if we use the Riemannian metric on $\mathbb{RP}^{N-1}$ induced by the Euclidean
metric on $\reals^N$).

\medskip
We may assume $b < 0$, in which case $a > 0$.
If all entries in $Z$ are nonzero, due to the projective nature of the
solution set, it seems reasonable to
say that
the partition of $V$ is defined by the signs of the entries in $Z$. 
Thus, $A$ will consist of nodes those $v_i$ for which $x_i > 0$.
Elements corresponding to zero entries can be assigned to either $A$
or $\overline{A}$, unless additional information is available.
In our implementation, they are assigned to $A$.

\begin{figure}[http]
  \begin{center}
 \includegraphics[height=1.5truein,width=1.5truein]{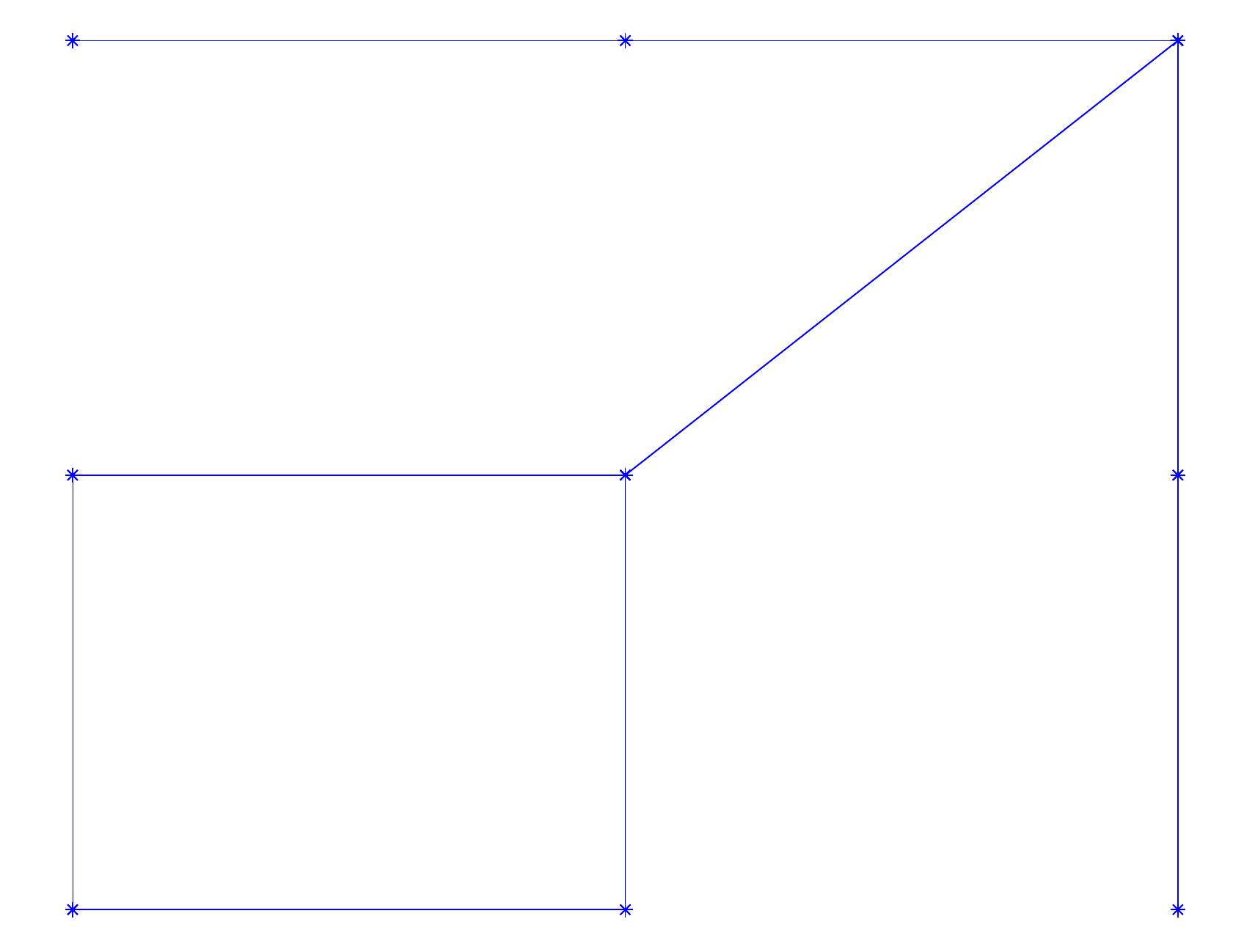}
\hspace{0.5cm}
 \includegraphics[height=1.5truein,width=1.5truein]{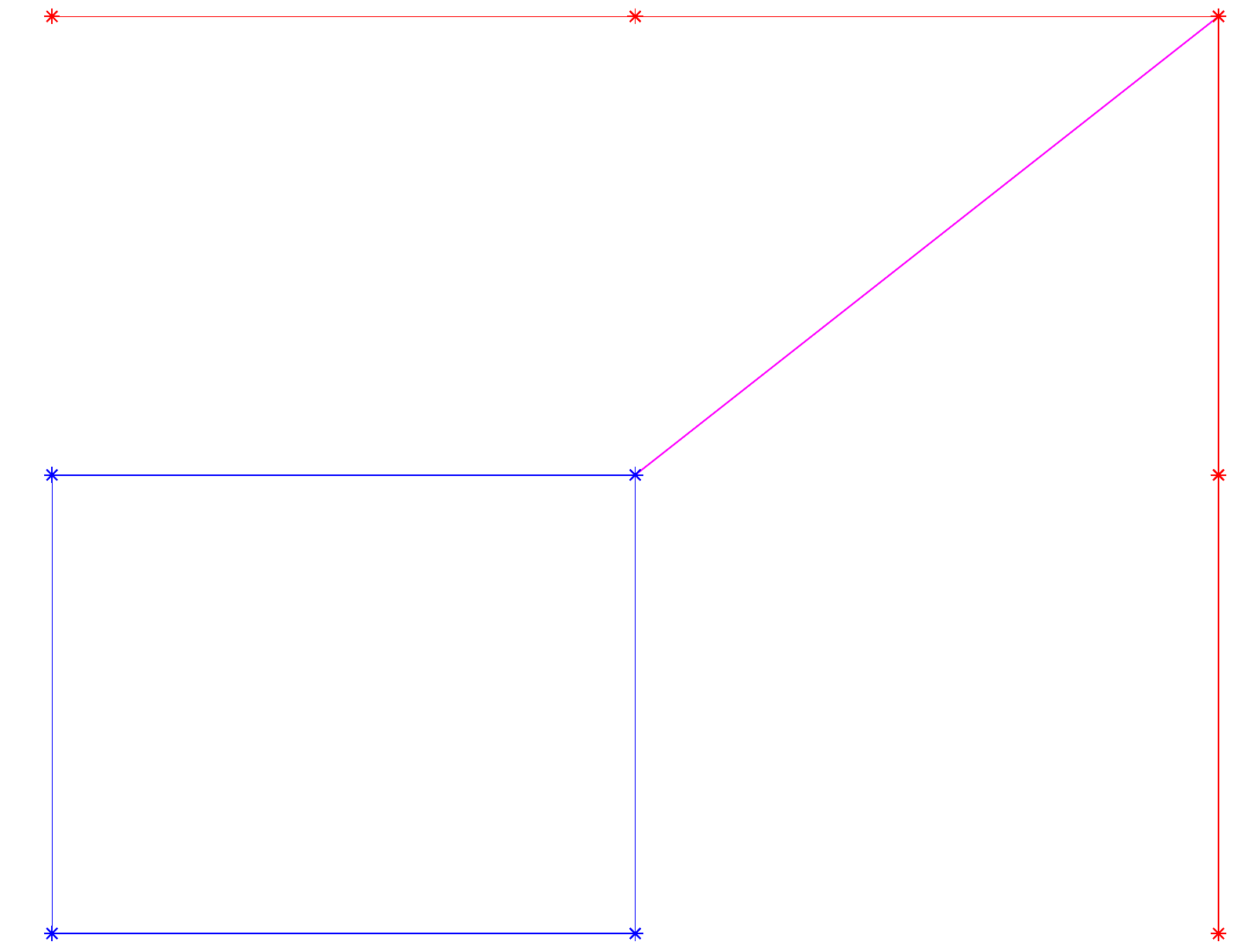}
\hspace{0.5cm}
 \includegraphics[height=1.5truein,width=1.5truein]{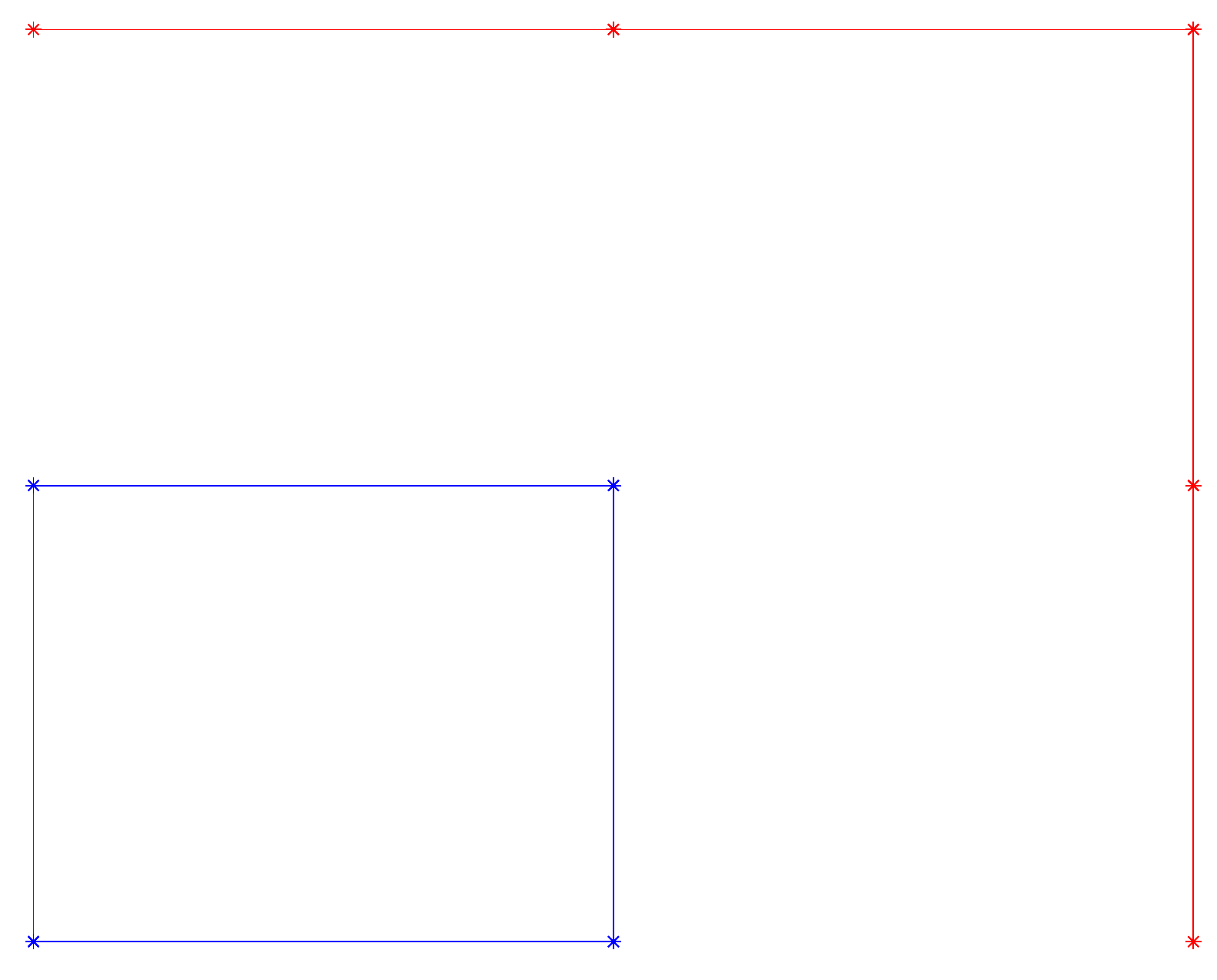}
  \end{center}
  \caption{Underlying graph of the matrix $W_1$ (left); normalized cut
    (middle); blocks of the cut (right).}
\label{gr1}
\end{figure}

\medskip
Here are some examples of normalized cuts found by a
fairly naive implementation of the method.
The weight matrix of the first example is
\[
W_1 = 
\begin{pmatrix}
0    & 1  &   0  &   1  &   0  &   0  &   0  &   0 &    0\\
     1  &   0  &   0  &   0   &  1  &   0  &   0  &   0  &   0\\
     0  &   0  &   0  &   0  &   0  &   1  &   0  &   0  &   0\\
     1  &   0  &   0   &  0  &   1  &   0  &   0  &   0  &   0\\
     0  &   1  &   0  &   1  &   0  &   0   &  0   &  0  &   1\\
     0  &   0  &   1   &  0  &   0  &   0  &   0  &   0  &   1\\
     0  &   0  &   0   &  0  &   0  &   0   &  0   &  1  &   0\\
     0  &   0  &   0   &  0  &   0  &   0   &  1   &  0  &   1\\
     0  &   0  &   0   &  0  &   1  &   1   &  0   &  1  &   0
\end{pmatrix}.
\]
Its underlying graph has $9$ nodes and $9$ edges and is shown in
Figure \ref{gr1} on the left. The normalized cut found by the
algorithm is shown in the middle; the edge of the cut is shown
in magenta, and the vertices of the blocks of the partition are
shown in blue and red.  The figure on the right shows the two 
disjoint subgraphs obtained after deleting the cut edge.

\begin{figure}[http]
  \begin{center}
 \includegraphics[height=1.5truein,width=1.5truein]{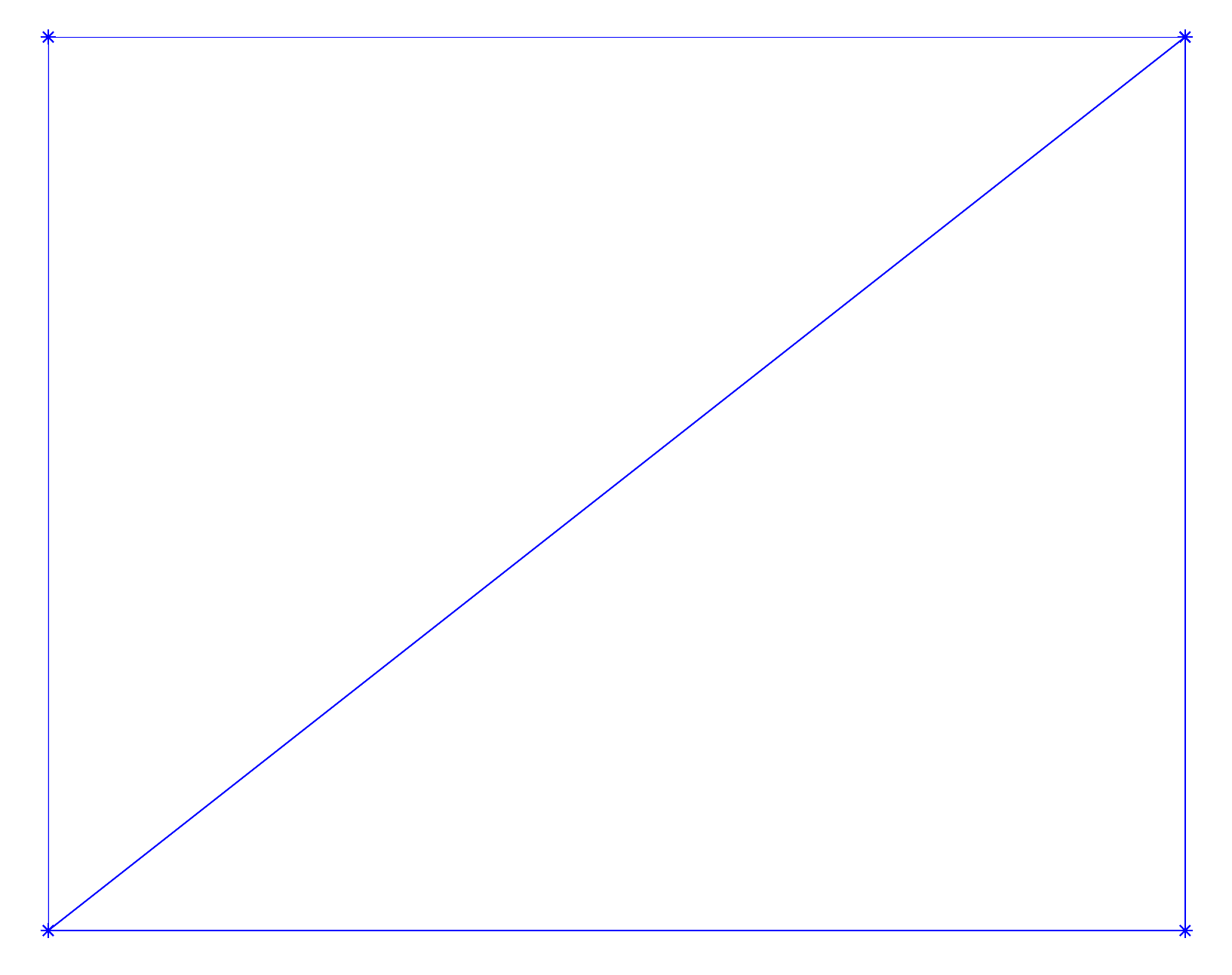}
\hspace{0.5cm}
 \includegraphics[height=1.5truein,width=1.5truein]{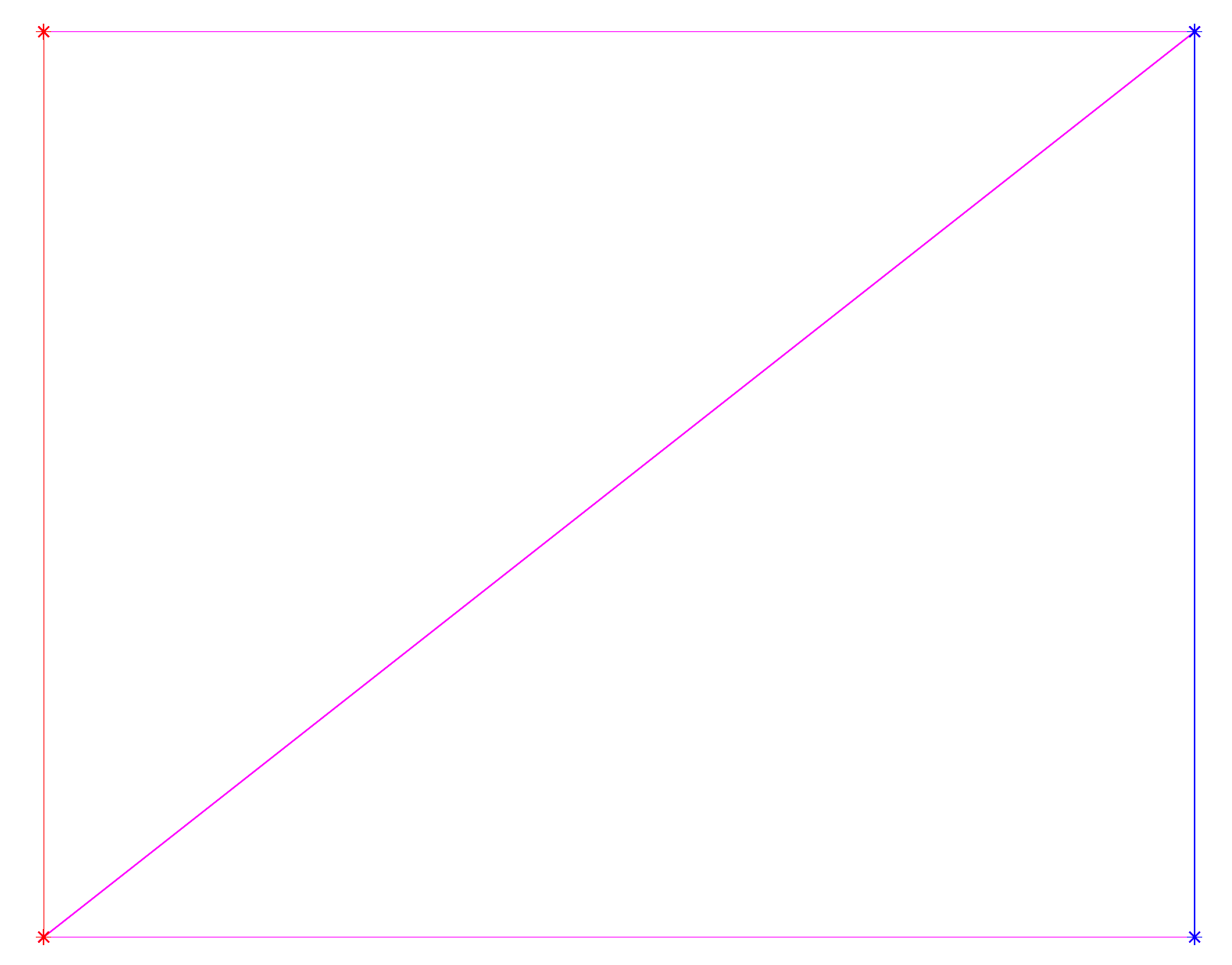}
\hspace{0.5cm}
 \includegraphics[height=1.5truein,width=1.5truein]{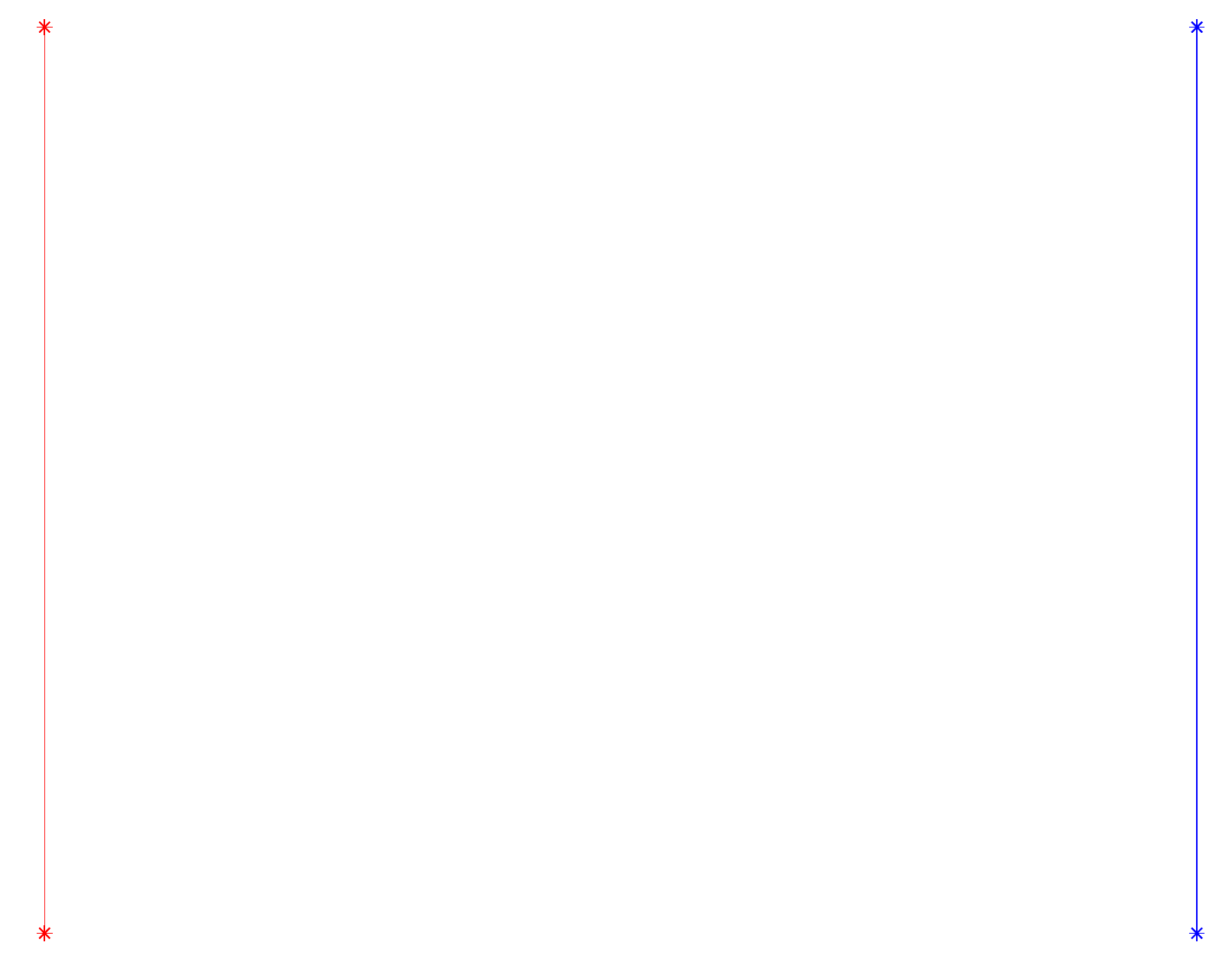}
  \end{center}
  \caption{Underlying graph of the matrix $W_2$ (left); normalized cut
    (middle); blocks of the cut (right).}
\label{gr2}
\end{figure}

The weight matrix of the second example is
\[
W_2 = 
\begin{pmatrix}
0  &   3  &   6  &   3\\
     3  &   0  &   0  &   3\\
     6  &   0  &   0  &   3\\
     3  &   3  &   3  &   0
\end{pmatrix} .
\]
Its underlying graph has $4$ nodes and $5$ edges and is shown in
Figure \ref{gr2} on the left. The normalized cut found by the
algorithm is shown in the middle; the edges of the cut are shown
in magenta, and the vertices of the blocks of the partition are
shown in blue and red.  The figure on the right shows the two 
disjoint subgraphs obtained after deleting the cut edges.

\begin{figure}[http]
  \begin{center}
 \includegraphics[height=1.5truein,width=1.5truein]{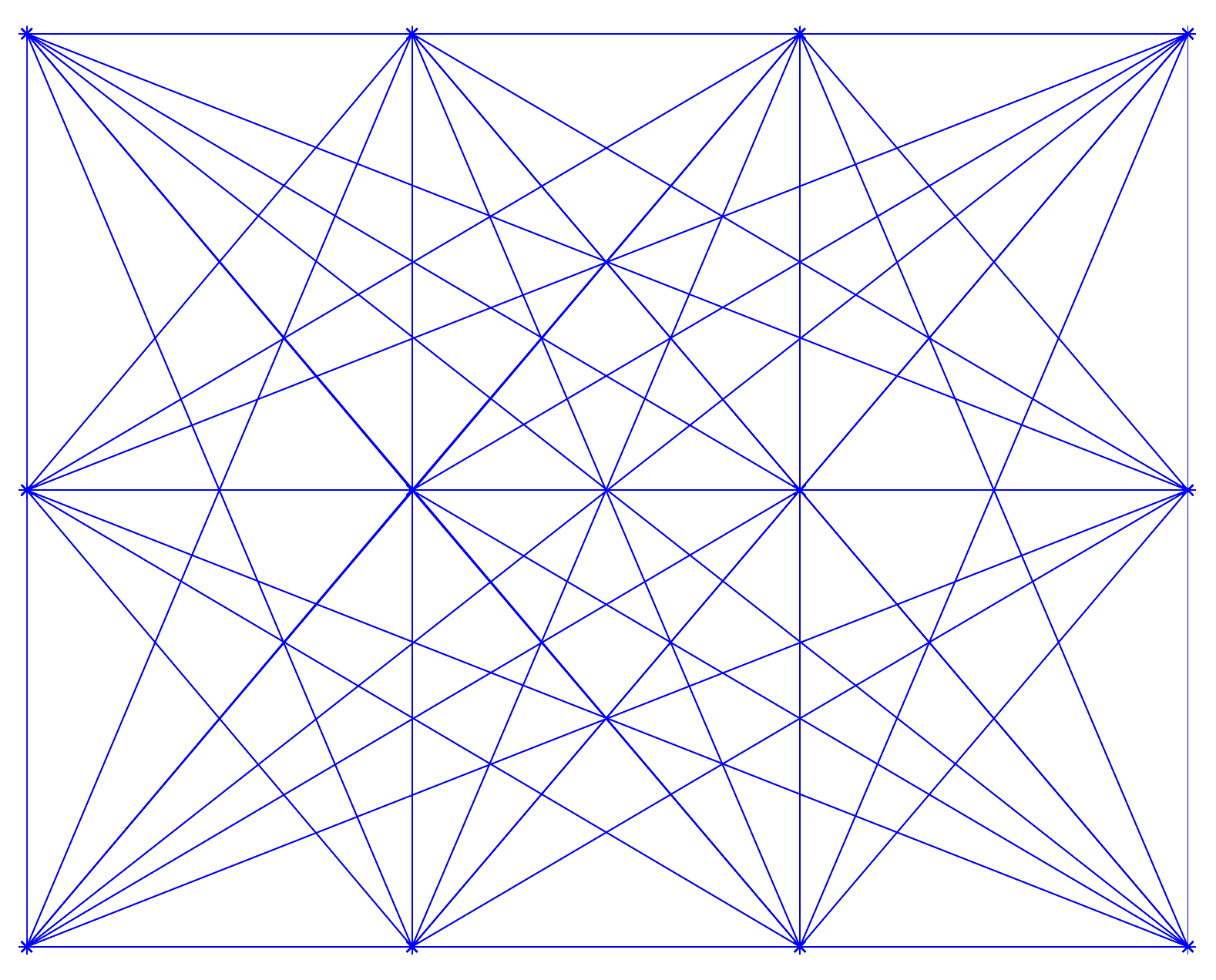}
\hspace{0.5cm}
 \includegraphics[height=1.5truein,width=1.5truein]{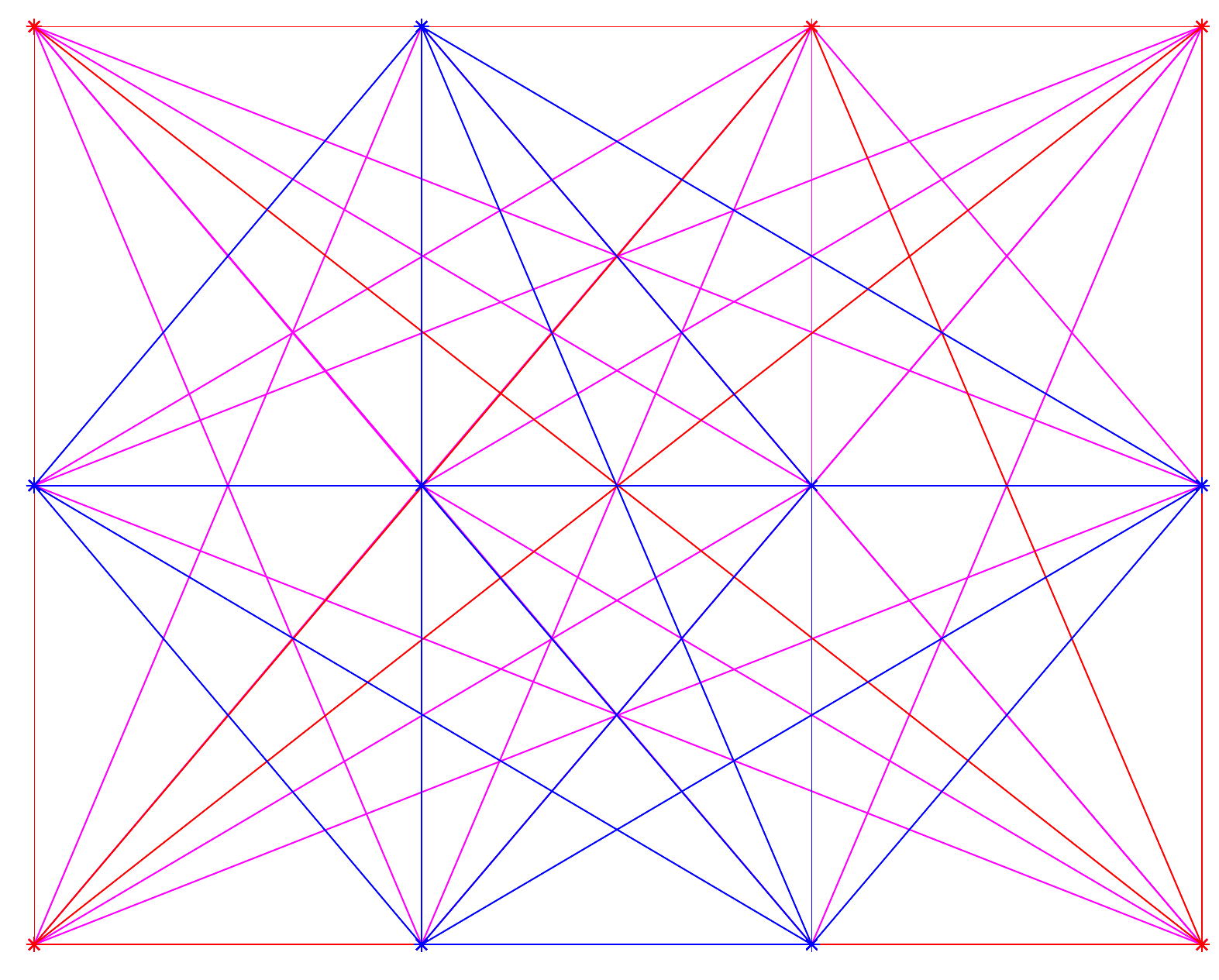}
\hspace{0.5cm}
 \includegraphics[height=1.5truein,width=1.5truein]{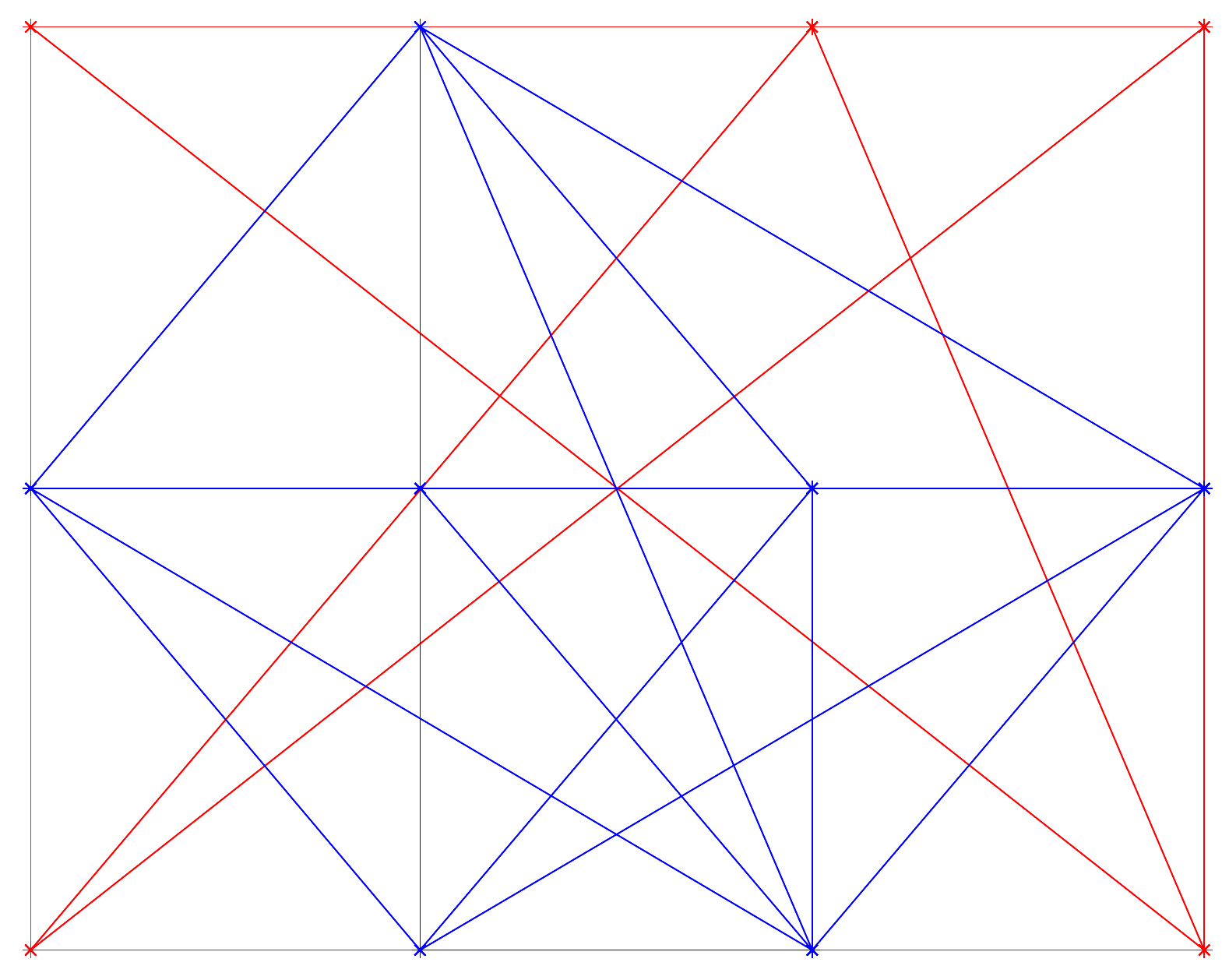}
  \end{center}
  \caption{Underlying graph of the matrix $W_3$ (left); normalized cut
    (middle); blocks of the cut (right).}
\label{gr3}
\end{figure}

The weight matrix $W_3$ of the third example is
the adjacency matrix of the complete graph on $12$ vertices.
All nondiagonal entries are equal to $1$, and the diagonal entries are
equal to $0$.
This graph has $66$ edges and is shown in
Figure \ref{gr3} on the left. 

The normalized cut found by the
algorithm is shown in the middle; the edges of the cut are shown
in magenta, and the vertices of the blocks of the partition are
shown in blue and red.  The figure on the right shows the two 
disjoint subgraphs obtained after deleting the cut edges.
Recall that $L_{\mathrm{sym}} = B_3\transpos{B_3}$ for  any incidence matrix $B_3$
associated with $W_3$, so that for any  SVD $U_3
\Sigma_3 V_3$ of $B_3$, the vectors in $U_3$ are eigenvectors of
$L_{\mathrm{sym}}$ for its eigenvalues listed in decreasing order.
The normalized Laplacian of this weight matrix has the eigenvalue
$1.0909$ with multiplicity $11$ (and any incidence matrix $B_3$
associated with $W_3$ has the singular value $1.0445$
with multiplicity $11$). 
Computing the SVD $U_3 \Sigma_3 V_3$ 
of $B_3$ and picking the next to the last eigenvector in $U_3$ yields
a partition consisting of $7$ and $5$ nodes. There are other
eigenvectors that yield partitions with an equal number of elements.
Since a complete graph has a lot of symmetries, it is not surprising
that there are many different solutions. In fact, examination of the
eigenvectors of $U_3$ reveal very unbalanced solutions.

\medskip
For graphs where the number $N$ of edges is very large and the number 
of edges is $O(N^2)$, computing the SVD of the incidence matrix $B$ is not 
practical.  Instead, we compute an SVD for $L_{\mathrm{sym}}$, which
appears to be more stable that  diagonalizing $L_{\mathrm{sym}}$.

\medskip
Our naive algorithm treated zero as a positive entry.
Now, using the fact that
\[
b = - \frac{\alpha a}{d - \alpha},
\]
a better solution is to look for a vector  $X\in \reals^N$ with $X_i
\in \{a, b\}$ which is closest to a  minimum $Z$ of the relaxed
problem (in the sense that $\norme{X - Z}$ is minimized)
and with $\norme{X} = \norme{Z}$.
We designed the following  algorithm. 

\medskip
A vector $X$ describing a partition $(A, \overline{A})$ is of the form
\[
X_i = 
\begin{cases}
a & \text{if $v_i\in A$} \\
-\beta a & \text{otherwise}, 
\end{cases}
\]
with 
\begin{align*}
\alpha & = \mathrm{vol}(\{v_i \mid v_i \in A\}), &
\beta & = \frac{\alpha}{d - \alpha} ,
\end{align*}
and where $a > 0$ is chosen so  
that $\norme{\overline{X}} =  \norme{Z}$.
For any  solution $Z$ of the relaxed problem,
let $I_{Z}^{+} = \{i \mid Z_i > 0\}$ be the set of indices of positive
entries in $Z$,  $I_{Z}^{-} = \{i \mid Z_i < 0\}$ the set of
indices of negative
entries in $Z$, $I_{Z}^{0} = \{i \mid Z_i = 0\}$  the set of indices
of zero entries in $Z$.  
Initially, it makes sense to form a discrete approximation $\overline{X}$ of
$Z$ such that all entries of index in $I_Z^+$ are assigned the value
$a > 0$ (to be determined later), and all other entries are assigned
the value $-\beta a$.
In order for $\overline{X}$ and $Z$ to have the same norm, since
\[
\norme{\overline{X}}^2 = n_a + \beta^2(N - n_a)
\]
with
\begin{align*}
n_a & = |I_Z^{+} |, &
\alpha & = \mathrm{vol}(\{v_i \mid i \in I_Z^{+}\}), &
\beta & = \frac{\alpha}{d - \alpha} ,
\end{align*}
we set
\[
a =\frac{\norme{Z}}{\sqrt{(n_a + \beta^2(N - n_a))}}.
\]

The problems is to determine whether an entry with an index
$i\in I_Z^0$ (which is initially assigned the value $-\beta a$)
should be reassigned the value $a$.
To make the decision, we form the new discrete solution
$\widetilde{X}$ obtained from $\overline{X}$ by
adding the index $i$ to $I_Z^+$, and updating $\alpha, \beta$ and $a$;
this is done in step (2). Then, we compare $\smnorme{\overline{X} - Z}$
and $\smnorme{\widetilde{X} - Z}$ and keep the vector that yields the
smallest norm. We delete $i$ from $I_Z^0$, and repeat step (2).
After a finite number of steps, $I_Z^0$ becomes empty and we obtain
a discrete solution $\overline{X}$ which is typically closer to
$X$ than the starting approximate solution.

\medskip
We also need to decide whether  to start with $Z$ or $-Z$
(remember that solution are determined up to a nonzero scalar).
We proceed  as follows.
Let $Z^+$ and $Z^-$ be the vectors given
by
\begin{align*}
Z^+_i & = 
\begin{cases}
Z_i & \text{if $i \in I_Z^+$} \\
0 & \text{if $i \notin I_Z^+$} 
\end{cases}
&
Z^-_i & = 
\begin{cases}
Z_i & \text{if $i \in I_Z^-$} \\
0 & \text{if $i \notin I_Z^-$} 
\end{cases}.
\end{align*}
Also let 
$n_a  = |I_Z^{+} |$, $n_b  = |I_Z^{-} |$, let $\overline{a}$ and $\overline{b}$ be the
average of the positive and negative entries in $Z$ respectively, that is,
\begin{align*}
\overline{a} &  = \frac{\sum_{i\in I_Z^+} Z_i}{n_a}  & 
\overline{b} & = \frac{\sum_{i\in I_Z^-} Z_i}{n_b}, 
\end{align*}
and let $\overline{Z^+}$ and $\overline{Z^-}$ be the vectors given by
\begin{align*}
(\overline{Z^+})_i & = 
\begin{cases}
\overline{a} & \text{if $i \in I_Z^+$} \\
0 & \text{if $i \notin I_Z^+$} 
\end{cases}
&
(\overline{Z^-})_i & = 
\begin{cases}
\overline{b} & \text{if $i \in I_Z^-$} \\
0 & \text{if $i \notin I_Z^-$} 
\end{cases}.
\end{align*}
If $\norme{\overline{Z^+} - Z^+} > \norme{\overline{Z^-} - Z^-}$, then 
replace $Z$ by $-Z$.

\medskip
Step 1 of the algorithm is to compute an initial
approximate discrete solution $\overline{X}$.
\begin{enumerate}
\item[(1)]
Let 
\begin{align*}
n_a & = |I_Z^{+} |, &
\alpha & = \mathrm{vol}(\{v_i \mid i \in I_Z^{+}\}), &
\beta & = \frac{\alpha}{d - \alpha} ,
\end{align*}
and form the vector $\overline{X}$ with
\[
\overline{X}_i = 
\begin{cases}
a & \text{if $i \in I_Z^{+}$} \\
-\beta a & \text{otherwise}, 
\end{cases}
\]
such that $\norme{\overline{X}} =  \norme{Z}$,
where the  scalar $a$ is
determined by
\[
a =\frac{\norme{Z}}{\sqrt{(n_a + \beta^2(N - n_a))}}.
\]

Next, pick some entry with index $i\in I_Z^0$ and see whether
we can impove the solution $\overline{X}$ by adding $i$ to $I_Z^+$.
\item[(2)]
While  $I_Z^0 \not= \emptyset$,  
pick the smallest index $i\in I_Z^0$, 
 compute
\begin{align*}
\widetilde{I}_{Z}^{+} & =  I_{Z}^{+} \cup \{i\} \\  
\widetilde{n}_a  & = n_a + 1 \\
\widetilde{\alpha} & =  \alpha + d(v_i) \\
\widetilde{\beta} & = \frac{\widetilde{\alpha}}{d - \widetilde{\alpha}},
\end{align*}
and then $\widetilde{X}$ with
\[
\widetilde{X}_j = 
\begin{cases}
\widetilde{a} & \text{if $j \in \widetilde{I}_Z^{+}$} \\
-\widetilde{\beta} \widetilde{a}  & \text{otherwise} ,
\end{cases}
\]
and
\[
\widetilde{a} = \frac{\norme{Z}}
{\sqrt{\widetilde{n}_a + \widetilde{\beta}^2(N - \widetilde{n}_a)}} .
\]
Set  $I_Z^0 = I_Z^0 - \{i\}$.
If $\smnorme{\widetilde{X} - Z} < \smnorme{\overline{X} - Z}$, 
then
let $\overline{X} = \widetilde{X}$, $I_Z^+ = \widetilde{I}_Z^+$,
$n_a = \widetilde{n}_a$,  $\alpha = \widetilde{\alpha}$.
Go back to (2).
\item[(3)]
The final answer if $\overline{X}$.
\end{enumerate}

I implemented this algorithm, and it seems to  do a god job 
dealing with zero entries in the continuous
solution $Z$.

\section{$K$-Way Clustering Using Normalized  Cuts}
\label{ch3-sec3}
We now consider the general case in which $K \geq 3$.
Two crucial  issues need  to be addressed
(to the best of our knowledge, these points are not clearly articulated 
in the literature).
\begin{enumerate}
\item
The choice of a matrix representation for partitions on the set of vertices.
It is important that such a representation be scale-invariant.
It is also necessary to state necessary and sufficient conditions
for such matrices to represent a partition.
\item
The choice of a metric to compare solutions.
It turns out that the space of discrete solutions can be
viewed as a subset of the $K$-fold product 
$(\mathbb{RP}^{N - 1})^K$ of the projective space $\mathbb{RP}^{N - 1}$.
Version 1 of the formulation of our minimization problem (PNC1)
makes this point clear. However, the relaxation $(*_2)$ of 
version 2 of  our minimization problem (PNC2), which is equivalent to
version 1, reveals that that the solutions of the relaxed problem
$(*_2)$ are members of the {\it Grassmannian\/} $G(K, N)$.
Thus, we have two choices of metrics: (1) a metric on
$(\mathbb{RP}^{N - 1})^K$; (2) a metric on $G(K, N)$.
We  discuss the first choice, which is the choice implicitly
adopted by Shi and Yu. Actually, it appears that 
it is difficult to deal with the
product metric on $(\mathbb{RP}^{N - 1})^K$ induced by a
metric on $\mathbb{RP}^{N - 1}$.
Instead, we approximate a metric on $(\mathbb{RP}^{N - 1})^K$ using
the Frobenius norm; see  Section 
\ref{ch3-sec5} for details.
\end{enumerate}

\medskip
We describe a  partition $(A_1, \ldots, A_K)$ of the set of nodes $V$ by
an $N\times K$ matrix  $X = [X^1 \cdots X^K]$
whose columns $X^1, \ldots,
X^K$ are indicator vectors of the partition $(A_1, \ldots, A_K)$.
Inspired by what we did in Section \ref{ch3-sec2}, 
we assume that the  vector $X^j$ is of the form
\[
X^j = (x_1^j, \ldots, x_N^j),
\]
where $x_i^j \in \{a_j, b_j\}$ for $j = 1, \ldots, K$ and
$i = 1, \ldots, N$, and
where $a_j, b_j$ are 
any  two distinct  real numbers.
The vector $X^j$  is an indicator vector for $A_j$ in
the sense that, for $i = 1, \ldots, N$,
\[
x_i^j =
\begin{cases}
a_j & \text{if $v_i \in A_j$} \\
b_j & \text{if $v_i \notin A_j$} .
\end{cases}
\]

When $\{a_j, b_j\} = \{0, 1\}$ for $j = 1, \ldots, K$,  such a matrix is called
a {\it partition matrix\/} by Yu and Shi. However, such a choice is
premature, since it is better to have a scale-invariant representation
to make the denominators of the Rayleigh ratios go away.

\medskip
Since the partition $(A_1, \ldots, A_K)$ consists of nonempty pairwise
disjoint blocks whose union is $V$,
some  conditions on $X$ are required to  reflect
these properties, but  we will worry about this later. 

\medskip
As in Section \ref{ch3-sec2}, we seek conditions on 
the $a_j$s and the $b_j$s in order to express the normalized cut 
$\mathrm{Ncut}(A_1, \ldots, A_K)$ as a sum of 
Rayleigh ratios. Then, we reformulate our
optimization problem 
in a more convenient form, by chasing the denominators in the Rayleigh ratios,
and by expressing the objective function in
terms of the {\it trace\/} of a certain matrix.
This will reveal the important fact that the solutions of the relaxed
problem are right-invariant under multiplication by a $K\times K$
orthogonal matrix.

\medskip
Let  $d = \transpos{\mathbf{1}} \Degsym \mathbf{1}$ and $\alpha_j =
\mathrm{vol}(A_j)$, so that $\alpha_1 + \cdots + \alpha_K = d$.
Then, $\mathrm{vol}(\overline{A_j}) = d - \alpha_j$,
and as in Section \ref{ch3-sec2},  we have
\begin{align*}
\transpos{(X^j)} L X^j & = 
(a_j - b_j)^2\, \mathrm{cut}(A_j, \overline{A_j}), \\
\transpos{(X^j)} \Degsym  X^j & = \alpha_j a_j^2 + (d - \alpha_j)b_j^2.
\end{align*}
When $K \geq 3$,  unlike the case $K = 2$,   in general 
we have $\mathrm{cut}(A_j, \overline{A_j})  \not=  \mathrm{cut}(A_k,
\overline{A_k}) $
if $j \not= k$, and  since 
\[
 \mathrm{Ncut}(A_1, \ldots, A_K) 
= \sum_{j = 1}^K 
\frac{\mathrm{cut}(A_j, \overline{A_j})}{\mathrm{vol}(A_j)},
\]
we would like to choose  $a_j , b_j$ so that 
\[
\frac{\mathrm{cut}(A_j, \overline{A_j})}{\mathrm{vol}(A_j)} 
= \frac{\transpos{(X^j)} L X^j}{\transpos{(X^j)}\Degsym X^j} \quad j =
1, \ldots, K, 
\]
because this implies that
\[
\mu(X) = \mathrm{Ncut}(A_1, \ldots, A_K) 
= \sum_{j = 1}^K 
\frac{\mathrm{cut}(A_j, \overline{A_j})}{\mathrm{vol}(A_j)}
= \sum_{j = 1}^K 
\frac{\transpos{(X^j)} L X^j}{\transpos{(X^j)}\Degsym X^j}.
\]

Since
\[
\frac{\transpos{(X^j)} L X^j}{\transpos{(X^j)}\Degsym X^j} = 
\frac{(a_j - b_j)^2\, \mathrm{cut}(A_j, \overline{A_j})}
{\alpha_j a_j^2 + (d - \alpha_j)b_j^2}
\]
and $\mathrm{vol}(A_j) = \alpha_j$, in order to have
\[
\frac{\mathrm{cut}(A_j, \overline{A_j})}{\mathrm{vol}(A_j)} 
= \frac{\transpos{(X^j)} L X^j}{\transpos{(X^j)}\Degsym X^j} \quad j =
1, \ldots, K, 
\]
we need to have
\[
\frac{(a_j - b_j)^2} 
{\alpha_j a_j^2 + (d - \alpha_j)b_j^2} = \frac{1}{\alpha_j}
\quad j = 1, \ldots, K.
\]
Thus, we must have
\[
(a_j^2 - 2a_jb_j + b_j^2)\alpha_j = \alpha_j a_j^2 + (d - \alpha_j)b_j^2,
\]
which yields
\[
2 \alpha_jb_j(b_j - a_j) = db_j^2.
\]
The above equation is trivially satisfied if $b_j = 0$. If
$b_j\not= 0$, then
\[
2 \alpha_j(b_j - a_j) = db_j,
\]
which yields
\[
a_j = \frac{2\alpha_j - d}{2\alpha_j} b_j.
\]
This choice seems more complicated that the choice $b_j = 0$, so
we will opt for the choice $b_j = 0$, $j = 1, \ldots, K$.
With this choice, we get
\[
\transpos{(X^j)} \Degsym  X^j  = \alpha_j a_j^2.
\]
Thus, it makes sense to pick
\[
a_j = \frac{1}{\sqrt{\alpha_j}} =
\frac{1}{\sqrt{\mathrm{vol}(A_j)}}, \quad j = 1, \ldots, K,
\]
which is the solution presented in von Luxburg \cite{Luxburg}.
This choice also corresponds to the scaled partition matrix used in
Yu \cite{Yu} and Yu and Shi \cite{YuShi2003}.

\medskip
When $N = 10$ and $K = 4$, 
an example  of a matrix $X$ 
representing the partition
of $V = \{v_1, v_2, \ldots, v_{10}\}$
into the four   blocks 
\[
\{A_1, A_2, A_3, A_4\} = 
\{\{v_2, v_4, v_6\}, \{v_1, v_5\}, \{v_3, v_8, v_{10}\}, \{v_7, v_9\}\} , 
\]
is shown below:
\[
X = 
\begin{pmatrix}
0  & a_2  & 0  & 0 \\
a_1 & 0  & 0  & 0  \\
0  & 0  & a_3  & 0  \\
a_1 & 0  & 0  & 0  \\
0   & a_2  & 0  & 0  \\
a_1 & 0  & 0  & 0  \\
0  & 0  & 0  & a_4  \\
0  & 0  & a_3  & 0  \\
0  & 0  & 0  & a_4  \\
0   & 0  & a_3   & 0  
\end{pmatrix}.
\]

\medskip
Let us now consider the problem of finding necessary and sufficient
conditions for a matrix $X$ to represent a partition of $V$.

\medskip
When $b_j = 0$, the pairwise disjointness of the $A_i$ is captured
by the orthogonality of the $X^i$:
\begin{equation}
\transpos{(X^i)}X^j = 0, \quad 1 \leq i, j \leq  K, \> i \not= j.
\tag{$*$}
\end{equation}
This is because,  for any matrix $X$ where the nonzero entries in each
column have the same sign,
for any $i\not= j$, the condition
\[
\transpos{(X^i)} X^j = 0
\]
says that for every $k = 1,\ldots, N$,
if $x^i_k \not= 0$ then  $x^j_k = 0$.

\medskip
When we formulate our minimization problem in terms of Rayleigh
ratios, conditions on the quantities $\transpos{(X^i)} \Degsym X^i$ show up, and
it is more convenient to express the orthogonality conditions
using the quantities $\transpos{(X^i)} \Degsym X^j$ instead of  
the $\transpos{(X^i)}  X^j$, because
these various conditions can  be combined into a single condition 
involving the matrix $\transpos{X} \Degsym X$.
Now, because $\Degsym$ is a diagonal
matrix with positive entries and because  the nonzero entries in each
column of $X$ have the same sign,
for any $i\not= j$, the condition
\[
\transpos{(X^i)} X^j = 0
\]
is equivalent to
\begin{equation}
\transpos{(X^i)} \Degsym X^j = 0,
\tag{$**$}
\end{equation}
since, as above, it means that for    $k = 1, \ldots, N$,
if $x^i_k \not= 0$ then  $x^j_k = 0$.
Observe that the orthogonality conditions $(*)$ (and $(**)$)
are equivalent to the fact
that every row of $X$ has at most one nonzero entry.

\medskip
\remark
The disjointness condition
\[
X \mathbf{1}_K= \mathbf{1}_N
\]
is used in Yu \cite{Yu}.
However, this condition does guarantee the disjointness of
the blocks. For example, it is satisfied by the matrix $X$ whose
first column is $\mathbf{1}_N$,  with $0$ everywhere else.

\medskip
Each $A_j$ is nonempty iff $X^j \not= 0$, and the fact that the union of
the $A_j$ is $V$
is captured by the fact that each row of $X$ must have some
nonzero entry
(every vertex appears in some block).
It is not immediately obvious how to state conveniently this condition in  matrix form.

\medskip
Observe that the diagonal entries of the matrix $X\transpos{X}$ are the
square Euclidean norms of the rows of $X$. Therefore, we can assert
that  these entries are all nonzero.   Let $\mathrm{DIAG}$ be the function which returns the
diagonal matrix (containing the diagonal of $A$),
\[
\mathrm{DIAG}(A) = \mathrm{diag}(a_{1\, 1}, \ldots, a_{n\, n}),
\]
for any square matrix $A = (a_{i \, j})$.
Then, the condition for the rows of $X$ to be nonzero can be stated as
\begin{equation*}
\det(\mathrm{DIAG}(X\transpos{X})) \not= 0. 
\end{equation*}

Since every row of any matrix $X$ representing a partition
has a single nonzero entry $a_j$, we have
\[
\transpos{X}X =
\mathrm{diag}\left(n_1a_1^2,  \ldots, 
n_K a_K^2\right),
\]
where $n_j$ is the number of elements in $A_j$, the $j$th block of the
partition. Therefore, an equivalent condition for the columns of $X$ to be nonzero is
\[
\det(\transpos{X}X) \not= 0.
\]

\remark
The matrix
\[
\mathrm{DIAG}(X\transpos{X})^{-1/2} X 
\]
is the result of normalizing the rows of $X$ so that they have
Euclidean norm $1$. This normalization step  is used by   Yu \cite{Yu}
in the search for  a discrete solution closest to a solution of 
a relaxation of our original problem.
For our special matrices representing  partitions, 
normalizing the rows will have the effect of rescaling the columns
(if row $i$ has $a_j$ in column $j$, then all nonzero entries
in column $j$ are equal to $a_j$), but for a more general matrix,
this is false.  Thus, in general,
$\mathrm{DIAG}(X\transpos{X})^{-1/2} X$
is not a solution of the original problem.
Still, as we will see in Section \ref{ch3-sec5},
this matrix is a pretty good approximation to a discrete solution.

\medskip
Another condition which does not involve explicitly a determinant and 
is scale-invariant  stems from the observation 
that not only
\[
\transpos{X}X =
\mathrm{diag}\left(n_1a_1^2,  \ldots, 
n_K a_K^2\right),
\]

but
\[
\transpos{X}\mathbf{1}_N =
\begin{pmatrix}
n_1a_1 \\
\vdots \\
n_K a_K
\end{pmatrix},
\]
and these equations imply that
\[
(\transpos{X} X)^{-1} \transpos{X} \mathbf{1}_N = 
\begin{pmatrix}
\frac{1}{a_1} \\
\vdots \\
\frac{1}{a_K}
\end{pmatrix},
\]  
and thus
\begin{equation}
X (\transpos{X} X)^{-1} \transpos{X} \mathbf{1}_N = \mathbf{1}_N.
\tag{$\dagger$}
\end{equation}
When $a_j = 1$ for $j = 1, \ldots, K$, we have
$(\transpos{X} X)^{-1} \transpos{X} \mathbf{1} = \mathbf{1}_K$, and 
condition $(\dagger)$ reduces to
\[
X\mathbf{1}_K = \mathbf{1}_N.
\]
Note that because the columns of $X$ are linearly independent,
$(\transpos{X} X)^{-1} \transpos{X}$ is the pseudo-inverse $X^+$ of
$X$. Consequently, if $\transpos{X} X$ is invertible,
condition $(\dagger)$ can also be written as
\[
XX^+ \mathbf{1}_N = \mathbf{1}_N.
\]
However, it is well known that $XX^+$ is the orthogonal
projection of $\reals^K$ onto the range of $X$
(see Gallier \cite{Gallbook2}, Section 14.1), 
so the condition
$XX^+ \mathbf{1}_N = \mathbf{1}_N$ is equivalent to the fact
that $\mathbf{1}_N$ belongs to the
range of $X$. In retrospect, this should have been obvious since
the columns of a solution $X$ satisfy the equation
\[
a_1^{-1} X^1 + \cdots + a_K^{-1} X^K  = \mathbf{1}_N.
\]

\medskip
We emphasize that it is important to use conditions that are invariant
under multiplication by a nonzero scalar, 
because the Rayleigh ratio is scale-invariant, and it is crucial to take
advantage of this fact to make the denominators go away.

\medskip
If we let
\[
\s{X}  = \Big\{[X^1\> \ldots \> X^K] \mid
X^j = a_j(x_1^j, \ldots, x_N^j) , \>
x_i^j \in \{1, 0\},
 a_j\in \reals, \> X^j \not= 0
\Big\}
\]
(note that the condition $X^j \not= 0$ implies that $a_j \not= 0$),
then the set of matrices representing partitions of $V$ into $K$
blocks is
\begin{align*}
& & &\s{K}  = \Big\{ X = [X^1 \> \cdots \> X^K] \quad \mid & &  X\in\s{X},  &&\\
         & & &  & & \transpos{(X^i)} \Degsym X^j = 0, \quad 1\leq i, j \leq K,\> 
i\not= j, && \quad\quad\quad\quad\quad\\
&  & & & & 
 X (\transpos{X} X)^{-1} \transpos{X} \mathbf{1} = \mathbf{1}\Big\}. && 
\end{align*}

Since for matrices in $\s{K}$, the orthogonality conditions
$\transpos{(X^i)} \Degsym X^j = 0$
are equivalent to the orthogonality conditions $\transpos{(X^i)} X^j =
0$, and since matrices in $\s{X}$ have nonzero columns,
$\transpos{X}X$ is  invertible, so the last condition makes sense.

\medskip
As in the case $K = 2$, to be rigorous, the solution are really
$K$-tuples of points in $\mathbb{RP}^{N-1}$, so our solution set is
really
\[
\mathbb{P}(\s{K})  = \Big\{(\mathbb{P}(X^1), \ldots, \mathbb{P}(X^K)) \mid
[X^1 \> \cdots \> X^K] \in \s{K} 
\Big\}.
\]

\medskip
In view of the above, we have our first formulation of $K$-way clustering
of a graph using normalized cuts, called problem PNC1 
(the notation PNCX  is used in  Yu \cite{Yu}, Section 2.1):

\medskip\noindent
{\bf $K$-way Clustering of a graph using Normalized Cut, Version 1: \\
Problem PNC1}

\begin{align*}
& \mathrm{minimize}     &  &  \sum_{j = 1}^K 
\frac{\transpos{(X^j)} L X^j}{\transpos{(X^j)}\Degsym X^j}& &  &  &\\
& \mathrm{subject\ to} &  & 
 \transpos{(X^i)} \Degsym X^j = 0, \quad 1\leq i, j \leq K,\> 
i\not= j,  & &  & & \\
& & & 
 X (\transpos{X} X)^{-1} \transpos{X} \mathbf{1} = \mathbf{1},  & & X\in \s{X}. & & 
\end{align*}

As in the case $K = 2$, the solutions that we are seeking are $K$-tuples 
$(\mathbb{P}(X^1), \ldots, \mathbb{P}(X^K))$ of points  in
$\mathbb{RP}^{N-1}$ determined by
their  homogeneous coordinates $X^1, \ldots, X^K$.

\medskip
\remark
Because
\[
\transpos{(X^j)} L X^j = 
\transpos{(X^j)} \Degsym X^j - \transpos{(X^j)} W X^j = 
\mathrm{vol}(A_j)   - \transpos{(X^j)} W X^j,
\]
Instead of minimizing
\[
\mu(X^1, \ldots, X^K) 
= \sum_{j = 1}^K 
\frac{\transpos{(X^j)} L X^j}{\transpos{(X^j)}\Degsym X^j},
\]
we can maximize
\[
\epsilon(X^1, \ldots, X^K)=  \sum_{j = 1}^K 
\frac{\transpos{(X^j)} W X^j}{\transpos{(X^j)}\Degsym X^j},
\]
since
\[
\epsilon(X^1, \ldots, X^K) = K - \mu(X^1, \ldots, X^K).
\]
This second option is the one chosen by
Yu \cite{Yu} and Yu and Shi \cite{YuShi2003} (actually, they work
with $\frac{1}{K}(K - \mu(X^1, \ldots, X^K))$, but this doesn't make any difference).
Theoretically, minimizing $\mu(X^1, \ldots, X^K)$ is equivalent to
maximizing $\epsilon(X^1, \ldots, X^K)$, but from a practical point of
view, it is preferable to maximize $\epsilon(X^1, \ldots, X^K)$. This
is because minimizing solutions of $\mu$ 
are obtained from (unit) eigenvectors corresponding to the $K$
{\it smallest\/} eigenvalues of $L_{\mathrm{sym}} =  \Degsym^{-1/2} L\Degsym^{-1/2}$ 
(by multiplying these eigenvectors by $\Degsym^{1/2}$). 
However, numerical methods for computing eigenvalues and eigenvectors
of a symmetric matrix  do much better at computing  largest
eigenvalues. Since   $L_{\mathrm{sym}}=  I - \Degsym^{-1/2}
W\Degsym^{-1/2}$, the  eigenvalues of  $L_{\mathrm{sym}}$ listed in increasing order
correspond to the eigenvalues of $I - L_{\mathrm{sym}} =
\Degsym^{-1/2} W\Degsym^{-1/2}$ listed in decreasing order.
Furthermore, $v$ is an eigenvector of $L_{\mathrm{sym}}$ for the $i$th
smallest  eigenvalue $\nu_i$ iff   $v$ is an eigenvector of $I -
L_{\mathrm{sym}}$ for the $(N + 1 - i)$th largest eigenvalue $\nu_i$.
Therefore, it is preferable to find the {\it largest\/} eigenvalues of
$I - L_{\mathrm{sym}} =  \Degsym^{-1/2} W\Degsym^{-1/2}$ and their eigenvectors.
In fact, since the eigenvalues of $L_{\mathrm{sym}}$ are in the range $[0,
2]$, the eigenvalues of $2I - L_{\mathrm{sym}} =  I + \Degsym^{-1/2}
W\Degsym^{-1/2}$ are also in the range $[0, 2]$ (that is, $I +
\Degsym^{-1/2} W  \Degsym^{-1/2}$ is positive semidefinite).

\medskip
Let us now show how our original formulation (PNC1) can be converted
to a more convenient form, by chasing the denominators in the Rayleigh ratios,
and by expressing the objective function in
terms of the {\it trace\/} of a certain matrix.

\medskip
For any $N\times N$ matrix $A$, 
because
\begin{align*}
\transpos{X} A X & = 
\begin{bmatrix}
 \transpos{(X^1)}  \\
 \vdots \\
\transpos{(X^K)}  
\end{bmatrix}
A [X^1 \cdots X^K]   \\
& = 
\begin{pmatrix}
\transpos{(X^1)} A X^1 & \transpos{(X^1)} A X^2 &  \cdots &
\transpos{(X^1)} A X^K \\
\transpos{(X^2)} A X^1 & \transpos{(X^2)} A X^2 &  \cdots &
\transpos{(X^2)} A X^K \\
\vdots & \vdots & \ddots & \vdots \\
\transpos{(X^K)} A X^1 & \transpos{(X^K)} A X^2 &  \cdots &
\transpos{(X^K)} A X^K 
\end{pmatrix},
\end{align*}
we have
\[
\mathrm{tr}(\transpos{X} A X) = \sum_{j = 1}^K \transpos{(X^j)} A X^j ,
\]
and the conditions
\[
 \transpos{(X^i)} A X^j = 0, \quad 1\leq i, j \leq K,\>  i\not= j, 
\]
are equivalent to
\[
\transpos{X} A X= \mathrm{diag}(\transpos{(X^1)} A X^1, \ldots, \transpos{(X^K)} A X^K).
\]
As a consequence, if we assume that
\[
\transpos{(X^1)} A X^1 = \cdots = \transpos{(X^K)} A X^K = \alpha^2,
\]
then we have
\[
\transpos{X} A X = \alpha^2 I,
\]
and if $R$ is any orthogonal $K\times K$ matrix, then
by multiplying on the left by $\transpos{R}$ and on the right by $R$,
we get
\[
\transpos{R}\transpos{X} A XR = \transpos{R} \alpha^2 I R = \alpha^2
\transpos{R} R = \alpha^2 I.
\]
Therefore,  if 
\[
\transpos{X} A X = \alpha^2 I,
\]
then 
\[
\transpos{(XR)} A (XR) = \alpha^2 I,
\]
for any orthogonal $K\times K$ matrix $R$.
Furthermore, because $\mathrm{tr}(AB)  =
\mathrm{tr}(BA)$ for all matrices $A, B$, we have 
\[
 \mathrm{tr}(\transpos{R}\transpos{X} A X R)  =  \mathrm{tr}(\transpos{X} A X).
\]
Since the Rayleigh ratios
\[
\frac{\transpos{(X^j)} L X^j}{\transpos{(X^j)}\Degsym X^j}
\]
are invariant under rescaling by a nonzero number, by replacing
$X^j$ by $(\transpos{(X^j)} \Degsym X^j)^{-1/2} X^j$, the denominators
become $1$, and we have
\begin{align*}
\mu(X) & = 
\mu(X^1, \ldots, X^K) 
 = \sum_{j = 1}^K 
\frac{\transpos{(X^j)} L X^j}{\transpos{(X^j)}\Degsym X^j} \\
& = \mu((\transpos{(X^1)} \Degsym X^1)^{-1/2} X^1, \ldots,
(\transpos{(X^K)} \Degsym X^K)^{-1/2} X^K) \\
& = \sum_{j = 1}^K (\transpos{(X^j)} \Degsym X^j)^{-1/2}
\transpos{(X^j)} L \, (\transpos{(X^j)} \Degsym X^j)^{-1/2} X^j \\
& = \mathrm{tr}(\Lambda^{-1/2}\transpos{X} L X \Lambda^{-1/2}) \\
& =  \mathrm{tr}(\Lambda^{-1}\transpos{X} L X ),
\end{align*}
where 
\[
\Lambda = \mathrm{diag}(
\transpos{(X^1)} \Degsym X^1, \ldots, \transpos{(X^K)} \Degsym X^K).
\]
If $\transpos{(X^1)} \Degsym X^1 =  \cdots =  
\transpos{(X^K)} \Degsym X^K = \alpha^2$, 
then $\Lambda = \alpha^2 I_K$,  so 
\[
\mu(X) 
=  \mathrm{tr}(\Lambda^{-1}\transpos{X} L X )
= \frac{1}{\alpha^2} \mathrm{tr}(\transpos{X} L X ),
\]
and for any orthogonal $K\times K$ matrix $R$,
\[
\mu(RX) =  \frac{1}{\alpha^2} \mathrm{tr}(\transpos{R}\transpos{X} L XR )
= \frac{1}{\alpha^2}\mathrm{tr}(\transpos{X} L X),
\]
and thus, 
\[
\mu(X) = \mu(XR).
\]

\medskip
The condition
\[
 X (\transpos{X} X)^{-1} \transpos{X} \mathbf{1} = \mathbf{1}
\]
is also invariant if  we replace $X$ by $XR$, where $R$ is any
invertible matrix, because
\begin{align*}
 XR (\transpos{(XR)} (XR))^{-1} \transpos{(XR)} \mathbf{1} & = 
 XR (\transpos{R}\transpos{X} X R)^{-1} \transpos{R}\transpos{X}
 \mathbf{1}\\
& =  XR R^{-1}(\transpos{X} X)^{-1}(\transpos{R})^{-1}  \transpos{R}\transpos{X}
 \mathbf{1} \\
& = X (\transpos{X} X)^{-1} \transpos{X} \mathbf{1} = \mathbf{1}.
\end{align*}
In summary we proved the following proposition:
\begin{proposition}
\label{Kway1}
For any orthogonal $K\times K$ matrix $R$, any symmetric $N\times N$
matrix $A$, and any $N\times K$ matrix $X = [X^1 \> \cdots \> X^K]$, the following properties hold:
\begin{enumerate}
\item[(1)]
$\mu(X) =  \mathrm{tr}(\Lambda^{-1}\transpos{X} L X)$,
where 
\[
\Lambda = \mathrm{diag}(
\transpos{(X^1)} \Degsym X^1, \ldots, \transpos{(X^K)} \Degsym X^K).
\]
\item[(2)]
If
$\transpos{(X^1)} \Degsym X^1 =  \cdots =  
\transpos{(X^K)} \Degsym X^K = \alpha^2$, then
\[
\mu(X) = \mu(XR) =  \frac{1}{\alpha^2}\mathrm{tr}(\transpos{X} L X).
\] 
\item[(3)]
The condition $\transpos{X} A X = \alpha^2 I$ is  
preserved if $X$ is replaced by $XR$.
\item[(4)]
 The condition $X (\transpos{X} X)^{-1} \transpos{X} \mathbf{1} =
 \mathbf{1}$ is preserved if $X$ is replaced by $XR$.
\end{enumerate}
\end{proposition}

\medskip
Now, by Proposition \ref{Kway1}(1)  and the
fact that the conditions in PNC1  are scale-invariant, 
we are led to the following  formulation of our problem:

\begin{align*}
& \mathrm{minimize}     &  &   
\mathrm{tr}(\transpos{X} L X) & &  &  &\\
& \mathrm{subject\ to} &  & 
\transpos{(X^i)} \Degsym X^j = 0, \quad 1\leq i, j \leq K,\> 
i\not= j,  & &  & & \\
& & & \transpos{(X^j)}\Degsym X^j = 1, \quad 1\leq j \leq K, & & & &\\
& & & 
 X (\transpos{X} X)^{-1} \transpos{X} \mathbf{1} = \mathbf{1},  & & X\in \s{X}. & & 
\end{align*}

Conditions on lines 2 and 3 can be combined in the equation
\[
\transpos{X} \Degsym X = I,
\]
and, we obtain the following  formulation of our  minimization
problem:

\medskip\noindent
{\bf $K$-way Clustering of a graph using Normalized Cut, Version 2: \\
Problem PNC2}

\begin{align*}
& \mathrm{minimize}     &  &  
\mathrm{tr}(\transpos{X} L X)& &  &  &\\
& \mathrm{subject\ to} &  & 
\transpos{X} \Degsym X = I, 
 & &  & & \\
& & & 
 X (\transpos{X} X)^{-1} \transpos{X} \mathbf{1} = \mathbf{1},  & & X\in \s{X}. & & 
\end{align*}

\medskip
Because  problem PNC2 requires the constraint 
$\transpos{X} \Degsym X =I$ to be satisfied,
it does not have the same set of solutions as  problem PNC1.
Nevertherless, 
problem PNC2 is equivalent to problem PNC1, in the sense that for every
minimal solution $(X^1, \ldots, X^K)$ of PNC1, 
$((\transpos{(X^1)} D X^1)^{-1/2} X^1, \ldots,
(\transpos{(X^K)} D X^K)^{-1/2} X^K)$
is a minimal solution of PNC2 (with the same minimum for the objective functions),
and that for every minimal solution $(Z^1, \ldots, Z^k)$ of PNC2,  
$(\lambda_1 Z^1, \ldots, \lambda_K Z^K)$ 
is a minimal solution of PNC1, for all  $\lambda_i \not= 0$,  $i = 1,
\ldots, K$ (with the same minimum for the objective functions).
In other words, problems PNC1   and PNC2 have the same
set of minimal solutions as $K$-tuples of points 
$(\mathbb{P}(X^1), \ldots, \mathbb{P}(X^K))$ in
$\mathbb{RP}^{N-1}$ determined by
their  homogeneous coordinates $X^1, \ldots, X^K$.

\medskip
Formulation PNC2 reveals that finding a minimum normalized cut 
has a geometric interpretation in terms of the graph drawings 
discussed in Section  \ref{ch2-sec1}. Indeed, PNC2 has the following
equivalent formulation: Find a minimal energy graph drawing $X$ in $\reals^K$ of the
weighted graph $G = (V, W)$ such that:
\begin{enumerate}
\item
The matrix $X$ is orthogonal with respect to the inner product
$\lag -, - \rag_{\Degsym}$ in $\reals^N$ induced by $\Degsym$, with
\[
\lag x, y \rag_{\Degsym} = \transpos{x} \Degsym y, \quad x, y \in \reals^N. 
\]
\item
The rows of $X$ are nonzero; this means that no vertex $v_i\in V$ is
assigned to the origin of $\reals^K$ (the zero vector $0_K$).
\item
Every vertex $v_i$ is assigned a point of the form 
$(0, \ldots, 0,a_j,0,\ldots,0)$  on some axis (in $\reals^K$).
\item
Every axis in  $\reals^K$ is assigned at least some vertex.
\end{enumerate}

\medskip
Condition 1 can be reduced to the standard condition for graph
drawings ($\transpos{R}R = I$)  by making the change of variable $Y = \Degsym^{1/2} X$ or equivalently
$X = \Degsym^{-1/2} Y$. Indeed, 
\[
\mathrm{tr}(\transpos{X} LX) = 
\mathrm{tr}
(\transpos{Y}\Degsym^{-1/2} L \Degsym^{-1/2} Y),
\]
so we use the normalized Laplacian $L_{\mathrm{sym}} = \Degsym^{-1/2} L \Degsym^{-1/2}$
instead of $L$, 
\[
\transpos{X} \Degsym X   = \transpos{Y} Y   = I,
\]
and conditions (2), (3), (4) are preserved under the change of
variable  $Y = \Degsym^{1/2} X$, since  $\Degsym^{1/2} $ is invertible.
However, conditions (2), (3), (4) are ``hard'' constraints, especially
condition (3). 
In fact, condition (3) implies that the columns of $X$ are  orthogonal
with respect to  both the Euclidean inner product and the inner
product $\lag -, - \rag_{\Degsym}$, so condition (1) is redundant,
except for the fact  that it prescribes the norm of the columns, but 
this is not essential due to the projective nature of the solutions.

\medskip
The main problem in finding a good relaxation of
problem PNC2 is that it is very difficult to enforce the condition
$X\in \s{X}$.
Also, the solutions $X$ are not preserved under arbitrary
rotations,
but only by very special rotations which leave $\s{X}$ invariant
(they exchange the axes).

\medskip
The first natural relaxation of problem PNC2 is to drop the condition
that $X\in \s{X}$, and we obtain the 

\medskip\noindent
{\bf Problem $(*_2)$}

\begin{align*}
& \mathrm{minimize}     &  &  
\mathrm{tr}(\transpos{X} L X)& &  &  &\\
& \mathrm{subject\ to} &  & 
\transpos{X} \Degsym X = I, 
 & &  & & \\
& & & 
X (\transpos{X} X)^{-1} \transpos{X} \mathbf{1} = \mathbf{1}.  & &  & & 
\end{align*}

Actually, since the discrete solutions $X\in \s{X}$ that we are ultimately seeking
are solutions of problem PNC1, the preferred relaxation 
is the one obtained from
problem PNC1 by dropping the condition $X\in \s{X}$,
and simply requiring that $X^j \not= 0$, for $j = 1, \ldots, K$:

\medskip\noindent
{\bf Problem $(*_1)$}

\begin{align*}
& \mathrm{minimize}     &  &  \sum_{j = 1}^K 
\frac{\transpos{(X^j)} L X^j}{\transpos{(X^j)}\Degsym X^j}& &  &  &\\
& \mathrm{subject\ to} &  & 
 \transpos{(X^i)} \Degsym X^j = 0, \quad X^j\not= 0& & 1\leq i, j \leq K,\> 
i\not= j,    & & \\
& & & 
 X (\transpos{X} X)^{-1} \transpos{X} \mathbf{1} = \mathbf{1}.  & &
 & & 
\end{align*}

\medskip
Now that we dropped the condition $X\in \s{X}$, it is not
clear that
$\transpos{X}X$ is invertible in $(*_1)$ and $(*_2)$.
However, since the columns of $X$ are nonzero and
$\Degsym$-orthogonal, they must be linearly independent,
so $X$ has rank $K$ and 
and $\transpos{X}X$ is invertible.

\medskip
As we explained before, every solution
$Z = [Z^1, \ldots, Z^K]$ of problem $(*_1)$ yields a solution 
of problem $(*_2)$ by normalizing each $Z^j$ by 
$(\transpos{(Z^j)}\Degsym Z^j)^{1/2}$, and conversely for every solution
$Z = [Z^1, \ldots, Z^K]$ of problem $(*_2)$, 
the $K$-tuple $[\lambda_1 Z^1, \ldots, \lambda_K Z^K]$
is a solution of problem $(*_1)$, where $\lambda_j\not= 0$ for
$j = 1, \ldots, K$. Furthermore,
by Proposition \ref{Kway1}, for every
orthogonal matrix $R\in \mathbf{O}(K)$  and 
for every solution $X$ of $(*_2)$, the matrix $XR$ is also a
solution of $(*_2)$. 
Since Proposition \ref{Kway1}(2)  requires that
all  $\transpos{(X^j)}\Degsym X^j$ have the same value
in order to have $\mu(X) = \mu(XR)$,
in general, if $X$  is a solution of $(*_1)$, the matrix $XR$ is not
necessarily a solution of $(*_1)$. 
However, every solution $X$ of 
$(*_2)$  is also a solution of  $(*_1)$, 
for every $R\in \mathbf{O}(K)$,
$XR$ is a solution  of both $(*_2)$ and $(*_1)$, and
since $(*_1)$ is scale-invariant,
for every diagonal invertible matrix $\Lambda$, the matrix $XR\Lambda$ 
is a solution of $(*_1)$. 

\medskip
In summary, every solution $Z$ of problem $(*_2)$ yields
a {\it family of solutions\/} of problem $(*_1)$; namely,
all matrices of the form $ZR\Lambda$, where $R \in \mathbf{O}(K)$
and $\Lambda$ is a  diagonal invertible matrix.
We will take advantage of this fact in looking for a
discrete solution $X$ ``close'' to a solution $Z$ of the relaxed problem
$(*_2)$. 

\medskip
Observe that a matrix is of the form
$R\Lambda$ with $R\in \mathbf{O}(K)$ and $\Lambda$
a diagonal invertible matrix
iff its columns are  nonzero and pairwise orthogonal.
First, we have
\[
\transpos{(R\Lambda)}R\Lambda = \transpos{\Lambda}\transpos{R}R \Lambda
= \Lambda^2,
\]
which implies that the columns of $R\Lambda$
are nonzero and pairwise orthogonal. Conversely, if
the  columns of $A$ are nonzero and pairwise orthogonal,
then 
\[
\transpos{A}A = \Lambda^2
\]
for some invertible diagonal matrix $\Lambda$, and then
$A = R \Lambda$, where $R = A\Lambda^{-1}$ is orthogonal.

\medskip
As a consequence of the invariance of solutions of $(*_2)$
under multiplication on the right by  matrices in $\mathbf{O}(K)$,
as explained below, 
we can view the solutions of
problem $(*_2)$ as elements of the {\it Grassmannian\/} $G(K, N)$.

\medskip
Recall that the {\it Stiefel manifold\/} $St(k, n)$ consists of the
set of orthogonal $k$-frames in $\reals^n$, that is, the 
$k$-tuples of orthonormal vectors $(u_1, \ldots, u_k)$ with $u_i\in \reals^n$.
For $k = n$, the manifold $St(n, n)$ is identical to the orthogonal
group $\mathbf{O}(n)$. For $1 \leq n \leq n - 1$,  the group
$\mathbf{SO}(n)$ acts transitively on $St(k, n)$, and $St(k, n)$ is
isomorphic to the coset manifold $\mathbf{SO}(n)/\mathbf{SO}(n - k)$.
The {\it Grassmann manifold\/} $G(k, n)$ consists of all (linear) 
$k$-dimensional subspaces of $\reals^n$. Again, 
the group $\mathbf{SO}(n)$ acts transitively on $G(k, n)$, and $G(k, n)$ is
isomorphic to the coset manifold $\mathbf{SO}(n)/S(\mathbf{O}(k)\times\mathbf{O}(n - k))$.
The  group $\mathbf{O}(k)$ acts on the right on the Stiefel manifold
$St(k, n)$ (by multiplication), and the orbit manifold
$St(k,n)/\mathbf{O}(k)$ is isomorphic to the Grassmann manifold $G(k,n)$. Furthermore, both $St(k, n)$ and $G(k, n)$ are {\it naturally
reductive homogeneous manifolds\/} (for the Stiefel manifold, when $n \geq 3$), and
$G(k, n)$ is even a {\it symmetric space\/}
(see O'Neill \cite{Oneill}). The upshot of all this is
that to a large extent, the differential geometry of these manifolds
is completely determined by some subspace $\mfrac{m}$ of the Lie algebra
$\mfrac{so}(n)$, such that we have a direct sum 
\[
\mfrac{so}(n) = \mfrac{m} \oplus \mfrac{h},
\]
where $\mfrac{h} = \mfrac{so}(n - k)$ in the case of the Stiefel
manifold, and $\mfrac{h} = \mfrac{so}(k) \times \mfrac{so}(n - k)$ in
the case of the Grassmannian manifold
(some additional condition on $\mfrac{m}$ is required). 
In particular,  the geodesics in both manifolds can be determined
quite explicitly, and thus we obtain closed form formulae for
distances, {\it etc}.

\medskip
The Stiefel manifold $St(k, n)$ can be viewed as the set of all $n
\times k$ matrices $X$ such that
\[
\transpos{X}  X = I_k.
\]
In our situation, we are considering $N\times K$ matrices $X$ such
that
\[
\transpos{X} D X = I.
\]
This is not quite the Stiefel manifold, but if we write 
$Y = D^{1/2} X$, then we have
\[
\transpos{Y} Y = I,
\]
so the space of matrices $X$ satisfying the condition 
$\transpos{X} D X = I$ is the image $\s{D}(St(K, N))$ of the
Stiefel manifold $St(K, N)$ under the linear map $\s{D}$ given by
\[
\s{D}(X) = D^{1/2} X.
\]
Now, the right action of $\mathbf{O}(K)$ on $\s{D}(St(K, N))$  yields a coset
manifold
$\s{D}(St(K, N))/\mathbf{O}(K)$ which is obviously isomorphic to the
Grassmann manidold $G(K, N)$. 

\medskip
Therefore, 
{\it the solutions of
problem $(*_2)$ can be viewed 
as elements of the {\it Grassmannian\/} $G(K, N)$\/}.
We can take advantage of this fact to find a discrete solution of our
original optimization problem PNC2 
approximated by  a continuous solution of $(*_2)$. 

\medskip
Recall that if $\transpos{X}X$ is invertible (which is the case),
condition
$X (\transpos{X} X)^{-1} \transpos{X} \mathbf{1} = \mathbf{1}$
is equivalent to $XX^+ \mathbf{1} = \mathbf{1}$, which is also
equivalent to the fact that $\mathbf{1}$ is in the range of $X$.
If we make the change of variable $Y = \Degsym^{1/2} X$ or equivalently
$X = \Degsym^{-1/2} Y$,  the condition that 
$\mathbf{1}$ is in the range of $X$ becomes the condition that
$\Degsym^{1/2} \mathbf{1}$ is in the range of $Y$, which is equivalent
to
\[
Y Y^+ \Degsym^{1/2} \mathbf{1} = \Degsym^{1/2} \mathbf{1}.
\]
However, since $\transpos{Y} Y = I$, we have
\[
Y^+ = \transpos{Y},
\]
so we get the equivalent problem

\medskip\noindent
{\bf Problem $(**_2)$}

\begin{align*}
& \mathrm{minimize}     &  &  
\mathrm{tr}(\transpos{Y}\Degsym^{-1/2} L \Degsym^{-1/2} Y)& &  &  &\\
& \mathrm{subject\ to} &  & 
\transpos{Y} Y = I, 
 & &  & & \\
& & & 
Y  \transpos{Y} \Degsym^{1/2} \mathbf{1}  = \Degsym^{1/2} \mathbf{1}.  & &  & & 
\end{align*}

\medskip
This time, the matrices $Y$ satisfying condition $\transpos{Y} Y = I$
do belong to the Stiefel manifold $St(K,  N)$, and again, {\it we view the
solutions of problem $(**_2)$ as elements of the Grassmannian $G(K, N)$\/}.
We pass from a solution $Y$ of problem $(**_2)$ in $G(K, N)$ to a
solution $Z$ of of problem $(*_2)$ in $G(K, N)$  by the linear map
$\s{D}^{-1}$; namely, $Z = \s{D}^{-1}(Y) = D^{-1/2} Y$.

\medskip
It is not a priori obvious that the minimum of
$\mathrm{tr}(\transpos{Y} L_{\mathrm{sym}} Y)$
over all $N\times K$ matrices $Y$ satisfying $\transpos{Y} Y = I$ is equal
to the sum $\nu_1 + \cdots + \nu_K$ of the first $K$ eigenvalues of 
$L_{\mathrm{sym}} = \Degsym^{-1/2} L \Degsym^{-1/2}$.
Fortunately, the Poincar\'e separation theorem
(Proposition \ref{interlace}) guarantees that 
the sum of the  $K$ smallest eigenvalues of $L_{\mathrm{sym}}$  is a lower bound for
$\mathrm{tr}(\transpos{Y} L_{\mathrm{sym}} Y)$. Furthermore,
if we temporarily ignore the second constraint, 
the minimum of  problem $(**_2)$
is achieved by any $K$ unit eigenvectors $(u_1, \ldots, u_K)$ associated with the smallest
eigenvalues
\[
0 = \nu_1\leq  \nu_2 \leq  \ldots \leq  \nu_K
\]
of $L_{\mathrm{sym}}$.%
\footnote{Other authors seem to accept this fact as obvious. This is
not quite so, and Godsil and Royle \cite{Godsil} provide a rigorous
proof using Proposition \ref{interlace}.}
We may assume that $\nu_2  > 0$, namely that the underlying graph is
connected (otherwise, we work with each connected component), in which
case $Y^1 = \Degsym^{1/2}\mathbf{1}/\norme{\Degsym^{1/2}\mathbf{1}}_2$, 
because $\mathbf{1}$ is in the nullspace of $L$.
Since $Y^1 =
\Degsym^{1/2}\mathbf{1}/\norme{\Degsym^{1/2}\mathbf{1}}_2$,
the vector $\Degsym^{1/2}\mathbf{1}$ is in the range of $Y$, so the
condition
\[
Y  \transpos{Y} \Degsym^{1/2} \mathbf{1}  = \Degsym^{1/2} \mathbf{1}
\]
is also satisfied.
Then,
$Z  = \Degsym^{-1/2}Y$ with
$Y = [u_1\> \ldots\>  u_K]$ yields a minimum of our
relaxed problem $(*_2)$ (the second constraint is satisfied because
$\mathbf{1}$ is in the range of $Z$).

\medskip
By Proposition \ref{Laplace3}, the vectors
$Z^j$  are
eigenvectors of $L_{\mathrm{rw}}$ associated with the eigenvalues
$0 = \nu_1 \leq  \nu_2 \leq  \ldots \leq  \nu_K$.
Recall that $\mathbf{1}$ is an eigenvector for the eigenvalue $\nu_1 = 0$,
and  
$Z^1 = \mathbf{1}/\norme{\Degsym^{1/2}\mathbf{1}}_2$.
Because, $\transpos{(Y^i)}Y^j = 0$ whenever $i \not= j$, 
we have
\[
\transpos{(Z^i)}\Degsym Z^j = 0, \quad\text{whenever $i \not= j$}.
\]
This implies that  $Z^2, \ldots, Z^K$ are all orthogonal to
$\Degsym\mathbf{1}$,
and thus, that each $Z^j$ has both some positive and some negative
coordinate, for $j = 2, \ldots, K$. 

\medskip
The conditions
$\transpos{(Z^i)}\Degsym Z^j = 0$ do not necessarily imply that
$Z^i$ and $Z^j$ are orthogonal (w.r.t. the Euclidean inner product),
but we can obtain a solution of Problems $(*_2)$  and $(*_1)$ 
achieving the same
minimum for which distinct columns $Z^i$ and $Z^j$ are simultaneously
orthogonal and $\Degsym$-orthogonal, 
by multiplying $Z$ by some
$K\times K$ orthogonal matrix $R$ on the right.
Indeed, if $Z$ is a solution of $(*_2)$ obtained as above,
the $K\times K$ symmetric matrix $\transpos{Z}Z$ can be
diagonalized by some orthogonal $K\times K$ matrix $R$ as  
\[
\transpos{Z}Z = R \Sigma \transpos{R},
\]
where $\Sigma$ is a diagonal matrix, 
and thus,
\[
\transpos{R}\transpos{Z} Z R = \transpos{(ZR)} ZR = \Sigma, 
\]
which shows that the columns of $ZR$ are orthogonal.
By Proposition \ref{Kway1}, $ZR$ also satisfies the 
constraints of $(*_2)$ and $(*_1)$, 
and
$\mathrm{tr}(\transpos{(ZR)} L (ZR)) = \mathrm{tr}(\transpos{Z} L Z)$.

\medskip

\remark
Since $Y$ has linearly independent columns (in fact,
orthogonal) and since $Z = D^{-1/2} Y$, the matrix $Z$ also has linearly
independent columns, so $\transpos{Z} Z$ is positive definite and 
the entries in $\Sigma$ are all positive. Also, instead of computing
$\transpos{Z} Z$ explicitly and diagonalizing it, the matrix $R$ can
be found  by computing an SVD of $Z$.

\medskip
In summary,   we  should look for a solution $Z$ of $(*_2)$
that corresponds to 
an element of the Grassmannian $G(K, N)$, and
hope that  for some suitable orthogonal matrix $R$
and some diagonal invertible matrix $\Lambda$,
the vectors in  $XR\Lambda$  are close to a true solution of the original problem.

\section[$K$-Way Clustering; Using The Dependencies Among $X^1, \ldots, X^K$]
{$K$-Way Clustering; Using The Dependencies \\
Among $X^1, \ldots, X^K$}
\label{ch3-sec4}
At this stage, it is interesting to reconsider the case $K = 2$ in the
light of what we just did when $K \geq 3$. When $K = 2$, $X^1$ and
$X^2$ are not independent, and it is convenient  to assume that the nonzero
entries in $X^1$ and $X^2$ are both equal to some positive real  $c\in
\reals$, so that 
\[
X^1 + X^2 = c \mathbf{1}.
\]
To avoid subscripts, write  $(A, \overline{A})$
for the partition of $V$ that we are seeking, and as before let
$d = \transpos{\mathbf{1}} \Degsym \mathbf{1}$ and $\alpha = \mathrm{vol}(A)$.
We know from Section 
\ref{ch3-sec2} that
\begin{align*}
\transpos{(X^1)} \Degsym X^1& = \alpha c^2 \\
\transpos{(X^2)} \Degsym X^2 &=  (d - \alpha) c^2,
\end{align*}
so we normalize $X^1$ and $X^2$ so that $\transpos{(X^1)} \Degsym X^1
= \transpos{(X^2)} \Degsym X^2 = c^2$, and we consider 
\[
X = \left[\frac{X^1}{\sqrt{\alpha}} \> \frac{X^2}{\sqrt{d - \alpha}}\right]. 
\]
Now, we claim that there is an orthogonal matrix $R$ so that if $X$ as above
is a solution to our discrete problem, then
$X R$ contains a multiple of $\mathbf{1}$ as a first column.
A similar observation is  made in Yu \cite{Yu} and Yu and Shi
\cite{YuShi2003}
(but beware that in these works $\alpha = \mathrm{vol}(A)/\sqrt{d}$).
In fact,  
\[
R = 
\frac{1}{\sqrt{d}} 
\begin{pmatrix}
 \sqrt{\alpha} & \sqrt{d - \alpha} \\[6pt]
\sqrt{d - \alpha} & - \sqrt{\alpha} 
\end{pmatrix}.
\]
Indeed, we have
\begin{align*}
X R & =  \left[\frac{X^1}{\sqrt{\alpha}} \> \frac{c \mathbf{1} - X^1}{\sqrt{d -
      \alpha}}\right] R \\
& =  \left[\frac{X^1}{\sqrt{\alpha}} \> \frac{c \mathbf{1} - X^1}{\sqrt{d -
      \alpha}}\right]
\frac{1}{\sqrt{d}} 
\begin{pmatrix}
 \sqrt{\alpha} & \sqrt{d - \alpha} \\[6pt]
\sqrt{d - \alpha} & - \sqrt{\alpha} 
\end{pmatrix} \\
& = \frac{1}{\sqrt{d}} 
\left[
c \mathbf{1} \>\> \sqrt{\frac{d - \alpha}{\alpha}}\, X^1 
- \sqrt{\frac {\alpha}{d   - \alpha}}\,(c \mathbf{1} - X^1)
\right].
\end{align*}
If we let
\[
a = c\sqrt{\frac{d - \alpha}{\alpha}}, \quad b = - c\sqrt{\frac {\alpha}{d   - \alpha}},
\]
then we check that
\[
\alpha a + b (d - \alpha) = 0, 
\]
which shows that the vector
\[
Z = \sqrt{\frac{d - \alpha}{d\alpha}}\, X^1 
- \sqrt{\frac {\alpha}{d(d   - \alpha)}}\,(c \mathbf{1} - X^1)
\]
is a potential solution of our discrete problem in the sense of Section
\ref{ch3-sec2}.
Furthermore, because $L \mathbf{1} = 0$, 
\[
\mathrm{tr}(\transpos{X} L X) = \mathrm{tr}(\transpos{(XR)} L (XR))  = \transpos{Z} L Z, 
\]
the vector $Z$ is indeed a solution of our discrete problem. 
Thus, we reconfirm the fact that the second eigenvector of
$L _{\mathrm{rw}} = \Degsym^{-1}  L$
is indeed a continuous approximation to the clustering problem when
$K = 2$. This can be generalized for any $K \geq 2$.

\medskip
Again, we may assume that the nonzero entries in $X^1, \ldots, X^K$
are some positive real $c\in \reals$, so that
\[
X^1 + \cdots + X^K = c \mathbf{1},
\]
and if $(A_1, \ldots, A_K)$ is
the partition of $V$ that we are seeking, write
$\alpha_j = \mathrm{vol}(A_j)$. We have
$\alpha_1 + \cdots + \alpha_K = d = \transpos{\mathbf{1}} \Degsym \mathbf{1}$. 
Since
\[
\transpos{(X^j)} \Degsym X^j = \alpha_j c^2,
\]
we normalize the  $X^j$ so that $\transpos{(X^j)} \Degsym X^j
= \cdots = \transpos{(X^K)} \Degsym X^K = c^2$, and we consider 
\[
X = \left[\frac{X^1}{\sqrt{\alpha_1}} \> \frac{X^2}{\sqrt{\alpha_2}}
  \>\cdots \> \> \frac{X^K}{\sqrt{\alpha_K}}\right]. 
\]
Then, we have the following result.

\begin{proposition}
\label{propdep1}
If $X = \left[\frac{X^1}{\sqrt{\alpha_1}} \> \frac{X^2}{\sqrt{\alpha_2}}
  \>\cdots \> \> \frac{X^K}{\sqrt{\alpha_K}}\right]$ is a solution of
our discrete problem, then there is an orthogonal matrix $R$ such that
its first column  $R^1$ is
\[
R^1 =
\frac{1}{\sqrt{d}}
\begin{pmatrix}
\sqrt{\alpha_1} \\ 
\sqrt{\alpha_2} \\
\vdots             \\
\sqrt{\alpha_K}
\end{pmatrix}
\] 
and
\[
XR = \left[\frac{c}{\sqrt{d}}  \mathbf{1} \> Z^2 \>\cdots \> Z^{K}\right]. 
\]
Furthermore, 
\[
\transpos{(XR)}\Degsym (XR) = c^2 I
\]
and
\[
\mathrm{tr} (\transpos{(XR)} L (XR)) = \mathrm{tr}(\transpos{Z} L Z),
\]
with $Z = [Z^2 \> \cdots \> Z^{K}]$.
\end{proposition}
\begin{proof}
Apply  Gram--Schmidt to $(R^1, e_2, \ldots, e_K)$ (where $(e_1,
\ldots, e_K)$ is the canonical basis of $\reals^K$) to
form an orthonormal basis. The rest follows from Proposition \ref{Kway1}.
\end{proof}

Proposition \ref{propdep1} suggests that if $Z = [\mathbf{1}\> Z^2 \>\cdots\> Z^K]$
is a solution of the relaxed problem $(*_2)$, then  there should be
an orthogonal matrix $R$ 
such that $Z\transpos{R}$ is an approximation of a 
solution of the discrete problem PNC1.

\section[Discrete Solution Close to a Continuous  Approximation]
{Finding a Discrete Solution Close to a Continuous  Approximation}
\label{ch3-sec5}
The next step is to find 
an exact solution  $(\mathbb{P}(X^1), \ldots, \mathbb{P}(X^K)) \in \mathbb{P}(\s{K})$
which is the closest (in a suitable sense) to our
approximate solution $(Z^1, \ldots, Z^K)\in G(K, N)$.
The set $\s{K}$ is  closed under very special  orthogonal
transformations
in $\mathbf{O}(K)$, so we can't view $\s{K}$ as a subset of the
Grassmannian $G(K, N)$. However, we can think of   $\s{K}$ as a
subset of  $G(K, N)$ by considering  the subspace 
spanned by $(X^1, \ldots, X^K)$ for every $[X^1\> \cdots X^K\> ] \in
\s{K}$.

\medskip
Recall from Section \ref{ch3-sec3}
that every solution $Z$ of problem $(*_2)$ yields
a {\it family of solutions\/} of problem $(*_1)$; namely,
all matrices of the form $ZQ$, where $Q$ is a $K\times K$
matrix with nonzero and pairwise orthogonal columns.
Since the solutions $ZQ$  of $(*_1)$ are all equivalent
(they yield the same minimum for the normalized cut),
it makes sense to look for a discrete solution $X$
closest to one of these $ZQ$.
Then, we have two choices of
distances.
\begin{enumerate}
\item
We view $\s{K}$ as a subset of $(\mathbb{RP}^{N-1})^K$.
Because $\s{K}$ is closed under the antipodal map, as
explained in Appendix \ref{ch3-sec6}, for every $j$ ($1\leq j \leq K$),
minimizing the distance
$d(\mathbb{P}(X^j), \mathbb{P}(Z^j))$ on $\mathbb{RP}^{N-1}$ is equivalent
to minimizing  $\norme{X^j - Z^j}_2$,
where $X^j$ and $Z^j$ are representatives of
$\mathbb{P} (X^j)$ and  $\mathbb{P}(Z^j)$  on the unit sphere
(if we use the Riemannian metric on $\mathbb{RP}^{N-1}$ induced by the Euclidean
metric on $\reals^N$).
Then, if we use the product distance on
$(\mathbb{RP}^{N-1})^K$ given by
\[
d\bigl((\mathbb{P}(X^1), \ldots, \mathbb{P}(X^K)), (\mathbb{P}(Z^1), \ldots, \mathbb{P}(Z^K))\bigr) 
= \sum_{j = 1}^K d(\mathbb{P}(X^j), \mathbb{P}(Z^j)),
\]
minimizing the distance 
$d\bigl((\mathbb{P}(X^1), \ldots, \mathbb{P}(X^K)), (\mathbb{P}(Z^1), \ldots, \mathbb{P}(Z^K))\bigr)$
in  $(\mathbb{RP}^{N-1})^K$ is equivalent to minimizing  
\[
\sum_{j = 1}^K \norme{X^j - Z^j}_2,
\quad \text{subject to}\quad
\norme{X^j}_2 = \norme{Z^j}_2 \> (j = 1, \ldots, K).
\]
We are not aware of any optimization method to solve the above
problem, which seems difficult to tackle due to constraints
$\norme{X^j}_2 = \norme{Z^j}_2$ ($j = 1, \ldots, K$).
Therefore, we drop these constraints and attempt to minimize 
\[
\norme{X - Z}_F^2 =  \sum_{j = 1}^K \norme{X^j - Z^j}_2^2,
\]
the Frobenius norm of $X - Z$.
This is implicitly the choice made by Yu. 
\item
We view $\s{K}$ as a subset of the Grassmannian $G(K, N)$. 
In this case, we need to pick a metric on the Grassmannian,
and we minimize the corresponding  Riemannian distance $d(X, Z)$.
A natural choice is the metric on $\mfrac{so}(n)$ given by
\[
\lag X, Y\rag = \mathrm{tr}(\transpos{X} Y).
\]
This choice remains to be explored.
\end{enumerate}

\medskip
Inspired by Yu \cite{Yu} and the previous discussion,
given a solution $Z$ of problem $(*_2)$, 
we  look for pairs
$(X, Q)$ with $X\in \s{X}$ and  where $Q$ is a $K\times K$ matrix with
nonzero and pairwise orthogonal columns,
with $\norme{X}_F = \norme{Z}_F$,
that minimize
\[
\varphi(X, Q) = \norme{X - ZQ}_F.
\]
Here, $\norme{A}_F$ is the Frobenius norm of $A$, with
$\norme{A}_F^2 = \mathrm{tr}(\transpos{A} A)$.
Yu \cite{Yu} and Yu and Shi \cite{YuShi2003} consider the special case
where $Q\in \mathbf{O}(K)$. 
We consider the more general case where
$Q = R\Lambda$, with
$R\in \mathbf{O}(K)$ and $\Lambda$ is a diagonal invertible matrix.

\medskip 
The key to minimizing $\norme{X - ZQ}_F$  rests on the following computation:
\begin{align*}
\norme{X - ZQ}_F^2 & = \mathrm{tr}(\transpos{(X - ZQ)}(X - ZQ)) \\
& = \mathrm{tr}((\transpos{X} -\transpos{Q}\transpos{Z})(X - ZQ)) \\
& = \mathrm{tr}(\transpos{X}X- \transpos{X}ZQ - \transpos{Q}\transpos{Z}X
+ \transpos{Q}\transpos{Z}ZQ)\\
& = \mathrm{tr}(\transpos{X}X)  -
\mathrm{tr}(\transpos{X}ZQ)   - \mathrm{tr}(\transpos{Q}\transpos{Z}X)  
+ \mathrm{tr}(\transpos{Q}\transpos{Z}ZQ)\\
& = \mathrm{tr}(\transpos{X}X)  -
\mathrm{tr}(\transpos{(\transpos{Q}\transpos{Z}X)})  
- \mathrm{tr}(\transpos{Q}\transpos{Z}X)  
+ \mathrm{tr}(\transpos{Z}ZQ \transpos{Q})\\
& = \norme{X}^2_F - 2\mathrm{tr}(\transpos{Q}\transpos{Z}X)   
+ \mathrm{tr}(\transpos{Z}ZQ \transpos{Q}).
\end{align*}
Therefore, since  $\norme{X}_F = \norme{Z}_F$ is fixed,   
minimizing $\norme{X - ZQ}_F^2$ is equivalent to minimizing
$ - 2\mathrm{tr}(\transpos{Q}\transpos{Z}X)   
+ \mathrm{tr}(\transpos{Z}ZQ \transpos{Q})$.

\medskip
This is a hard problem because it is a nonlinear optimization problem 
involving two matrix unknowns $X$ and $Q$. To simplify the problem,
we proceed by alternating steps during which we minimize 
$\varphi(X, Q) = \norme{X - ZQ}_F$ with respect to $X$ holding $Q$
fixed, and steps during which we minimize 
$\varphi(X, Q) = \norme{X - ZQ}_F$ with respect to $Q$ holding $X$
fixed. 

\medskip
This second step in which $X$ is held fixed has been studied, but it
is still a hard problem for which no closed--form solution is known.
Consequently, we further simplify the problem.
Since $Q$ is of the form $Q = R\Lambda$ where
$R\in \mathbf{O}(K)$ and $\Lambda$ is a diagonal invertible matrix,
we minimize $\norme{X - ZR\Lambda}_F$ in two stages.
\begin{enumerate}
\item
We set $\Lambda = I$ and find $R\in \mathbf{O}(K)$
that minimizes  $\norme{X - ZR}_F$.
\item
Given $X$, $Z$, and $R$,  find a 
diagonal invertible matrix $\Lambda$ that
minimizes  $\norme{X - ZR\Lambda}_F$.
\end{enumerate}

The matrix $R\Lambda$ is not a minimizer of
$\norme{X - ZR\Lambda}_F$ in general, but it is an improvement
on $R$ alone, and both stages can be solved quite easily.

\medskip
In stage 1, the matrix $Q=R$ is orthogonal, so $Q\transpos{Q} = I$, and
since $Z$ and $X$ are given, 
the problem reduces to minimizing
$ - 2\mathrm{tr}(\transpos{Q}\transpos{Z}X)$; that is,
maximizing   $\mathrm{tr}(\transpos{Q}\transpos{Z}X)$.
To solve this problem, we  need the following
proposition. 

\begin{proposition}
\label{Kway2}
For any $n \times n$ matrix $A$ and  any orthogonal matrix $Q$, we
have
\[
\max\{\mathrm{tr}(QA) \mid Q\in \mathbf{O}(n)\} 
= \sigma_1  + \cdots + \sigma_n,
\]
where $\sigma_1 \geq \cdots \geq \sigma_n$ are the singular values of
$A$.
Furthermore, this maximum is achieved by  $Q =
V\transpos{U}$,
where $A = U \Sigma \transpos{V}$ is any SVD for $A$.
\end{proposition}

\begin{proof}
Let $A = U \Sigma \transpos{V}$ be any SVD for $A$. Then we have
\begin{align*}
\mathrm{tr}(QA) &  = \mathrm{tr}(Q U \Sigma \transpos{V})\\
&  = \mathrm{tr}(\transpos{V}  Q U \Sigma). 
\end{align*}
The matrix $Z = \transpos{V}  Q U$ is an orthogonal matrix 
so $|z_{i j}| \leq 1$ for $1\leq i, j \leq n$, and
$\Sigma$ is a diagonal matrix, so we have
\[
\mathrm{tr}(Z\Sigma) = z_{1 1} \sigma_1 + \cdots + z_{n n } \sigma_n
\leq \sigma_1 + \cdots + \sigma_n, 
\]
which proves the first statement of the proposition.
For $Q = V\transpos{U}$, we get
\begin{align*}
\mathrm{tr}(QA)  &  = \mathrm{tr}(Q U \Sigma \transpos{V}) \\
 &  = \mathrm{tr}(V\transpos{U} U \Sigma \transpos{V}) \\
&  = \mathrm{tr}(V\Sigma \transpos{V}) = \sigma_1  + \cdots + \sigma_n,
\end{align*}
which proves the second part of the proposition.
\end{proof}

As a corollary of Proposition \ref{Kway2} (with $A = \transpos{Z} X$
and $Q = \transpos{R}$), we get the following
result (see  Golub and Van Loan \cite{Golub}, Section 12.4.1):

\begin{proposition}
\label{Kway3}
For any two fixed $N\times K$ matrices $X$ and $Z$,  the minimum of
the set
\[
\{\norme{X - ZR}_F \mid R\in \mathbf{O}(K)\}
\]
is achieved by $R = U \transpos{V}$, for any SVD decomposition 
$U \Sigma \transpos{V} = \transpos{Z} X$ of $\transpos{Z} X$.
\end{proposition}

The following proposition takes care of stage 2.

\begin{proposition}
\label{Kway4}
For any two fixed $N\times K$ matrices $X$ and $Z$, where $Z$ has no
zero column, there is a unique diagonal matrix
$\Lambda  = \mathrm{diag}(\lambda_1, \ldots,  \lambda_K)$
minimizing $\norme{X - Z\Lambda}_F$ given by
\[
\lambda_j = \frac{(\transpos{Z}X)_{j j}}{\norme{Z^j}_2^2}\quad
j = 1, \ldots, K.
\]
\end{proposition}

\begin{proof}
Since $\Lambda$ is a diagonal matrix,  we have
\begin{align*}
\norme{X - Z\Lambda}^2 &  = 
\norme{X}_2^2   - 2\mathrm{tr}(\transpos{\Lambda}\transpos{Z}X)   
+ \mathrm{tr}(\transpos{Z}Z\Lambda \transpos{\Lambda}) \\
& =  \norme{X}_2^2  - 2\mathrm{tr}(\transpos{Z}X \Lambda)   
+ \mathrm{tr}(\transpos{Z}Z\Lambda^2) \\
& =  \norme{X}_2^2 
-2\sum_{j = 1}^K (\transpos{Z}X)_{j j}\lambda_j
+ \sum_{j = 1}^K \norme{Z^j}_2^2\lambda_j^2.
\end{align*}
The above functional has a critical point obtained by setting the
partial derivatives with respect to the $\lambda_j$ to $0$, which
gives
\[
- 2(\transpos{Z}X)_{j j} + 2 \norme{Z^j}_2^2\lambda_j = 0;
\]
that is,
\[
\lambda_j = \frac{(\transpos{Z}X)_{j j}}{\norme{Z^j}_2^2}.
\]
Since the functional is a sum of quadratic functions
and the coefficients $\norme{Z^j}_2^2$
of the $\lambda_j^2$ are positive, this critical point is indeed a minimum.
\end{proof}

It should be noted that Proposition \ref{Kway4}
does not guarantee that $\Lambda$ is invertible.
For example, for 
\[
X = 
\begin{pmatrix}
1 & 0 \\
0 & 1\\
1 & 0
\end{pmatrix},
\quad
Z = 
\begin{pmatrix}
1 & 1 \\
1 & 0\\
1 & -1
\end{pmatrix},
\]
we have
\[
\transpos{Z} X = 
\begin{pmatrix}
1 & 1 & 1 \\
1 & 0 & -1
\end{pmatrix}
\begin{pmatrix}
1 & 0 \\
0 & 1\\
1 & 0
\end{pmatrix}
= 
\begin{pmatrix}
2 & 1  \\
0 & 0 
\end{pmatrix},
\]
so $\lambda_2 = 0$.
When Proposition \ref{Kway4} yields a singular matrix,
we skip stage 2 (we set $\Lambda = I$).

\medskip
We now deal with step 1, where $Q = R\Lambda$ is held fixed.
For fixed $Z$ and $Q$, we would like to find some $X\in \s{K}$
with $\norme{X}_F = \norme{Z}_F$
so that $\norme{X - ZQ}_F$ is minimal. 
Without loss of generality, we may assume that the entries
$a_1, \ldots, a_K$ occurring in the matrix $X$ are positive
and all equal to some common value $a\not= 0$.
Recall that a matrix $X\in \s{X}$ has the property that
every row contains exactly one nonzero entry, and that every column
is nonzero.

\medskip
To find $X\in \s{K}$, first we find the shape $\widehat{X}$ of $X$, which
is the matrix obtained from $X$ by rescaling the columns of $X$ so
that $\widehat{X}$ has entries $+1,  0$. 
The problem is to decide for each row, which column contains
the nonzero entry.
After having found $\widehat{X}$, 
we rescale its columns 
so that $\norme{X}_F = \norme{Z}_F$.

\medskip
Since
\[
\norme{X - ZQ}_F^2 = \norme{X}_F^2  -
2\mathrm{tr}(\transpos{Q}\transpos{Z}X)
 + \mathrm{tr}(\transpos{Z}ZQ \transpos{Q}),
\]
minimizing $\norme{X - ZQ}_F$ is equivalent to maximizing 
\[
\mathrm{tr}(\transpos{Q}\transpos{Z}X) =
\mathrm{tr}(\transpos{(ZQ)}X) =\mathrm{tr}(X\transpos{(ZQ)}),
\]
and since the $i$th row of $X$ contains a single nonzero entry 
$a$ in column $j_i$ ($1\leq j_i \leq K$), if we write $Y = ZQ$, then 
\begin{equation}
\mathrm{tr}(X \transpos{Y}) = a\sum_{i = 1}^N   y_{i\, j_i }. 
\tag{$*$}
\end{equation}
By $(*)$, since $a > 0$, the quantity 
$\mathrm{tr}(X \transpos{Y})$ is maximized iff
$y_{i  j_i }$ is 
maximized for $i = 1, \ldots, N$; 
this is achieved if for the  $i$th
row of $X$, we pick a column index $\ell$ such that $y_{i \ell}$ is
maximum.

\medskip
To find the shape
$\widehat{X}$ of $X$, we first find a matrix $\overline{X}$ 
by chosing a single nonzero entry  $\overline{x}_{i j} = 1$ on row $i$
in such a way that $y_{i j}$ is maximum according
to the following method.
If we let
\begin{align*}
\mu_i & = \max_{1 \leq j \leq K} y_{i j} \\
J_i & = \{j \in \{1, \ldots, K\} \mid y_{i j} = \mu_i\},
\end{align*}
 for $i = 1, \ldots, N$, 
then 
\[
\overline{x}_{i j} =
\begin{cases}
+1 & \text{for some chosen  $j \in J_i$,} \\
0 & \text{otherwise}.
\end{cases}
\]
Of course, a single column index is chosen for each
row. In our implementation, we pick the smallest index in $J_i$.

\medskip
Unfortunately, the matrix
$\overline{X}$ may not be a correct solution, because the above prescription does
not guarantee that every column of $\overline{X}$ is
nonzero. When this happens, we 
reassign certain nonzero entries in columns having ``many''
nonzero entries to zero columns, so that we get a matrix in $\s{K}$.

\medskip
Suppose column $j$ is zero. Then, 
we pick the leftmost index $k$ of a column with a maximum number of
$1$, and if $i$ the smallest index for which $\overline{X}_{i k} = 1$, then
we set $\overline{X}_{i k} = 0$ and $\overline{X}_{i j} = 1$. We repeat this reallocation
scheme until every column is nonzero.

\medskip
We obtain a new matrix   $\widehat{X}$ in $\s{X}$, and finally we
normalize  $\widehat{X}$ to obtain $X$, so that
$\norme{X}_F = \norme{Z}_F$.

\medskip
A practical way to deal with zero
columns in $\overline{X}$ is to simply decrease $K$. 
Clearly, further work is needed to justify the soundness of such a method.

\medskip
The above method is essentially the method described in 
Yu \cite{Yu} and Yu and Shi \cite{YuShi2003}, 
except that in these works
(in which  $X, Z$ and $Y$ are denoted by $X^*,
\widetilde{X}^*$, and $\widetilde{X}$, respectively)
the entries in $X$
belong to $\{0, 1\}$;  as described above,
 for row $i$, the index $\ell$ 
corresponding to the entry $+1$ is given by
\[
\arg \max_{1\leq j \leq K} \widetilde{X}(i, j).
\]
The fact that  $\overline{X}$ may have zero columns is  not
addressed by Yu.
Furthermore, it is important to make sure that $X$ has
the same norm as $Z$, but this normalization step
is not performed in the above works. On the other hand, the rows of $Z$ are
normalized and   the resulting matrix may no longer
be a correct solution of the relaxed problem.
In practice, it appears to be a good approximation of a
discrete solution; see option (3) of the initialization methods 
for $Z$ described below.

\medskip
Any matrix obtained by
flipping the signs of some of the columns of a solution
$ZR$ of problem $(*_2)$ is still a solution.
Moreover, all entries in $X$ are nonnegative. It follows that a
``good'' solution $ZQ_p$ (that is, close to a discrete solution)
should have the property 
that the average of each of its column is nonnegative.
We found that the following heuristic is quite helpful in finding a
better discrete solution $X$. 
Given a solution $ZR$ of problem $(*_2)$,
we compute $ZQ_p$, defined such that if the average of column $(ZR)^j$ is
negative, then $(ZQ_p)^j = -(ZR)^j$, else $(ZQ_p)^j = (ZR)^j$.
It follows that the average of every column in $ZQ_p$ is nonnegative.
Then, we apply the above procedure to find discrete solutions
$X$ and $X_p$ closest to $ZR$ and $ZQ_p$ respectively, and we
pick the solution corresponding to
$\min\{\norme{X - ZR}_F, \norme{X_p - ZQ_p}_F\}$.
Flipping signs of columns of $ZR$ correspond to a
diagonal matrix $R_p$
with entries $\pm 1$,  a very special
kind of orthogonal matrix. In summary, the procedure 
for finding a discrete $X$ close to a continuous $ZR$ also updates 
$R$ to $Q_p = RR_p$. 
This step appears to be very effective for finding a good
initial $X$.

\medskip
The method due to Yu and Shi (see Yu \cite{Yu} and Yu and Shi
\cite{YuShi2003})  
to find $X\in \s{K}$ and $Q =R \Lambda$ with
$R\in \mathbf{O}(K)$ and $\Lambda$ diagonal invertible
that minimize
$\varphi(X, Q) = \norme{X - ZQ}_F$ is 
to alternate steps during which either $Q$ is held fixed (step PODX) or $X$ is
held fixed (step PODR), except that  Yu and Shi
consider the special case where $\Lambda = I$.  

\begin{enumerate}
\item[(1)]
In step PODX, the next discrete solution $X^*$ is obtained fom the previous
pair $(Q^*, Z)$ by computing $\overline{X}$ and then $X^* =
\widehat{X}$ from $Y = ZQ^*$,
as just explained above.
\item[(2)]
In step PODR, the next matrix $Q^* = R\Lambda$ is obtained from the previous pair
$(X^*, Z)$ by first computing
\[
R =  U \transpos{V},
\]
for any SVD decomposition 
$U \Sigma \transpos{V}$ of $\transpos{Z} X^*$, and then
computing $\Lambda$  from $X^*$ and $ZR$ 
using Proposition \ref{Kway4}. If $\Lambda$ is singular, then set
$\Lambda = I$.  
\end{enumerate}

We keep track of the progress of the procedure by computing
$\varphi(X^*,Q^*) = \norme{X^* - ZQ^*}_F$ after every step
and checking that $X^*$ or $\varphi(X^*,Q^*)$
stops changing, whichever comes first.
We observed that after a small number of steps, up to machine
precision,  $\varphi(X^*,Q^*)$ stops decreasing,  and when this occurs the
procedure halts (we also set a maximum number of steps in case
$\varphi(X^*,Q^*)$ decreases for a very long time).
Moreover, looking for  $Q = R\Lambda$ where $R\in \mathbf{O}(K)$ and
$\Lambda$ is obtained using the method of Proposition \ref{Kway4}
speeds up the convergence and yields a better discrete
solution $X$.

\medskip
The process of searching for $X$ and $Q$ has an illuminating geometric
interpretation in terms of graph drawings. We may assume that
the entries in the discrete solution  $X$ are $0$ or $1$. Then
the rows of the discrete solutions $X$ correspond to the tips of the unit vectors
along the coordinate axes in $\reals^K$. Every axis contains at least
such a point, and the multiplicity of the point along the $j$th axis is
the number of nodes in the $j$th block of the partition.
Similarly, the rows of $Z$ are the nodes of a graph drawing
of the weighted graph $(V, W)$. Multiplying $Z$ on the right
by a  $K\times K$ matrix  $Q$ (obtaining $ZQ$)
is equivalent to multiplying $\transpos{Z}$
on the left by $\transpos{Q}$ (obtaining $\transpos{Q}\transpos{Z}$).
This means that the points in $\reals^K$ representing the rows
of $ZQ$ are obtained by applying the linear transformation
$\transpos{Q}$ to the columns of $\transpos{Z}$.
Thus, $ZR$ amounts to applying the rigid motion $\transpos{R}$
to the graph drawing $Z$, and $Z\Lambda$ (where $\Lambda$
is a diagonal invertible matrix) amounts to stretching or shrinking 
the graph drawing $Z$
in the directions of the axes. 

\medskip
Then, in step 2 (PODR), we are trying
to deform the graph drawing given by $Z$ using a linear map
$\transpos{(R\Lambda)}$, so that the deformed graph drawing 
$ZR\Lambda$  is as close as
possible to  $X$ (in the sense that 
$\norme{X - ZR\Lambda}_F$ is minimized).

\medskip
In step 1 (PODX), we are trying to approximate the
deformed graph drawing $ZR\Lambda$ by a discrete graph 
drawing $X$ (whose nodes are the tips of the unit vectors),
so that $\norme{X - ZR\Lambda}_F$ is minimized.

\medskip
If we are willing to give up  the requirement that
the deformed $Z$ is still a solution of problem $(*_1)$,
we have quite a bit of freedom in step 2.
For example, we may allow normalizing the rows.
This seems reasonable to obtain an initial transformation $Q$.
However, we feel uncomfortable in allowing intermediate
deformed $Z$ that are not solutions of $(*_1)$ 
during the iteration process.
This point should be investigated further.

\medskip
In some sense, we have less freedom in step 1, since
the $i$th row of $ZR\Lambda$ is assigned to the $j$th unit vector
iff the index of the leftmost largest coordinate of this row is $j$.
If some axis has not been assigned any row of $R$, then we reallocate
one of the points on an axis with a maximum number of points.

\begin{figure}[http]
  \begin{center}
 \includegraphics[height=2.3truein,width=2.3truein]{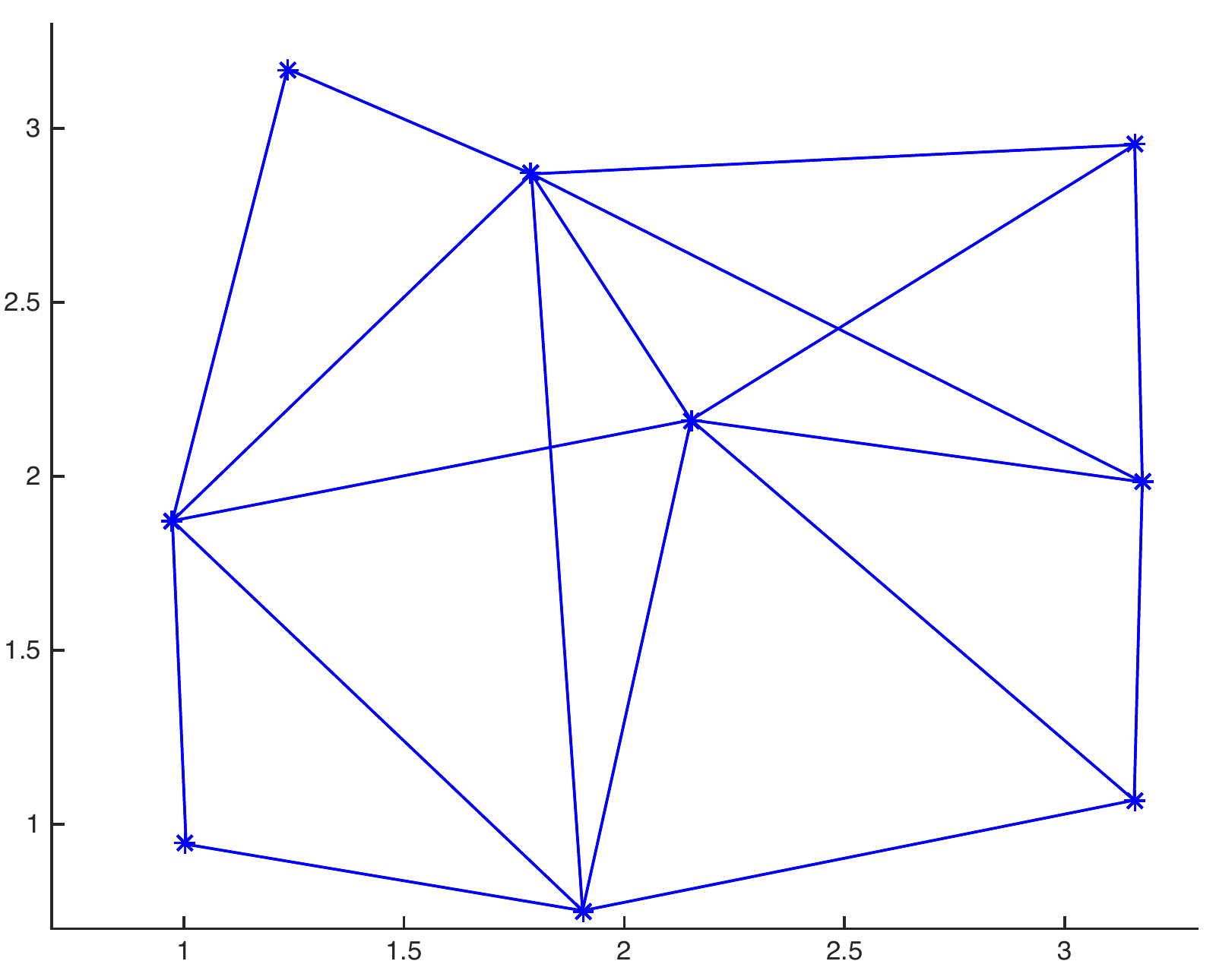}
\hspace{0.5cm}
 \includegraphics[height=3truein,width=3truein]{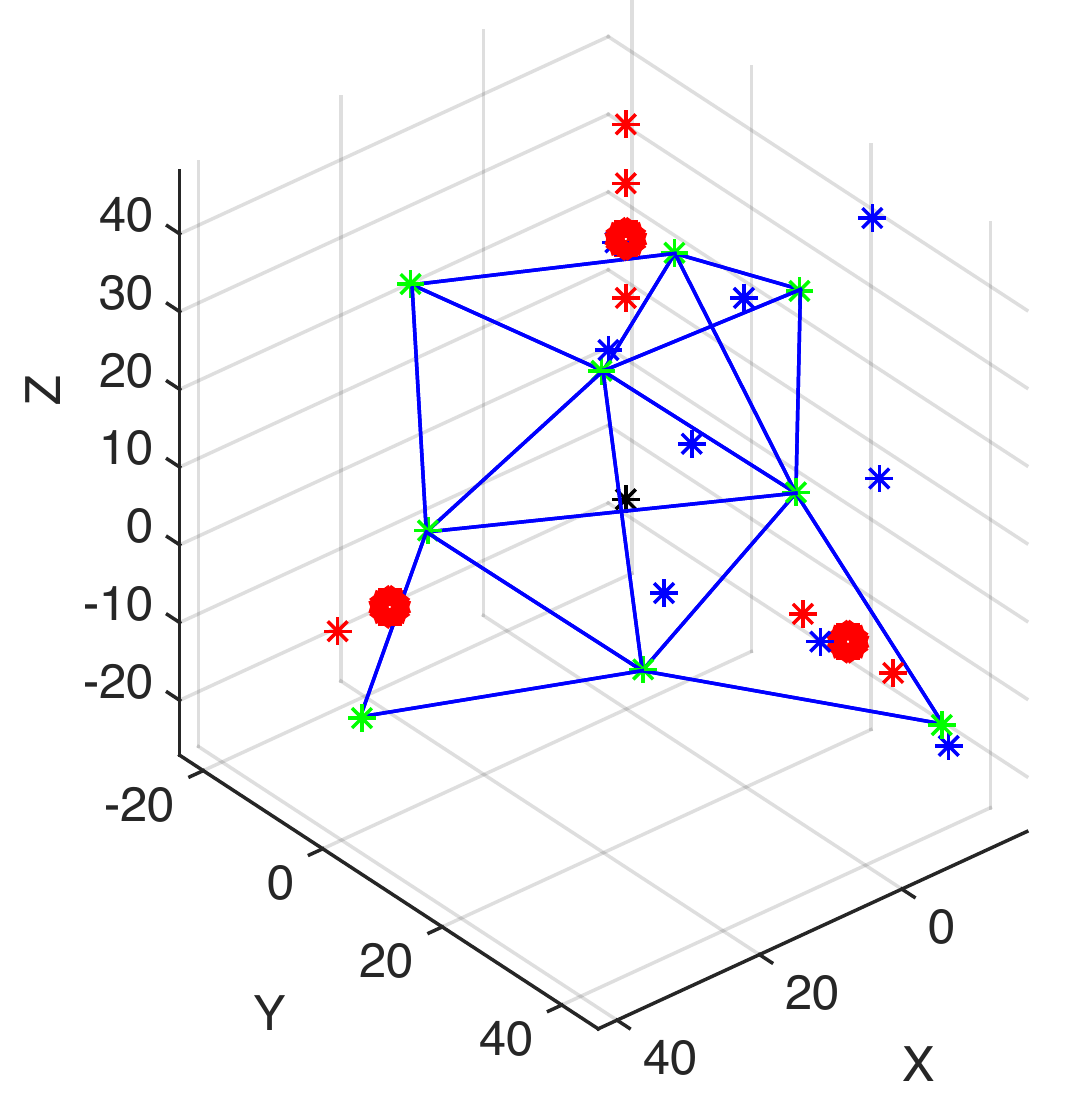}
  \end{center}
  \caption{A graph and its drawing to find $3$ clusters.}
\label{gdr1}
\end{figure}

\medskip
Figure \ref{gdr1} shows a graph (on the left) and
the graph drawings $X$ and $Z*R$ obtained by 
applying our method for three clusters.
The rows of $X$ are represented 
by the red points along the axes, and the rows of $Z*R$
by the green points (on the right).
The original vertices corresponding to the rows of 
$Z$ are represented in blue.
We can see how the two red points correspond
to an edge, the three red points correspond to a triangle, and the
four red points to a quadrangle. 
These constitute the clusters.

\medskip
It remains to initialize $Q^*$ to start the process, and then steps
(1) and (2) are iterated, starting with step (1).
Actually, what we really need is a ``good'' initial $X^*$, but to find
it, we need an initial $R^*$.

\medskip
{\it Method 1.} 
One method is to use an orthogonal matrix denoted $R_1$,
such that
distinct columns of $ZR_1$ are simultaneously
orthogonal and $\Degsym$-orthogonal. The matrix $R_1$ can be found by
diagonalizing $\transpos{Z}Z$ as 
$\transpos{Z}Z = R_1\Sigma\transpos{R_1}$, as we explained at the
end of Section  \ref{ch3-sec3}. We write
$Z_2= ZR_1$.

\medskip
{\it Method 2.} 
The method advocated by Yu \cite{Yu} is to pick $K$ rows of $Z$ that
are as orthogonal to each other as possible and to make
a matrix $R$ whose columns consist of these rows normalized to have
unit length.
The intuition behind this method is that if a continuous
solution $Z$ can be sent close
to a discrete solution $X$ by a rigid motion, then
many  rows of $Z$ viewed as vectors in $\reals^K$ should
be nearly orthogonal. This way,
$ZR$ should contain at least $K$ rows well aligned with 
the canonical basis vectors, and these rows are good candidates
for  some of the rows of the discrete solution $X$.

\medskip
The algorithm given in Yu \cite{Yu} needs a small correction,
because rows are not removed from $Z$ when they are added to $R$,
which may cause the same row to be added several times to $R$.

\medskip
Given the $N\times K$ matrix $Z$ (whose columns all have the same
norm), we compute a matrix $R$ whose columns are  certain rows of $Z$.
We use a vector $c\in \reals^N$ to keep track of the inner products
of all rows of $Z$ with the columns $R^1, \ldots, R^{k-1}$ that have been
constructed so far, and initially when $k = 1$, we set $c = 0$.

\medskip
The first column $R^1$ of $R$ is any chosen row of $Z$.

\medskip
Next, for $k = 2, \ldots, K$, we compute all the inner products of
$R^{k-1}$ with all rows in $Z$, which are recorded in the vector
$ZR^{k-1}$, and  we update $c$ as follows:
\[
c = c + \mathtt{abs}(ZR^{k - 1}).
\]
We take the  absolute values of the entries in $ZR^{k - 1}$
so that the $i$th entry in $c$ is  a score of how orthogonal
is the $i$th row  of $Z$ to $R^1, \ldots, R^{k-1}$.
Then, we  choose $R^k$ as any row $Z_i$ of $Z$ for which $c_i$ is minimal
(the customary (and ambiguous) $i = \arg \min c$), and we delete
this row from $Z$. The process is repeated (with the updated $Z$)
until $k = K$.

\medskip
At the end of the above process, we normalize the columns of $R$,
to obtain a matrix that we denote $R_2$.

\medskip
After some experimentation, we found that to obtain a better
initial $X^*$, it is may desirable to start from a variant of
the continuous solution $Z$  obtained  by solving problem $(*_2)$.
We have implemented three methods.
\begin{enumerate}
\item
We attempt to rescale the columns of $Z$ by some
diagonal invertible matrix 
$\Lambda = \mathrm{diag}(\lambda_1, \ldots, \lambda_K)$, 
so that
the rows of $Z\Lambda$ sum to $1$ as much as possible in the 
least-squares sense.
Since the vector of sums of rows of $Z\Lambda$ is
$Z\Lambda\mathbf{1}_K = Z\lambda$, with
$\transpos{\lambda} = (\lambda_1, \ldots, \lambda_K)$, the
least-squares problem is to minimize
\[
\norme{Z\lambda - \mathbf{1}_N}_2^2,
\]
and since $Z$ has rank $K$, the solution
is $\lambda = (\transpos{Z}Z)^{-1}\transpos{Z}\mathbf{1}_N$, and thus,
\[
\Lambda = \mathrm{diag}((\transpos{Z}Z)^{-1}\transpos{Z}\mathbf{1}_N).
\]
The matrix $\Lambda$  is singular if some of the columns of $Z$ sum
to $0$. This happens for regular graphs, where the
degree matrix is a multiple of the identity. There are also cases
where 
some of the $\lambda_j$ are very small, so we use a tolerance factor
to prevent this, and in case of failure, we set $\Lambda = I$.
In case of failure,
we  may also use $ZR_1$ instead of $Z$, where $R_1$ is the
orthogonal matrix that makes $ZR_1$ both $\Degsym$-orthogonal and
orthogonal.
\item
We attempt to rescale the columns of  $Z$  by some
diagonal invertible matrix 
$\Lambda = \mathrm{diag}(\lambda_1, \ldots, \lambda_K)$, 
so that
the rows of $Z\Lambda$ have unit length as much as possible in the
least-squares sense. Since the square-norm of the
$i$th row of $Z\Lambda$ is
\[
\sum_{j = 1}^K z_{i j}^2\lambda_j^2,
\]
if we write $Z\circ Z$ for the matrix $(z_{i j}^2)$
of square entries of elements in $Z$ (the Hadamard product of $Z$ with
itself), the least-squares problem is to mimimize
\[
\norme{Z\circ Z\lambda^2 - \mathbf{1}_N}_2^2,
\]
where $\transpos{(\lambda^2)} = (\lambda_1^2, \ldots, \lambda_K^2)$. 
The matrix $Z\circ Z$ may not have rank $K$, so the least-squares
solution for $\lambda^2$ is given by the pseudo-inverse of
$Z\circ Z$, as
\[
\lambda^2 = (Z\circ Z)^+ \mathbf{1}_N.
\]
There is no guarantee that the vector on the right-hand side
has all positive entries, so the method may fail.
It may also fail when some of the $\lambda_j$ are very small.
We use a tolerance factor to prevent this, and in case of failure, we 
set $\Lambda = I$.
\item
We use a method more drastic than (2), which consists in normalizing
the rows of $Z$. Thus, we form the matrix
\[
NZ = \mathrm{diag}((Z\transpos{Z})_{1 1}^{-1/2}, \ldots,
(Z\transpos{Z})_{N N}^{-1/2}),
\]
and we return $NZ* Z$.
Unlike the methods used in (1) and (2), this method does not guarantee
that $NZ*Z$ is a solution of problem $(*_1)$.
However, since the rows of $Z$ can be interpreted as
vectors in $\reals^K$ that should align as much as possible
with the canonical basis  vectors of $\reals^K$, this method makes sense
as a way to come closer to a discrete solution. In fact, 
we found that it does well in most cases.
\end{enumerate}

\medskip
We implemented a computer program that prompts the user 
for various options. To avoid confusion, let us denote the
original solution of problem $(*_2)$ by $Z_1$, and
let $Z_2 = Z_1R_1$, as obtained by initialization method 1.
The four options are:
\begin{enumerate}
\item
Use the original solution $Z_1$ of problem $(*_2)$, as well as 
$Z_2$.
\item
Apply method 1 to $Z_1$ and $Z_2$.
\item 
Apply method 2 to $Z_1$ and $Z_2$.
\item
Apply method 3 to $Z_1$ and $Z_2$.
\end{enumerate}
Then, for each of these options,
if we denote by $Zinit_1$ and $Zinit_2$
the solutions returned by the method, 
our program computes initial solutions $X_1, X_2, X_3, X_4$
as follows:
\begin{enumerate}
\item
Use $Zinit_1$ and $R = I$.
\item
Use $Zinit_1$ and $R = R2a$, the matrix given by initialization method 2.
\item
Use $Zinit_2$ and $R = I$.
\item
Use $Zinit_2$ and $R = R2b$, the matrix given by initialization method 2.
\end{enumerate}
After this, the program picks the discrete solution $X = X_i$ which
corresponds to the minimum of
\[
\norme{X1 - Zinit1}, \> \norme{X2 - Zinit1*R2a}, \> 
\norme{X3 - Zinit2}, \> \norme{X4 - Zinit2*R2b}. 
\]

Our experience is that options (3) and (4) tend to give better
results. However, it is harder to say whether any of the
$X_i$ does a better job than the others, although (2) and (4)
seem to do slightly better than (1) and (3).
We also give the user the option in step PODR
to only compute $R$ and set $\Lambda = I$.
It appears that the more general method is hardly more expansive
(because finding $\Lambda$ is cheap) and always gives better results.

\medskip
We also found that we obtain better results if we rescale $Z$ (and $X$)
so that$\norme{Z}_F = 100$.

\begin{figure}[http]
  \begin{center}
 \includegraphics[height=1.2truein,width=1.2truein]{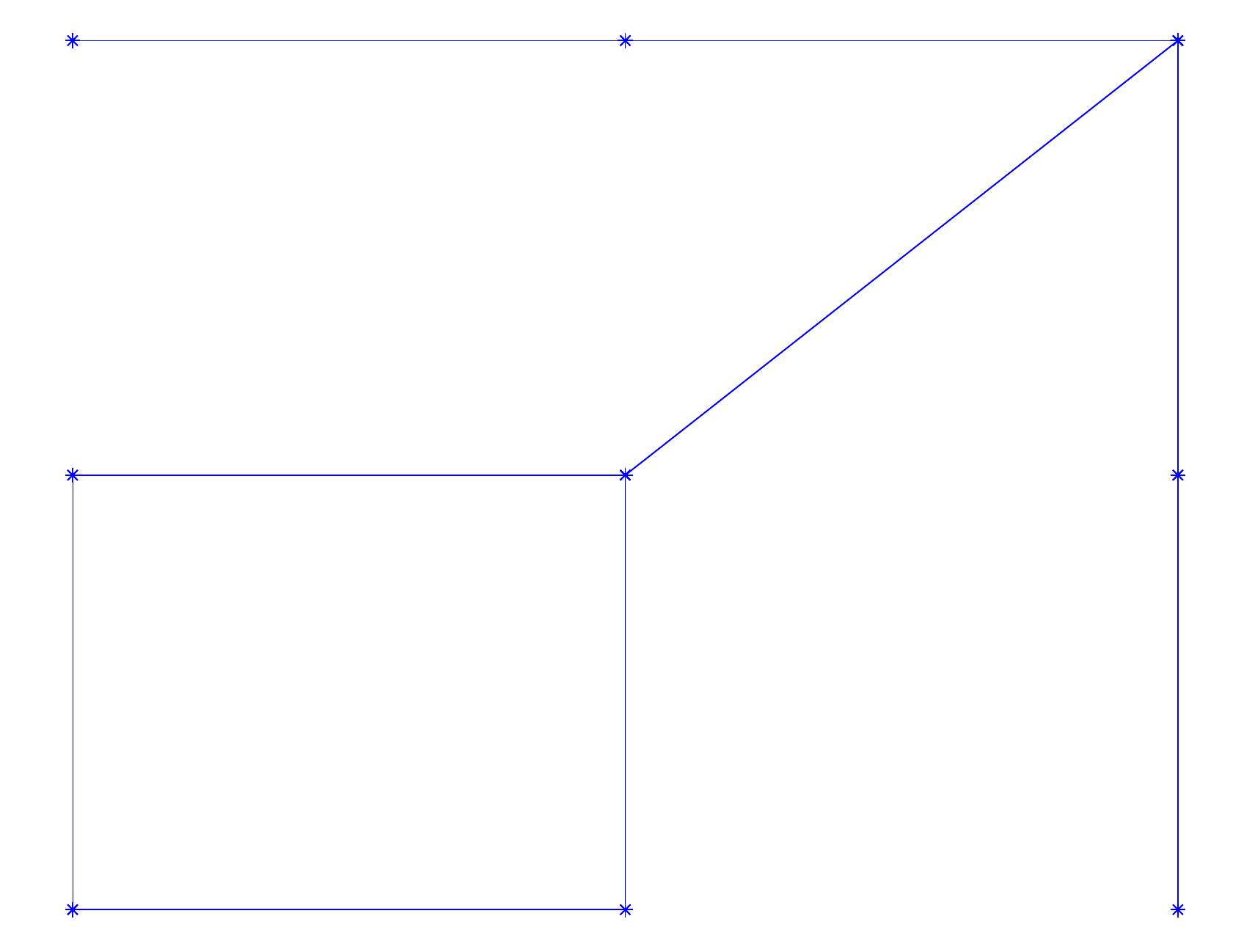}
\end{center}
  \caption{Underlying graph of the matrix $W_1$.}
\label{gr1b}
\end{figure}

\begin{figure}[http]
  \begin{center}
 \includegraphics[height=1.2truein,width=1.2truein]{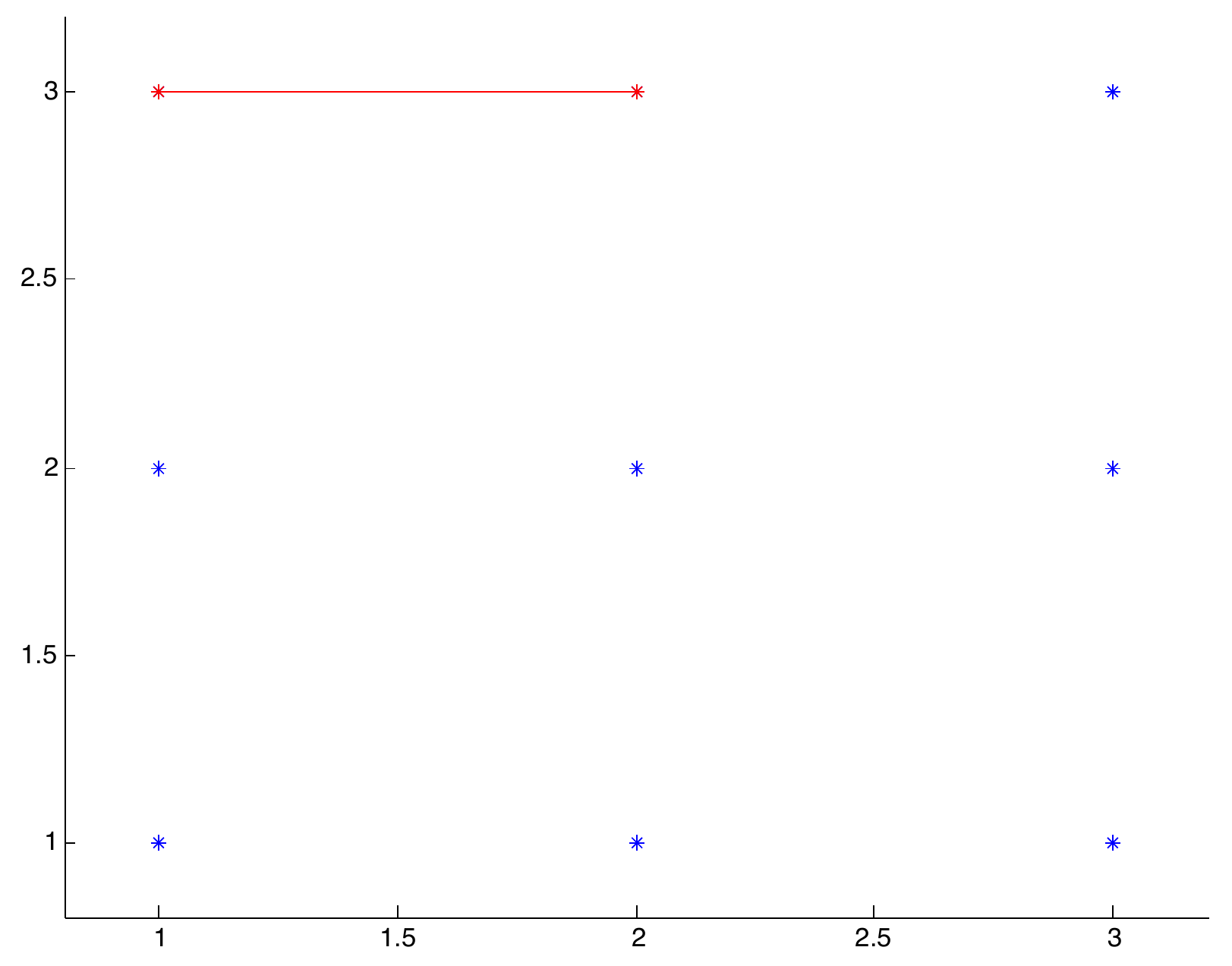}
\hspace{0.5cm}
 \includegraphics[height=1.2truein,width=1.2truein]{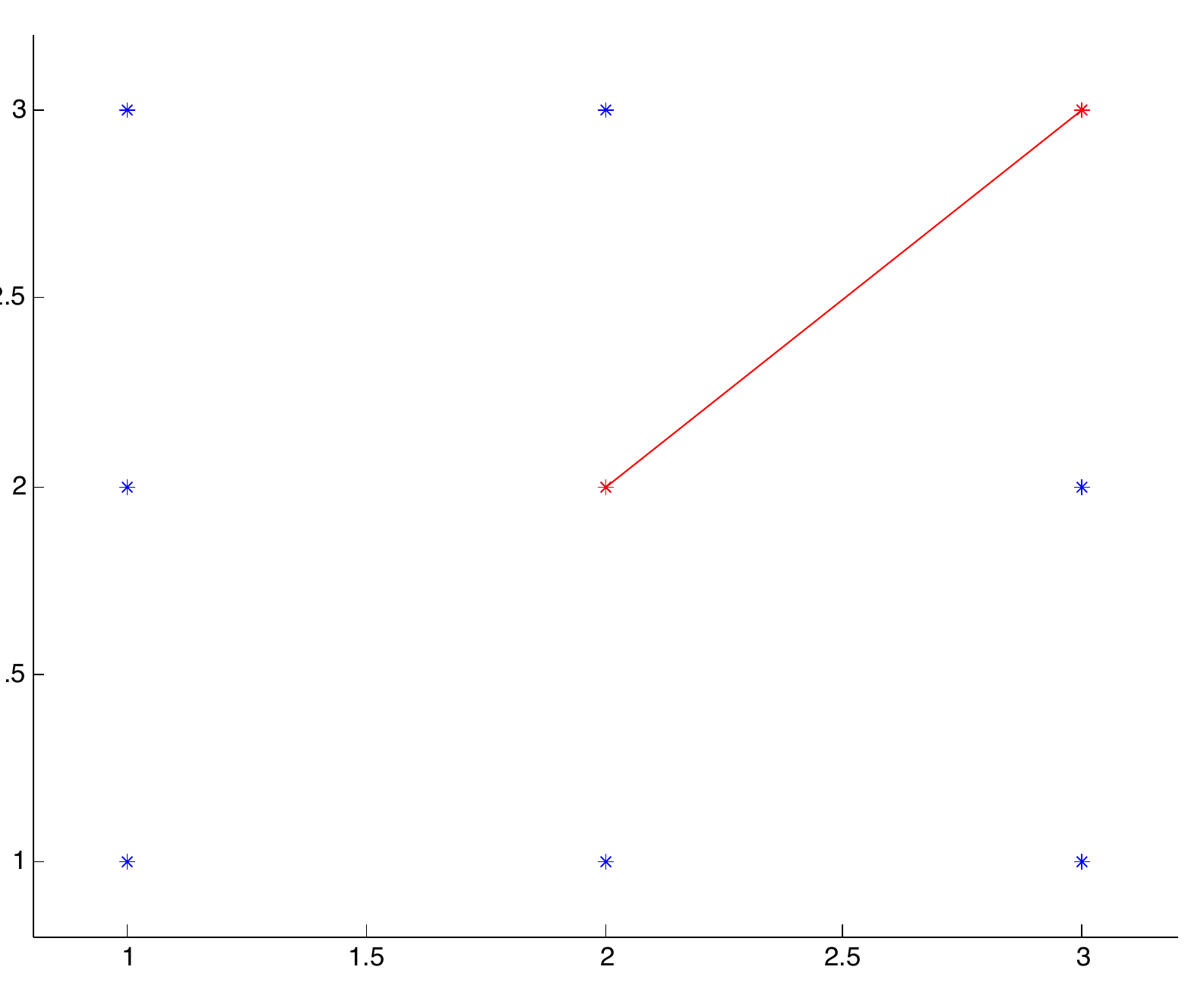}
\hspace{0.5cm}
 \includegraphics[height=1.2truein,width=1.2truein]{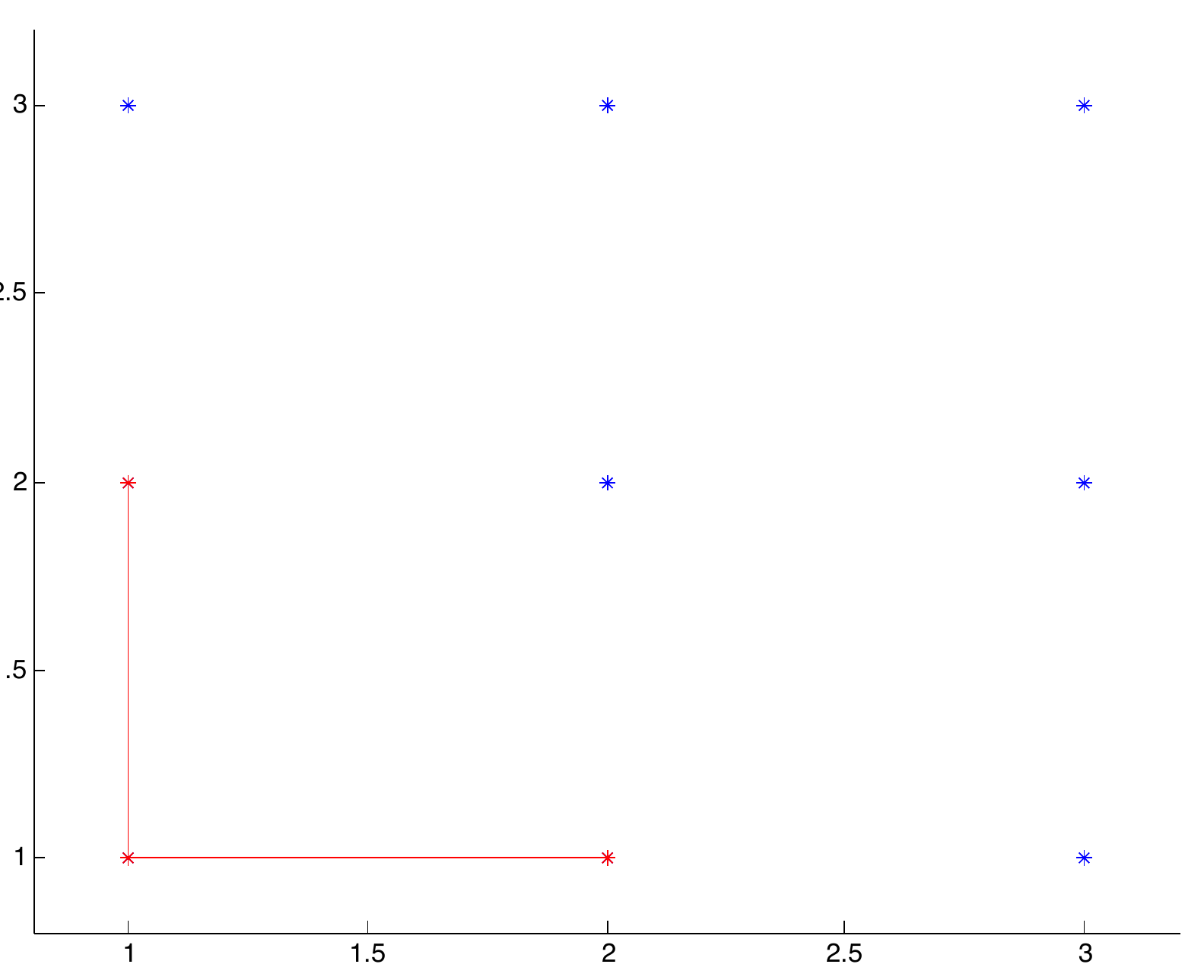}
\hspace{0.5cm}
 \includegraphics[height=1.2truein,width=1.2truein]{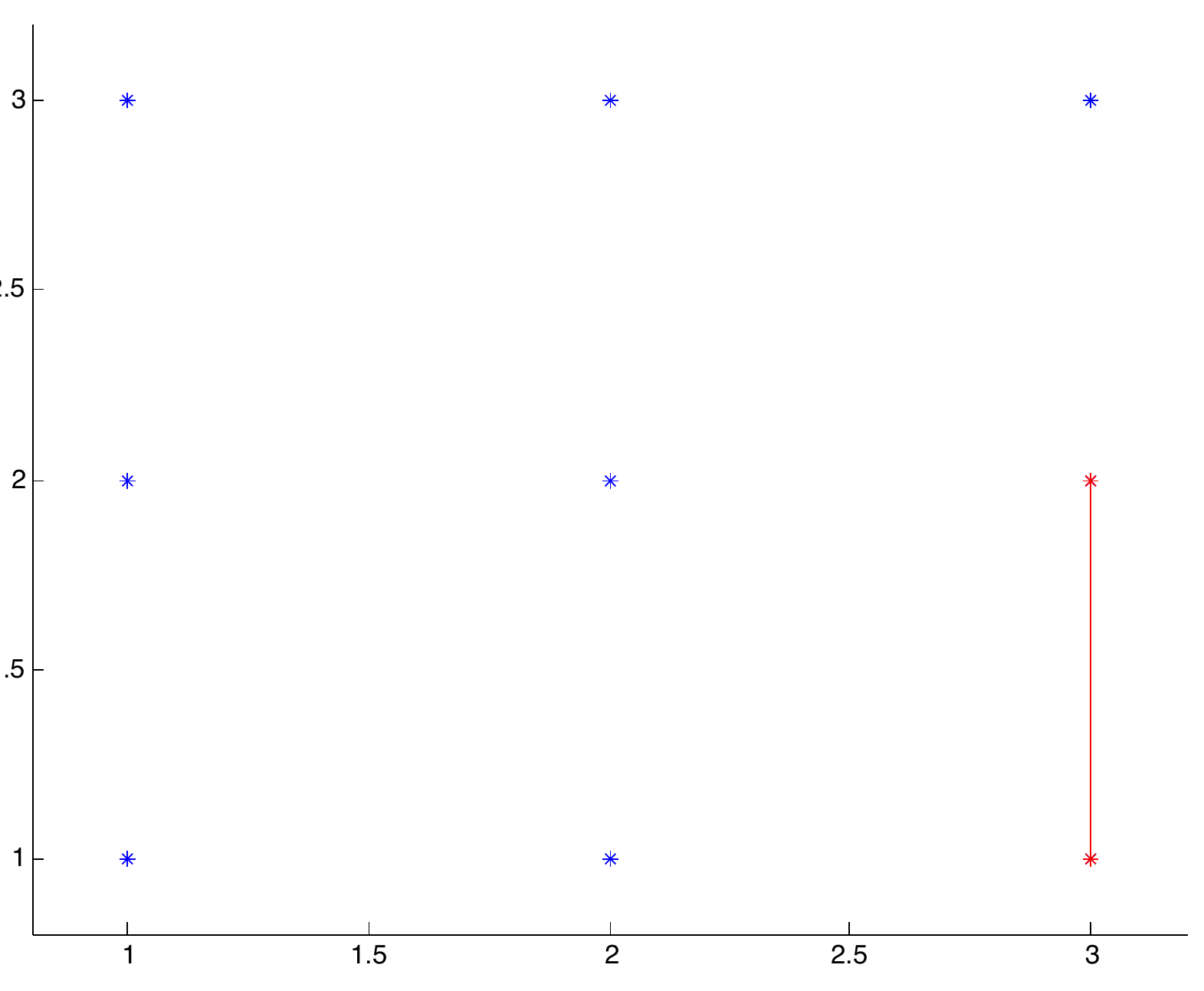}
  \end{center}
  \caption{Four blocks of a  normalized cut for the  graph associated
    with $W_1$ .}
\label{gr2b}
\end{figure}

\medskip
If we apply the method (using method 3 to find the initial $R$)
to the graph associated with  the the matrix $W_1$
shown in Figure \ref{gr1b} for $K = 4$ clusters, 
the algorithm converges in $3$ steps and
we find the clusters
shown in Figure \ref{gr2b}.

The solution $Z$ of the relaxed
problem is
\[
Z =
\begin{pmatrix}
-21.3146  & -0.0000  & 19.4684 & -15.4303 \\
   -4.1289 &   0.0000  & 16.7503  & -15.4303 \\
  -21.3146 &  32.7327 & -19.4684 & -15.4303\\
   -4.1289 &  -0.0000 &  16.7503 & -15.4303\\
   19.7150 &   0.0000  &  9.3547 &  -15.4303\\
   -4.1289 &  23.1455 & -16.7503 &  -15.4303\\
  -21.3146 &  -32.7327 & -19.4684 & -15.4303\\
   -4.1289 & -23.1455 & -16.7503 & -15.4303\\
   19.7150 &  -0.0000 &  -9.3547 & -15.4303
\end{pmatrix} .
\]

We find the following sequence for $Q, Z*Q, X$:
\[
Q =
\begin{pmatrix}
         0  &  0.6109 &  -0.3446 &  -0.7128\\
   -1.0000  &  0.0000  &  0.0000  & -0.0000\\
    0.0000 &   0.5724  &  0.8142 &   0.0969\\
   -0.0000 &   0.5470  & -0.4672 &   0.6947
\end{pmatrix},
\]
which is the initial $Q$ obtained by method 1;
\[
Z*Q =
\begin{pmatrix}
  0.0000 & -10.3162 &  30.4065  &  6.3600\\
    0.0000 &  -1.3742  & 22.2703 &  -6.1531\\
  -32.7327 & -32.6044 &  -1.2967 &   2.5884\\
    0.0000 &  -1.3742 &  22.2703 &  -6.1531\\
    0.0000 &   8.9576  &  8.0309 & -23.8653\\
  -23.1455 & -20.5505  & -5.0065 &  -9.3982\\
   32.7327 & -32.6044 &  -1.2967  &  2.5884\\
   23.1455 & -20.5505 &  -5.0065  & -9.3982\\
   -0.0000 &  -1.7520 &  -7.2027 & -25.6776
\end{pmatrix}
\quad
X = 
\begin{pmatrix}
     0   &  0 &    1 &    0\\
     0   &  0 &    1 &    0\\
     0   &  0 &    0 &    1\\
     0   &  0 &    1  &   0\\
     0   &  1 &    0  &   0\\
     0   &  0 &    1  &   0\\
     1   &  0 &    0  &   0\\
     1   &  0 &    0  &   0\\
     1   &  0 &    0  &   0
\end{pmatrix};
\]

\[
Q =
\begin{pmatrix}
-0.0803  &  0.8633 &  -0.4518  & -0.2102\\
   -0.6485 &   0.1929  &  0.1482 &   0.7213\\
   -0.5424  &  0.0876 &   0.5546 &  -0.6250\\
   -0.5281 &  -0.4581  & -0.6829 &  -0.2119
\end{pmatrix}
\]
\[
Z*Q =
\begin{pmatrix}
-0.6994  & -9.6267 &  30.9638  & -4.4169\\
   -0.6051  &  4.9713 &  21.6922 &  -6.3311\\
   -0.8081 &  -6.7218 &  14.2223  & 43.5287\\
   -0.6051  &  4.9713  & 21.6922  & -6.3311\\
    1.4913  & 24.9075 &   6.8186 &  -6.7218\\
    2.5548  &  6.5028 &   6.5445 &  31.3015\\
   41.6456  & -19.3507  &  4.5190  & -3.6915\\
   32.5742 &  -2.4272 &  -0.3168  & -2.0882\\
   11.6387 &  23.2692 &  -3.5570 &   4.9716
\end{pmatrix}
\quad
X =
\begin{pmatrix}
     0  &   0  &   1  &   0\\
     0  &   0  &   1  &   0\\
     0  &   0  &   0  &   1\\
     0  &   0  &   1  &   0\\
     0  &   1  &   0  &   0\\
     0  &   0  &   0  &   1\\
     1  &   0  &   0  &   0\\
     1  &   0  &   0  &   0\\
     0  &   1  &   0  &   0
\end{pmatrix};
\]
\[
Q =
\begin{pmatrix}
-0.3201 &   0.7992  & -0.3953 &  -0.3201\\
   -0.7071  & -0.0000 &   0.0000  &  0.7071\\
   -0.4914 &  -0.0385 &   0.7181 &  -0.4914\\
   -0.3951  & -0.5998  & -0.5728  & -0.3951
\end{pmatrix}
\]
\[
Z*Q =
\begin{pmatrix}
 3.3532 &  -8.5296  & 31.2440  &  3.3532\\
   -0.8129  &   5.3103 &  22.4987 &  -0.8129\\
   -0.6599  & -7.0310  &  3.2844 &  45.6311\\
   -0.8129 &   5.3103 &  22.4987 &  -0.8129\\
   -4.8123 &  24.6517 &   7.7629 &  -4.8123\\
   -0.7181  &  6.5997 &  -1.5571 &  32.0146\\
   45.6311  & -7.0310 &   3.2844 &  -0.6599\\
   32.0146 &   6.5997  & -1.5571 &  -0.7181\\
    4.3810 &  25.3718  & -5.6719 &   4.3810
\end{pmatrix}
\quad
X=
\begin{pmatrix}
     0   &  0  &   1  &   0\\
     0   &  0  &   1  &   0\\
     0   &  0  &   0  &   1\\
     0   &  0  &   1  &   0\\
     0   &  1  &   0  &   0\\
     0   &  0  &   0  &   1\\
     1   &  0  &   0  &   0\\
     1   &  0  &   0  &   0\\
     0   &  1  &   0  &   0
\end{pmatrix} .
\]
During the next round, the exact same matrices are obtained and the
algorithm stops.
Comparison of the matrices $Z*Q$ and $X$ makes it clear that $X$ is
obtained from $Z*Q$ by retaining on every row the leftmost largest
value and setting the others to $0$ (non-maximum supression).

\begin{figure}[http]
  \begin{center}
 \includegraphics[height=1.1truein,width=1.1truein]{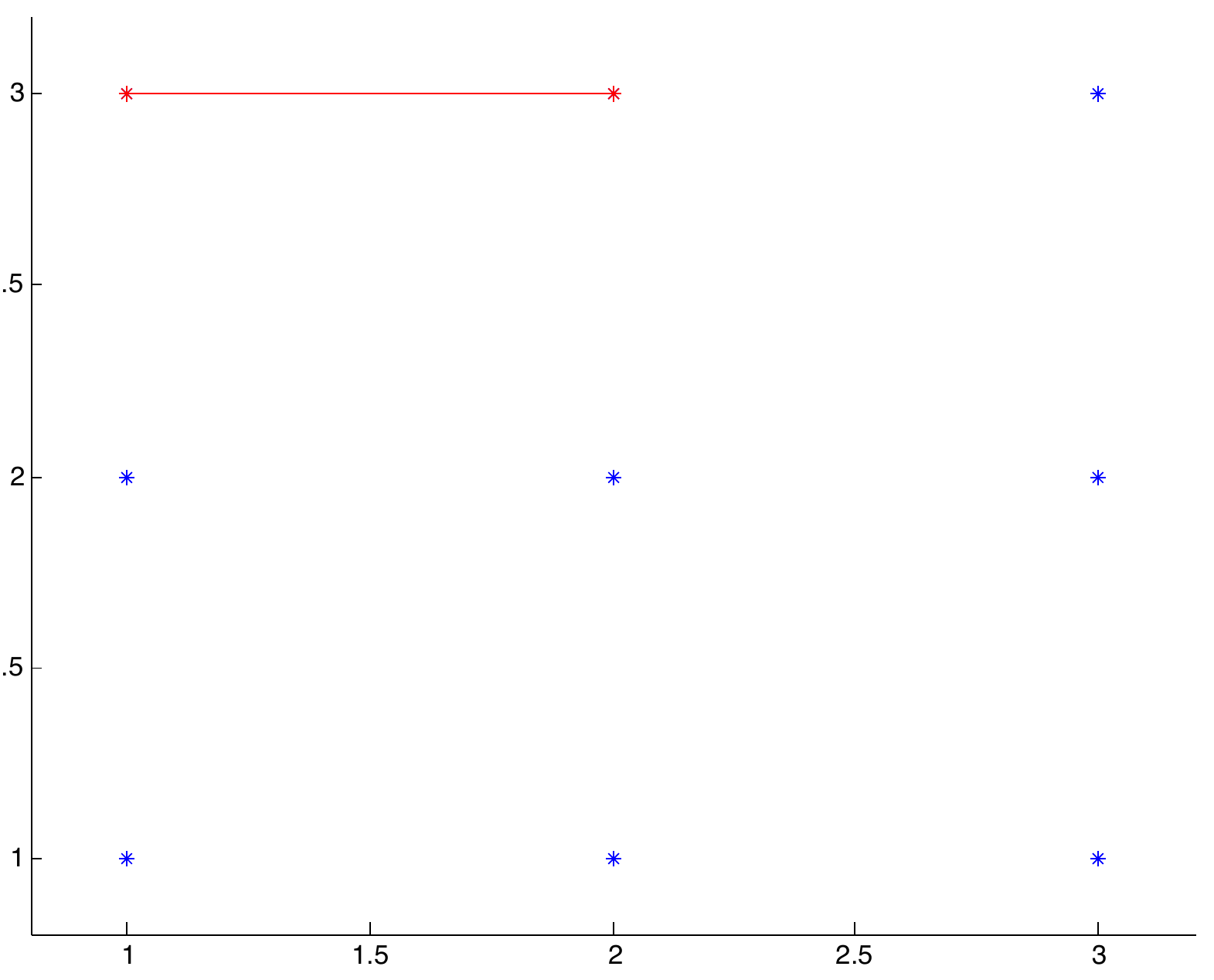}
\hspace{0.25cm}
 \includegraphics[height=1.1truein,width=1.1truein]{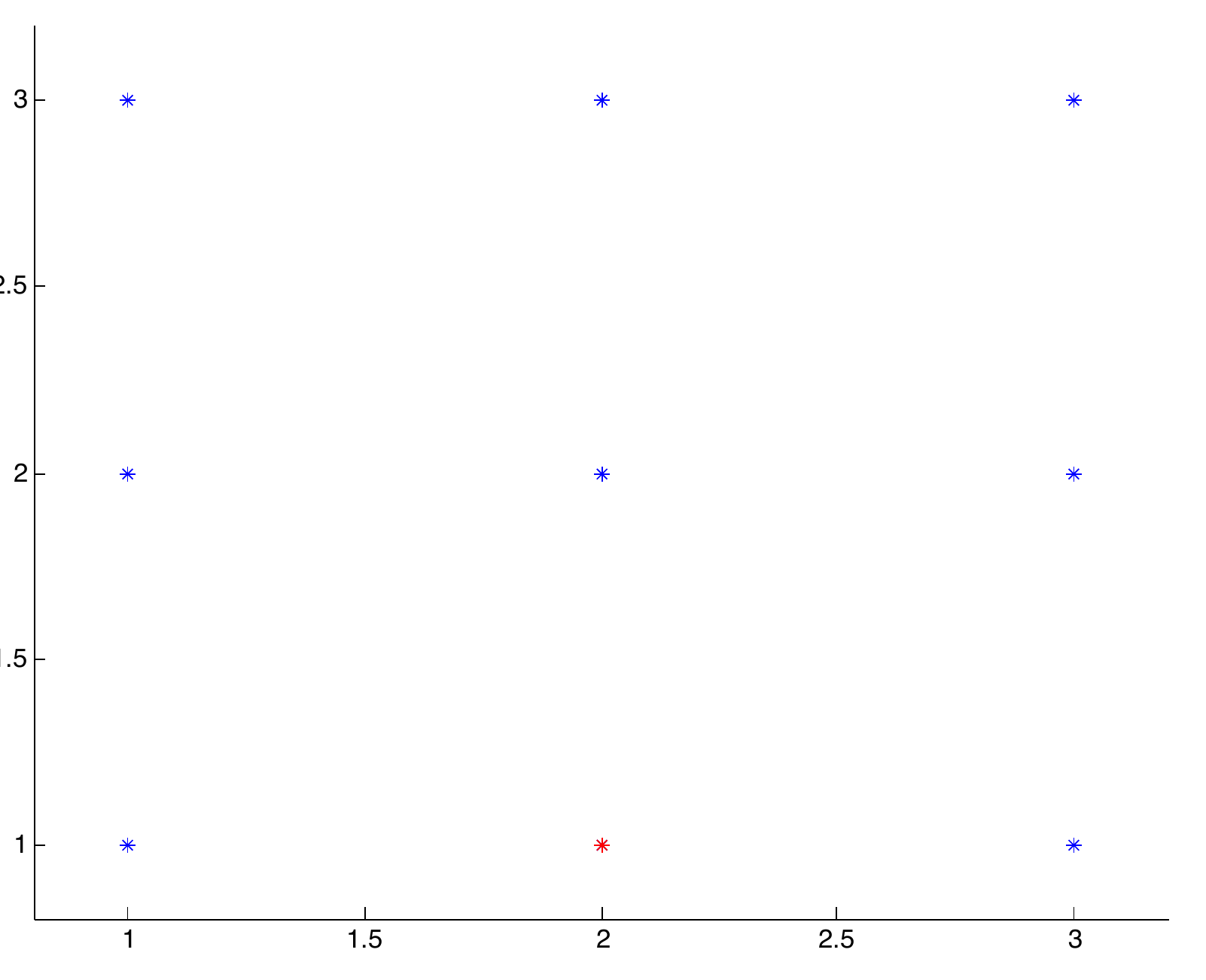}
\hspace{0.25cm}
 \includegraphics[height=1.1truein,width=1.1truein]{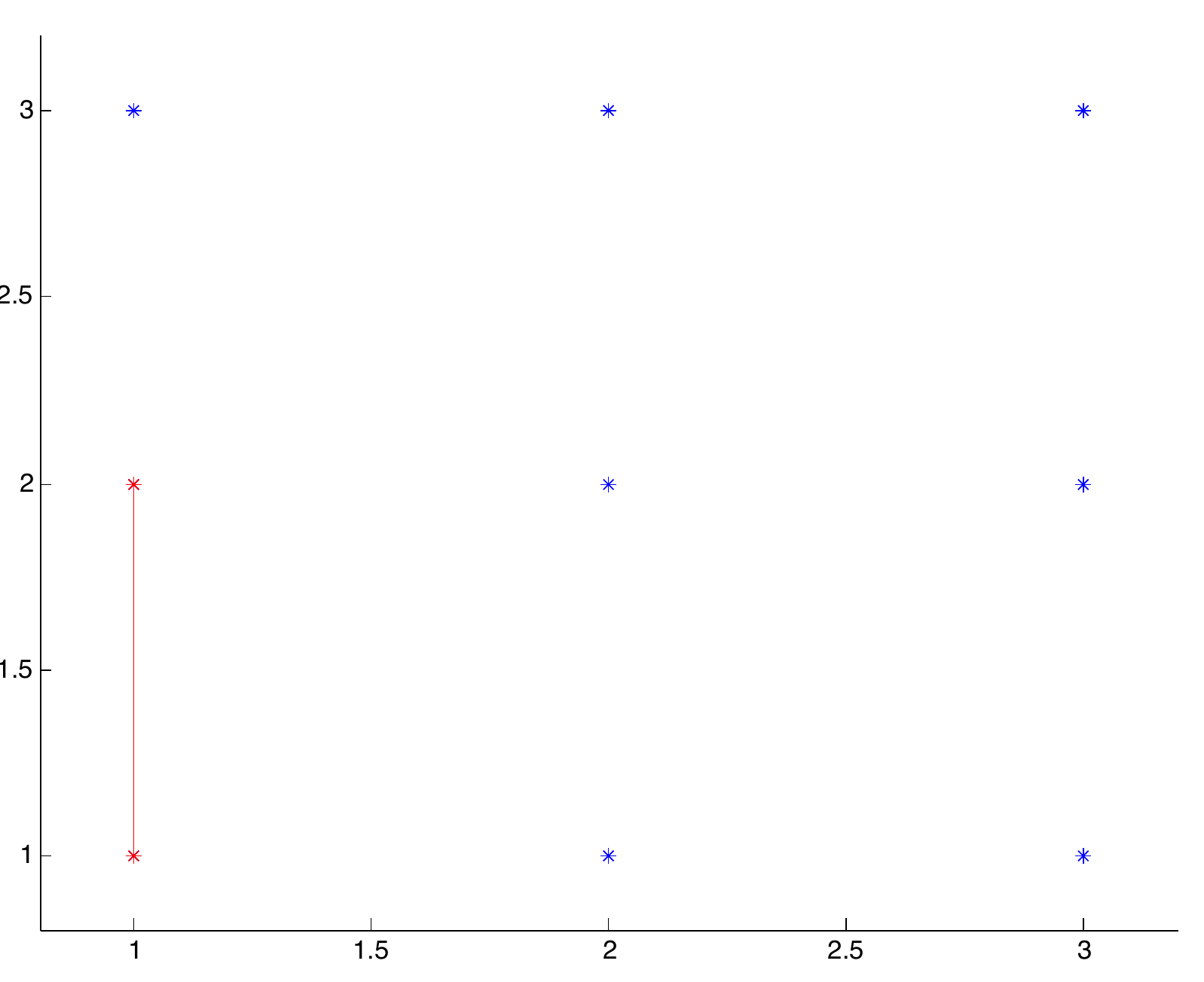}
\hspace{0.25cm}
 \includegraphics[height=1.1truein,width=1.1truein]{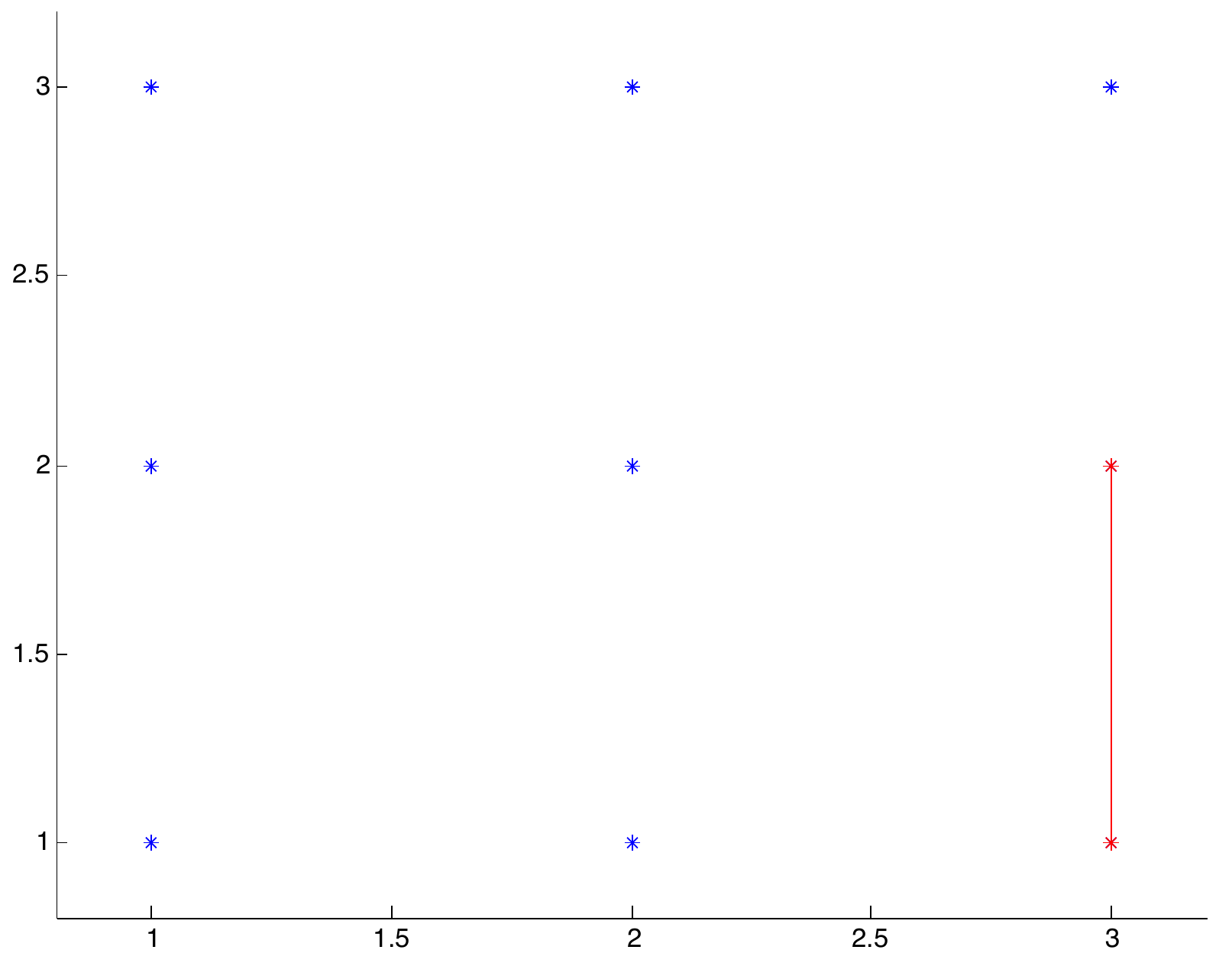}
\hspace{0.25cm}
 \includegraphics[height=1.1truein,width=1.1truein]{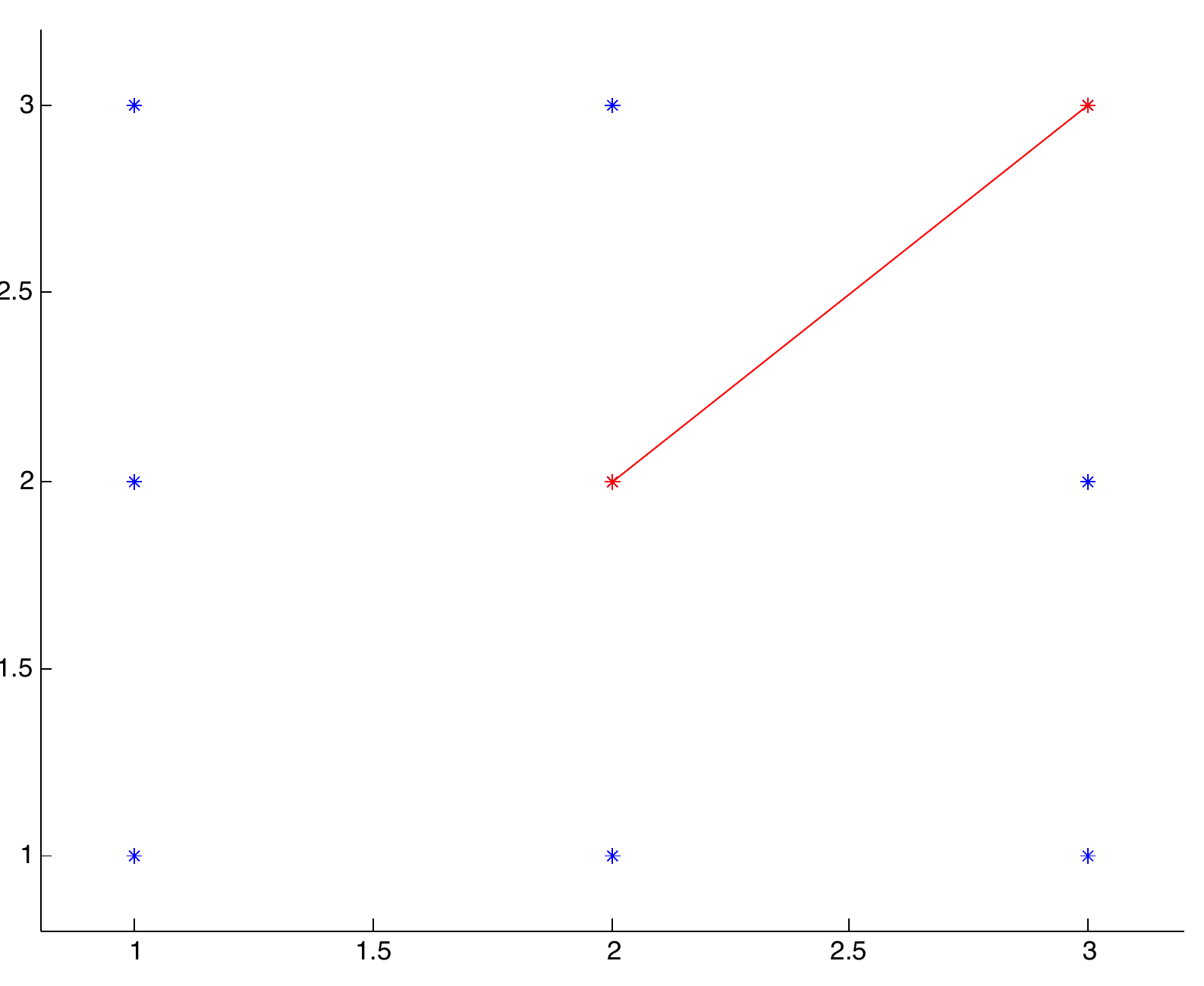}
  \end{center}
  \caption{Five blocks of a  normalized cut for the  graph associated
    with $W_1$ .}
\label{gr2c}
\end{figure}

\medskip
In this example, the columns of all $X$ were nonzero, but this may happen,
for example when we apply the algorithm to the graph of Figure \ref{gr1b}
to find $K = 5$ clusters shown in Figure \ref{gr2c}.

We find that the initial value for $Z*Q$ is

\[
Z*Q =
\begin{pmatrix}
-5.7716  & -27.5934 &   0.0000  & -9.3618  & -0.0000 \\
    5.5839 & -20.2099  & -29.7044  & -1.2471 &  -0.0000\\
   -2.3489 &   1.1767  & -0.0000 & -29.5880  & -29.7044\\
    5.5839 & -20.2099 &  29.7044  & -1.2471 &   0.0000\\
   21.6574 &  -7.2879 &   0.0000 &   8.1289  &  0.0000\\
    8.5287  &  4.5433 &  -0.0000 & -18.6493 & -21.0042\\
   -2.3489 &   1.1767 &  -0.0000 & -29.5880 &  29.7044\\
    8.5287 &   4.5433  & -0.0000 & -18.6493 &  21.0042\\
   23.3020 &   6.5363 &  -0.0000 &  -1.5900 &  -0.0000
\end{pmatrix}.
\]
The matrix $X1$ given by the above method in which we pick the
leftmost largest entry  on every row has a 
fourth row  equal to $0$.  The matrix $X1$ is repaired by migrating a $1$
from the second entry of the first column, which contains the largest
number of $1$'s, yielding the matrix $X2$; see below.

\[
X1 =
\begin{pmatrix}
0  &     0 &    1 &    0 &    0\\
   {\red  1}  &   0  &   0  &   0  &   0\\
     0  &   1  &   0  &   0  &   0\\
     0  &   0  &   1  &   0  &   0\\
     1  &   0  &   0  &   0  &   0\\
     1  &   0  &  0   &  0   &  0\\
     0  &   0  &   0  &   0  &   1\\
     0  &   0  &   0  &   0  &   1\\
     1  &   0  &   0  &   0  &   0
\end{pmatrix}
\qquad
X2 =
\begin{pmatrix}
0  &    0  &   1  &   0  &   0\\
     0  &   0  &   0  &  {\blue 1}  &   0\\
     0   &  1  &   0  &   0  &   0\\
     0   &  0  &   1  &   0  &   0\\
     1   &  0  &   0  &   0  &   0\\
     1   &  0  &   0  &   0  &   0\\
     0   &  0  &   0  &   0  &   1\\
     0   &  0  &   0  &   0  &   1\\
     1   &  0  &   0  &   0  &   0
\end{pmatrix}
\]

\chapter{Signed Graphs}
\label{chap4}
\section{Signed Graphs and Signed  Laplacians}
\label{ch4-sec1}
Intuitively, in a weighted graph, an edge with  a positive weight denotes similarity or
proximity of its endpoints. For many reasons, it is desirable  to allow
edges labeled with negative weights, the intuition being  that a
negative weight indicates dissimilarity or distance. 

\medskip
Weighted graphs
for which the weight matrix is a symmetric matrix in which negative
and positive entries are allowed are called {\it signed graphs\/}.
Such graphs (with weights $(-1, 0, +1)$) were introduced as early as
1953 by Harary \cite{Harary53}, to model social relations involving disliking,
indifference, and liking.  The problem of clustering the nodes
of a signed graph arises naturally as a generalization of the
clustering problem for weighted graphs. From our perspective, 
we would like to know whether clustering  using normalized cuts can be
extended to signed graphs. 

\medskip
Given a signed graph $G = (V, W)$ (where $W$ is a symmetric matrix
with zero diagonal entries), the {\it underlying graph\/} of $G$ is
the graph with node set $V$ and set of (undirected) edges
$E = \{\{v_i, v_j\} \mid w_{i j} \not= 0\}$.

\medskip
The first obstacle is that the degree matrix may now contain zero or negative
entries. As a consequence, the Laplacian $L$  may no longer be positive
semidefinite, and worse, $\Degsym^{-1/2}$ may not exist.

\medskip
A simple remedy is to use the absolute values of the  weights in the degree
matrix!  This idea applied to signed graph with weights $(-1, 0, 1)$ occurs in Hou \cite{Hou}. 
Kolluri,  Shewchuk and  O'Brien \cite{Kolluri:2004:SSR} take the 
natural step of  using absolute values of weights in
the degree matrix in their original work on surface reconstruction
from noisy point clouds.  Given a Delaunay tetrahedralization, they
build a graph with positive and negative edges and use the normalized
cut method for two clusters to decide which tetrahedra are inside or
outside the original object. The triangulated surface (called the {\it eigencrust\/})
consists of the triangles where an inside and an outside tetrahedron meet. 
The authors  state  that the Lapacians arising from
such graphs   are always positive definite, which is not quite correct
since this is only true for unbalanced graphs (see Section \ref{ch4-sec3}). 
Kunegis et al. \cite{kunegis:spectral} appear to be the first to make
a systematic study of spectral methods applied to signed graphs. 
In fact, many results in this section originate from
Kunegis et al.  \cite{kunegis:spectral}.  However, it should be noted that
only $2$-clustering is considered in the above papers.

\medskip
As we will see, the trick of using absolute values of weights in the
degree matrix
allows the whole machinery that we
have presented to be used to attack the problem of clustering signed
graphs using normalized cuts.  This requires a modification of
the notion of normalized cut. This new notion it is quite reasonable,
as we will see shortly.

\medskip
If $(V, W)$ is a signed graph, where $W$ is an $m\times m$ symmetric
matrix with zero diagonal entries and with the other entries
$w_{i j}\in \reals$ arbitrary, for any node $v_i \in V$, the {\it signed degree\/} of $v_i$ is defined as
\[
\overline{d}_i = \overline{d}(v_i) = \sum_{j = 1}^m |w_{i j}|,
\]
and the {\it signed degree matrix \/} $\overline{\Degsym}$ as
\[
\overline{\Degsym} = \mathrm{diag}(\overline{d}(v_1) , \ldots, \overline{d}(v_m)).
\]
For any subset $A$ of the set of nodes
$V$, let
\[
\mathrm{vol}(A) = \sum_{v_i\in A} \overline{d}_i = 
 \sum_{v_i\in A} \sum_{j = 1}^m |w_{i j}|.
\]
For any two subsets $A$ and $B$ of $V$, 
define $\mathrm{links}^+(A,B)$, 
$\mathrm{links}^-(A,B)$, and $\mathrm{cut}(A,\overline{A})$ by
\begin{align*}
\mathrm{links}^+(A, B) & = 
\sum_{\begin{subarray}{c}
v_i \in A, v_j\in B \\
w_{i j} > 0
\end{subarray}
}
w_{i j}  \\
\mathrm{links}^-(A,B) & = 
\sum_{\begin{subarray}{c}
v_i \in A, v_j\in B \\
w_{i j} < 0
\end{subarray}
}
- w_{i j} \\
\mathrm{cut}(A,\overline{A}) & = 
\sum_{\begin{subarray}{c}
v_i \in A, v_j\in \overline{A} \\
w_{i j} \not=  0
\end{subarray}
}
|w_{i j}| .
\end{align*}
Note that
$\mathrm{links}^+(A, B) = \mathrm{links}^+(B, A)$,  $\mathrm{links}^-(A,
B) = \mathrm{links}^-(B, A)$,  and 
\[
\mathrm{cut}(A,\overline{A})  =  \mathrm{links}^+(A,\overline{A}) + \mathrm{links}^-(A,\overline{A}) .
\]
Then, the {\it signed Laplacian\/} $\overline{L}$ is defined by
\[
\overline{L} = \overline{\Degsym} - W, 
\]
and its normalized version $\overline{L}_{\mathrm{sym}}$ by
\[
\overline{L}_{\mathrm{sym}} =  \overline{\Degsym}^{-1/2}\, \overline{L}\,
\overline{\Degsym}^{-1/2}
= I - \overline{\Degsym}^{-1/2} W \overline{\Degsym}^{-1/2}.  
\]
For a graph without isolated vertices, we have $\overline{d}(v_i) > 0$
for $i = 1, \ldots, m$, so $\overline{\Degsym}^{-1/2}$ is well defined.

\medskip
The signed Laplacian is symmetric positive semidefinite. As for the
Laplacian of a weight matrix (with nonnegative entries), this can be shown
in two ways. The first method consists in defining a notion of
incidence matrix for a signed graph, and appears in Hou \cite{Hou}. 

\begin{definition}
\label{incidence-matrix-sw}
Given a signed graph $G = (V, W)$, with $V = \{v_1, \ldots, v_m\}$, 
if $\{e_1, \ldots, e_n\}$ are the edges of the underlying graph of $G$
(recall that $\{v_i, v_j\}$ is an edge of this graph iff $w_{i j}
\not= 0$), 
for any oriented graph $G^{\sigma}$ obtained by giving an
orientation to 
the underlying graph of $G$,
the {\it incidence matrix\/} $B^{\sigma}$ of
$G^{\sigma}$ is the $m\times n$
matrix whose entries $b_{i\, j}$ are given by
\[
b_{i\, j} = 
\begin{cases}
+\sqrt{w_{i j}} & \text{if $w_{i j} > 0$ and  $s(e_j) = v_i$}\\
-\sqrt{w_{i j}} & \text{if $w_{i j} > 0$ and $t(e_j) = v_i$} \\
\sqrt{-w_{i j}} & \text{if $w_{i j} < 0$ and  $(s(e_j) = v_i$ or  $t(e_j) = v_i)$} \\
0 & \text{otherwise}.
\end{cases}
\]
\end{definition}

Then, we have the  following proposition whose proof is easily adapted
from the proof of Proposition \ref{adjp2}.

\begin{proposition}
\label{adjp2ws}
Given any signed   graph $G = (V, W)$  with $V = \{v_1, \ldots, v_m\}$, 
if $B^{\sigma}$ is the incidence
matrix of any oriented graph $G^{\sigma}$ obtained from the underlying
graph of $G$ and $\overline{\Degsym}$ is the signed degree matrix of $W$,
then
\[
B^{\sigma} \transpos{(B^{\sigma})}= \overline{\Degsym} - W = \overline{L}.
\]
Consequently, $B^{\sigma}\transpos{(B^{\sigma})}$ is 
independent of the orientation of the underlying graph of  $G$ and
$\overline{L} = \overline{\Degsym} -W$ is symmetric
and positive semidefinite; that is, the eigenvalues of
$\overline{L} = \overline{\Degsym} - W$ are real and nonnegative.
\end{proposition}

\medskip
Another way to prove that $\overline{L}$ is positive semidefinite
is to evaluate the
quadratic form $\transpos{x} \overline{L} x$. We will need this  computation
to figure out what is the new notion of normalized cut.
For any real $\lambda\in \reals$, define $\mathrm{sgn}(\lambda)$ by
\[
\mathrm{sgn}(\lambda) =
\begin{cases}
+1 & \text{if $\lambda  > 0$} \\
 -1 & \text{if $\lambda  < 0$}  \\
 0  & \text{if $\lambda  = 0$} .
\end{cases}
\]

\begin{proposition}
\label{Laplace1s}
For any  $m\times m$ symmetric matrix $W = (w_{i j})$, if we let $\overline{L} = \overline{\Degsym} - W$
where $\overline{\Degsym}$ is the signed degree matrix associated with $W$,
then  we have
\[
\transpos{x} \overline{L} x =
\frac{1}{2}\sum_{i, j = 1}^m |w_{i  j}| (x_i - \mathrm{sgn}(w_{i j}) x_j)^2
\quad\mathrm{for\ all}\> x\in \reals^m.
\]
Consequently, $\overline{L}$  is positive semidefinite.
\end{proposition}
\begin{proof}
We have
\begin{align*}
\transpos{x}\overline{L} x & = \transpos{x} \overline{\Degsym} x - \transpos{x} W x \\
& = \sum_{i = 1}^m \overline{d}_i x_i^2 - \sum_{i, j = 1}^m w_{i  j} x_i x_j \\
& = \sum_{i, j = 1}^m ( |w_{i j}| x_i^2  - w_{i  j} x_i x_j) \\
& = \sum_{i, j = 1}^m  (|w_{i j}|( x_i^2 - \mathrm{sgn}(w_{i  j}) x_i x_j) \\
& = \frac{1}{2}\left( \sum_{i,  j = 1}^m |w_{i j}|( x_i^2 - 2 \mathrm{sgn}(w_{i  j}) x_i x_j
+  x_j^2)
\right) \\
& = 
\frac{1}{2}\sum_{i, j = 1}^m |w_{i  j}| (x_i -  \mathrm{sgn}(w_{i j})x_j)^2,
\end{align*}
and  this quantity is nonnegative.
\end{proof}

\section{Signed Normalized Cuts}
\label{ch4-sec2}
As in Section \ref{ch3-sec3}, given a partition of $V$ into $K$
clusters $(A_1, \ldots, A_K)$, if we represent the $j$th block of
this partition by a vector $X^j$ such that
\[
X^j_i = 
\begin{cases}
a_j & \text{if $v_i \in A_j$} \\
0 &  \text{if $v_i \notin A_j$} ,
\end{cases}
\]
for some $a_j \not= 0$, then we have the following result.

\begin{proposition}
\label{scut}
For any vector $X^j$ representing the $j$th block of a partition 
$(A_1, \ldots, A_K)$ of $V$, we have
\[
\transpos{(X^j)} \overline{L} X^j =
a_j^2(\mathrm{cut}(A_j, \overline{A_j}) + 2 \mathrm{links}^-(A_j, A_j)).
\]
\end{proposition}

\begin{proof}
Using Proposition  \ref{Laplace1s}, we have
\[
\transpos{(X^j)} \overline{L} X^j =
\frac{1}{2}\sum_{i, k = 1}^m |w_{i k}| (X^j_i - \mathrm{sgn}(w_{i k}) X^j_k)^2.
\]
The sum on the righthand side splits into four parts:
\begin{enumerate}
\item[(1)]
$S_1 = \frac{1}{2}\sum_{i, k  \in A_j}|w_{i k}| (X^j_i - \mathrm{sgn}(w_{i k}) X^j_k)^2$.
In this case, $X^j_i = X^j_k = a_j$, so only negative edges have a
nonzero contribution, and we have
\[
S_1 = \frac{1}{2}\sum_{i, k  \in A_j, w_{i k } < 0 }|w_{i k}|(a_j + a_j)^2
=  2a_j^2 \mathrm{links}^-(A_j, A_j).
\]
\item[(2)]
$S_2 = \frac{1}{2}\sum_{i \in A_j, k\in \overline{A_j}}|w_{i k}| (X^j_i - \mathrm{sgn}(w_{i k}) X^j_k)^2$.
In this case, $X^j_i = a_j$ and $X^j_k = 0$, so
\[
S_2  = \frac{1}{2} a_j^2\sum_{i \in A_j, k\in \overline{A_j}}|w_{i  k}| 
=   \frac{1}{2} a_j^2 \mathrm{cut}(A_j, \overline{A_j}).
\]
\item[(3)]
$S_3 = \frac{1}{2}\sum_{i \in \overline{A_j}, k\in A_j}|w_{i k}| (X^j_i - \mathrm{sgn}(w_{i k}) X^j_k)^2$.
In this case, $X^j_i = 0$ and $X^j_k = a_j$, so
\[
S_3  = \frac{1}{2} a_j^2\sum_{i \in \overline{A_j}, k \in A_j}|w_{i  k}| 
=   \frac{1}{2} a_j^2 \mathrm{cut}(\overline{A_j}, A_j)
=  \frac{1}{2} a_j^2 \mathrm{cut}(A_j, \overline{A_j}).
\]
\item[(4)]
\end{enumerate}
$S_4 = \frac{1}{2}\sum_{i, k  \in \overline{A_j}}|w_{i k}| (X^j_i - \mathrm{sgn}(w_{i k}) X^j_k)^2$.
In this case, $X^j_i = X^j_k = 0$, so 
\[
S_4 = 0.
\]
In summary,
\[
\transpos{(X^j)} \overline{L} X^j = S_1 + S_2 + S_3 + S_4 =
 2a_j^2 \mathrm{links}^-(A_j, A_j) + a_j^2
 \mathrm{cut}(A_j, \overline{A_j}),
\]
as claimed.
\end{proof}

Since with the revised definition of $\mathrm{vol}(A_j)$,   we also have
\[
\transpos{(X^j)} \overline{\Degsym} X^j = a_j^2\sum_{v_i\in A_j}
\overline{d}_i = a_j^2 \mathrm{vol}(A_j), 
\]
we deduce that
\[
\frac{\transpos{(X^j)} \overline{L} X^j} 
  {\transpos{(X^j)} \overline{\Degsym} X^j}
= \frac{\mathrm{cut}(A_j, \overline{A_j}) + 2 \mathrm{links}^-(A_j, A_j)}{\mathrm{vol}(A_j)}.
\]

The calculations of the previous paragraph suggest  the following definition.

\begin{definition}
\label{sncut}
The {\it signed normalized cut\/}
$\mathrm{sNcut}(A_1, \ldots, A_K)$ of the
partition $(A_1, \ldots, A_K)$ is defined as
\[
\mathrm{sNcut}(A_1, \ldots, A_K) = \sum_{j = 1}^K
\frac{\mathrm{cut}(A_j, \overline{A_j})}{\mathrm{vol}(A_j)} + 
2 \sum_{j = 1}^K\frac{\mathrm{links}^-(A_j, A_j)}{\mathrm{vol}(A_j)}.
\] 
\end{definition}

\remark
Kunegis et al.  \cite{kunegis:spectral} deal with a different notion
of cut, namely ratio cut (in which $\mathrm{vol}(A)$ is replaced by
the size $|A|$ of $A$),
and only for two clusters. In this case, 
by a clever choice of indicator vector, they obtain a notion of signed
cut that only takes into account the positive edges between 
$A$ and $\overline{A}$, and  the negative edges among nodes in $A$
and nodes in $\overline{A}$.  This trick does not seem to generalize
to more than two clusters, and this is why we use our representation
for partitions. Our definition of a signed normalized cut appears to
be novel.

\medskip
Based on previous computations, we have
\[
\mathrm{sNcut}(A_1, \ldots, A_K) = 
\sum_{j = 1}^K \frac{\transpos{(X^j)} \overline{L} X^j} 
  {\transpos{(X^j)} \overline{\Degsym} X^j}.
\]
where $X$ is the $N\times K$ matrix whose $j$th column is $X^j$.
Therefore, this is the same problem as in Chapter \ref{chap3}, with
$L$ replaced by $\overline{L}$ and $\Degsym$ replaced by
$\overline{\Degsym}$.

\medskip
Observe that minimizing $\mathrm{sNcut}(A_1, \ldots, A_K)$ amounts to 
minimizing the number of positive and negative edges between clusters,
and also minimizing the number of negative edges within clusters.
This second minimization captures the intuition that nodes connected
by a negative edge should not be together  (they do not ``like''
each other; they should be far from each other). 

\medskip
The $K$-clustering problem for signed graphs is related but not equivalent
to another problem known as {\it correlation clustering\/}.
In correlation clustering, in our terminology and notation,
given a graph $G = (V,W)$ with positively and
negatively weighted  edges, one seeks a clustering of $V$ that minimizes the
sum $\mathrm{links}^-(A_j,A_j)$ of the absolute values of the negative
weights of  the edges 
within each cluster $A_j$,  
and minimizes the sum $\mathrm{links}^+(A_j,\overline{A}_j)$ 
of the positive weights of the
edges between distinct clusters. In contrast to $K$-clustering,
the number $K$ of clusters is not given in advance, and
there is no normalization with respect to size of volume.
Furthermore, in correlation clustering, 
only the contribution  $\mathrm{links}^+(A_j,\overline{A}_j)$ of positively
weighted edges is minimized, but our method only allows us to
minimize  $\mathrm{cut}(A_j,\overline{A}_j)$, which also takes into account
negatively weighted edges between distinct clusters.
Correlation clustering was first introduced and studied 
for complete graphs by Bansal, Blum and Chawla \cite{Bansal}.
They prove that this problem is NP-complete and give several
approximation algorithms, including a PTAS for maximizing agreement.
Demaine and Immorlica \cite{demaine} 
consider the same problem for arbitrary weighted graphs, and they give
an $O(\log n)$-approximation algorithm based on linear programming.
Since correlation clustering does not assume that $K$ is given and not 
not include nomalization by size or volume, it is
not clear whether algorithms for correlation clustering can be applied to
normalized $K$-clustering, and conversely.

\section{Balanced  Graphs}
\label{ch4-sec3}
Since 
\[
\mathrm{sNcut}(A_1, \ldots, A_K) = 
\sum_{j = 1}^K \frac{\transpos{(X^j)} \overline{L} X^j} 
  {\transpos{(X^j)} \overline{\Degsym} X^j},
\]
the whole machinery of Sections \ref{ch3-sec3} and \ref{ch3-sec5}  can
be applied with $\Degsym$ replaced by $\overline{\Degsym}$
and $L$ replaced by $\overline{L}$. However, there is a new
phenomenon, which is that $\overline{L}$ may be positive definite.
As a consequence, $\mathbf{1}$ is not always an eigenvector of
$\overline{L}$. 
As observed by Kunegis et al.  \cite{kunegis:spectral},
it is also possible to characterize for which signed
graphs the Laplacian $\overline{L}$ is positive definite.
Such graphs are ``cousins'' of bipartite graphs and were introduced
by Harary \cite{Harary53}. 
Since a graph is the union of its connected components, we restrict
ourselves to connected graphs.

\begin{definition}
\label{balance}
Given a signed graph $G = (V, W)$ with negative weights whose
underlying graph is connected, we say that $G$ is {\it balanced\/}
if there is a partition of its set of nodes $V$ into two blocks
$V_1$ and $V_2$ such that all positive edges connect nodes
within $V_1$ or $V_2$, and negative edges connect nodes 
between $V_1$ and $V_2$.
\end{definition}

\begin{figure}[hhtp]
  \begin{center}
 \includegraphics[height=1.9truein,width=2.3truein]{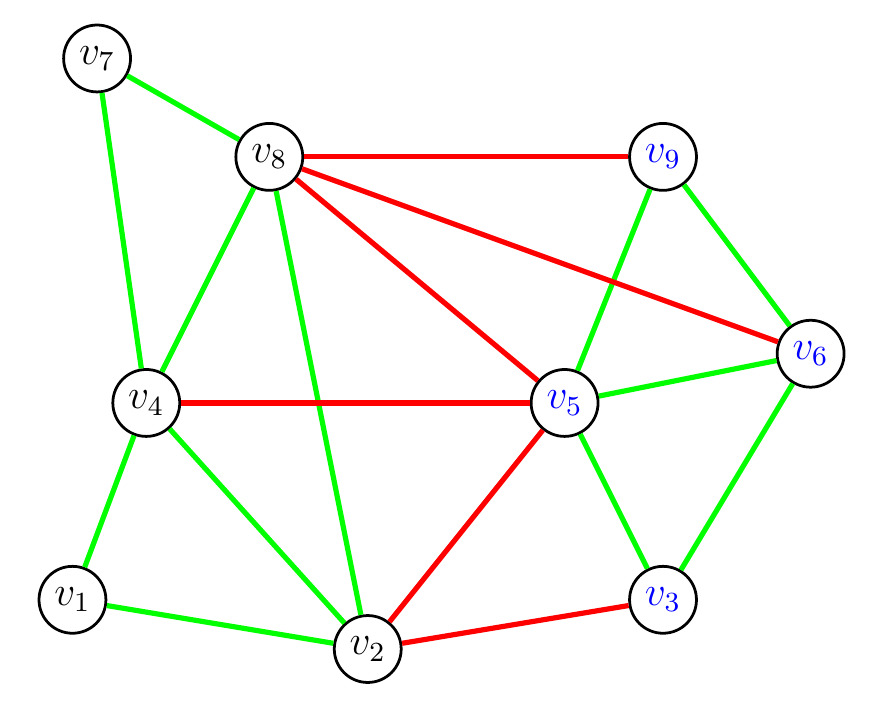}
\hspace{0.25cm}
 \includegraphics[height=1.9truein,width=2.3truein]{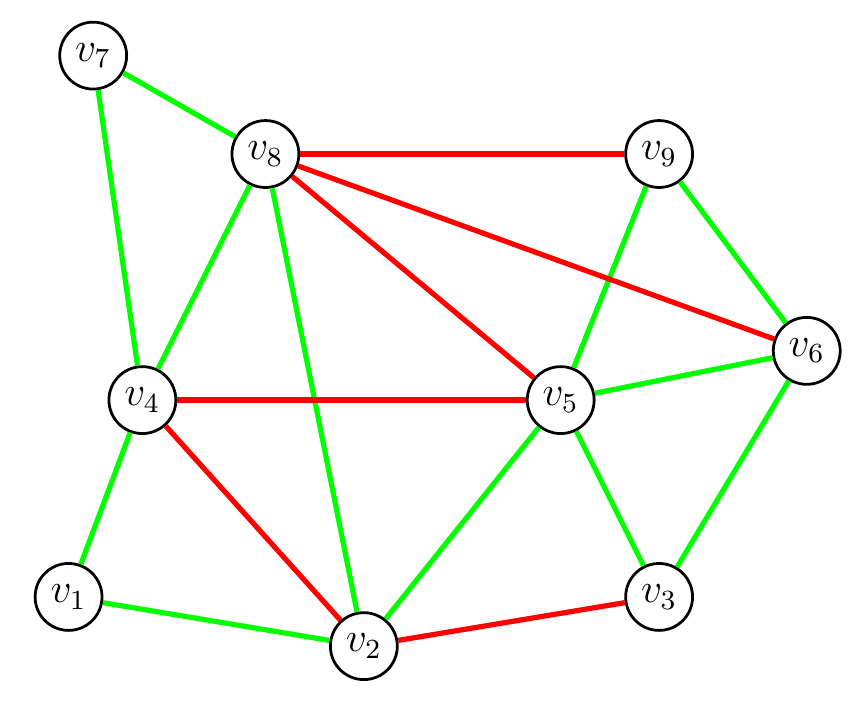}
  \end{center}
  \caption{A balanced signed graph $G_1$ (left). An unbalanced signed graph
    $G_2$ (right).}
\label{sgraphfig1}
\end{figure}

An example of a balanced graph is shown in Figure \ref{sgraphfig1} on
the left, in
which positive edges are colored green and negative edges are colored
red. This graph admits the partition
\[
(\{v_1, v_2, v_4, v_7, v_8\}, \{v_3, v_5, v_6, v_9\}).
\]
On the other hand, the graph shown  on the right
contains  the cycle $(v_2, v_3, v_6, v_5, v_4, v_2)$ with
an odd number of negative edges ($3$), and thus is not balanced.

Observe that if we delete all positive edges in a balanced graph, then the resulting graph
is bipartite. Then, it is not surprising that connected balanced graphs can be
characterized as signed graphs in which every cycle has an even
number of negative edges. This is analogous to the characterization
of a connected bipartite graph as a graph in which every cycle has
even length. The following proposition was first proved by Harary
\cite{Harary53}. We give a more direct proof.

\begin{proposition}
\label{balancep1}
If $G = (V, W)$ is a connected signed graph with negative weights,
then $G$ is balanced iff every cycle contains an even number of
negative edges.
\end{proposition}

\begin{proof}
If $G$ is balanced, then every cycle must switch from a node in $V_1$
to a node in $V_2$ (or from a node in $V_2$ to a node in $V_1$) an
even number of times. Therefore, it contains an even number of
negative edges.

\medskip
Conversely, assume that $G$ contains no cycle with an odd number of
negative edges.  Since $G$ is connected, pick some some $v_0$ in $V$,
and let $V_1$ be the set of node reachable from $v_0$ by a path with
an odd number of negative edges, and let $V_2$ be the set of node reachable from $v_0$ by a path with
an even number of negative edges. Clearly, $(V_1, V_2)$ is a partition
of $V$. Assume that there is a negative edge $\{u, v\}$ between two nodes within
$V_1$ (or $V_2$). Then, using the paths from $v_0$ to $u$ and $v$,
where the parity of the number of negative edges is the same, we would
obtain a cycle with an odd number of negative edges, a contradiction.
Therefore, edges between nodes in $V_1$ (or $V_2$) are positive, and
negative edges connect nodes in $V_1$ and $V_2$.
\end{proof}

\medskip
We can also detect whether a connected signed graph is balanced
in terms of the kernel of the transpose of any of its incidence matrices.

\begin{proposition}
\label{balancep2}
If $G = (V, W)$ is a connected signed graph with negative weights and with
$m$ nodes, for any orientation of its underlying graph, let  $B$ be the
corresponding incidence matrix. 
The underlying graph of  $G$ is balanced iff
$\mathrm{rank}(B) = m - 1$.
Furthermore, if $G$ is balanced, then there is a vector $u$ with
$u_i \in \{-1, 1\}$ such that $\transpos{B} u = 0$, and the
sets of nodes $V_1 = \{v_i \mid u_i = -1\}$ and $V_2 = \{v_i \mid u_i =
  +1\}$ form a partition of $V$ for which $G$ is balanced.
 \end{proposition}

\begin{proof}
Assume that  $\mathrm{rank}(B) =  m-1$; this implies that $\Ker(\transpos{B})
\not= (0)$. For any $u\not= 0$,
we have $\transpos{B} u = 0$ iff $\transpos{u} B = 0$ iff $u$ is
orthogonal to every column of $B$. By definition of $B$, we have
\[
u_i = \mathrm{sgn}(w_{i j}) u_j
\]
iff there is an edge between $v_i$ and $v_j$.

\medskip
Pick node $v_1$ in $V$ and define $V_1$
and $V_2$ as in the proof of Proposition \ref{balancep1}.
The above equation shows that $u$ has the same value on nodes
connected by a path with an even number of negative edges, and
opposite values on nodes connected by a path with an odd number of
negative edges. Since $V_1$ consists of all nodes connected to $v_1$
by a path with an odd number of negative edges and $V_2$
consists of all nodes connected to $v_1$
by a path with an even number of negative edges,
it follows that $u$ has the same value  $c = u_1$ on all nodes in
$V_1$, and the value $-c$ on all nodes in  $V_2$. Then, there is no
negative edge between any two nodes in $V_1$ (or $V_2$), 
since otherwise $u$ would take opposite values on theses two nodes, contrary to
the fact that $u$ has a constant value on $V_1$ (and $V_2$).
This implies that 
$(V_1, V_2)$ is a partition of $V$ making $G$ a balanced graph.

\medskip
Conversely, if $G$ is balanced, then there is a partition $(V_1, V_2)$ of $V$ 
such that positive edges connect nodes within $V_1$ or $V_2$, and
negative edges connect nodes between $V_1$ and $V_2$. Then,
if $u$ is the vector   with $u_i \in \{-1, 1\}$ defined so that $u_i = +1$ iff
$v_i \in V_1$ and $u_i = -1$ iff $v_i \in V_2$, we have
\[
u_i = \mathrm{sgn}(w_{i j}) u_j,
\]
and so  $\transpos{B} u = 0$, which shows that $u\in
\Ker(\transpos{B})$. 
Furthermore, the argument in the first part of the proof shows that
every  vector in $\Ker(\transpos{B})$ must have 
the same value  $c$ on all nodes in
$V_1$, and the value $-c$ on all nodes in  $V_2$, so it must be a multiple of
the vector $u$ given by $u_i = +1$ iff $v_i \in V_1$ and $u_i = -1$
iff $v_i \in V_2$. Therefore,  $\mathrm{dim}(\Ker(\transpos{B})) = 1$,
and $\mathrm{rank} = m - 1$.
The third part of the proposition has already been shown.
\end{proof}

\remark
A simple modification
of the proof of Proposition \ref{balancep2} shows that if there are
$c_1$ components containing only positive edges, 
$c_2$ components that are balanced graphs, and $c_3$ components
that are not balanced (and contain some negative edge), then
\[
c_1 + c_2  = m -\mathrm{rank}(B). 
\]

\medskip
Since by Proposition \ref{adjp2ws} we have $\overline{L} =
B\transpos{B}$ for any incidence matrix $B$  associated with an orientation
of the underlying graph of $G$, we obtain the following important
result (which is proved differently in
Kunegis et al.  \cite{kunegis:spectral}). 

\begin{theorem}
\label{balancep3}
The signed Laplacian $\overline{L}$ of a connected signed  graph $G$ is positive
definite iff $G$ is not balanced (possesses some cycle with an odd
number of negative edges).
\end{theorem}

\medskip
If $G = (V, W)$ is a balanced graph, then there is a partition 
$(V_1,  V_2)$ of $V$ such that for every edge $\{v_i, v_j\}$,  if
$w_{i j} >0$, then  $v_i, v_j\in V_1$ or $v_i, v_j\in V_2$, and if
$w_{i j} < 0$, then $v_i\in V_1$ and $v_j\in V_2$. It follows that if
we define the vector $x$ such that $x_i = +1$ iff $v_i \in V_1$ and
$x_i = -1$ iff $v_i \in V_2$, then for every edge $\{v_i, v_j\}$ we
have
\[
\mathrm{sgn}(w_{i j}) = x_i x_j.
\]
We call $x$ a  {\it bipartition\/} of $V$.

\medskip
The signed Laplacian of the balanced graph $G_1$ is given by
\[
\overline{L}_1 = 
\begin{pmatrix}
     2   & -1  &   0  &  -1  &   0  &   0  &   0  &   0  &   0\\
    -1  &   5   &  1  &  -1  &   1  &   0  &   0  &   -1  &   0\\
     0   &  1  &   3  &   0  &  -1   & -1  &   0  &   0  &   0\\
    -1  &  -1  &   0  &   5  &   1  &   0   & -1   & -1  &   0\\
     0  &   1   & -1  &   1  &   6  &  -1  &   0  &   1  &  -1\\
     0  &   0  &  -1  &   0  &  -1  &   4  &   0   &  1  &  -1\\
     0  &   0  &   0   & -1  &   0   &  0  &   2  &  -1  &   0\\
     0  &   -1  &   0  &  -1  &   1  &   1  &  -1  &   6  &   1\\
     0  &   0  &   0  &   0  &  -1  &  -1  &   0  &   1  &   3
\end{pmatrix}
\]
Using {\tt Matlab}, we find that its eigenvalues are
\[
0,\> 1.4790,\> 1.7513, \> 2.7883,\> 4.3570, \> 4.8815,\> 6.2158,\> 7.2159,\> 7.3112.
\]
The eigenvector corresponding to the eigenvalue $0$ is
\[
(0.3333, \> 0.3333,\> -0.3333,\> 0.3333, \> -0.3333, \> -0.3333, \>
0.3333, \> 0.3333, \> -0.3333)
\]
It gives us the bipartition 
\[(\{v_1, v_2, v_4, v_7, v_8\}, \{v_3, v_5,
v_6, v_9\}),
\]
as guaranteed by  Proposition \ref{balancep2}.

\medskip
The signed Laplacian of the unbalanced graph $G_2$ is given by
\[
\overline{L}_2 = 
\begin{pmatrix}
 2  &  -1  &   0  &  -1   &  0  &   0  &   0  &   0 &    0\\
    -1  &   5  &   1  &   1  &  -1  &   0  &   0  &   -1  &   0\\
     0  &   1  &   3  &   0  &  -1  &  -1  &   0  &   0  &   0\\
    -1  &   1  &   0  &   5  &   1  &   0  &  -1  &  -1  &   0\\
     0  &  -1  &  -1  &   1  &   6  &  -1  &   0  &   1  &  -1\\
     0  &   0  &  -1  &   0  &  -1  &   4  &   0  &   1  &  -1\\
     0  &   0   &  0  &  -1  &   0  &   0  &   2  &  -1   &  0\\
     0  &   -1  &   0  &  -1  &   1  &   1  &  -1  &   6  &   1\\
     0  &   0   &  0  &   0  &  -1   & -1   &  0  &   1  &   3
\end{pmatrix}
\]
The eigenvalues of $\overline{L}_2$ are
\[
0.5175,\> 1.5016, \> 1.7029, \> 2.7058, \> 3.7284, \> 4.9604, \>
5.6026, \> 7.0888, \> 8.1921.
\]
The matrix $\overline{L}_2$ is indeed positive definite (since $G_2$
is unbalanced).  Hou \cite {Hou} gives bounds on the smallest
eigenvalue of an unbalanced graph. The lower bound involves a measure
of how unbalanced the graph is (see Theorem 3.4  in Hou \cite {Hou}).

\medskip
Following Kunegis et al., we can prove the following result
showing that the eigenvalues and the eigenvectors of $\overline{L}$
and its unsigned counterpart $\s{L}$ are strongly related. 
Given a symmetric signed matrix $W$, we define the unsigned matrix
$\s{W}$ such that $\s{W}_{i j} = |w_{i j}|$ ($1\leq i, j \leq m$).
We let $\s{L}$ be the Laplacian associated with $\s{W}$.
Note that
\[
\s{L} = \overline{\Degsym} - \s{W}.
\]
The following proposition is shown in
Kunegis et al.  \cite{kunegis:spectral}). 

\begin{proposition}
\label{balancep4}
Let $G = (V, W)$ be a signed graph and let $\s{W}$ be the unsigned
matrix associated with $W$. If $G$ is balanced, and $x$ is a
bipartition of $V$, then for any  diagonalization
$\overline{L} = P \Lambda \transpos{P}$ of $\overline{L}$,
where $P$ is an orthogonal
matrix of eigenvectors of $\overline{L}$,  if we define the matrix $\s{P}$ so that
\[
\s{P}_{i} = x_i P_{i},
\]
where $\s{P}_i$ is the $i$th row of $\s{P}$ and $P_i$ is the $i$th row
of $P$, then $\s{P}$ is orthogonal and 
\[
\s{L} = \s{P} \Lambda \transpos{\s{P}}
\]
is a diagonalization of $\s{L}$. In particular, $\overline{L}$ and
$\s{L}$
have the same eigenvalues with the same multiplicities.
\end{proposition}

\begin{proof}
Observe that if we let 
\[
X = \mathrm{diag}(x_1, \ldots, x_m),
\]
then
\[
\s{P} = X P.
\]
It follows that
\[
\s{P} \Lambda \transpos{\s{P}} = X P \Lambda \transpos{P} \transpos{X}
= X \overline{L} \transpos{X}= X \overline{L} X,
\]
since $X$ is a diagonal matrix.
As a consequence, for diagonal entries, we have
\[
x_i^2  \overline{L}_{i i} =  \overline{D}_{i i} = \s{L}_{i i},
\]
and for $i \not= j$, we have
\[
x_i x_j \overline{L}_{i j}  
 = \mathrm{sgn}(w_{i j}) \overline{L}_{i j}  
= -\mathrm{sgn}(w_{i j}) w_{i j} = -|w_{i j}| = - \s{W}_{i j} =
\s{L}_{i j},  
 \]
which proves that $\s{L} = \s{P} \Lambda \transpos{\s{P}}$.
It remains to prove that $\s{P}$ is orthogonal.
Since $X$ is a diagonal matrix whose
entries  are $\pm 1$, we have $\transpos{X}  X = I$, so
\[
\transpos{\s{P}} \s{P} = \transpos{(XP)} XP = 
\transpos{P} \transpos{X} X P = 
\transpos{P} I P = I, 
\]
since $P$ is orthogonal. Thus, $\s{P}$ is indeed orthogonal.
\end{proof}

\section{$K$-Way Clustering of Signed Graphs}
\label{ch4-sec4}
Using the signed Laplacians $\overline{L}$ and
$\overline{L}_{\mathrm{sym}}$, we can define the optimization problems
as in Section \ref{ch3-sec3} and solve them as in Section \ref{ch3-sec5}, except that we drop the
constraint
\[
X (\transpos{X} X)^{-1} \transpos{X} \mathbf{1} = \mathbf{1},
\]
since $\mathbf{1}$ is not necessarily an eigenvector of
$\overline{L}$.
By  Proposition \ref{interlace},
the sum of the  $K$ smallest eigenvalues of $\overline{L}_{\mathrm{sym}}$  is a lower bound for
$\mathrm{tr}(\transpos{Y} \overline{L}_{\mathrm{sym}} Y)$, and 
the minimum of  problem $(**_2)$
is achieved by any $K$ unit eigenvectors $(u_1, \ldots, u_k)$ associated with the smallest
eigenvalues
\[
0 \leq \nu_1\leq  \nu_2 \leq  \ldots \leq  \nu_K
\]
of $\overline{L}_{\mathrm{sym}}$. The difference with
unsigned graphs is that $\nu_1$ may be strictly positive.
Here is the result of applying this method to various examples.

\begin{figure}[http]
  \begin{center}
 \includegraphics[height=1.5truein,width=1.5truein]{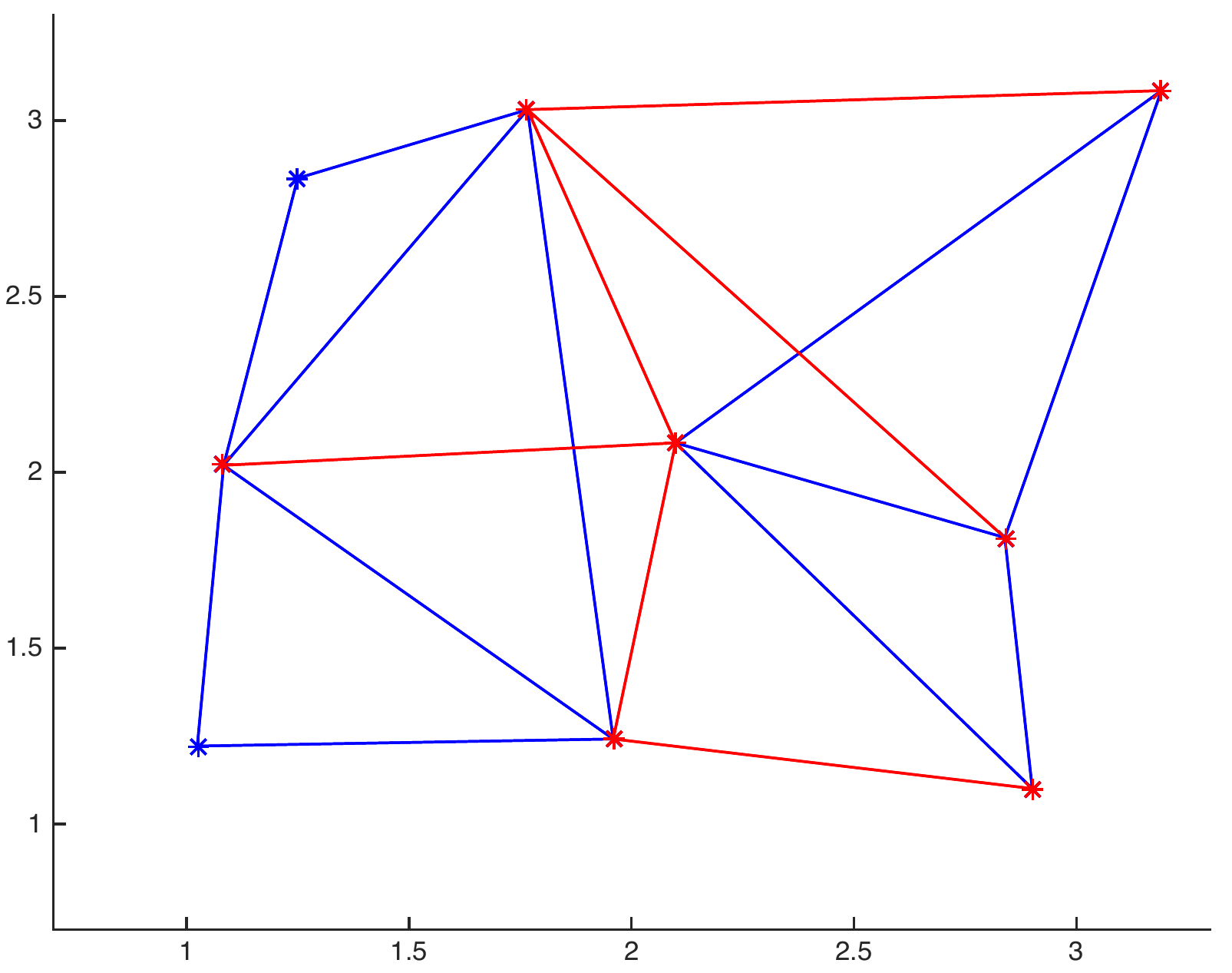}
\end{center}
  \caption{The balanced graph  $G_1$.}
\label{sgr1fig}
\end{figure}

\begin{figure}[http]
  \begin{center}
 \includegraphics[height=1.2truein,width=1.2truein]{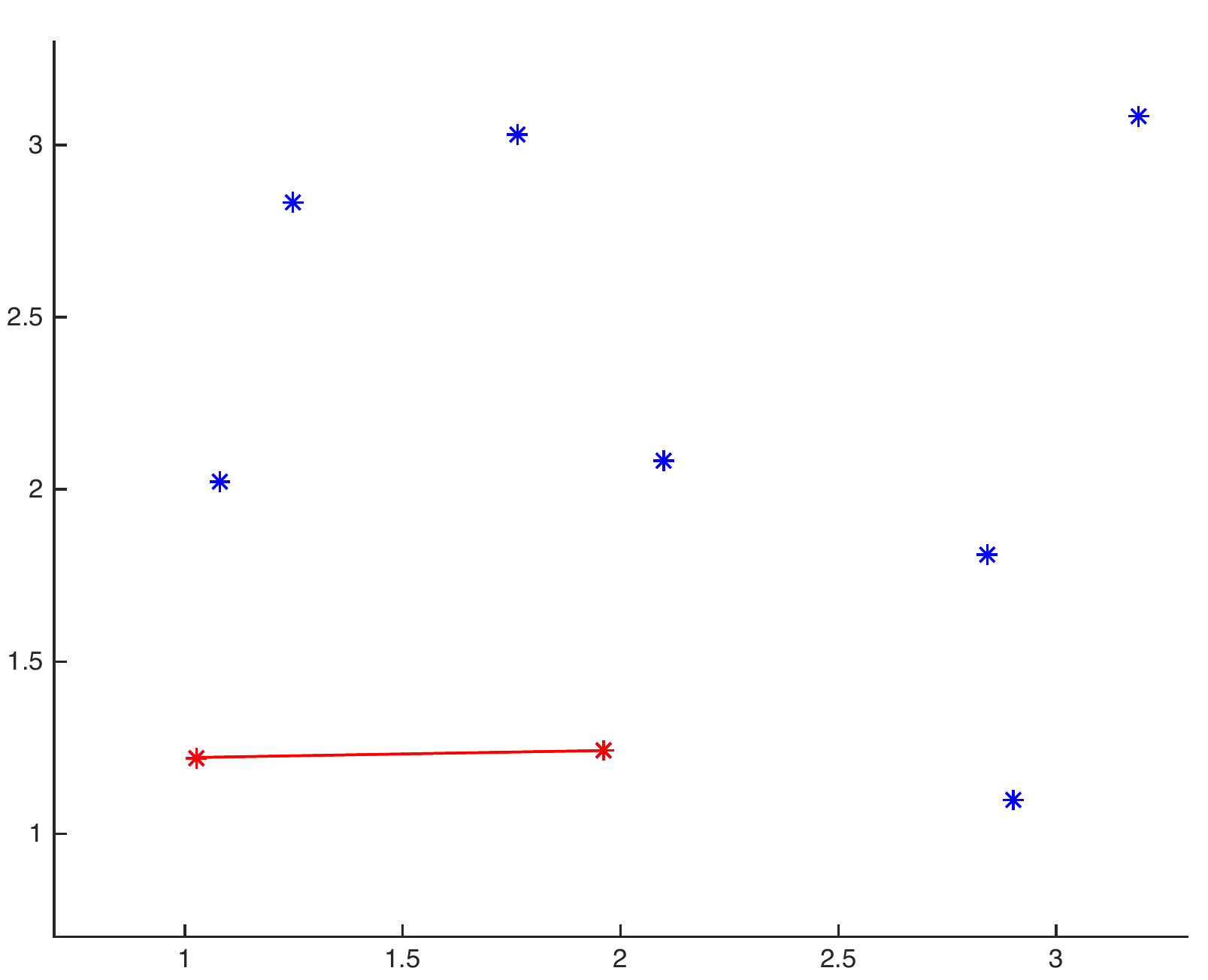}
\hspace{0.5cm}
 \includegraphics[height=1.2truein,width=1.2truein]{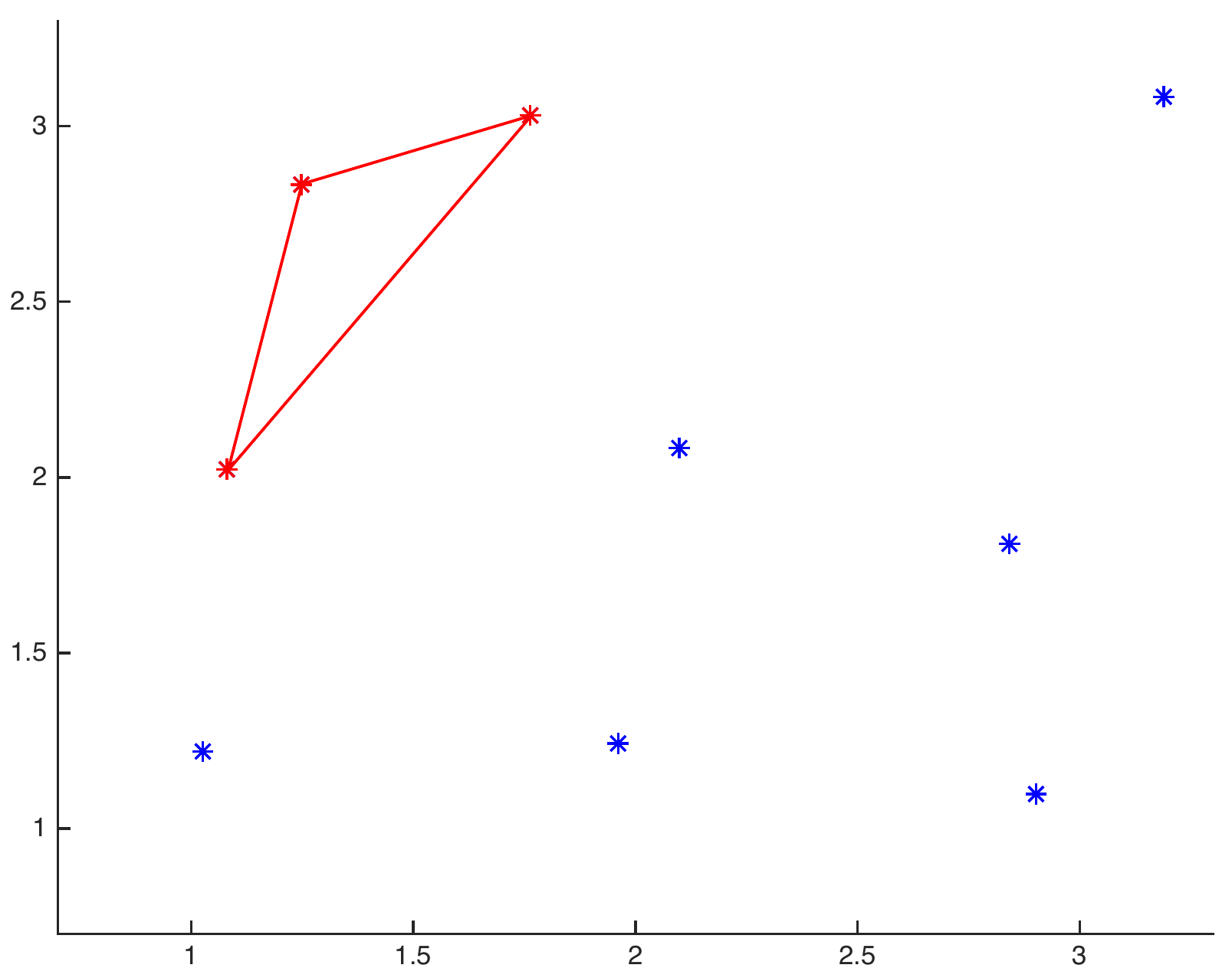}
\hspace{0.5cm}
 \includegraphics[height=1.2truein,width=1.2truein]{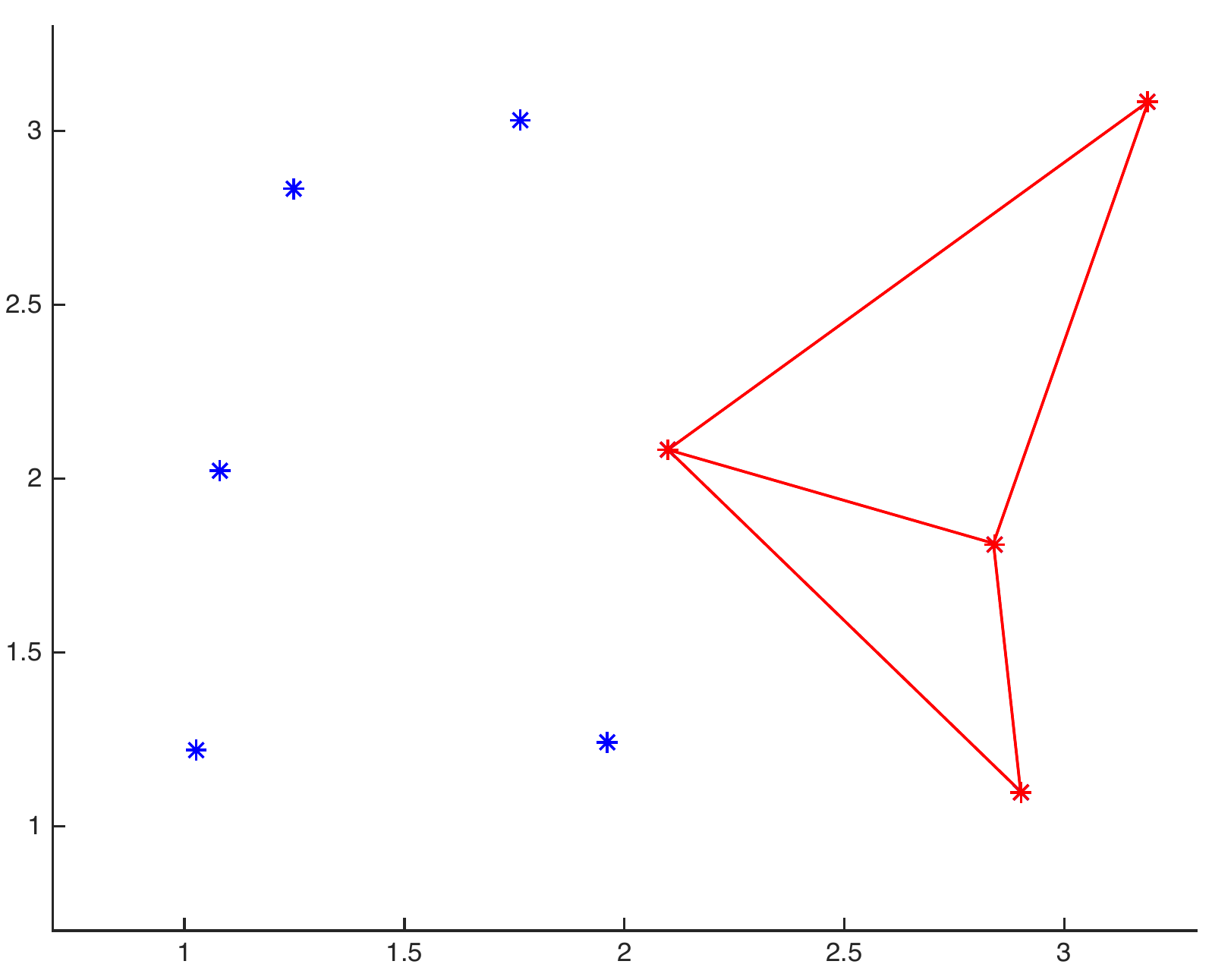}
  \end{center}
  \caption{Three blocks of a  normalized cut for the  graph associated
    with $G_1$.}
\label{sgr1b}
\end{figure}

\medskip
First, we apply our algorithm to find three clusters for the balanced
graph $G_1$. The graph $G_1$ as outputted by the algorithm is shown in Figure
\ref{sgr1fig} and the three clusters are shown in Figure \ref{sgr1b}.
As desired, these clusters do not contain negative edges.


By the way, for two clusters, the algorithm finds the bipartition of
$G_1$, as desired.

\medskip
Next, we apply our algorithm to find three clusters for the unbalanced
graph $G_2$. The graph $G_2$ as outputted by the algorithm is shown in Figure
\ref{sgr1fig} and the three clusters are shown in Figure \ref{sgr1b}.
As desired, these clusters do not contain negative edges.

\begin{figure}[http]
  \begin{center}
 \includegraphics[height=1.5truein,width=1.5truein]{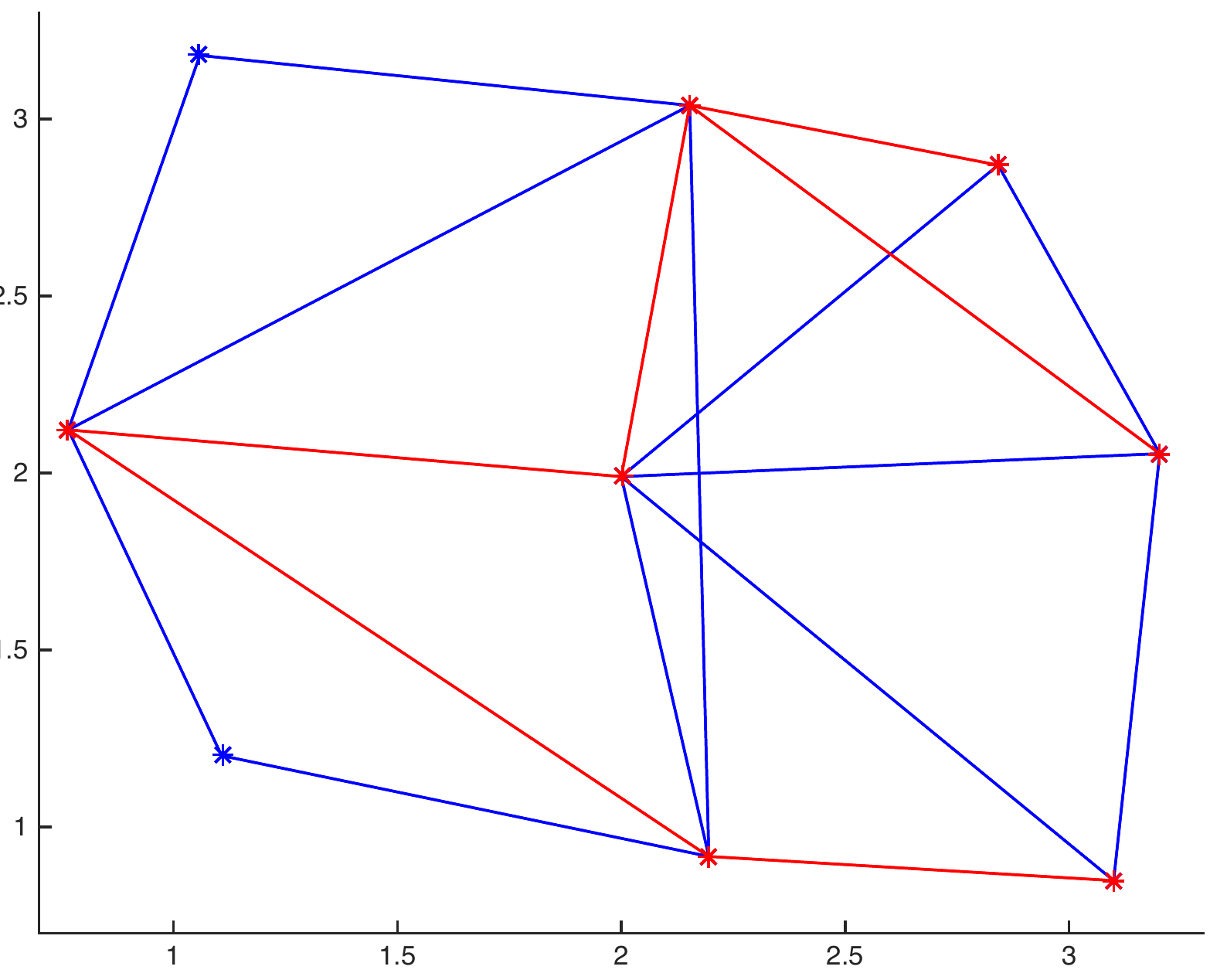}
\end{center}
  \caption{The unbalanced graph  $G_2$.}
\label{sgr2fig}
\end{figure}
\begin{figure}[http]
  \begin{center}
 \includegraphics[height=1.2truein,width=1.2truein]{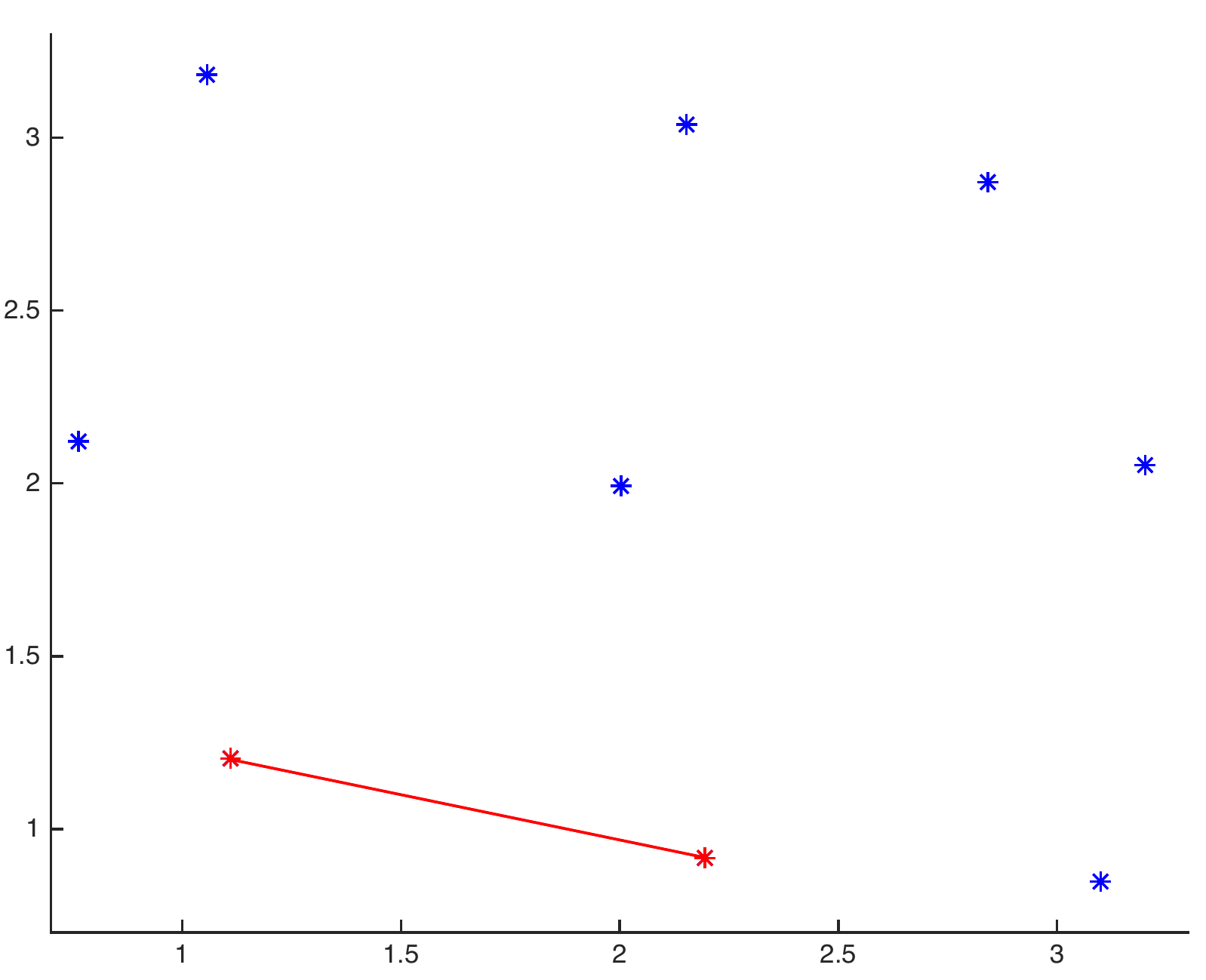}
\hspace{0.5cm}
 \includegraphics[height=1.2truein,width=1.2truein]{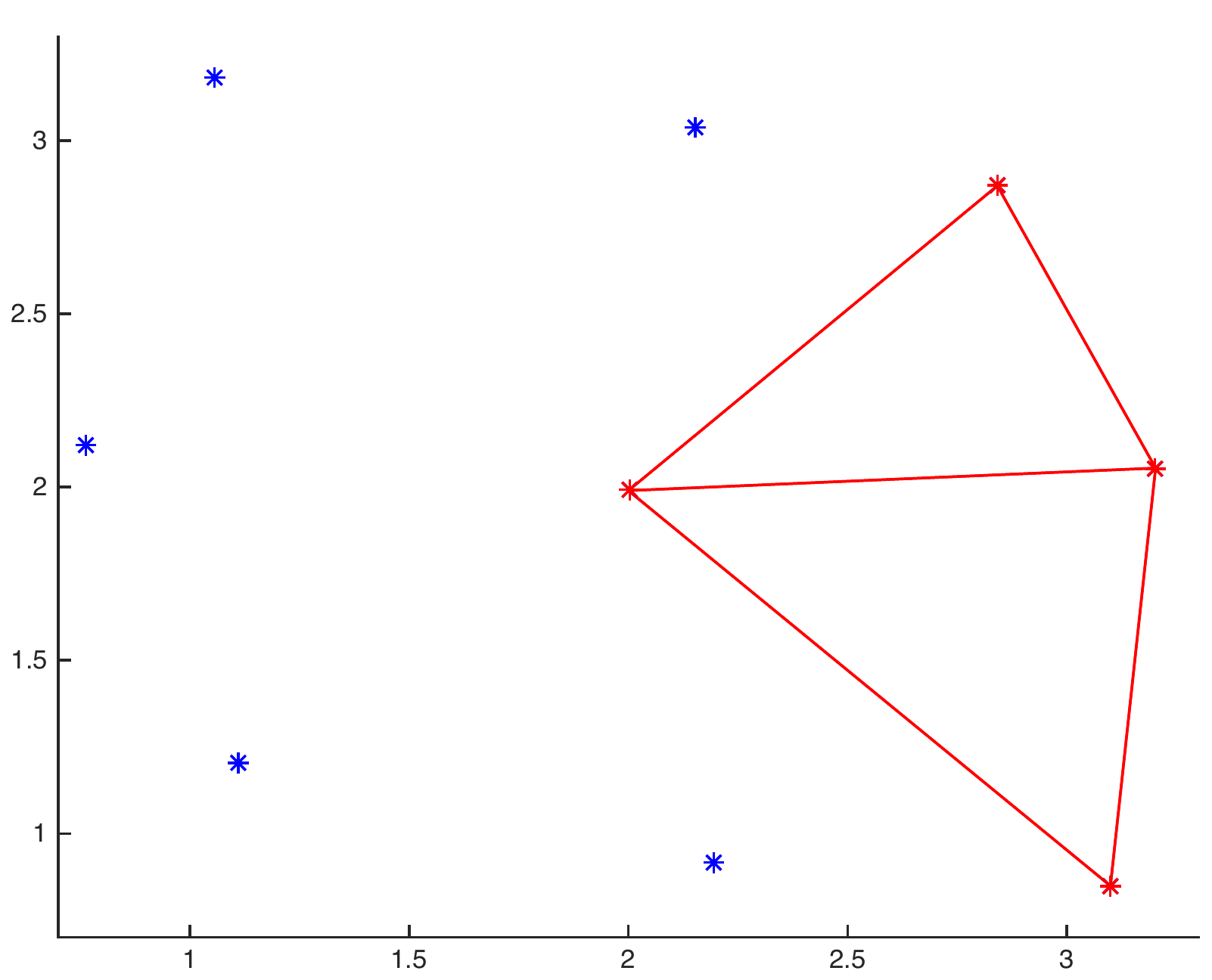}
\hspace{0.5cm}
 \includegraphics[height=1.2truein,width=1.2truein]{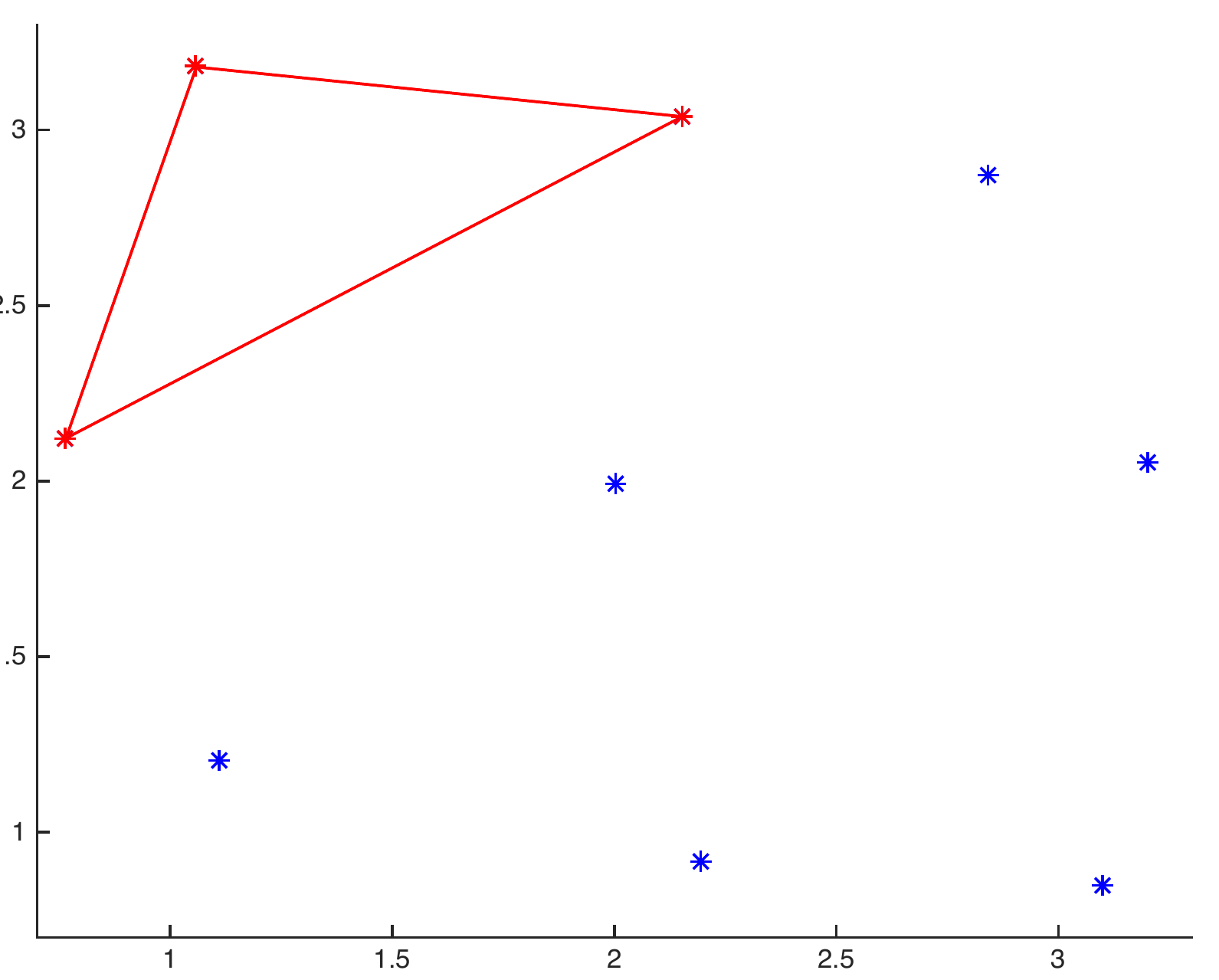}
  \end{center}
  \caption{Three blocks of a  normalized cut for the  graph associated
    with $G_2$.}
\label{sgr2b}
\end{figure}

The algorithm finds the same clusters, but this is probably due to the
fact that $G_1$ and $G_2$ only differ by the signs of two edges.

\section{Signed Graph Drawing}
\label{ch4-sec5}
Following Kunegis et al.  \cite{kunegis:spectral}, if our goal is to
draw a signed graph $G = (V, W)$ with $m$ nodes, 
a natural way to interpret negative weights
is to assume that the endpoints $v_i$ and $v_j$ of an edge with a negative weight
should be placed far apart, which can be achieved if instead of
assigning the point $\rho(v_j)\in \reals^n$ to $v_j$, we assign the
point $-\rho(v_j)$. Then, if $R$ is the $m\times n$ matrix of a graph
drawing of $G$ in $\reals^n$, the energy function $\s{E}(R)$ is
redefined to be
\[
\s{E}(R) = \sum_{\{v_i, v_j\}\in E} |w_{i j}| \norme{\rho(v_i) -
  \mathrm{sgn}(w_{i j})\rho(v_j)}^2.
\]
We obtain the following version of Proposition \ref{energyprop1}. 

\begin{proposition}
\label{energyprop1s}
Let $G = (V, W)$ be a signed graph, with $|V| = m$
and with $W$ a $m\times m$ symmetric matrix, and let $R$ be the
matrix of a graph drawing $\rho$ of  $G$ in $\reals^n$ 
(a $m\times n$ matrix). Then, we have
\[
\s{E}(R) = \mathrm{tr}(\transpos{R} \overline{L}  R).
\]
\end{proposition}

\begin{proof}
Since $\rho(v_i)$ is the $i$th row of $R$ (and $\rho(v_j)$ is the
$j$th row of $R$), if we denote the $k$th column
of $R$ by $R^k$, using Proposition \ref{Laplace1s},
we have
\begin{align*}
\s{E}(R)  & = \sum_{\{v_i, v_j\}\in E} |w_{i j}| \norme{\rho(v_i) -
  \mathrm{sgn}(w_{i j})\rho(v_j)}^2\\
 & = \sum_{k = 1}^n\sum_{\{v_i, v_j\}\in E} |w_{i j}| (R_{i k} -
  \mathrm{sgn}(w_{i j})R_{j k})^2\\
 & = \sum_{k = 1}^n\frac{1}{2}\sum_{i, j = 1}^m  |w_{i j}| (R_{i k} -
  \mathrm{sgn}(w_{i j})R_{j k})^2\\
 & = \sum_{k = 1}^n \transpos{(R^k)}\overline{L} R^k =  \mathrm{tr}(\transpos{R} \overline{L}  R), 
\end{align*}
as claimed.
\end{proof}

Then, as in Chapter \ref{chap2}, we look for a graph drawing $R$ that
minimizes $\s{E}(R) =  \mathrm{tr}(\transpos{R} \overline{L}  R)$
subject to $\transpos{R}R = I$.
The new ingredient is that $\overline{L}$ is positive definite iff $G$
is not a balanced graph. Also, in the case of a signed graph,
$\mathbf{1}$ does not belong to the kernel of $\overline{L}$, so we do
not get a balanced graph drawing.  

\medskip
If $G$ is a signed balanced graph,
then $\Ker L$ is nontrivial, and if $G$ is connected, then
$\Ker L$ is spanned by a vector whose components are either $+1$ or
$-1$.
Thus, if we use the first $n$ unit eigenvectors
$(u_1, u_2, \ldots, u_n)$ associated with the $n$ smallest eigenvalues
$0 = \lambda_1 < \lambda_2 \leq\cdots \leq \lambda_n$ of $\overline{L}$, we
obtain a drawing for which the nodes are partitionned into
two sets living in two hyperplanes corresponding to the value of their
first coordinate. Let us call such a drawing a 
{\it bipartite drawing\/}. However, if $G$ is connected, the vector
$u_2$ does not belong to $\Ker \overline{L}$, so if $m \geq 3$, 
it must have at least three coordinates with distinct absolute values,
and using $(u_2, \ldots, u_{n+1})$ we obtain a nonbipartite graph.
Then, the  following version of  Theorem \ref{graphdraw} is easily shown.

\begin{theorem}
\label{graphdraws}
Let $G = (V, W)$ be a signed graph with $|V| = m\geq 3$,
assume that $G$ has some negative edge and  is connected, and let
$\overline{L} = \overline{\Degsym} - W$ be the signed
Laplacian of  $G$.
\begin{enumerate}
\item[(1)]
If $G$ is not balanced
and if the eigenvalues of $L$ are
$0 < \lambda_1 \leq \lambda_2 \leq \lambda_3 \leq \ldots \leq \lambda_m$, 
then the minimal energy of any orthogonal  graph drawing of
$G$ in $\reals^n$ is equal to $\lambda_1 + \cdots + \lambda_{n}$
The $m \times n$ matrix 
$R$ consisting of any unit eigenvectors $u_1, \ldots, u_{n}$
associated with $\lambda_1 \leq \ldots \leq \lambda_{n}$ 
yields an orthogonal graph drawing of minimal energy.
\item[(2)]
If $G$ is  balanced
and if the eigenvalues of $L$ are
$0 = \lambda_1 < \lambda_2 \leq \lambda_3 \leq \ldots \leq \lambda_m$, 
then the minimal energy of any orthogonal nonbipartite graph drawing of
$G$ in $\reals^n$ is equal to $\lambda_2 + \cdots + \lambda_{n+1}$
(in particular, this implies that $n < m$).  
The $m \times n$ matrix 
$R$ consisting of any unit eigenvectors $u_2, \ldots, u_{n+1}$
associated with $\lambda_2 \leq \ldots \leq \lambda_{n+1}$ 
yields an orthogonal nonbipartite graph drawing of minimal energy.
\item[(3)]
If $G$ is  balanced, for $n = 2$, a graph drawing of $G$ as a
bipartite graph (with positive edges only withing the two blocks of vertices)
is obtained from the $m\times 2$ matrix consisting
of any two unit eigenvectors $u_1$ and $u_2$ associated with $0$ and $\lambda_2$.
\end{enumerate}
In all cases, the graph drawing $R$  satisfies the
condition $\transpos{R} R = I$ (it is an orthogonal graph drawing).
\end{theorem}

\begin{figure}[http]
  \begin{center}
 \includegraphics[height=1.5truein,width=1.5truein]{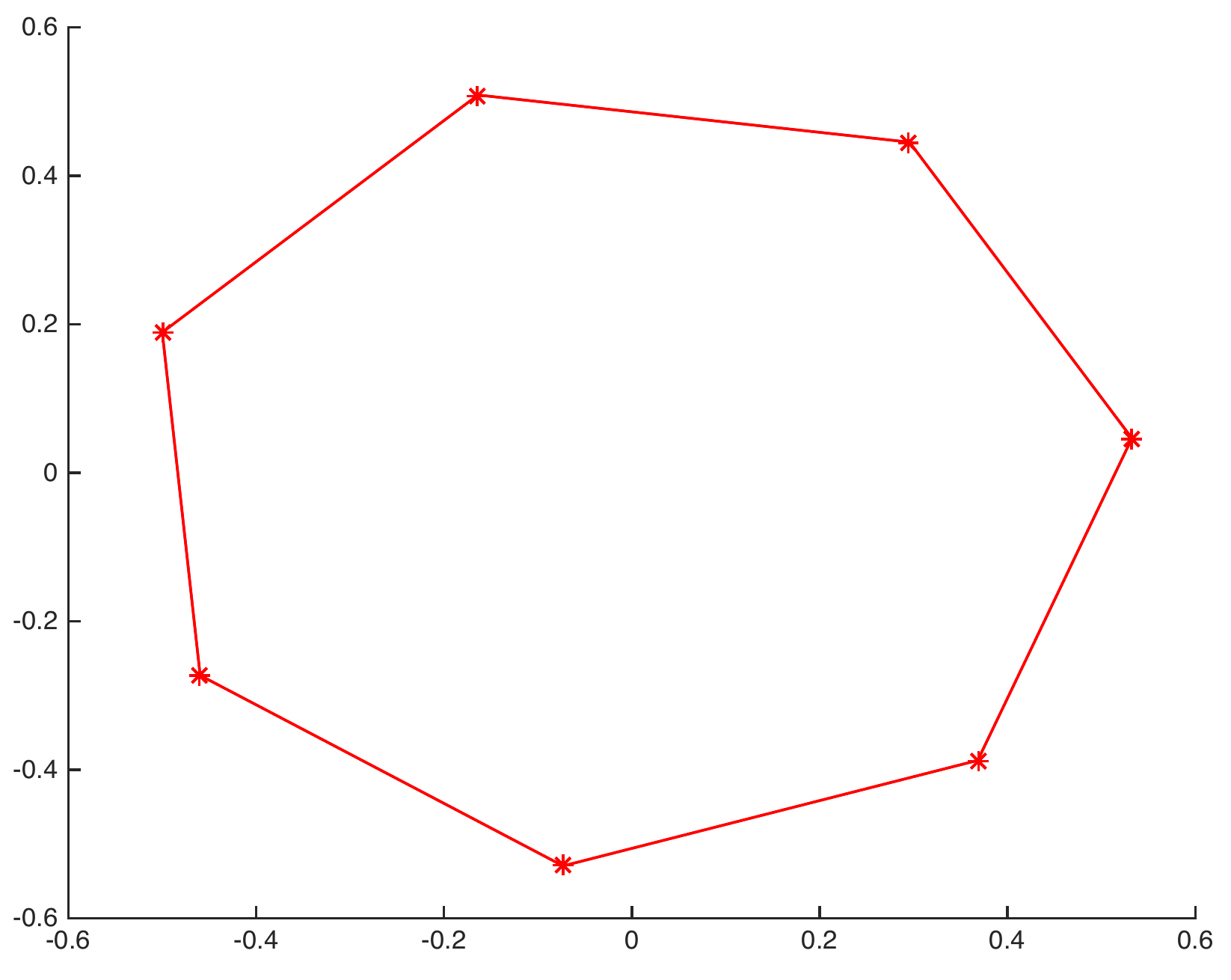}
\hspace{1cm}
 \includegraphics[height=1.5truein,width=1.5truein]{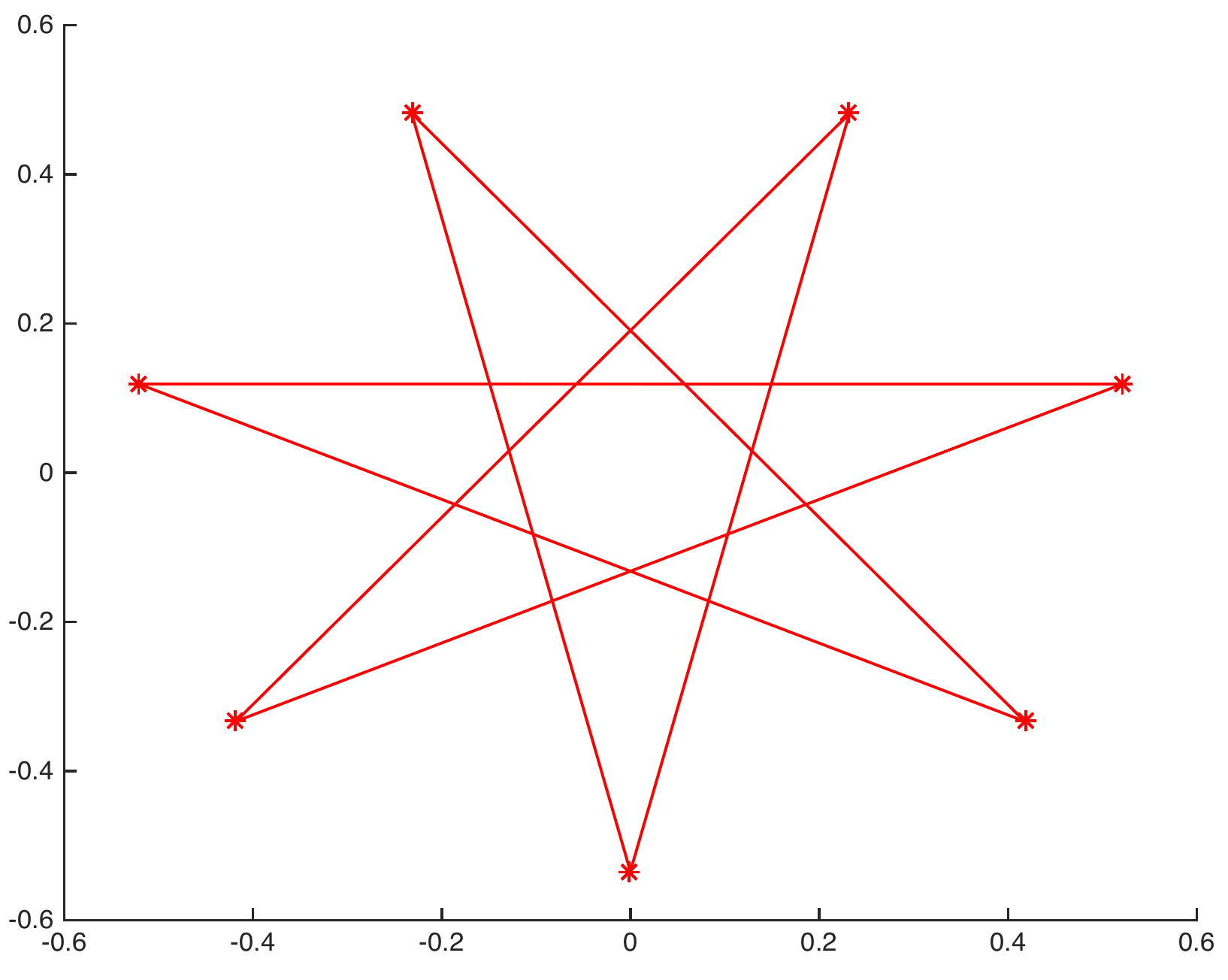}
  \end{center}
  \caption{The signed graph $G4$.}
\label{sgrG4}
\end{figure}

\medskip
Our first example is the signed graph $G4$ defined by the weight
matrix given by the following {\tt Matlab} program:
\begin{verbatim}
nn = 6; G3 = diag(ones(1,nn),1); G3 = G3 + G3';
G3(1,nn+1) = 1; G3(nn+1,1) = 1; G4 = -G3;
\end{verbatim}
All edges of this graph are negative. The graph obtained by using $G3$
is shown on the left and the graph obtained by using the signed
Laplacian of $G4$ is shown on the right in Figure \ref{sgrG4}.

\begin{figure}[H]
  \begin{center}
 \includegraphics[height=1.5truein,width=1.5truein]{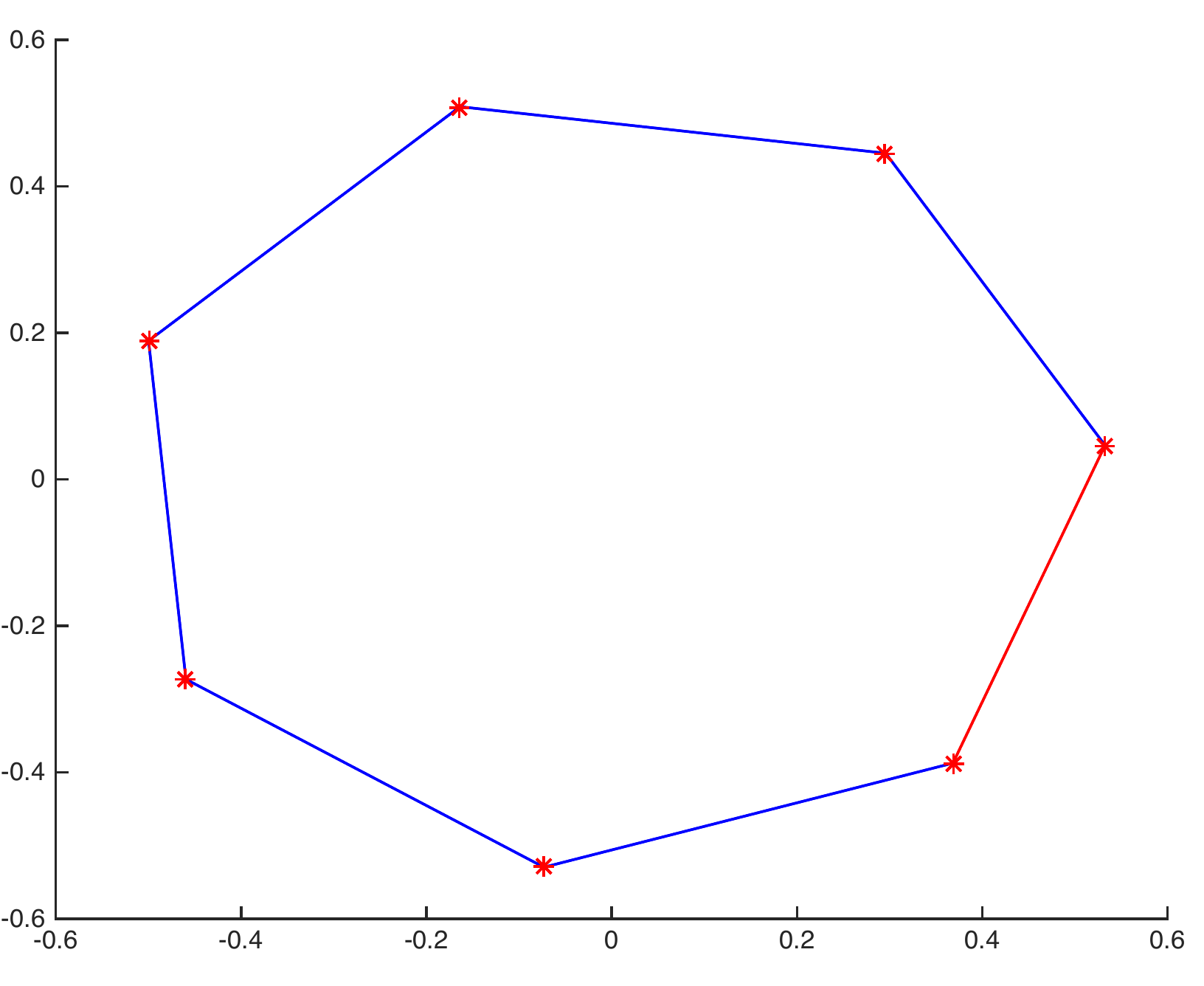}
\hspace{1cm}
 \includegraphics[height=1.5truein,width=1.5truein]{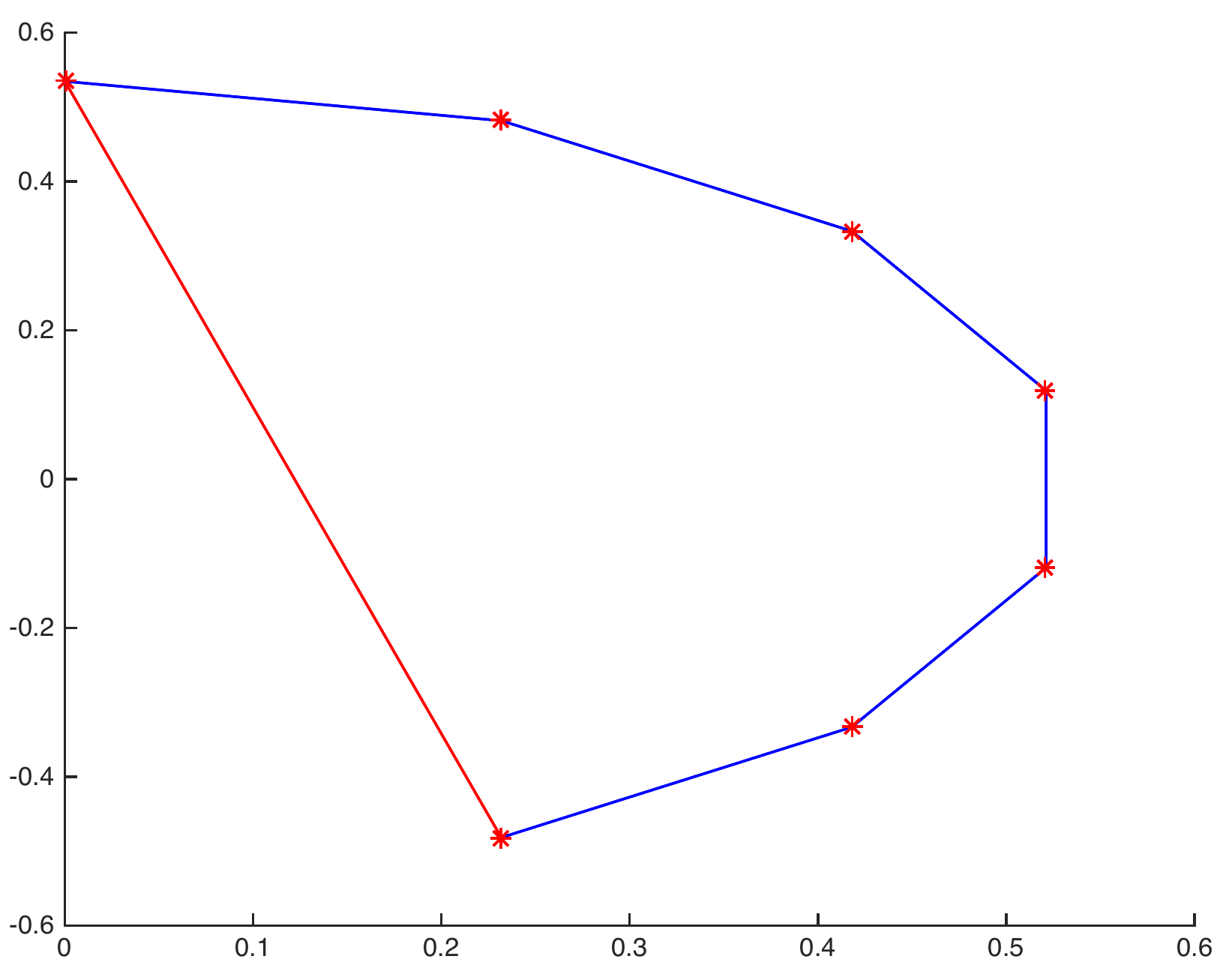}
  \end{center}
  \caption{The signed graph $G5$.}
\label{sgrG5}
\end{figure}

The second example is the signed graph $G5$ obtained from $G3$ by
making a single edge negative:

\begin{verbatim}
G5 = G3; G5(1,2) = -1; G5(2,1) = -1;
\end{verbatim}
The graph obtained by using $G3$
is shown on the left and the graph obtained by using the signed
Laplacian of $G5$ is shown on the right in Figure \ref{sgrG5}.
Positive edges are shown in blue and negative edges are shown in red.

\begin{figure}[http]
  \begin{center}
 \includegraphics[height=1.5truein,width=1.5truein]{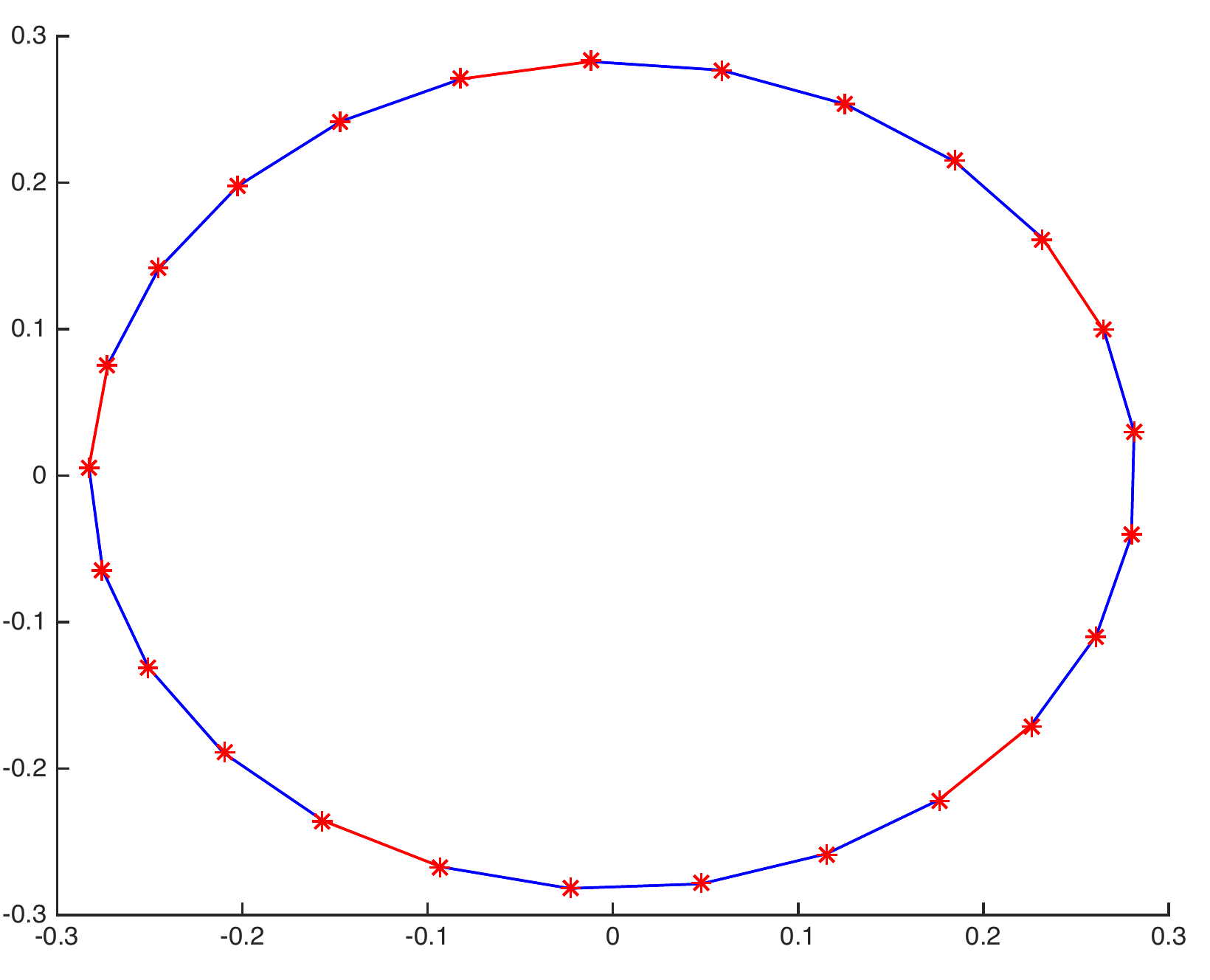}
\hspace{1cm}
 \includegraphics[height=1.5truein,width=1.5truein]{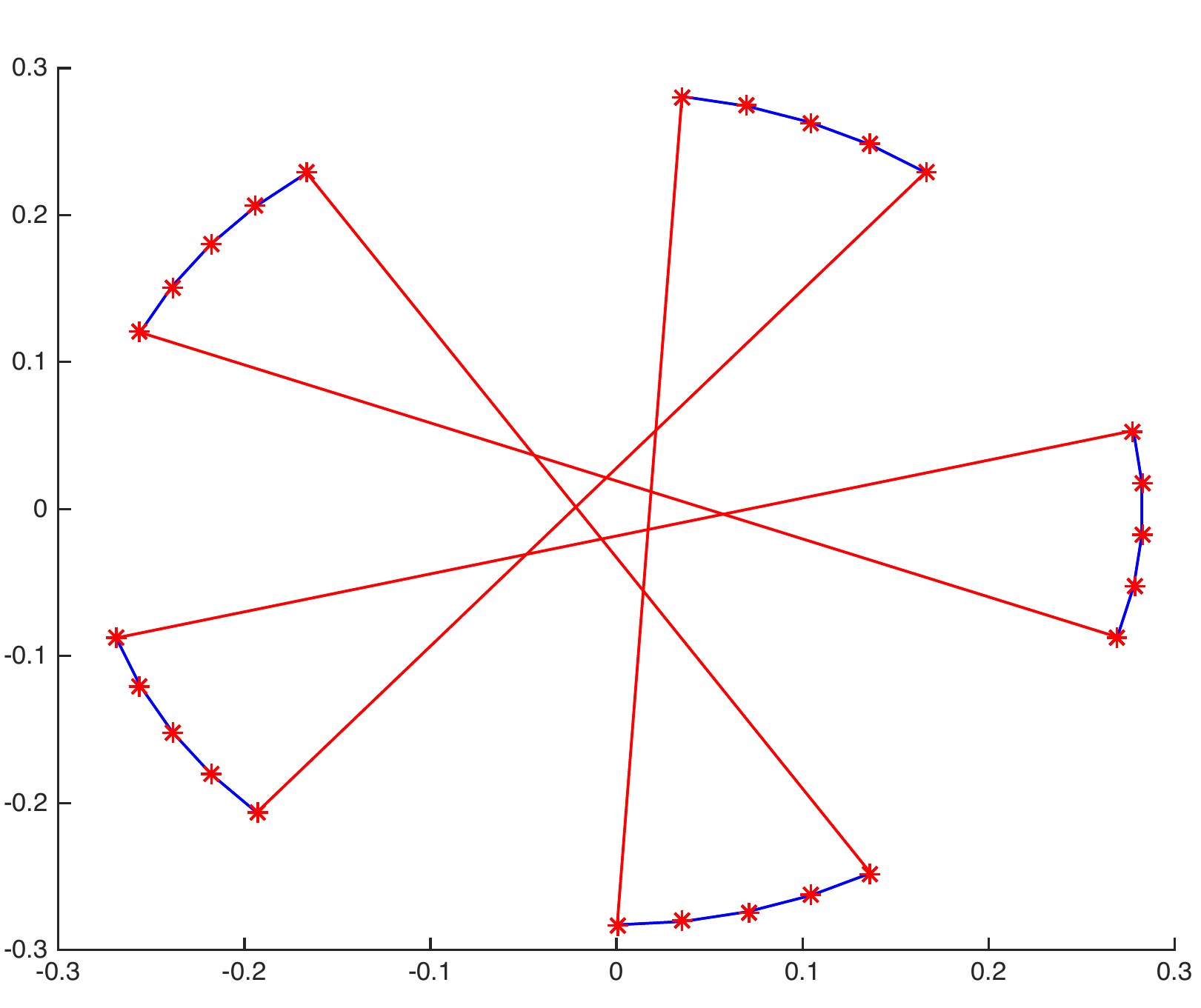}
  \end{center}
  \caption{The signed graph $G6$.}
\label{sgrG6}
\end{figure}

The third example is the signed graph $G6$ defined by the weight
matrix given by the following {\tt Matlab} program:
\begin{verbatim}
nn = 24; G6 = diag(ones(1,nn),1); G6 = G6 + G6';
G6(1,nn+1) = 1; G6(nn+1,1) = 1;
G6(1,2) = -1; G6(2,1) = -1; G6(6,7) = -1; G6(7,6) = -1;
G6(11,12) = -1; G6(12,11) = -1; G6(16,17) = -1; G6(17,16) = -1;
G6(21,22) = -1; G6(22,21) = -1;
\end{verbatim}
The graph obtained by using absolute values in $G6$
is shown on the left and the graph obtained by using the signed
Laplacian of $G6$ is shown on the right in Figure \ref{sgrG6}.

\begin{figure}[H]
  \begin{center}
 \includegraphics[height=1.5truein,width=1.5truein]{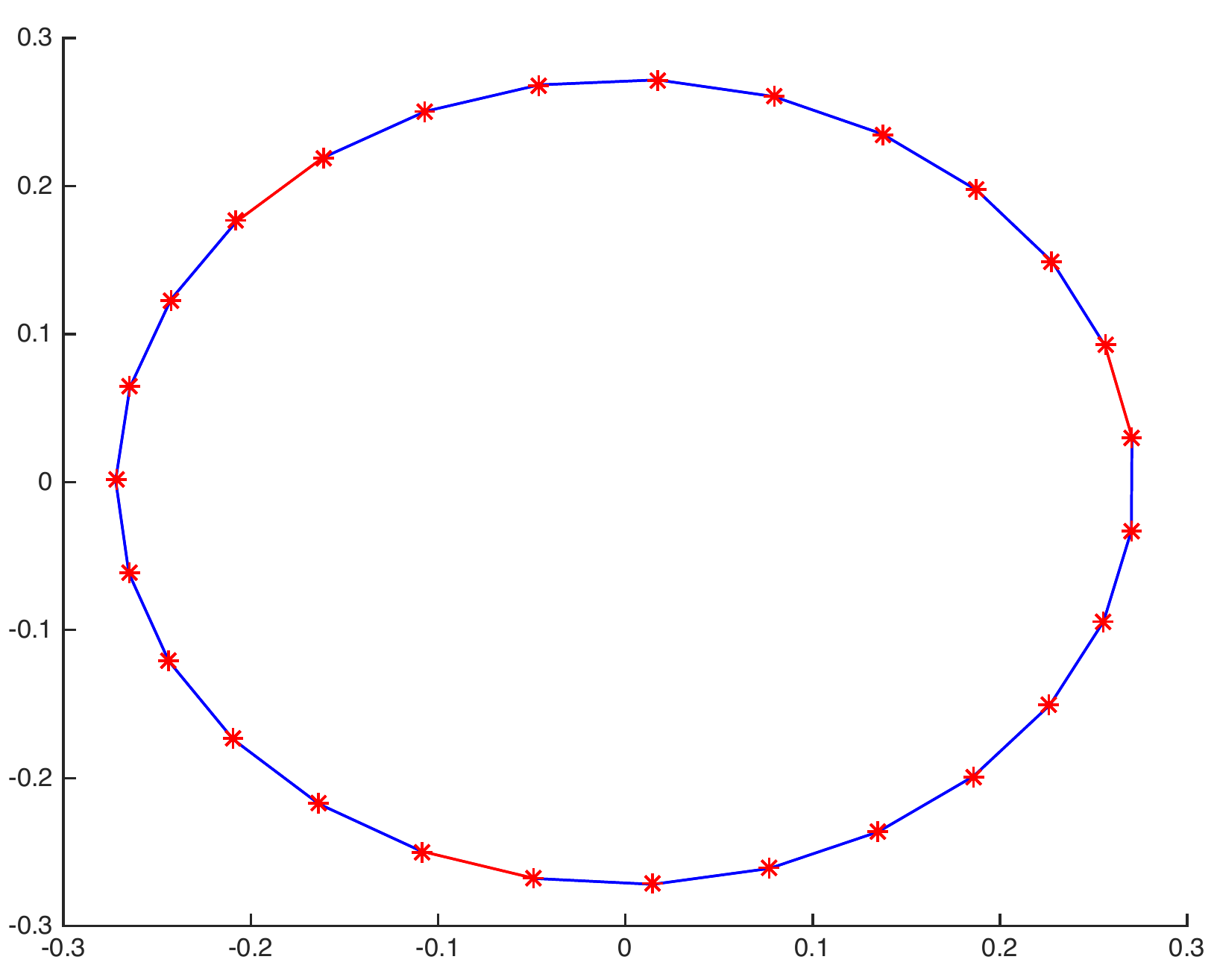}
\hspace{1cm}
 \includegraphics[height=1.5truein,width=1.5truein]{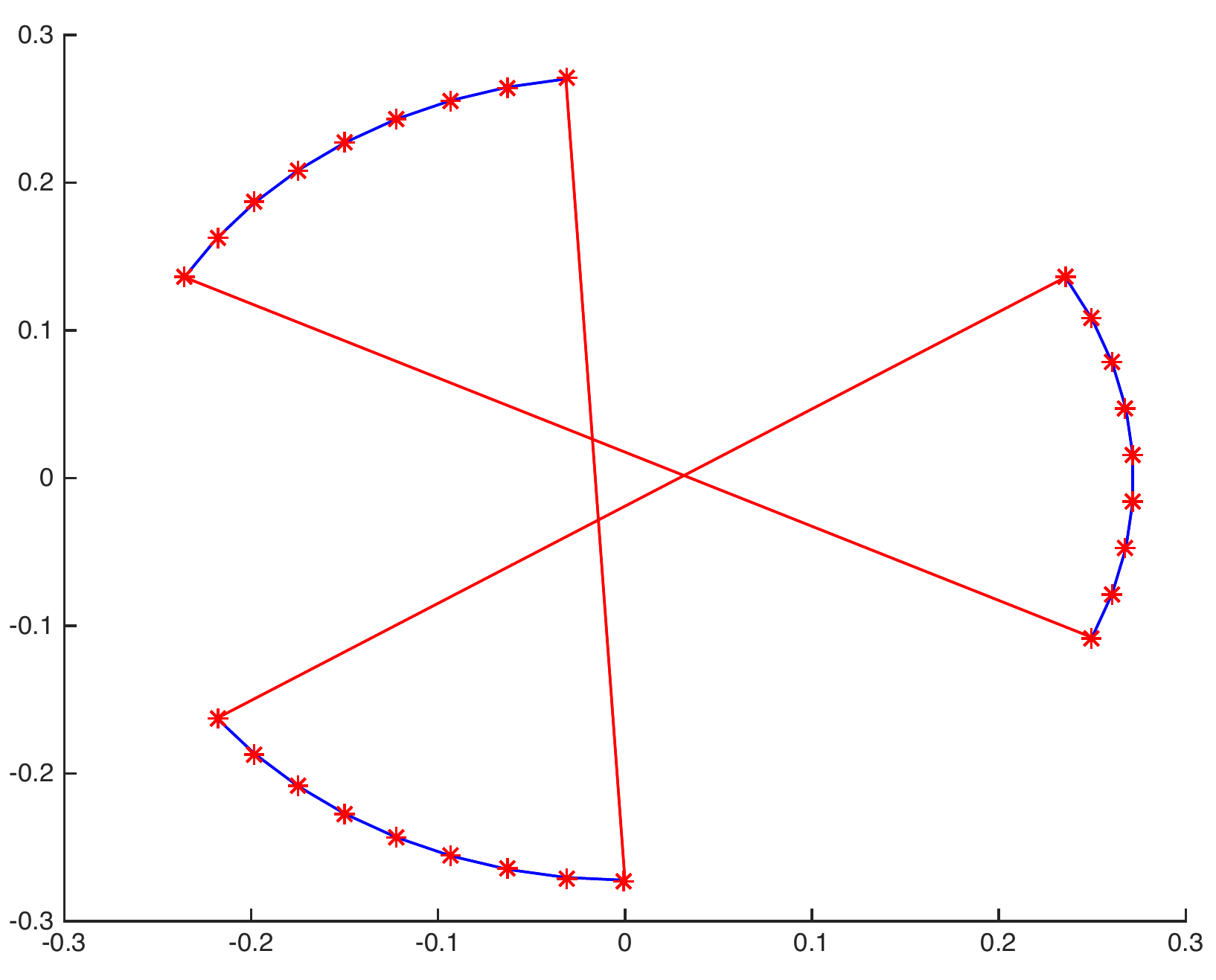}
  \end{center}
  \caption{The signed graph $G7$.}
\label{sgrG7}
\end{figure}

The fourth example is the signed graph $G7$ defined by the weight
matrix given by the following {\tt Matlab} program:
\begin{verbatim}
nn = 26; G7 = diag(ones(1,nn),1); G7 = G7 + G7';
G7(1,nn+1) = 1; G7(nn+1,1) = 1;
G7(1,2) = -1; G7(2,1) = -1; G7(10,11) = -1; G7(11,10) = -1;
G7(19,20) = -1; G7(20,19) = -1;
\end{verbatim}
The graph obtained by using absolute values in $G7$
is shown on the left and the graph obtained by using the signed
Laplacian of $G7$ is shown on the right in Figure \ref{sgrG7}.
These graphs are all unbalanced.  As predicted, nodes linked by
negative edges are far from each other.

\begin{figure}[H]
  \begin{center}
 \includegraphics[height=1.5truein,width=1.5truein]{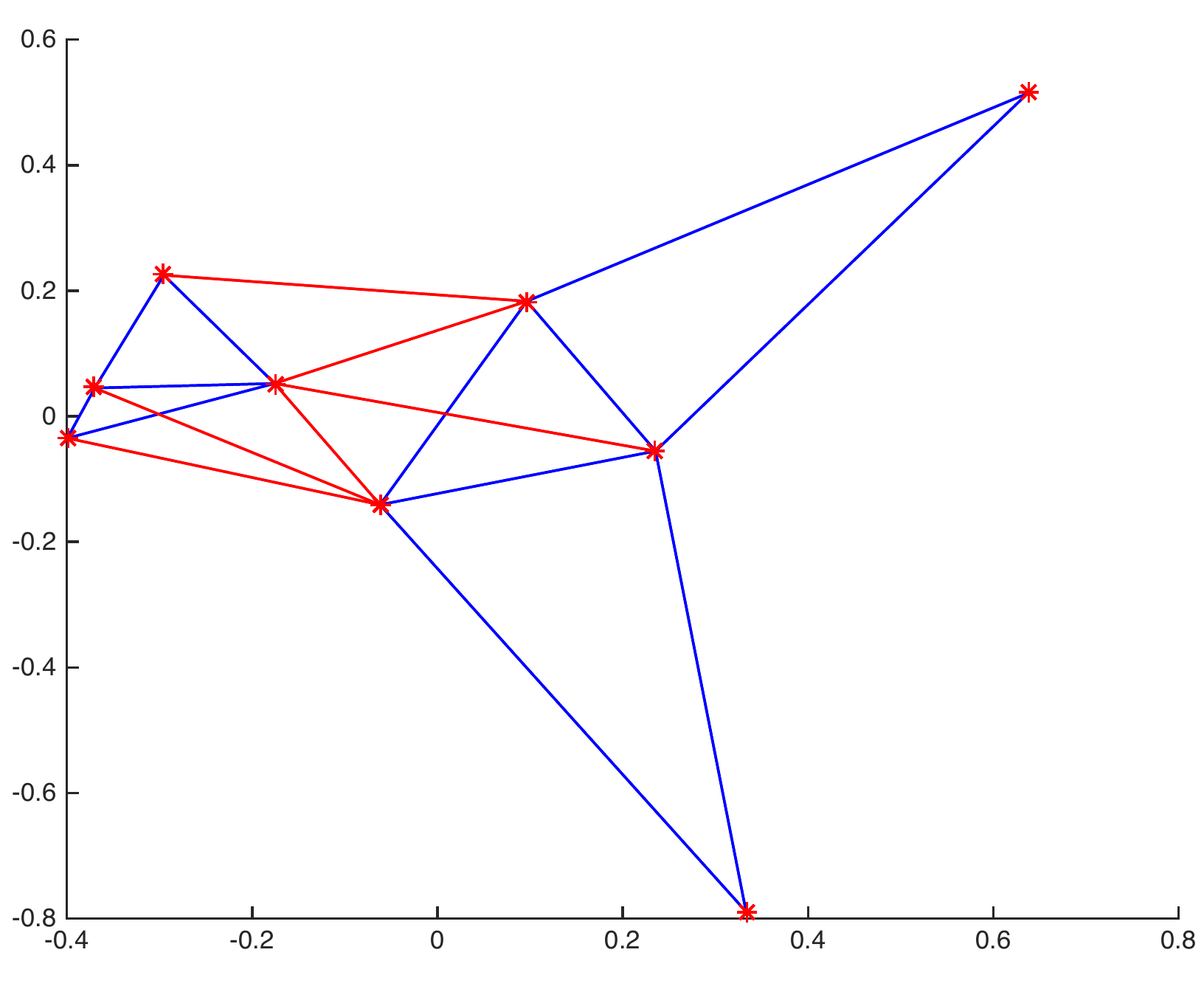}
\hspace{1cm}
 \includegraphics[height=1.5truein,width=1.5truein]{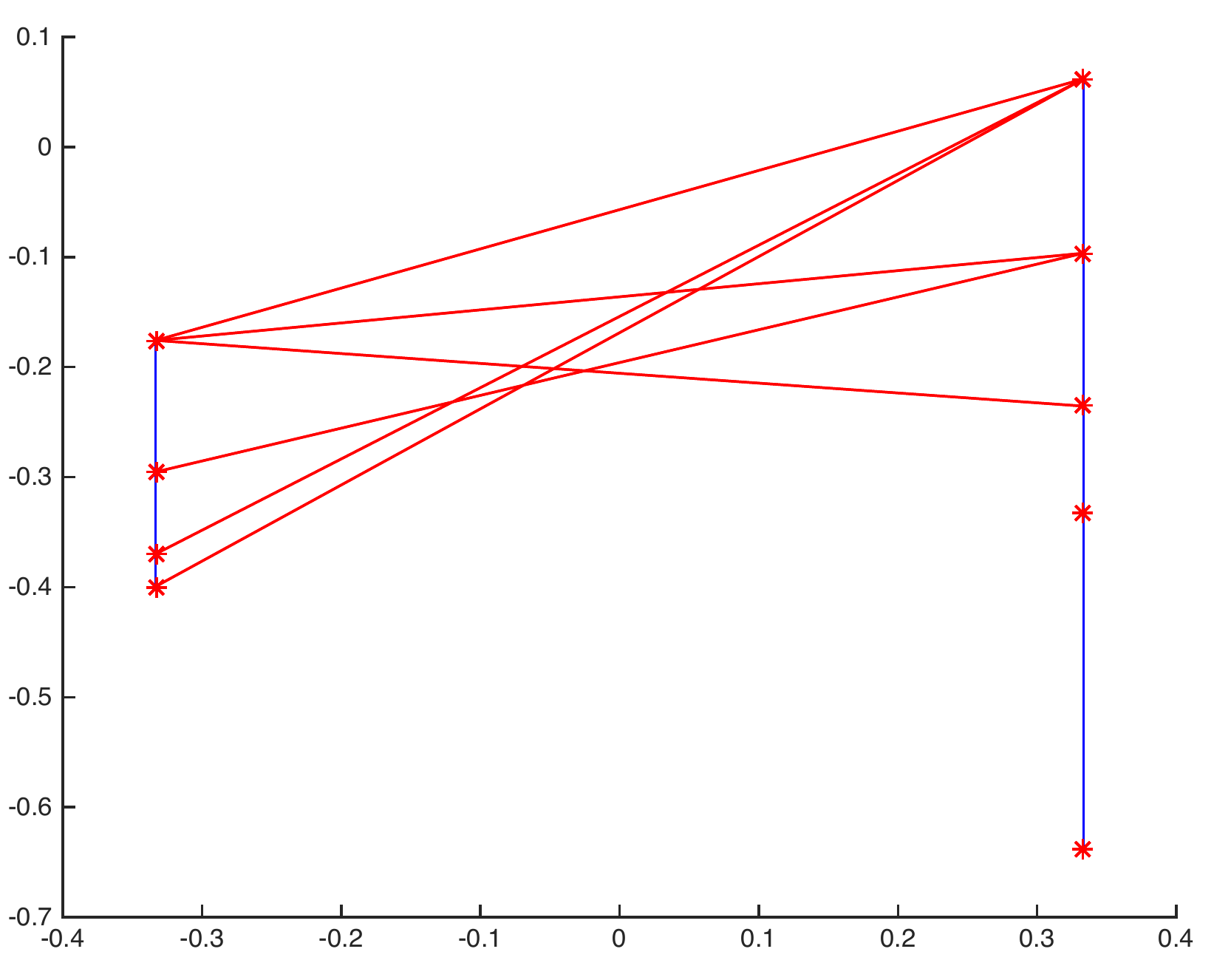}
  \end{center}
  \caption{The balanced  graph $G1$.}
\label{G1fig}
\end{figure}

\medskip
Our last example is the balanced graph $G1$ from Figure
\ref{sgraphfig1}. 
The graph obtained by using absolute values in $G1$
is shown on the left and the bipartite graph obtained by using the signed
Laplacian of $G1$ is shown on the right in Figure \ref{G1fig}.

\chapter{Graph Clustering Using Ratio Cuts}
\label{chap-ratio-cut}
In this short chapter, we consider the alternative to
normalized cut, called ratio cut, and show that
the methods of Chapters \ref{chap3} and \ref{chap4}
can be trivially adapted to solve the clustering problem
using ratio cuts.  All that needs to be done is to replace
the normalized Laplacian $L_{\mathrm{sym}}$ by the unormalized
Laplacian $L$, and omit  the step of considering Problem $(**_2)$. 
In particular, there is no need to  multiply the continuous
solution $Y$ by $\Degsym^{-1/2}$.
The idea of ratio cut is to replace the volume $\mathrm{vol}(A_j)$ of
each block $A_j$ of the partition by its size, $|A_j|$
(the number of nodes in $A_j$).
First, we deal with unsigned graphs, the case where the entries
in the symmetric weight matrix $W$ are nonnegative.

\begin{definition}
\label{ratiocut}
The {\it ratio cut\/}  $\mathrm{Rcut}(A_1, \ldots, A_K)$ of the 
partition $(A_1, \ldots, A_K)$ is defined as
\[
\mathrm{Rcut}(A_1, \ldots, A_K) = 
\sum_{i = 1}^K \frac{\mathrm{cut}(A_j, \overline{A}_j)}{|A_j|}.
\]
\end{definition}

As in Section \ref{ch3-sec3}, given a partition of $V$ into $K$
clusters $(A_1, \ldots, A_K)$, if we represent the $j$th block of
this partition by a vector $X^j$ such that
\[
X^j_i = 
\begin{cases}
a_j & \text{if $v_i \in A_j$} \\
0 &  \text{if $v_i \notin A_j$} ,
\end{cases}
\]
for some $a_j \not= 0$, then 
\begin{align*}
\transpos{(X^j)} L X^j & =
a_j^2(\mathrm{cut}(A_j, \overline{A_j})  \\
\transpos{(X^j)}  X^j& = a_j^2 |A_j|.
\end{align*}
Consequently, we have
\[
\mathrm{Rcut}(A_1, \ldots, A_K) = 
\sum_{i = 1}^K \frac{\mathrm{cut}(A_j, \overline{A}_j)}{|A_j|}
= \sum_{i = 1}^K \frac{\transpos{(X^j)} L X^j}{\transpos{(X^j)}  X^j}.
\]
On the other hand, the normalized cut is given by
\[
\mathrm{Ncut}(A_1, \ldots, A_K) = 
\sum_{i = 1}^K \frac{\mathrm{cut}(A_j, \overline{A}_j)}{\mathrm{vol}(A_j)}
= \sum_{i = 1}^K \frac{\transpos{(X^j)} L X^j}{\transpos{(X^j)}\Degsym  X^j}.
\]
Therefore, ratio cut is the special case of normalized cut where
$\Degsym = I$.
If we let
\[
\s{X}  = \Big\{[X^1\> \ldots \> X^K] \mid
X^j = a_j(x_1^j, \ldots, x_N^j) , \>
x_i^j \in \{1, 0\},
 a_j\in \reals, \> X^j \not= 0
\Big\}
\]
(note that the condition $X^j \not= 0$ implies that $a_j \not= 0$),
then the set of matrices representing partitions of $V$ into $K$
blocks is
\begin{align*}
& & &\s{K}  = \Big\{ X = [X^1 \> \cdots \> X^K] \quad \mid & &  X\in\s{X},  &&\\
         & & &  & & \transpos{(X^i)}  X^j = 0, \quad 1\leq i, j \leq K,\> 
i\not= j \Big\}.\\
\end{align*}

Here is our first formulation of $K$-way clustering
of a graph using ratio cuts, called problem PRC1 :

\medskip\noindent
{\bf $K$-way Clustering of a graph using Ratio Cut, Version 1: \\
Problem PRC1}
\begin{align*}
& \mathrm{minimize}     &  &  \sum_{j = 1}^K 
\frac{\transpos{(X^j)} L X^j}{\transpos{(X^j)}X^j}& &  &  &\\
& \mathrm{subject\ to} &  & 
 \transpos{(X^i)} X^j = 0, \quad 1\leq i, j \leq K,\> 
i\not= j,  & &  & & \\
& &  & X\in \s{X}. & & & & 
\end{align*}

The solutions that we are seeking are $K$-tuples 
$(\mathbb{P}(X^1), \ldots, \mathbb{P}(X^K))$ of points  in
$\mathbb{RP}^{N-1}$ determined by
their  homogeneous coordinates $X^1, \ldots, X^K$.
As in Chapter \ref{chap3},  chasing denominators and introducing a trace,
we obtain the following  
formulation of our  minimization
problem:

\medskip\noindent
{\bf $K$-way Clustering of a graph using Ratio Cut, Version 2: \\
Problem PRC2}

\begin{align*}
& \mathrm{minimize}     &  &  
\mathrm{tr}(\transpos{X} L X)& &  &  &\\
& \mathrm{subject\ to} &  & 
\transpos{X} X = I,
 & &  & & \\
& &  & X\in \s{X} .& & & & 
\end{align*}

The  natural relaxation of problem PRC2 is to drop the condition
that $X\in \s{X}$, and we obtain the 

\medskip\noindent
{\bf Problem $(R*_2)$}

\begin{align*}
& \mathrm{minimize}     &  &  
\mathrm{tr}(\transpos{X} L X)& &  &  &\\
& \mathrm{subject\ to} &  & 
\transpos{X}  X = I.
 & &  & & \\
\end{align*}
This time, since the normalization condition is
$\transpos{X}  X = I$, we can use the eigenvalues and the eigenvectors
of $L$, and by Proposition \ref{PCAlem1}, the minimum
is achieved by any $K$ unit eigenvectors
$(u_1, \ldots, u_K)$ associated with the smallest $K$ eigenvalues
\[
0 = \lambda_1 \leq \lambda_2 \leq \ldots \leq \lambda_K 
\]
of $L$. The matrix $Z = Y =  [u_1, \ldots, u_K]$ yields a minimum of
our relaxed problem  $(R*_2)$.
The rest of the algorithm is as before; we try to find $Q = R\Lambda$
with $R\in \mathbf{O}(K)$, $\Lambda$ diagonal invertible, 
and $X\in \s{X}$
such that $\norme{X - ZQ}$ is minimum.

\medskip
In the case of signed graphs, we define the
{\it signed ratio cut\/}
$\mathrm{sRcut}(A_1, \ldots, A_K)$ of the
partition $(A_1, \ldots, A_K)$  as
\[
\mathrm{sRcut}(A_1, \ldots, A_K) = \sum_{j = 1}^K
\frac{\mathrm{cut}(A_j, \overline{A_j})}{|A_j|} + 
2 \sum_{j = 1}^K\frac{\mathrm{links}^-(A_j, A_j)}{|A_j|}.
\] 
 Since we still have
\[
\transpos{(X^j)} \overline{L} X^j =
a_j^2(\mathrm{cut}(A_j, \overline{A_j}) + 2 \mathrm{links}^-(A_j, A_j)),
\]
we obtain
\[
\mathrm{sRcut}(A_1, \ldots, A_K) = 
\sum_{j = 1}^K \frac{\transpos{(X^j)} \overline{L} X^j} 
  {\transpos{(X^j)}  X^j} .
\]
Therefore, this is similar to the case of unsigned graphs, with $L$
replaced with $\overline{L}$. The same algorithm applies, but as in
Chapter \ref{chap4}, the signed Laplacian $\overline{L}$
is positive definite iff $G$ is unbalanced.
Modifying the computer program implementing
normalized cuts to deal with ratio cuts is trivial
(use $\overline{L}$ instead of $\overline{L}_{\mathrm{sym}}$
and don't multiply $Y$ by $\overline{\Degsym}^{-1/2}$).
Generally, normalized cut seems to yield ``better clusters,''  but 
this is not a very satisfactory statement since we haven't
defined  precisely in which sense a clustering is better than another.
We leave this point as further research.

\appendix
\chapter{Rayleigh Ratios and the Courant-Fischer Theorem}
\label{Rayleigh-Ritz} 
The most important property of symmetric matrices is that they
have real eigenvalues and that they can be diagonalized with respect
to an orthogonal matrix. Thus, if $A$ is an $n\times n$ symmetric
matrix, then it has $n$ real eigenvalues $\lambda_1, \ldots,
\lambda_n$  (not necessarily distinct), and there is
an orthonormal basis of eigenvectors $(u_1, \ldots, u_n)$
(for a proof, see Gallier \cite{Gallbook2}).
Another  fact that is used frequently in optimization problem is that
the eigenvalues of a symmetric matrix are characterized in terms of
what is known as the {\it Rayleigh ratio\/}, defined by
\[
R(A)(x) = \frac{\transpos{x} A x}{\transpos{x} x},\quad x\in \reals^n,
x\not= 0.
\]

The following proposition is often used to prove 
the correctness of various
optimization or approximation problems
(for example PCA).

\begin{proposition} ({\it Rayleigh--Ritz})
\label{PCAlem1a} 
If $A$ is a symmetric  $n\times n$ matrix with eigenvalues
$\lambda_1 \leq \lambda_2 \leq \cdots \leq \lambda_n$ and if
$(u_1, \ldots, u_n)$ is any orthonormal basis of eigenvectors
of $A$, where $u_i$ is a unit eigenvector associated with $\lambda_i$,
then
\[
\max_{x\not= 0} \frac{\transpos{x} A x}{\transpos{x}{x}} = \lambda_n
\]
(with the maximum attained for $x = u_n$),  and
\[
\max_{x\not= 0, x \in \{u_{n - k + 1}, \ldots, u_n\}^{\perp}} 
\frac{\transpos{x} A x}{\transpos{x}{x}} = \lambda_{n - k}
\]
 (with the maximum attained for $x = u_{n - k}$), where 
$1 \leq k \leq n - 1$. 
Equivalently, if $V_k$ is the subspace spanned by 
$(u_1, \ldots, u_{k})$, then
\[
\lambda_{k} = 
\max_{x\not= 0, x \in V_{k}} 
\frac{\transpos{x} A x}{\transpos{x}{x}},
\quad k = 1, \ldots, n. 
\]
\end{proposition}

\begin{proof}
First, observe that
\[
\max_{x\not= 0} \frac{\transpos{x} A x}{\transpos{x}{x}} =
\max_{x} \{ \transpos{x} A x \mid  \transpos{x}{x} = 1\},
\]
and similarly,
\[
\max_{x\not= 0, x \in \{u_{n - k + 1}, \ldots, u_n\}^{\perp}} 
\frac{\transpos{x} A x}{\transpos{x}{x}} =
\max_{x} \left\{\transpos{x} A x \mid 
(x \in \{u_{n - k + 1}, \ldots, u_n\}^{\perp}) \land (\transpos{x}{x} = 1)\right\}.
\]
Since $A$ is a symmetric matrix, its eigenvalues are real and
it can be diagonalized with respect
to an orthonormal basis of eigenvectors, so let
$(u_1, \ldots, u_n)$ be such a basis. If we write
\[ 
x = \sum_{i = 1}^n x_i u_i, 
\]
a simple computation shows that
\[
\transpos{x} A x = \sum_{i = 1}^n \lambda_i x_i^2.
\]
If $\transpos{x} x = 1$, then $\sum_{i = 1}^n  x_i^2 = 1$,
and since we assumed that 
$\lambda_1 \leq \lambda_2 \leq \cdots \leq \lambda_n$, we get
\[
\transpos{x} A x  =  \sum_{i = 1}^n \lambda_i x_i^2 
 \leq   \lambda_n \biggl(\sum_{i = 1}^n  x_i^2\biggr)
 =  \lambda_n.
\]
Thus, 
\[
\max_{x} \left\{ \transpos{x} A x \mid  \transpos{x}{x} = 1\right\} 
\leq \lambda_n,
\]
and since this maximum is achieved for $e_n =  (0, 0, \ldots, 1)$,
we conclude that
\[
\max_{x} \left\{ \transpos{x} A x \mid  \transpos{x}{x} = 1\right\} 
= \lambda_n.
\]
Next, observe that $x\in \{u_{n - k + 1}, \ldots, u_n\}^{\perp}$ 
and $\transpos{x}{x} = 1$ iff
$x_{n - k+ 1} = \cdots = x_n = 0$ and $\sum_{i = 1}^{n -k} x_i^2 = 1$.
Consequently, for such an $x$, we have
\[
\transpos{x} A x  =  \sum_{i = 1}^{n -k} \lambda_i x_i^2 
 \leq   \lambda_{n - k} \biggl(\sum_{i = k + 1}^n  x_i^2\biggr)
 =  \lambda_{n - k}.
\]
Thus, 
\[
\max_{x} \left\{ \transpos{x} A x \mid  
(x \in \{u_{n - k + 1}, \ldots, u_n\}^{\perp}) \land (\transpos{x}{x} = 1)\right\}
\leq \lambda_{n - k},
\]
and since this maximum is achieved for 
$e_{n - k} =  (0, \ldots, 0, 1, 0, \ldots, 0)$ with a $1$ in position
$n - k$,
we conclude that
\[
\max_{x} \left\{ \transpos{x} A x \mid  
(x \in \{u_{n - k + 1}, \ldots, u_n\}^{\perp}) \land (\transpos{x}{x} = 1)\right\}
= \lambda_{n - k},
\]
as claimed.
\end{proof} 

\medskip
For our purposes, we also need the version of Proposition \ref{PCAlem1a}
applying to $\min$ instead of $\max$, whose proof is obtained by
a trivial modification of the proof of Proposition \ref{PCAlem1a}.

\begin{proposition} ({\it Rayleigh--Ritz})
\label{PCAlem1}
If $A$ is a symmetric  $n\times n$ matrix with eigenvalues
$\lambda_1 \leq \lambda_2 \leq \cdots \leq \lambda_n$ and if
$(u_1, \ldots, u_n)$ is any orthonormal basis of eigenvectors
of $A$, where $u_i$ is a unit eigenvector associated with $\lambda_i$,
then
\[
\min_{x\not= 0} \frac{\transpos{x} A x}{\transpos{x}{x}} = \lambda_1
\]
(with the minimum attained for $x = u_1$),  and
\[
\min_{x\not= 0, x \in \{u_{1}, \ldots, u_{i-1}\}^{\perp}} 
\frac{\transpos{x} A x}{\transpos{x}{x}} = \lambda_{i}
\]
 (with the minimum attained for $x = u_{i}$), where 
$2 \leq i \leq n$. 
Equivalently, if $W_k = V_{k-1}^{\perp}$ denotes the subspace spanned by
$(u_{k}, \ldots, u_n)$ (with $V_{0} = (0)$), then
\[
\lambda _k =
\min_{x\not= 0, x \in W_k} 
\frac{\transpos{x} A x}{\transpos{x}{x}}
= \min_{x\not= 0, x \in V_{k-1}^{\perp}} 
\frac{\transpos{x} A x}{\transpos{x}{x}}, 
\quad k = 1, \ldots, n.
\]
\end{proposition}

\medskip
Propositions  \ref{PCAlem1a}  and \ref{PCAlem1} together 
are known as  the
{\it Rayleigh--Ritz theorem\/}.

\medskip
As an application of Propositions \ref{PCAlem1a} and \ref{PCAlem1}, we give a proof of
a proposition which is the key to the proof of Theorem
\ref{graphdraw}. First, we need a definition. Given an $n\times n$
symmetric matrix $A$ and an $m\times m$ symmetric $B$, with $m \leq
n$, if $\lambda_1 \leq \lambda_2 \leq \cdots \leq \lambda_n$
are the eigenvalues of $A$ and $\mu_1  \leq \mu_2 \leq \cdots \leq
\mu_m$ are the eigenvalues of $B$, then we say that the
 eigenvalues of $B$  {\it interlace\/} the eigenvalues of $A$ if
\[
\lambda_{i} \leq \mu_{i} \leq \lambda_{n - m + i}, \quad i = 1, \dots, m.
\]
The following proposition is known as the
{\it Poincar\'e separation theorem\/};
see Horn and Johnson \cite{HornJohn},
Section 4.3, Corollary 4.3.16.

\begin{proposition}
\label{interlace}
Let $A$ be an $n\times n$ symmetric matrix, $R$ be an $n\times m$
matrix such that $\transpos{R} R = I$ (with $m \leq n$), and let $B =
\transpos{R} A R$ (an $m\times m$ matrix). The following properties hold:
\begin{enumerate}
\item[(a)]
The eigenvalues of $B$ interlace the eigenvalues of $A$. 
\item[(b)]
If $\lambda_1 \leq \lambda_2 \leq \cdots \leq \lambda_n$
are the eigenvalues of $A$ and $\mu_1  \leq \mu_2 \leq \cdots \leq
\mu_m$ are the eigenvalues of $B$, and if $\lambda_i = \mu_i$, then
there is an eigenvector $v$ of $B$ with eigenvalue $\mu_i$  such that
$R v$ is an eigenvector of $A$ with eigenvalue $\lambda_i$.
\end{enumerate}
\end{proposition}

\begin{proof}
(a)
Let $(u_1, \ldots, u_n)$ be an orthonormal basis of eigenvectors for
$A$,
and let $(v_1, \ldots,v_m)$ be an orthonormal basis of eigenvectors
for $B$. Let $U_j$ be the subspace spanned by $(u_1, \ldots, u_j)$ and
let $V_j$  be the subspace spanned by $(v_1, \ldots, v_j)$.
For any $i$, the subpace $V_i$ has dimension $i$ and the subspace
$\transpos{R} U_{i - 1}$ has dimension at most $i - 1$. Therefore,
there is some nonzero vector 
$v\in V_i \cap (\transpos{R} U_{i -  1})^{\perp}$, and since
\[
\transpos{v} \transpos{R} u_j =  \transpos{(Rv)}  u_j  = 0, \quad j = 1, \ldots, i - 1,
\]
we have $Rv \in (U_{i - 1})^{\perp}$.  By Proposition \ref{PCAlem1} 
and using the fact that $\transpos{R} R = I$,  we have
\[
\lambda_i \leq 
 \frac{\transpos{(R v)} A R v}{\transpos{(R v)}  R v} =
 \frac{\transpos{v}B v}{\transpos{v} v}.
\]
On the other hand, by Proposition \ref{PCAlem1a},
\[
\mu_i =
\max_{x\not= 0, x \in \{v_{i + 1}, \ldots, v_n\}^{\perp}} 
\frac{\transpos{x} B x}{\transpos{x}{x}} = 
\max_{x\not= 0, x \in \{v_{1}, \ldots, v_i\}} 
\frac{\transpos{x} B x}{\transpos{x}{x}},  
\]
so 
\[
 \frac{\transpos{w}B w}{\transpos{w} w} \leq \mu_i
\quad
\hbox{for all $w\in V_{i}$},
\] 
and since $v\in V_i$,  we have
\[
\lambda_i \leq \frac{\transpos{v}B v}{\transpos{v} v}
\leq 
\mu_i, \quad i = 1, \ldots, m.
\]
We can apply the  same argument to the symmetric matrices $-A$ and
$-B$, to conclude that 
\[
- \lambda_{n - m + i} \leq - \mu_{i},
\]
that is,
\[
\mu_i \leq \lambda_{n - m + i}, \quad i = 1, \ldots, m.
\]
Therefore, 
\[
\lambda_{i} \leq \mu_i \leq \lambda_{n - m + i}, \quad i = 1, \ldots, m,
\]
as desired.

\medskip
(b)
If $\lambda_i = \mu_i$, then 
\[
\lambda_i  = \frac{\transpos{(R v)} A R v}{\transpos{(R v)}  R v} =
\frac{\transpos{v}B v}{\transpos{v} v} = \mu_i,
\]
so $v$ must be an eigenvector for $B$ and $Rv$ must be an eigenvector
for $A$, both for the eigenvalue  $\lambda_i = \mu_i$. 
\end{proof}

\medskip
Observe that Proposition \ref{interlace} implies that
\[
\lambda_1 + \cdots + \lambda_{m} \leq 
\mathrm{tr}(\transpos{R} A R) \leq \lambda_{n - m + 1} + \cdots + \lambda_n.
\]
The left inequality is used to prove Theorem \ref{graphdraw}.

\medskip
For the sake of completeness, we also prove the Courant--Fischer
characterization
of the eigenvalues of a symmetric matrix.

\begin{theorem} ({\it Courant--Fischer\/})
\label{Courant-Fischer}
Let  $A$ be  a symmetric  $n\times n$ matrix  with eigenvalues
$\lambda_1 \leq \lambda_2 \leq \cdots \leq \lambda_n$ and let
$(u_1, \ldots, u_n)$ be any orthonormal basis of eigenvectors
of $A$, where $u_i$ is a unit eigenvector associated with $\lambda_i$.
If $\s{V}_k$ denotes the set of subspaces of $\reals^n$ of dimension
$k$, then
\begin{align*}
\lambda_k & = \max_{W\in \s{V}_{n - k + 1}} \min_{x\in W, x\not= 0} 
\frac{\transpos{x} A x}{\transpos{x} x} \\
\lambda_k & =\min_{W\in \s{V}_{k}} \max_{x\in W, x\not= 0} 
\frac{\transpos{x} A x}{\transpos{x} x} .
\end{align*}
\end{theorem}

\begin{proof}
Let us consider the second equality, the proof of the first
equality being similar.
Observe that the space $V_k$ spanned by $(u_1, \ldots, u_k)$ has
dimension $k$, and by Proposition \ref{PCAlem1a}, we have
\[
\lambda_k = 
\max_{x\not= 0, x \in V_k} 
\frac{\transpos{x} A x}{\transpos{x}{x}}
\geq \min_{W\in \s{V}_{k}} \max_{x\in W, x\not= 0} 
\frac{\transpos{x} A x}{\transpos{x} x}. 
\]
Therefore, we need to prove the reverse inequality; that is, we have
to show that
\[
\lambda _k \leq \max _{x\not= 0, x\in W} 
\frac{\transpos{x} A  x}{\transpos{x} x},
\quad \hbox{for all}\quad W\in \s{V}_{k}.
\]
Now, for any  $W\in \s{V}_{k}$, if we can prove that
$W\cap V_{k-1}^{\perp} \not= (0)$, then for any nonzero $v\in W\cap
V_{k-1}^{\perp}$,
by Proposition \ref{PCAlem1} , we have
\[
\lambda _k 
= \min_{x\not= 0, x \in V_{k-1}^{\perp}} 
\frac{\transpos{x} A x}{\transpos{x}{x}} 
\leq \frac{\transpos{v} A v}{\transpos{v}{v}}
\leq \max_{x\in W, x\not= 0}  \frac{\transpos{x} A x}{\transpos{x}{x}}.
\]
It remains to prove that $ \mathrm{dim}(W\cap   V_{k-1}^{\perp}) \geq 1$. 
However, $\mathrm{dim}(V_{k-1}) = k - 1$, so
$\mathrm{dim}(V_{k-1}^{\perp}) = n - k + 1$, and by hypothesis
$\mathrm{dim}(W) = k$. By the Grassmann relation,
\[
\mathrm{dim}(W) + \mathrm{dim}(V_{k-1}^{\perp}) =
\mathrm{dim}(W\cap   V_{k-1}^{\perp}) + \mathrm{dim}(W + V_{k-1}^{\perp}), 
\]
and since $ \mathrm{dim}(W + V_{k-1}^{\perp}) \leq
\mathrm{dim}(\reals^n) = n$, we get
\[
k + n - k + 1 \leq \mathrm{dim}(W\cap   V_{k-1}^{\perp}) + n;
\]
that is, $1 \leq  \mathrm{dim}(W\cap   V_{k-1}^{\perp})$, as claimed.
\end{proof}

\chapter{Riemannian Metrics on  Quotient Manifolds}
\label{ch3-sec6} 
In order to define a metric on the projective space $\mathbb{RP}^n$, we
need to review a few notions of differential geometry. First, we need to
define the quotient $M/G$ of a manifold by a group acting on $M$.
This section relies heavily on Gallot, Hulin, Lafontaine \cite{Gallot}
and Lee \cite{Lee}, which contain thorough expositions and should be 
consulted for details.

\begin{definition}
\label{actdef}
Recall that an {\it action\/} of a group $G$ (with identity element
$1$) on a set $X$ is a map
$\mapdef{\gamma}{G\times X}{X}$ satisfying the following properties:
\begin{enumerate}
\item[(1)]
$\gamma(1, x) = x$, for all $x\in X$.
\item[(2)]
$\gamma(g_1,\gamma(g_2, x)) = \gamma(g_1g_2, x)$,
for all $g_1, g_2\in G$, and all $x\in X$.
\end {enumerate}
We usually abbreviate $\gamma(g, x)$ by $g\cdot x$.

\medskip
If $X$ is a topological space and $G$ is a topological group,
we say that the action is {\it
  continuous\/} iff the map $\gamma$ is continuous. In this case,
for every $g\in G$, the  map $x \mapsto
g\cdot x$ is a homeomorphism.
If $X$ is a smooth manifold and $G$ is a Lie group,
we say that the action is {\it
  smooth\/} iff the map $\gamma$ is smooth.
In this case, for every $g\in G$, the  map $x \mapsto
g\cdot x$ is a diffeomorphism.
\end{definition}

\remark
To be more precise, what we have defined in Definition \ref{actdef}
is a {\it left action\/} of the group $G$ on the set $X$.
There is also a notion of a {\it right action\/}, but we won't need it.

\medskip
The {\it quotient of $X$ by $G$\/}, denoted $X/G$, is the set of
orbits of $G$; that is, the set of equivalences classes of the
equivalence relation $\simeq$ defined such that, for any $x, y\in X$,
\[
x \simeq y \quad\hbox{iff}\quad  (\exists g\in G )(y = g\cdot x).
\] 
The {\it orbit\/} of $x\in X$ (the equivalence class of $x$) is the set
\[
O(x) = \{g\cdot x \mid g\in G\}, 
\] 
also denoted by $G\cdot x$.
If $X$ is a topological space, we give $X/G$ the quotient topology.

\medskip
For any subset $V$ of $X$ and for any $g\in G$, we denote by $gV$ the set
\[
gV = \{g\cdot x \mid x\in V\}.
\]

\medskip
One problem is that  even if $X$ is Hausdorff, $X/G$ may not be.
Thus, we need to find conditions to ensure that $X/G$ is Hausdorff.

\medskip
By a {\it discrete group\/}, we mean a group equipped with the
discrete topology (every subset is open). In other words, we don't
care about the topology of $G$!
The following conditions prove to be useful.

\begin{definition}
\label{properfree}
Let $\mapdef{\cdot}{G\times X}{X}$ be the action of a group $G$ on a
set $X$. We say that $G$ acts {\it freely\/} (or that the action is
{\it free\/})  iff  for all $g\in G$ and  all $x\in X$, if $g\not= 1$ then
$g\cdot x \not= x$.

\medskip
If $X$ is a locally compact space and $G$ is a discrete group acting
continuously on $X$,
we say that $G$ acts {\it
  properly\/} (or that the action is {\it proper\/}) iff
\begin{enumerate}
\item[(i)]
For every $x\in X$, there is some open subset $V$ with $x\in V$ such
that $gV\cap V \not= \emptyset$ for only finitely many $g\in G$.  
\item[(ii)]
For all $x, y\in X$, if $y\notin G\cdot x$ ($y$ is not in the orbit of $x$),
then there exist some open sets $V, W$ with $x\in V$ and $y\in W$ such
that $gV \cap W = 0$ for all $g\in G$.
\end{enumerate}
\end{definition}

\medskip
The following proposition gives  necessary and sufficient conditions
for a discrete group to act freely and properly often found in the
literature
(for instance, O'Neill \cite{Oneill},  Berger and Gostiaux
\cite{BergerGos},
and do Carmo \cite{DoCarmo}, but beware that in this last reference
Hausdorff separation is not required!).

\begin{proposition}
\label{freep1}
If $X$ is a locally compact space and $G$ is a discrete group, then
a smooth action of $G$ on $M$ is free and proper iff the following
conditions hold:
\begin{enumerate}
\item[(i)]
For every $x\in X$, there is some open subset $V$ with $x\in V$ such
that $gV\cap V = \emptyset$ for all  $g\in G$ such that $g\not= 1$.  
\item[(ii)]
For all $x, y\in X$, if $y\notin G\cdot x$ ($y$ is not in the orbit of $x$),
then there exist some open sets $V, W$ with $x\in V$ and $y\in W$ such
that $gV \cap W = 0$ for all $g\in G$.
\end{enumerate}
\end{proposition}

\begin{proof}
Condition (i) of Proposition \ref{freep1} 
implies condition (i) of  Definition \ref{properfree}, and
condition (ii) is the same in  Proposition \ref{freep1} 
and  Definition \ref{properfree}.
If (i) holds, then the action must be free since if $g\cdot x = x$,
then $gV\cap V \not= \emptyset$, which implies that $g =1$.

\medskip
Conversely, we just have to prove that the conditions of  Definition
\ref{properfree}
imply condition (i) of Proposition \ref{freep1}.
By (i) of Definition \ref{properfree}, there is some open
subset $U$ containing $x$ and a finite number of elements of $G$,
say $g_1, \ldots, g_m$, with $g_i \not= 1$, such that
\[
g_iU\cap U \not= \emptyset, \quad i = 1, \ldots, m.
\]
Since our action is free and $g_i \not= 1$, we have $g_i \cdot x \not
= x$, so by Hausdorff separation, there exist some open subsets
$W_i, W_i'$, with $x\in W_i$ and  $g_i\cdot x\in W_i'$, such that $W_i\cap W_i' =
\emptyset$, $i = 1, \ldots, m$.
Then, if we let
\[
V = W \cap \bigg(\bigcap_{i = 1}^m (W_i \cap g_i^{-1} W_i')\bigg),
\]
we see that $V\cap g_iV = \emptyset$, and since $V \subseteq W$,
we also have $V\cap gV = \emptyset$ for all other $g\in G$. 
\end{proof}

\remark
The action of a discrete group satisfying the properties of Proposition
\ref{freep1}
is often called ``properly discontinuous.''  However, as pointed out
by Lee (\cite{Lee}, just before Proposition 9.18), this term is
self-contradictory
since such actions are smooth, and thus continuous!

\medskip
We also need covering maps.

\begin{definition}
\label{covermap}
Let $X$ and $Y$ be two topological spaces. A map $\mapdef{\pi}{X}{Y}$
is a {\it covering map\/} iff the following conditions hold:
\begin{enumerate}
\item[(1)]
The map $\pi$ is continuous and surjective.
\item[(2)]
For every $y\in Y$, there is some open subset $W\subseteq Y$ with
$y\in W$, such that
\[
\pi^{-1}(W) = \bigcup_{i\in I} U_i,
\]
where the $U_i \subseteq X$ are pairwise disjoint open subsets such
that the restriction of $\pi$ to $U_i$ is a homeomorphism for every
$i\in I$.
\end{enumerate}
If $X$ and $Y$ are smooth manifolds, we assume that $\pi$ is smooth
and that the restriction of $\pi$ to each $U_i$ is a diffeomorphism.
\end{definition}

\medskip
Then, we have the following useful result.

\begin{theorem}
\label{quotman}
Let $M$ be a smooth manifold and let $G$ be discrete  group
acting smoothly, freely and properly on $M$.
Then there is a unique structure of smooth manifold on $M/G$ such that
the projection map $\mapdef{\pi}{M}{M/G}$ is a covering map.
\end{theorem}

For a proof, see  Gallot, Hulin, Lafontaine \cite{Gallot}
(Theorem 1.88) or Lee \cite{Lee} (Theorem 9.19).

\medskip
Real projective spaces are illustrations of Theorem \ref{quotman}.
Indeed, if  $M$ is the unit $n$-sphere $S^n \subseteq \reals^{n+1}$
and $G = \{I, -I\}$, where $-I$ is the antipodal map, then the
conditions of Proposition \ref{freep1} are easily checked (since $S^n$
is compact), and consequently the quotient
\[
\mathbb{RP}^n = S^n/G
\]
is a smooth manifold and the projection map
$\mapdef{\pi}{S^n}{\mathbb{RP}^n}$
is a covering map. The fiber $\pi^{-1}([x])$ of every point 
$[x] \in \mathbb{RP}^n$ consists of two antipodal points: $x, -x\in S^n$.

\medskip
The next step is see how a Riemannian metric on $M$ 
induces a Riemannian metric on the quotient manifold $M/G$.

\begin{definition}
\label{localisom}
Given any two Riemmanian manifolds $(M, g)$ and $(N, h)$ 
a smooth map $\mapdef{f}{M}{N}$ is a {\it local isometry\/} iff
for all $p\in M$, the tangent map
$\mapdef{df_p}{T_p M}{T_{f(p)} N}$ is an orthogonal transformation
of the Euclidean spaces $(T_p M, g_p)$ and $(T_{f(p)} N, h_{f(p)}))$.
Furthermore, if $f$ is a diffeomorphism, we say that $f$ is an {\it isometry\/}. 
\end{definition}

\medskip
The Riemannian version of a covering map is the following:

\begin{definition}
\label{covermap2}
Let $(M, g)$ and $(N, h)$ be two Riemannian manifolds. A map $\mapdef{\pi}{M}{N}$
is a {\it Riemannian covering map\/} iff the following conditions hold:
\begin{enumerate}
\item[(1)]
The map $\pi$ is a smooth covering.
\item[(2)]
The map $\pi$ is a local isometry.
\end{enumerate}
\end{definition}

\medskip
The following theorem is the Riemannian version of Theorem \ref{quotman}.

\begin{theorem}
\label{quotman2}
Let $(M, h)$ be a Riemannian manifold and let $G$ be discrete  group
acting smoothly, freely and properly on $M$, and such that
the map $x \mapsto \sigma\cdot x$ is an isometry for  all $\sigma\in G$.
Then there is a unique structure of Riemannian manifold on $N = M/G$ such that
the projection map $\mapdef{\pi}{M}{M/G}$ is a Riemannian  covering map.
\end{theorem}

\begin{proof}[Proof sketch]
For a complete proof see  Gallot, Hulin, Lafontaine \cite{Gallot}
(Proposition 2.20). To define a Riemannian metric $g$ on $N = M/G$
we need to define an inner product $g_p$ on the tangent space $T_p N$
for every $p\in N$. Pick any $q_1\in \pi^{-1}(p)$ in the fibre of
$p$. Because $\pi$ is a Riemannian covering map, it is a local
diffeomorphism, and thus  $\mapdef{d\pi_{q_1}}{T_{q_1} M}{T_p M}$ is an
isometry. Then, given any two tangent vectors $u, v\in T_p N$,  we
define their inner product $g_p(u, v)$ by
\[
g_p(u, v) = h_{q_1}(d\pi_{q_1}^{-1}(u), d\pi_{q_1}^{-1}(v)). 
\]
Now, we need to show that $g_p$ does not depend on the choice of $q_1\in \pi^{-1}(p)$.
So, let $q_2\in \pi^{-1}(p)$ be any other point in the fibre of
$p$. By definition of $N = M/G$, we have $q_2  = g\cdot q_1$ for some $g\in
G$, and we know that the map $f\co q \mapsto g \cdot q$ is an isometry
of $M$. Now, since $\pi = \pi \circ f$ we have
\[
d\pi _{q_1}=   d \pi_{q_2} \circ df_{q_1},
\]
and since   $\mapdef{d\pi_{q_1}}{T_{q_1} M}{T_p M}$ and
$\mapdef{d\pi_{q_2}}{T_{q_2} M}{T_p M}$ 
are isometries, we get 
\[
d\pi _{q_2}^{-1}= df_{q_1}  \circ d \pi_{q_1}^{-1}.
\]
But $\mapdef{df_{q_1}}{T_{q_1}M}{T_{q_2} M}$ is also an isometry, 
so 
\[
h_{q_2}(d\pi_{q_2}^{-1}(u), d\pi_{q_2}^{-1}(v)) = 
h_{q_2}(df_{q_1}(d\pi_{q_1}^{-1}(u)), df_{q_1}(d\pi_{q_2}^{-1}(v))) = 
h_{q_1}(d\pi_{q_1}^{-1}(u), d\pi_{q_1}^{-1}(v)). 
\]
Therefore, the inner product $g_p  $ is well defined on $T_p N$.
\end{proof}

\medskip
Theorem \ref{quotman2} implies that every Riemannian metric $g$ on the
sphere $S^n$ induces a Riemannian metric $\widehat{g}$ on the projective
space $\mathbb{RP}^n$, in such a way that the projection 
$\mapdef{\pi}{S^n}{\mathbb{RP}^n}$ is a Riemannian covering.
In particular, if $U$ is an open hemisphere obtained by removing its
boundary $S^{n-1}$ from a closed hemisphere, then
$\pi$ is an isometry between $U$ and its image 
$\mathbb{RP}^n - \pi(S^{n-1}) \approx \mathbb{RP}^n -
\mathbb{RP}^{n-1}$.  

\medskip
We also observe that for any two points
$p= [x]$ and $q= [y]$ in $\mathbb{RP}^n$, where $x, y\in S^n$, 
if $x\cdot y = \cos\theta$, with  $0 \leq \theta \leq \pi$, 
then there are two possibilities:
\begin{enumerate}
\item
 $x\cdot y \geq 0$, which means that $0 \leq \theta \leq \pi/2$, or
\item
 $x\cdot y < 0$, which means that $\pi/2 < \theta \leq \pi$.
\end{enumerate}
In the second case, since $[-y] = [y]$ and $x\cdot (-y) = -x\cdot y$,
we can replace the representative $y$ of $q$ by $-y$, and we have
$x\cdot (-y) = \cos(\pi - \theta)$, with $0 \leq  \pi - \theta < \pi/2$.
Therefore, in all cases, for any two points $p, q\in \mathbb{RP}^n$,
we can find an open hemisphere $U$ such that $p = [x], q = [y]$, 
$x, y\in U$, and $x\cdot y \geq 0$; that is, the angle $\theta\geq 0$
between $x$ and $y$ is at most $\pi/2$.
This fact together with the following simple proposition will allow us
to figure out the distance (in the sense of Riemannian geometry)
between two points in $\mathbb{RP}^n$.

\begin{proposition}
\label{geolem1}
Let $\mapdef{\pi}{M}{N}$ be a  Riemannian covering map between two Riemannian manifolds  
 $(M, g)$ and $(N, h)$. Then, the geodesics of $(N,h)$ are the
 projections of geodesics in $(M,g)$ (i.e., curves $\pi\circ \gamma$ in
 $(N, h),$ where
 $\gamma$ is a geodesic in $(M, g)$),
and the geodesics of $(M,g)$ are the
 liftings of geodesics in $(N,h)$ (i.e., curves $\gamma$ of $(M, g)$,
 such that $\pi\circ \gamma$ is a geodesic in $(N, h)$).
\end{proposition}

The proof of Proposition \ref{geolem1} can be found in 
Gallot, Hulin, Lafontaine \cite{Gallot} (Proposition 2.81).

\medskip
Now, if $(M, g)$ is a connected Riemannian manifold, recall that we
define the distance $d(p, q)$ between two points $p, q\in M$ as 
\[
d(p, q) = \inf\{ L(\gamma) \mid \mapdef{\gamma}{[0, 1]}{M}\},
\]
where $\gamma$ is any piecewise $C^1$-curve from $p$ to $q$,
and 
\[
L(\gamma) = \int_0^1 \sqrt{g(\gamma'(t), \gamma'(t))}\, dt
\]
is the length of $\gamma$.
It is well known that $d$ is a metric on $M$.
The Hopf-Rinow Theorem (see Gallot, Hulin, Lafontaine \cite{Gallot},
Theorem 2.103) says among other things that $(M, g)$ is geodesically
complete
(which means that every geodesics $\gamma$ of $M$ can be extended to a
geodesic   $\widetilde{\gamma}$ defined on all of  $\reals$)
iff any two points of $M$ can be joined by a minimal geodesic iff $(M, d)$ is
a complete metric space. 
Therefore, in a complete (connected) manifold
\[
d(p, q) = \inf\{ L(\gamma) \mid \mapdef{\gamma}{[0, 1]}{M} \quad\text{is a
    geodesic}\}.
\]
In particular, compact manifolds are complete, so
the distance between two points is the infimum of the length of
minimal geodesics joining these points.

\medskip
Applying this to $\mathbb{RP}^n$ and the canonical Euclidean 
metric induced by $\reals^{n+1}$, since geodesics of $S^n$ are great
circles, by the discussion above, for any two points
$p = [x]$ and $q = [y]$ in $\mathbb{RP}^n$, with $x, y\in S^n$,
the distance between them is given by
\[
d(p, q) = d([x], [y]) =
\begin{cases}
\cos^{-1}(x\cdot y)& \text{if $x\cdot y \geq 0$} \\
\cos^{-1}(- x\cdot y) & \text{if $x\cdot y < 0$}.
\end{cases}
\]
Here $\cos^{-1}(z) = \arccos(z)$ is the unique angle $\theta\in
[0,\pi]$
such that $\cos(\theta) = z$. Equivalently,
\[
d([x], [y])  = \cos^{-1}(|x\cdot y|),
\]
and
\[
d([x], [y]) =\min\{\cos^{-1}(x\cdot y), \pi - \cos^{-1}(x\cdot y)\}.
\]
If the representatives $x, y\in \reals^{n + 1}$ of $p = [x]$ and $q =
[q]$ are not unit vectors, then
\[
d([x], [y])  = \cos^{-1}\left(\frac{|x\cdot y|}{\norme{x}\norme{y}}\right).
\]
Note that $0 \leq d(p, q) \leq \pi/2$.

\medskip
Now, the Euclidean distance between $x$ and $y$ on $S^n$ is given by
\[
\norme{x - y}^2_2 = \norme{x}^2_2 + \norme{y}^2_2 - 2 x\cdot y
= 2 - 2 \cos\theta = 4\sin^2(\theta/2).
\]
Thus,
\[
\norme{x- y}_2 = 2 \sin(\theta/2), \quad 0 \leq \theta \leq \pi.
\]
It follows that for any $x\in S^n$, and for any subset $A \subseteq
S^n$, a point $a\in A$ minimizes the distance $d_{S^n}(x, a) =
\cos^{-1}(x\cdot a) = \theta$ on $S^n$ iff it minimizes the Euclidean distance
$\norme{x - a}_2 = 2\sin(\theta/2)$  (since $0 \leq
\theta\leq \pi$).  Then, on $\mathbb{RP}^n$, 
for any point $p = [x]\in \mathbb{RP}^n$ and any 
$A \subseteq \mathbb{RP}^n$, a point  $[a]\in A$ minimizes the distance
$d([x], [a])$ on $\mathbb{RP}^n$ iff it minimizes 
$\min\{\norme{x - a}_2, \norme{x + a}_2\}$.
So, we are looking for $[b]\in A$ such that
\begin{align*}
\min\{\norme{x - b}_2, \norme{x + b}_2\} & = 
\min_{[a] \in A} \min\{\norme{x - a}_2, \norme{x + a}_2\} \\
&  = \min\{ \min_{[a] \in A} \norme{x - a}_2, \min_{[a] \in A} \norme{x + a}_2\}. 
\end{align*}
If the subset $A\subseteq S^n$ is closed under the antipodal map (which
means that if $x\in A$, then $-x\in A$), then finding 
$\min_{a\in A} d([x], [a])$  on $\mathbb{RP}^n$ is equivalent to
finding $\min_{a\in A} \norme{x - a}_2$, the minimum of the Euclidean
distance. This is the case for the set $\s{X}$ in Section
\ref{ch3-sec2} and the set $\s{K}$ in
Section \ref{ch3-sec3}.

\medskip\noindent
{\bf Acknowlegments}: 
First, it must be said that the seminal and highly original work of
Jianbo Shi and Stella Yu on normalized cuts, 
was the source of inspiration for this document.
I also wish to thank Katerina Fragkiadaki for pointing out a number of
mistakes in an earlier version of this paper. Roberto Tron
also made several suggestions that contributed to improving this report.
Katerina, Ryan Kennedy, 
Andrew Yeager, and Weiyu Zhang
made many useful comments and suggestions.
Special thanks to Jocelyn  Quaintance,  Joao Cedoc  and Marcelo Siqueira who
proofread my manuscript with an eagle's eye, and made many comments that
helped me improve it.  Finally, 
thanks to Dan Spielman for making available his lovely survey on
spectral graph theory, and to
Kostas for giving me the opportunity to
hold hostage a number of people for three Friday afternoons  in a row.

   %


\addcontentsline{toc}{chapter}{Bibliography}
\begin{thebibliography}{10}

\bibitem{Bansal}
Nikhil Bansal, Avrim Blum, and Shuchi Chawla.
\newblock Correlation clustering.
\newblock {\em Machine Learning}, 56:89--113, 2004.

\bibitem{Belkin-Niyogi}
Mikhail Belkin and Partha Niyogi.
\newblock Laplacian eigenmaps for dimensionality reduction and data
  representation.
\newblock {\em Neural Computation}, 15:1373--1396, 2003.

\bibitem{BergerGos}
Marcel Berger and Bernard Gostiaux.
\newblock {\em G\'eom\'etrie diff\'erentielle: vari\'et\'es, courbes et
  surfaces}.
\newblock Collection Math\'ematiques. Puf, second edition, 1992.
\newblock English edition: Differential geometry, manifolds, curves, and
  surfaces, GTM No. 115, Springer Verlag.

\bibitem{Chung}
Fan R.~K. Chung.
\newblock {\em Spectral Graph Theory}, volume~92 of {\em Regional Conference
  Series in Mathematics}.
\newblock AMS, first edition, 1997.

\bibitem{demaine}
Eric~D. Demaine and Nicole Immorlica.
\newblock Correlation clustering with partial information.
\newblock In S.~Arora et~al., editor, {\em Working Notes of the 6th
  International Workshop on Approximation Algorithms for Combinatorial
  Problems}, LNCS Vol. 2764, pages 1--13. Springer, 2003.

\bibitem{DoCarmo}
Manfredo~P. do~Carmo.
\newblock {\em Riemannian Geometry}.
\newblock Birkh\"auser, second edition, 1992.

\bibitem{GallDiscmath}
Jean~H. Gallier.
\newblock {\em Discrete Mathematics}.
\newblock Universitext. Springer Verlag, first edition, 2011.

\bibitem{Gallbook2}
Jean~H. Gallier.
\newblock {\em {Geometric Methods and Applications, For Computer Science and
  Engineering}}.
\newblock TAM, Vol. 38. Springer, second edition, 2011.

\bibitem{Gallot}
S.~Gallot, D.~Hulin, and J.~Lafontaine.
\newblock {\em Riemannian Geometry}.
\newblock Universitext. Springer Verlag, second edition, 1993.

\bibitem{Godsil}
Chris Godsil and Gordon Royle.
\newblock {\em Algebraic Graph Theory}.
\newblock GTM No. 207. Springer Verlag, first edition, 2001.

\bibitem{Golub}
H.~Golub, Gene and F.~Van~Loan, Charles.
\newblock {\em Matrix Computations}.
\newblock The Johns Hopkins University Press, third edition, 1996.

\bibitem{Harary53}
Frank Harary.
\newblock On the notion of balance of a signed graph.
\newblock {\em Michigan Math. J.}, 2(2):143--146, 1953.

\bibitem{HornJohn}
Roger~A. Horn and Charles~R. Johnson.
\newblock {\em Matrix Analysis}.
\newblock Cambridge University Press, first edition, 1990.

\bibitem{Hou}
Jao~Ping Hou.
\newblock Bounds for the least laplacian eigenvalue of a signed graph.
\newblock {\em Acta Mathematica Sinica}, 21(4):955--960, 2005.

\bibitem{Kolluri:2004:SSR}
Ravikrishna Kolluri, Jonathan~R. Shewchuk, and James~F. O'Brien.
\newblock Spectral surface reconstruction from noisy point clouds.
\newblock In {\em Symposium on Geometry Processing}, pages 11--21. ACM Press,
  July 2004.

\bibitem{kunegis:spectral}
J\'er\^ome Kunegis, Stephan Schmidt, Andreas Lommatzsch, J\"urgen Lerner,
  Ernesto William~De Luca, and Sahin Albayrak.
\newblock Spectral analysis of signed graphs for clustering, prediction and
  visualization.
\newblock In {\em SDM'10}, pages 559--559, 2010.

\bibitem{Lee}
John~M. Lee.
\newblock {\em Introduction to Smooth Manifolds}.
\newblock GTM No. 218. Springer Verlag, first edition, 2006.

\bibitem{Oneill}
Barrett O'Neill.
\newblock {\em Semi-Riemannian Geometry With Applications to Relativity}.
\newblock Pure and Applies Math., Vol 103. Academic Press, first edition, 1983.

\bibitem{Petersen}
Peter Petersen.
\newblock {\em Riemannian Geometry}.
\newblock GTM No. 171. Springer Verlag, second edition, 2006.

\bibitem{ShiMalik}
Jianbo Shi and Jitendra Malik.
\newblock Normalized cuts and image segmentation.
\newblock {\em Transactions on Pattern Analysis and Machine Intelligence},
  22(8):888--905, 2000.

\bibitem{Spielman}
Daniel Spielman.
\newblock Spectral graph theory.
\newblock In Uwe Naumannn and Olaf Schenk, editors, {\em Combinatorial
  Scientific Computing}. CRC Press, 2012.

\bibitem{Luxburg}
von Luxburg~Ulrike.
\newblock A tutorial on spectral clustering.
\newblock {\em Statistics and Computing}, 17(4):395--416, 2007.

\bibitem{Yu}
Stella~X. Yu.
\newblock {\em Computational Models of Perceptual Organization}.
\newblock PhD thesis, Carnegie Mellon University, Pittsburgh, PA 15213, USA,
  2003.
\newblock Dissertation.

\bibitem{YuShi2003}
Stella~X. Yu and Jianbo Shi.
\newblock Multiclass spectral clustering.
\newblock In {\em 9th International Conference on Computer Vision, Nice,
  France, October 13-16}. IEEE, 2003.

\end{thebibliography}
\bibliographystyle{plain} 
\end{document}